%% file: main.tex
\documentclass[10pt]{article}
\usepackage[T1]{fontenc}
\usepackage{amsmath}
\usepackage{amssymb}
\usepackage{amsthm}
\usepackage{booktabs}
\usepackage{longtable,array}
\usepackage[round,authoryear]{natbib}
\usepackage{xcolor,hyperref}
\hypersetup{pdfauthor={Bowen Xu, Boyu Chen, Jiaying Li, Haiyi Li}}
\usepackage{pgfplots}
\pgfplotsset{compat=1.18}
\usepgfplotslibrary{groupplots}
\usepackage{placeins}
\usepackage{style/biowm_preprint}
\theoremstyle{plain}
\newtheorem{proposition}{Proposition}

\theoremstyle{definition}
\newtheorem{definition}{Definition}
\newtheorem{assumption}{Assumption}
\theoremstyle{remark}
\newtheorem{remark}{Remark}
\input{macros/all_macros.tex}
\title{Stand-In Effects and Pooling Amplification: Lessons from Biological World Models for Testing Action Effects}
\author{%
  \begin{minipage}{\textwidth}
    \centering\normalsize
    \textbf{Bowen Xu}\textsuperscript{$*\dagger$}\quad
    {\small\itshape bxu666@stanford.edu}\\
    {\small\itshape Department of Biology, Stanford University, Stanford, CA, USA}\\[0.8em]
    \textbf{Boyu Chen}\textsuperscript{$*$}\\
    {\small\itshape Institute of Health Informatics, University College London, London, UK}\\[0.8em]
    \textbf{Jiaying Li}\\
    {\small\itshape Beijing Tongren Eye Center, Beijing Tongren Hospital, Capital Medical University, Beijing, China}\\[0.8em]
    \textbf{Haiyi Li}\\
    {\small\itshape Google, Mountain View, CA, USA}\\
    {\small\itshape\textsuperscript{$*$}Equal contribution.\quad
      \textsuperscript{$\dagger$}Corresponding author.}
  \end{minipage}%
}
\date{}
\begin{document}
\maketitle
\input{sections/00_abstract.tex}
\input{sections/01_intro.tex}
\input{sections/02_problem_setup.tex}
\input{sections/03_method.tex}
\input{sections/04_theory.tex}
\input{sections/05_synthetic_validation.tex}
\input{sections/06_pooling_amplification.tex}
\input{sections/07_standin_trap.tex}
\input{sections/08_real_data.tex}
\input{sections/08b_published_model.tex}
\input{sections/08c_two_models.tex}
\input{sections/11_robot_data.tex}
\input{sections/12_overbetting.tex}

\input{sections/13_partial_execution.tex}
\input{sections/14_reading_evidence.tex}

\input{sections/15_trajectory_evidence.tex}
\FloatBarrier
\input{sections/09_discussion.tex}
\input{sections/10_statements.tex}
\clearpage
\appendix
\input{appendix/A_biowm_eval.tex}
\input{appendix/B_proofs.tex}
\input{appendix/B_v5_proofs.tex}
\input{appendix/C_assignment_laws.tex}
\input{appendix/D_tables_disclosures.tex}
\input{appendix/E_osf_hashes.tex}
\input{appendix/F_erk_field_diag.tex}
\input{appendix/G_pixel_full_grid.tex}
\input{appendix/H_baselines.tex}
\input{appendix/I_published_model.tex}
\input{appendix/J_two_models.tex}
\input{appendix/K_exploratory.tex}
\input{appendix/L_robot_data.tex}
\input{appendix/M_fourth_registration.tex}
\input{appendix/N_matched_comparators.tex}
\input{appendix/O_fitted_stakes.tex}
\input{appendix/P_degraded_action.tex}
\input{appendix/Q_thirty_tasks.tex}
\input{appendix/R_references.tex}
\input{appendix/S_occupancy.tex}
\input{appendix/T_fresh_trajectories.tex}
\input{appendix/U_trajectory_uncertainty.tex}
\FloatBarrier
\input{sections/16_use_of_ai.tex}
\bibliographystyle{plainnat}
\bibliography{refs,refs_v3,refs_v4,refs_v5}
\end{document}

%% file: sections/00_abstract.tex
\begin{abstract}
Action-conditioned world models are increasingly trained on data from randomized experiments, and whether their predictions respond to actions is usually checked by replacing or removing the actions and measuring the loss in accuracy. In autocorrelated designs, future actions can stand in for history the model does not see, so replacement diagnostics can report effects even under a sharp null. A predictor that learns this dependence can strengthen the false signal. We express a conditional randomization test as an e-value. When the conditional design law is known and the score, timing, eligibility and decoy conditions of Proposition 1 hold, the test gives finite-sample validity for any admissibly frozen model. When many sites share an action sequence, pooling can amplify a shared score error until the e-value is near zero, despite a real effect. In a pixel world, its median log e-value fell from \PxTTwoCzeroTwoModelAvgNoneDHiPixelsSixtyFourTrainSixtyFourGridGOneMedianLogEDisp{} to \PxTTwoCzeroTwoModelAvgNoneDHiPixelsFortyNinetySixTrainSixtyFourGridGOneMedianLogEDisp{} as frames grew from $64$ to $4096$ pixels, while the oracle's rose. Fixed admissible tempering preserves the design-based validity guarantee. In our preregistered reanalysis of optogenetic ERK data, the pooled statistic failed its positive controls. A tempered statistic selected on synthetic data and registered before we obtained the confirmatory data rejected on one previously unopened field under the reconstructed assignment law ($E\approx\MfiveEHtenScientific$). Every single tempering factor we tried---estimated from residuals, or tuned on validation data---collapsed or lost most of the evidence in at least one setting of our first three registered studies and their development work, whereas mixtures over a fixed grid of factors did not collapse in any of them; they were weaker where effects were weak, and when actions were executed only part of the time a statistic with a bounded stake rejected in more units (Section 13). For a one-step or single-residue product, a fraction of a profitable temperature retains at least that fraction of its log e-value. Temperatures above the product's break-even point give negative log e-values. We tested the mixtures in three registered studies on world models trained and released by others. On a diffusion world model for three Atari games, the emulator executed the uniformly drawn recorded actions in the 96 ACTIVE test units. The pooled test and Kish tempering each rejected in 0 units, contrast tempering in 1, and the mixture in all 96. On IRIS, a token-based world model for 26 Atari games, and TD-MPC2, a latent world model for six control tasks, the pooled test reached the registered power bar of 24 of 32 units in 10 of the 26 games and in 1 of the 6 tasks; the registered validation-weighted mixture reached it in 24 games and all 6 tasks. On recorded robot video scored by iVideoGPT, the weighted mixture rejected in 155 of 159 trajectories and the pooled statistic in 143 of the same 159 trajectories. Both counts describe sensitivity with the recorded commands under a reconstructed assignment law and undocumented command and frame timing. The constructed label-null arm used assignments we controlled. Under constructed exact nulls the statistics of the three studies made 14 rejections in 26,664 tests of several statistics on shared units. The DIAMOND and IRIS/TD-MPC2 fits were condition-specific, with some NULL-selected temperatures near the bottom of the grid. A fourth registered study repeated an exploratory design on new seeds using one TD-MPC2 model family. At one step its bounded-stake mixture SW rejected in 376 of 576 partial-execution units and ordinary temperature scaling T in 248. In a separate exploratory analysis, fitted stakes were within a factor of two of the execution probability in 56 of 61 cells above a prespecified full-execution evidence threshold. With the score calibrated, the mean log e-value per step measures how well a scorer tells the recorded action label from design draws. In exploratory studies the largest of five released models had more of this evidence than the smallest in all 30 control tasks. Under uniform actions the original inverse-dynamics reference exceeded the 317M model in all 30 tested tasks. Under policy-centred actions a tested reference exceeded it in every task across the original evaluation and a follow-up on fresh units for the focal tasks. The evidence also depended on where a trajectory had been: with the conditional law of the tested action, the configuration and the reference held fixed, two released world models kept only 6\% to 33\% of their evidence at policy-centred steps in four pole tasks when each step of the trajectory was drawn uniformly with probability one half, while a reference critic kept 73\% to 80\%. The fall of one model was forecast and that of the other was larger than forecast. The pattern recurred on fresh trajectories using the frozen procedure. For each world model, the pointwise trajectory-bootstrap interval for its midpoint retention minus the critic's lay below zero in each of the four pole tasks. Every analysis is preregistered, fixed in a dated protocol before it was run, or labelled exploratory, and every reported result is generated from hashed result files.
\end{abstract}

%% file: sections/01_intro.tex
\section{Introduction}\label{sec:intro}

World models that predict how a system will evolve under a sequence of actions are now trained on data from experiments as well as from simulators. In biology the actions are often stimulation programs---light pulses delivered to optogenetically engineered cells, for example---and the hope is that a model trained on such data will answer interventional questions: what would this cell population do under a different program? Before such a model is trusted for that purpose, one wants to know whether its action channel reflects a causal effect of the actions at all. In machine learning this is usually checked with diagnostics that replace the actions by zeros, by another unit's actions or by a shuffled sequence, or retrain the model without them, and report an effect if predictions get worse. Recent benchmarks for action following in world models instead compare generated rollouts with simulator replays of expert and off-expert actions~\citep{worldecho2026,worldsimprobe2026}, or score rollouts under perturbed or commanded actions with trajectory estimators, vision-language judges or human annotation~\citep{mirabench2026,worldexam2026}; \citet{admissibility2026} propose a ladder of checks that a world-model simulator must pass before its verdicts are trusted. None of them reports a test with finite-sample error control. Neither route is available, or reliable, when the data come from a real experiment that cannot be replayed.

\begin{figure}[t]
\centering
\includegraphics[width=\linewidth]{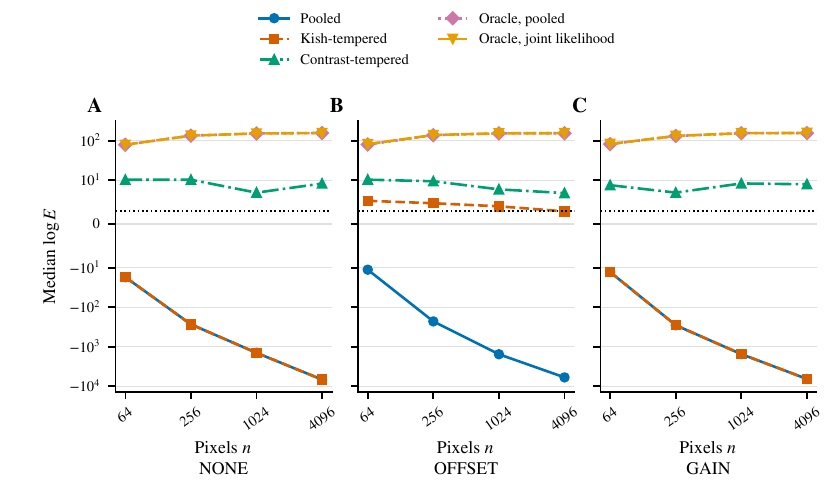}
\caption{Pooling amplification in a pixel world model. Median $\log E$ over test episodes at the larger effect size as frames grow from $64$ to $4096$ pixels, without shared noise (A), with a frame-wide additive offset (B) and with a multiplicative gain on the blob (C). The pooled, Kish-tempered and contrast-tempered statistics are averages over the two learned models; the oracles use the true dynamics with independence pooling or with the exact joint likelihood. In A and C the Kish factor is close to one and the Kish-tempered curve coincides with the pooled one; the two oracle curves nearly coincide in every panel. Dotted line: $\log 20$, the rejection threshold. Symmetric-log axis, linear for $|\log E|\le10$.}
\label{fig:pooling}
\end{figure}

\paragraph{The stand-in trap.} Replacement diagnostics fail in a simple way. Experimental designs are often autocorrelated: the next stimulus depends on the stimuli before it. A model that sees only a window of recent history can then use the future actions as a stand-in for earlier history that still drives the outcome. Changing the future actions can then worsen prediction even when those actions have no effect. In the construction of Proposition~\ref{prop:P4}, the three specified replacement and retraining diagnostics report an effect with probability tending to one for the ideal predictors. In development simulations the replacement diagnostics reported effects in up to \MtwoaDevProxyTrapPooledPerModelAcrossDesignsDynamicsBswapRejectionsMax{} replicates of such a null with twelve units, and in \MtwoaDevProxyTrapMarkovLinearUnitsFiveTwelvePooledPerModelRangeBswapRejectionsMin{} for every model with $512$ units; in a pixel world in which the logging policy depends on state the model cannot see they did so in \PxcheckTrapDiagRatePerModelLGNoneAcrossTrainingSizesHfourPixelsTwoFiftySixMin{} to \PxcheckTrapDiagRatePerModelCNNNoneAcrossTrainingSizesHfourPixelsTwoFiftySixMax{} of test episodes (\S\ref{sec:trap}, Figure~\ref{fig:trap}).

\paragraph{A design-based test.} A known conditional assignment law supplies the resampling distribution. Under the sharp null and the conditions of Proposition~\ref{prop:P1}, the recorded block and the resampled blocks are exchangeable in the Monte Carlo bag. The world-model score supplies the statistic for the conditional randomization test~\citep{candes2018panning}. The e-value construction multiplies along nonoverlapping time grids and, under the stated conditional product law, across the permitted units. Prechosen averages across grids and models are valid at the fixed analysis time (\S\ref{sec:method}). The conditioning matters: in a post hoc analysis, a resampler that draws each action from its marginal frequency rejected the same pixel-world null in \PosthocNcWrongRateTwoModelAvgAcrossNoiseAcrossTrainingSizesHfourPixelsTwoFiftySixMin{} to \PosthocNcWrongRateTwoModelAvgAcrossNoiseAcrossTrainingSizesHfourPixelsTwoFiftySixMax{} of episodes, while the design-law resampler rejected in none for the two-model average (\S\ref{sec:trap:nc}).

\paragraph{Pooling amplification.} World models raise a second difficulty that does not arise for a single outcome. In an ERK optogenetics experiment, each microscope field receives one light program and records hundreds of cells; in a pixel world, every pixel of a frame shares the frame's actions. The natural statistic sums the sites' log predictive densities and bets on the result. If a common alternative block has a score advantage per site bounded below by a positive constant as sites are added, its total advantage grows at least linearly. The exact log factor then has an upper bound that decreases at least linearly if that block's design mass stays bounded away from zero. The analogous upper bound holds in fixed-size Monte Carlo bags containing the block (\S\ref{sec:method:free}). We met this failure in our own registered analysis: on the ERK positive controls, where light is known to drive the reporter, the pooled statistic gave $\log E=\MfourLogESlotOneDisp$ (\S\ref{sec:real:m4}). In the pixel world, its median $\log E$ fell from \PxTTwoCzeroTwoModelAvgNoneDHiPixelsSixtyFourTrainSixtyFourGridGOneMedianLogEDisp{} to \PxTTwoCzeroTwoModelAvgNoneDHiPixelsFortyNinetySixTrainSixtyFourGridGOneMedianLogEDisp{} as frames grew from $64$ to $4096$ pixels, while the oracle's rose from \PxTTwoOracleCzeroOracleNoneDHiPixelsSixtyFourTrainSixtyFourGridGOneMedianLogEDisp{} to \PxTTwoOracleCzeroOracleNoneDHiPixelsFortyNinetySixTrainSixtyFourGridGOneMedianLogEDisp{} (\S\ref{sec:pooling}, Figure~\ref{fig:pooling}).

\paragraph{A repair that cannot break validity.} Because the design, not the statistic, controls the error rate, the pooled score may be tempered by any factor fixed on training information that meets the conditions of Proposition~\ref{prop:P1}. We temper by an inverse design effect: the statistic registered for our confirmatory test uses the Kish factor of the residual correlation, and an exploratory variant measures the design effect along the action contrast, which is what governs amplification (\S\ref{sec:method:tempering}). The Kish-tempered statistic, selected on synthetic data, gave $\log E=\XLogEHtenDisp$ on the same ERK test fields in an exploratory reanalysis. In a confirmatory test registered before we obtained the field's data, it rejected on one ERK field that no one involved had opened, under the reconstructed assignment law, with $E\approx\MfiveEHtenScientific$ (\S\ref{sec:real:m5}). In the pixel world, where the Kish factor does not capture the relevant correlation, contrast tempering restored power under every noise condition we simulated, at the larger of the two effect sizes. Estimated factors are fragile, however. The Kish correlation factor and the contrast factor do not change when all per-site precisions are multiplied by one common factor, with predictions, residuals and contrasts held fixed, whereas for unfloored Gaussian scores the tempered log-e curve shifts along the log-temperature axis (Proposition~\ref{prop:blind}). A mixture over $G$ fixed tempering factors needs no estimated factor and loses at most $\log G$ in log e-value relative to the best member of that grid chosen in hindsight (Proposition~\ref{prop:mixture}); in an exploratory comparison on the saved pixel and ERK-like runs it matched contrast tempering in the pixel world and Kish tempering on ERK-like data (\S\ref{sec:pooling:baselines}). A weighted version that puts most of its weight on the factor chosen on validation data keeps most of validation tuning's power when that choice transfers, and the mixture's protection when it does not (\S\ref{sec:method:robust}, \S\ref{sec:regime}, \S\ref{sec:erkxm}). For a one-step or single-residue product, a fraction of a profitable temperature retains at least that fraction of its log e-value. Temperatures above the product's break-even point give negative log e-values.

\paragraph{Published world models.} If pooling amplification is a property of pooled scores and not of our models, it should appear on world models trained by others. We registered three confirmatory tests on released checkpoints that we neither trained nor modified. In the DIAMOND study (\citealp{alonso2024diamond}), the emulator executed the uniformly drawn recorded actions in 96 ACTIVE test units across Boxing, Pong and Breakout. The pooled score and Kish tempering each rejected in 0 units, contrast tempering in 1, and the primary mixture in all 96 (\S\ref{sec:published}). The second test covered IRIS, a token-based Transformer world model for 26 Atari games (\citealp{micheli2023transformers}), and TD-MPC2, a latent world model for continuous control (\citealp{hansen2024tdmpc2}), with a preregistered validation-weighted mixture rule as the primary statistic. Under IRIS's own token likelihood the pooled score reached the registered power bar, 24 of 32 units, in only 10 of the 26 games; under TD-MPC2's latent score it did so in 1 of 6 tasks. The weighted mixture reached it in 24 of 26 games and in all 6 tasks (\S\ref{sec:p4}). The third test used recorded video of a real robot, which cannot be replayed: BAIR robot pushing (\citealp{ebert2017self}), scored by iVideoGPT (\citealp{wu2024ivideogpt}). Its null arm is exact by construction; its arm with the recorded commands assesses detection under a law reconstructed from the command records. Under that reconstructed law, and with the timing of commands relative to frames undocumented, the weighted mixture rejected in 155 of 159 trajectories; this count is a sensitivity result. The pooled score did not collapse there: it rejected in 143, and failed badly only in a few trajectories, where the mixture's loss stayed bounded (\S\ref{sec:p5}). Across the three studies, the reported statistics rejected 14 times in 26,664 constructed-null tests of several statistics on shared units, which are not independent trials. The DIAMOND and IRIS/TD-MPC2 validation fits were specific to each condition, and some factors selected on NULL validation units were near the bottom of the grid, where a statistic hardly moves. The robot study therefore also ran NULL checks of V and W(3/4) with the factors selected on ACTIVE units; neither rejected under either law.

\paragraph{Stakes, and reading the evidence.} A temperature mixture limits the loss relative to the best temperature on its grid. Under partial execution a bounded stake also limits the loss at each origin. In a registered repeat of an exploratory design on new TD-MPC2 seeds, SW rejected in 376 of 576 one-step partial-execution units and T in 248. Separate exploratory fits had stakes within a factor of two of the execution probability in 56 of 61 cells above a prespecified full-execution evidence threshold. Once the score is calibrated, the mean log e-value per step is a number about the scorer, and it can be read against reference scorers that use no world model (\S\ref{sec:v5-evidence}). In exploratory studies with released TD-MPC2 models, the six-task sum fell under the tested action-input attenuation, with exceptions in individual tasks. The evidence rate was higher for the largest model than for the smallest in all 30 tasks tested. It also depended on where the trajectory had been: with the conditional law of the tested action, the configuration and the reference held fixed, the evidence of two released world models at policy-centred steps fell to between 6\% and 33\% of its value in four pole tasks when each step was drawn uniformly with probability one half, while a reference critic kept 73\% to 80\% (\S\ref{sec:v5-trajectories}). The fall of the single-task model was forecast before the experiment, the fall of the multitask model was larger than forecast, and the pattern recurred on fresh trajectories with every fitted quantity frozen.

\paragraph{Contributions.}
\begin{enumerate}
\item We describe the stand-in trap for action diagnostics and give a construction in which the specified replacement and retraining diagnostics detect a nonexistent effect with probability tending to one (Proposition~\ref{prop:P4}). In that construction, resampling from the correct conditional assignment law controls size. For feedback designs, Proposition~\ref{prop:closedloop} extends validity under Assumption~\ref{ass:design-cl} and gives a sufficient delayed-policy construction.
\item We identify pooling amplification, a failure of pooled scores when many sites share an action sequence. A common competing block with a persistent positive per-site average score advantage can drive the factor down under the conditions in \S\ref{sec:method:free}. Our validation check missed this failure (\S\ref{sec:synthetic:pilot}). Across site counts, the realised trend followed this direction in 29 of 36 groups, not in all of them as forecast (Table~\ref{tab:v5-regime-predictions}).
\item We show that within the randomization test, tempering is merely a choice of statistic, fixed on admissible training information and, under the conditions of Proposition~\ref{prop:P1}, incapable of affecting validity, and we propose measuring the design effect along the action contrast; when the contrast is shared across sites this reduces to a weighted Kish factor (Proposition~\ref{prop:kish}). We show that the Kish and contrast factors are invariant under a common rescaling of the per-site precisions that moves the tempered log-e curve, and that a mixture over a fixed grid of factors loses at most the logarithm of the grid size in log e-value relative to the best of them (Propositions~\ref{prop:growth}--\ref{prop:mixture}); a version that puts most of its weight on the factor chosen on validation data loses at most log(4/3) relative to that factor (\S\ref{sec:method:robust}).
\item We give pathwise bounds on the evidence a statistic can achieve. For known open-loop assignment and a fixed analysis schedule, the command-path law gives a bound on rejection probability (Proposition~\ref{prop:P2}).
\item The two registered simulator studies on DIAMOND, IRIS and TD-MPC2 document pooling amplification and repairs on released checkpoints. The registered iVideoGPT study on robot video provides a constructed-null check and a working-law sensitivity analysis with the recorded commands (\S{}\S{}9–11). On the robot video and in three Atari games the pooled score of a model's own token likelihood did not collapse, and we say where it does and does not (\S\ref{sec:discussion}).
\item We show that, for a one-step product and for each residue-class product at longer horizons, the log e-value is concave in the tempering factor and zero at zero (Proposition~\ref{prop:overbet}); this need not hold for the logarithm of an average over residue classes. Hence every negative tempered log e-value of such a product is an over-bet, a fraction of a profitable temperature retains at least that fraction of its log e-value, and a mixture over a dyadic grid keeps at least half of the log e-value of any factor in its range whose log e-value is non-negative, minus the logarithm of the grid size; relative to the best member of the grid it loses at most the logarithm of the grid size. In 4512 saved units, every observed collapse was an over-bet (\S\ref{sec:overbet}).
\item We report a complete, preregistered validation pipeline for world-model action tests on real experiments, including its failures, with every reported number generated from committed, hashed result files.
\item On six TD-MPC2 tasks, SW rejected in 376 of 576 one-step partial-execution units against 248 for T in a registered repeat on new seeds. Matched exploratory comparisons on the same stored scores found more rejections when a stake was allowed. On IRIS tokens the partial-execution differences were small (\S{}13).
\item We read the calibrated evidence per step as a property of the scorer. In exploratory studies on a latent world-model family, the six-task sum fell under the tested action-input attenuation, with exceptions in individual tasks. The evidence rate was higher for the largest model than for the smallest in all 30 tasks tested. Under uniform actions the original inverse-dynamics reference exceeded the 317M model in all 30 tested tasks. Under policy-centred actions a tested reference exceeded it in every task across the original evaluation and a follow-up on fresh units for the focal tasks. Under uniform actions, 7 of the 30 tasks, all pendulum-like, had a model with near-best control return and at most a quarter of the reference's evidence (\S\ref{sec:v5-evidence}).
\item We show that this evidence depends on where the trajectory has been. With the conditional law of the tested action, the configuration and the reference held fixed, the evidence of two released world models at policy-centred steps fell to between 6\% and 33\% of its value in four pole tasks once each step was uniform with probability one half; a reference critic kept most of its evidence, and in two locomotion tasks the models kept at least four fifths. The single-task model's fall was forecast and the multitask model's was larger than forecast; the pattern recurred on fresh trajectories with a frozen procedure. In an exploratory analysis of four pole tasks the mixed-training forward scorer had higher midpoint evidence than either released TD-MPC2 scorer, with different training and score definitions. On the fresh trajectories, pointwise trajectory-bootstrap intervals for the midpoint change lay below zero for both released models in each pole task. The interval for the single-task model's change minus the critic's also lay below zero in each of the four tasks (\S\ref{sec:v5-trajectories}).
\end{enumerate}

\paragraph{Related work.} Randomization tests for experiments in which treatments act over time are established: exact tests of the sharp null of no temporal causal effect, with lag-$p$ causal estimands, for time-series experiments~\citep{bojinov2019time}; conditional randomization tests of partially sharp nulls indexed by a lag in stepped-wedge trials, and by a distance for spillover~\citep{zhang2025randomization}; and tests of the carryover horizon and of non-anticipation in switchback experiments~\citep{liu2026switchback}. The use of machine-learning predictions as randomization-test statistics is also established~\citep{tansey2022holdout,ham2024conjoint,guo2025ml}; we use both and claim neither. E-values and testing by betting~\citep{shafer2021testing,vovk2021evalues,ramdas2023game,grunwald2024safe} give the aggregation we rely on; model-X conditional-independence tests with e-values~\citep{grunwald2024anytime} and sequential Monte Carlo tests~\citep{fischer2024sequential} are close relatives. Tempering likelihoods under misspecification is established in generalized and safe Bayesian inference~\citep{grunwald2017inconsistency}, and learning or weighting a grid of learning rates in online learning~\citep{koolen2014learning,vanerven2016metagrad}; we make no priority claim for tempering, for mixtures or for rank calibrators. The tempering formulas are standard; \S\ref{sec:method:tempering} and \S\ref{sec:method:betting} give their sources, and the contrast factor resembles the score-based rescalings used in composite likelihood. Criticism of permutation-based feature importance for forcing extrapolation~\citep{hooker2021unrestricted,borgonovo2026} and work on causal confusion, copycat behaviour and policy confounding in imitation and reinforcement learning identify related shortcuts; those works ask whether a policy has learned the wrong cue and repair it with targeted interventions~\citep{dehaan2019causal}, adversarial removal of the nuisance information~\citep{wen2020fighting}, or a broader distribution of trajectories~\citep{suau2024bad}, whereas we ask whether a world model's action channel reflects an effect and use only the known design, with finite-sample error control. Design-respecting world models~\citep{yu2026design} and advantage-style action channels that contrast a world model's predictions across actions~\citep{dueling2026} are conceptually adjacent. In biology, randomized optogenetic stimulation has been used for causal inference~\citep{loewinger2024optogenetics}; we analyze two published optogenetics datasets~\citep{naleczjawecki2023erk,lugagne2024ccas}, released as \citet{naleczjawecki2022erkdata} and \citet{lugagne2023ccasdata}.

\paragraph{Organization.} Section~\ref{sec:setup} sets up units, sites, designs and nulls; \S\ref{sec:method} gives the test and the tempered statistics; \S\ref{sec:theory} states the theory. Sections~\ref{sec:synthetic}--\ref{sec:trap} report synthetic validation, pooling amplification and its repair, and the stand-in trap; \S\ref{sec:real} reports the real-data evidence, and \S\ref{sec:published}--\S\ref{sec:p5} the registered tests on published world models, the last on recorded robot video; \S\ref{sec:overbet} gives one account of every collapse; \S\ref{sec:v5-stakes} treats partial execution and bounded stakes, with a fourth registered study; \S\ref{sec:v5-evidence} reads the calibrated evidence against reference scorers; \S\ref{sec:v5-trajectories} follows it as trajectories leave the policy's own; \S\ref{sec:discussion} compares all settings and discusses limitations. The appendices give proofs, design laws, all tables and disclosures, the registrations and hashes, the baseline study, the full records of the three published-model tests, and the exploratory studies run after the second version of this paper, and, new in the fifth version, the record of the fourth registered study and of seven further exploratory analyses.

%% file: sections/02_problem_setup.tex
\section{Problem setup}\label{sec:setup}

\subsection{Units, sites and actions}\label{sec:setup:units}

We observe assignment units $i=1,\dots,N$ at discrete times $t$. At each time unit~$i$ receives an action $A_{i,t}\in\mathcal{A}$, and outcomes $Y_{i,c,t}$ are recorded at each of its measurement sites $c\in\mathcal{C}_i$. All sites of a unit share the unit's action sequence: actions are assigned to units, not to sites. This distinction matters throughout the paper. In the CcaS/CcaR optogenetic experiment a unit is one cell lineage with a single site; in the ERK experiment a unit is a microscope field, and its sites are the hundreds of cells that receive the same light program; in a pixel world model a unit is an episode and its sites are pixels.

Write $a=(a_{j,s})$ for a complete assignment of actions to all units and times, and $Y_{i,c,s}(a)$ for the corresponding potential outcome. We assume no anticipation: $Y_{i,c,s}(a)$ depends on $a$ only through $(a_{j,r})_{r\le s}$. Observed outcomes are $Y_{i,c,s}=Y_{i,c,s}(A)$. For an origin $t$ and a horizon $h\ge 1$, the \emph{future block} of unit~$i$ is $A^{F}_{i,t}=(A_{i,t+1},\dots,A_{i,t+h})$ and the \emph{targets} are $Y_{i,c,t+h}$. Let $\mathcal{F}_t$ denote the observed joint history up to $t$: all units' outcomes and actions up to $t$, the clock, and any further covariates that are functions of the past.

\subsection{Randomized open-loop design}\label{sec:setup:design}

\begin{assumption}[Design]\label{ass:design}
For every origin $t$, conditionally on $\mathcal{F}_t$ and on all potential outcomes,
\[
A^{F}_{\cdot,t}\;\sim\;\textstyle\prod_{i} Q_{i,t},
\]
where $Q_{i,t}$ is a known law over future blocks that depends only on unit~$i$'s own past actions $A_{i,\le t}$ and on a design stratum fixed in advance.
\end{assumption}

Assumption~\ref{ass:design} states that actions were generated by an open-loop randomized program that did not look at outcomes. We call $Q_{i,t}$ the \emph{design law}. In the experiments we analyze it is reconstructed from the published protocol and acquisition code (Appendix~\ref{app:assignment}); every validity statement is conditional on that reconstruction, which we call a working law, and approximate laws are treated in \S\ref{sec:theory}. Two features of the design law matter below. First, $Q_{i,t}$ may depend on the unit's \emph{entire} action history, typically far more of it than a predictive model conditions on. Second, a unit whose design is deterministic at $t$ (a zero-entropy law) carries no randomization information, and contributes an e-value of exactly one.

\paragraph{Designs that read past outcomes.} Assumption~\ref{ass:design} describes open-loop programs. Some experiments, including the trap experiment of \S\ref{sec:trap}, use a logging policy that reads earlier outcomes or states. The validity results extend to such designs under the following assumption.

\begin{assumption}[Design with delayed feedback]\label{ass:design-cl}
For every origin $t$, conditionally on $\mathcal{F}_t$, on all potential outcomes and on the frozen training and design information,
\[
A^{F}_{\cdot,t}\;\sim\;\textstyle\prod_{i} Q_{i,t},
\]
where each $Q_{i,t}$ is known and $\mathcal{F}_t$-measurable. For the own-channel null of \S\ref{sec:setup:nulls} with a tested independent set $I$, the same product form holds for the blocks of the units in $I$ conditionally also on the future blocks of all units outside $I$.
\end{assumption}

Assumption~\ref{ass:design} is the special case in which $Q_{i,t}$ depends only on the unit's own past actions and its stratum. The product form is essential: a known marginal law for each unit does not justify multiplying unit-wise factors when one unit's policy can react to another unit's actions inside the block. Proposition~\ref{prop:closedloop} shows that the validity results hold under Assumption~\ref{ass:design-cl}, and gives a sufficient condition: each unit's policy reads observed history only with a lag of at least $h$, never another unit's actions inside the block, and uses fresh randomization independent of everything else.

\subsection{Partial sharp nulls}\label{sec:setup:nulls}

\begin{definition}[Global sharp-in-future null]\label{def:H0G}
$H_0^{G}(h)$: for every unit $i$, site $c$ and origin $t$, $Y_{i,c,t+h}(a)=Y_{i,c,t+h}(a')$ whenever $a$ and $a'$ differ only in the future blocks $a^{F}_{\cdot,t}$ of any units.
\end{definition}

\begin{definition}[Own-channel null]\label{def:H0own}
Given a spillover graph $\mathcal{N}$ on units, $H_0^{\mathrm{own}}(h;\mathcal{N})$: for every unit $i$, site $c$ and origin $t$, $Y_{i,c,t+h}(a)$ is unchanged when $a^{F}_{i,t}$ alone is changed. Future actions of $\mathcal{N}$-neighbours may affect $Y_{i,c,t+h}$; units that are not neighbours have no effect.
\end{definition}

Both nulls are \emph{sharp}: they constrain every unit's potential outcomes, not an average effect. A rejection therefore means that some site responds to some future action, not that the response is large or typical. Both are \emph{partial}: they concern only the effect of the next $h$ actions on the outcome at $t+h$. Actions before the window may have arbitrary effects---this is the dependence that makes action-conditioned prediction possible. At the same target times, the null for horizon $h$ therefore implies the null for every shorter horizon, but not for a longer one, whose window includes earlier actions. Shared noise of any form, across sites or across units, is allowed. Rejecting $H_0^{G}$ means only that global future-action invariance fails; attributing an effect to a unit's own actions requires $H_0^{\mathrm{own}}$ and its layout assumptions. Conditional randomization tests of partially sharp nulls are not new: \citet{zhang2025randomization} test nulls indexed by a time lag for lagged treatment effects in stepped-wedge trials, assuming no interference and no anticipation, and nulls indexed by a distance threshold for the range of spillover effects; our contribution is not the null, but the analysis of what goes wrong when it is tested with learned models (\S\ref{sec:setup:standin} and \S\ref{sec:method:free}).

\paragraph{Admissible statistics.} A statistic for unit~$i$ at origin $t$ may use any $\mathcal{F}_t$-measurable covariate, unit~$i$'s own targets $Y_{i,c,t+h}$, and unit~$i$'s candidate future block, and nothing else. In particular, no statistic may use other units' future actions, and under $H_0^{\mathrm{own}}$ none may use other units' targets, which can respond to unit~$i$'s future actions through spillover. This is a requirement to be verified for each dataset, not an assumption: in the CcaS release, for example, the neighbour-exposure covariate is aligned one step ahead of the neighbours' commands and must be replaced by its lag (Appendix~\ref{app:assignment}). Which sites contribute at an origin is determined by an observation mask (finite own history and target). That this mask does not respond to future actions under the null is a stated assumption about the measurement process; formally, the measurement process, including any randomness in it, is part of the potential outcomes on which Assumption~\ref{ass:design} conditions.

\subsection{Stand-in effects}\label{sec:setup:standin}

An action-conditioned world model predicts $Y_{i,c,t+h}$ from a finite window of the past together with the future block. Under an autocorrelated design the future block is correlated with actions that precede the window, and those earlier actions may affect the target through the system's memory. The future block then carries information about the target that the window lacks, even when $H_0^{G}$ holds: future actions act as a \emph{stand-in} for omitted history. A model trained to predict will use this information, and its predictions will change when the future block is changed.

Common action diagnostics read this change as an action effect. They compare predictions under the observed block with predictions under inputs that break the stand-in dependence---an all-zero block, a block taken from another unit or another time, or a model retrained without actions---and report an effect when the prediction loss rises. We call all of them \emph{replacement diagnostics} (the pixel protocol of \S\ref{sec:trap} and Appendix~\ref{app:pixel-grid} calls them common diagnostics). Nothing in these comparisons controls how often they do so when no effect exists, and a model that predicts better learns the stand-in better. Proposition~\ref{prop:P4} gives a construction in which the zero-block, other-unit and retraining diagnostics report an ``effect'' with probability tending to one under $H_0^{G}$; blocks taken from another time of the same unit are not covered by it.

The design law removes the problem. Blocks drawn from the full history-measurable design law $Q_{i,t}$ retain the dependence on the recorded design state that the observed block has, so they carry the same stand-in information; under Assumption~\ref{ass:design} this state reduces to the unit's past actions and its stratum. Under $H_0^{G}$ and Assumption~\ref{ass:design} or~\ref{ass:design-cl}, the observed block is exchangeable with such draws given $\mathcal{F}_t$ and the targets, and no statistic can separate them except by chance. A resampler that changes the conditional assignment law can break the stand-in dependence. In \S\ref{sec:trap}, resampling from marginal action frequencies reproduced false effects.

%% file: sections/03_method.tex
\section{Method}\label{sec:method}

A predictor $f$ is trained and frozen, on units whose data do not depend on the test units' tested actions, before any test outcome is used, and it enters the test only through a statistic. The design law, not the model, controls the error rate; the model determines power.

\subsection{Scores and per-origin e-values}\label{sec:method:evalue}

For unit $i$ and origin $t$, let $\mathcal{C}_{i,t}\subseteq\mathcal{C}_i$ be the sites whose own history and target are observed, and let $n_{i,t}=|\mathcal{C}_{i,t}|$. For a candidate block $a$, define the site log score and the \emph{pooled} unit score
\begin{equation}\label{eq:pooled}
\ell_{i,c,t}(a)=\log\max\bigl\{\varepsilon,\;p_f\bigl(Y_{i,c,t+h}\mid H_{i,c,t},\,a\bigr)\bigr\},
\qquad
S_{i,t}(a)=\sum_{c\in\mathcal{C}_{i,t}}\ell_{i,c,t}(a),
\end{equation}
where $H_{i,c,t}$ is the site's admissible history and $p_f$ is the model's predictive density. All our reference models use a Gaussian density whose variance, like the floor $\varepsilon$, is fixed on training data; the floor is applied per site before summing. The pooled score is the independence log-likelihood of the unit's sites~\citep{chandler2007inference}. It treats sites as independent, but validity does not rest on that.

For any admissible statistic $T_{i,t}(a)$---the pooled score, or the tempered scores of \S\ref{sec:method:tempering}---the exact and Monte Carlo (MC) e-values at $(i,t)$ are
\begin{align}
e^{\mathrm{ex}}_{i,t}&=\frac{\exp T_{i,t}(A^{F}_{i,t})}{\sum_{a}Q_{i,t}(a)\,\exp T_{i,t}(a)},\label{eq:e-exact}\\[2pt]
e^{\mathrm{mc}}_{i,t}&=\frac{(B+1)\,\exp T_{i,t}(A^{F}_{i,t})}{\exp T_{i,t}(A^{F}_{i,t})+\sum_{b=1}^{B}\exp T_{i,t}(A^{(b)}_{i,t})},
\qquad A^{(1)}_{i,t},\dots,A^{(B)}_{i,t}\overset{\mathrm{iid}}{\sim}Q_{i,t}.\label{eq:e-mc}
\end{align}
The numerator scores the observed future actions; the denominator averages the score over the design. This is the conditional randomization test~\citep{candes2018panning}, with the design law in the role of the known conditional law, written as an e-value. Under $H_0^{G}$ and Assumption~\ref{ass:design}, the observed block is distributed as $Q_{i,t}$ given $\mathcal{F}_t$ and the targets (under $H_0^{\mathrm{own}}$: given $\mathcal{F}_t$, the future blocks of units outside the independent set, and the targets of the units in it), and the statistic is a fixed function of the candidate block given these. Both e-values therefore have conditional mean one (Proposition~\ref{prop:P1}). The MC form is valid because it includes the observed score in its denominator, which also caps it at $B+1$. Identical candidate blocks may be merged with their multiplicities, and a zero-entropy design gives $e\equiv1$, as does an origin at which no site is eligible, since the pooled and tempered scores are then identically zero (the tempered score is read as zero there even where its factor would be undefined); statements about the field mean and the site average concern nonempty site sets. We use the exact form when the block space can be enumerated and the MC form otherwise. The exact form requires $0<\sum_{a}Q_{i,t}(a)\exp T_{i,t}(a)<\infty$, with integration in place of summation for a continuous block space; this is automatic for finite support and finite scores. The MC form requires only that the $B+1$ candidate scores be finite.

\subsection{Aggregation}\label{sec:method:aggregation}

Origins are grouped into residue classes $r\in\{0,\dots,h-1\}$ by $t\equiv r\pmod h$. Within a class, consecutive origins are $h$ apart, so their future blocks do not overlap and each factor is determined by the history at the next origin of the class and the Monte Carlo draws used at its own origin; the factors therefore multiply into an e-value along the class (Proposition~\ref{prop:P1}). Under $H_0^{G}$ the joint design also allows multiplication across units:
\begin{equation}\label{eq:aggregate}
E_{r}=\prod_{i}\;\prod_{k:\,t_k\equiv r\ (\mathrm{mod}\ h)} e_{i,t_k},
\qquad
E=\frac{1}{h}\sum_{r=0}^{h-1}E_{r}.
\end{equation}
Averages of e-values are e-values without any independence assumption~\citep{vovk2021evalues}. For $H_0^{\mathrm{own}}$ the product over units is taken within each independent set of the spillover graph, and the set products are averaged; without a known graph, singletons are used.

With $K$ reference models we aggregate each model completely and then average the models' e-values, $E=K^{-1}\sum_{k}E^{(k)}$. Because $E\ge\max_k E^{(k)}/K$, the average loses at most $\log K$ in $\log E$ relative to the best model chosen in hindsight, so no model selection is needed. We reject at level $\alpha$ iff $E\ge 1/\alpha$ and report $p=\min(1,1/E)$; for $m$ prespecified hypotheses we use the e-Bonferroni rule $E\ge m/\alpha$ (Markov's inequality and a union bound). Where the registration specifies it, we also report a secondary batch randomization p-value at one prespecified common origin, resampling all units jointly under $H_0^{G}$, or the units of one prespecified independent set under $H_0^{\mathrm{own}}$.

\subsection{The statistic is a free choice}\label{sec:method:free}

The validity argument uses only two facts: the observed block is distributed as $Q_{i,t}$ given $\mathcal{F}_t$ and the targets (under $H_0^{\mathrm{own}}$: given $\mathcal{F}_t$, the future blocks of units outside the independent set, and the targets of the units in it), and the statistic is fixed once these and the training data are fixed. Every admissible statistic therefore yields a valid test. Choosing the statistic is a question of power alone, and it may be settled by any procedure that uses neither test outcomes nor the test units' realised future actions---for example, selection on training units or on synthetic data.

\paragraph{Pooling amplification.} The pooled score is the natural choice, and it fails in a characteristic way. Since the denominator of \eqref{eq:e-exact} is at least any one of its terms, for every block $a'$ with $Q_{i,t}(a')>0$,
\begin{equation}\label{eq:loge-bound}
\log e^{\mathrm{ex}}_{i,t}\;\le\;-\bigl[T_{i,t}(a')-T_{i,t}(A^{F}_{i,t})\bigr]-\log Q_{i,t}(a'),
\end{equation}
and the MC form obeys $\log e^{\mathrm{mc}}_{i,t}\le\log(B+1)-\log K(a')-[T_{i,t}(a')-T_{i,t}(A^{F}_{i,t})]$ for every drawn block $a'$, where $K(a')\ge1$ is the number of times $a'$ occurs among the $B+1$ blocks. For the pooled score the bracket is a sum over the $n_{i,t}$ sites. If a common alternative block has a score advantage per site bounded below by a positive constant as sites are added, its total advantage grows at least linearly. The exact log factor then has an upper bound that decreases at least linearly if that block's design mass stays bounded away from zero. The analogous upper bound holds in fixed-size Monte Carlo bags containing the block. Independent site errors partly cancel in the sum; common errors do not. Under these conditions, pooling can drive the log factor far below zero as sites are added. We call this failure \emph{pooling amplification}.

Errors common to the sites along the contrast arise in two ways. A misspecified model can err in its predicted response to the action, which all sites of a unit share; and noise shared by the sites can lie along the contrast, as when field-wide fluctuations move all cells in the direction an action would. Shared noise orthogonal to the contrast does not amplify. In the pixel world of \S\ref{sec:pooling}, a correctly specified model is unaffected by both forms of shared noise we simulate, while the fitted models collapse under every noise condition; the cause there is misspecification.

\subsection{Tempered scores}\label{sec:method:tempering}

We temper the pooled score,
\begin{equation}\label{eq:tempered}
T_{i,t}(a)=\lambda_{i,t}\,S_{i,t}(a),\qquad 0<\lambda_{i,t}\le 1,
\end{equation}
with $\lambda_{i,t}$ a function of training data and of the site count $n_{i,t}$ only. The count is determined by the observation mask. Under the stated assumption that the mask does not respond to the tested blocks under the null (\S\ref{sec:setup:nulls}), $\lambda_{i,t}$ is fixed given the conditioning field of Proposition~\ref{prop:P1}, so \eqref{eq:tempered} is admissible and Proposition~\ref{prop:P1} applies unchanged. (The same assumption is already needed for the pooled score, whose site set is the mask; if the mask responded to the tested blocks, neither score would be covered. The restriction $\lambda\le1$ is not needed for validity.) Setting $\lambda=1$ recovers the pooled score, and $\lambda\to0$ gives $e\to1$; $1/\lambda_{i,t}$ can be read as the number of sites that together carry one site's worth of independent evidence about the action.

Tempering a log-likelihood in this way has a long history: Kish's design effect for clustered samples~\citep{kish1965survey}, adjustments that rescale independence and composite log-likelihoods so that likelihood-ratio statistics recover their usual reference distribution~\citep{chandler2007inference,varin2011overview,pace2011adjusting}, and learning rates or power likelihoods for misspecified Bayesian models~\citep{grunwald2017inconsistency,holmes2017assigning}. We claim no novelty for the formulas below.

\paragraph{Kish tempering (registered).} With $\hat\rho$ an estimate of the correlation between sites' residuals,
\begin{equation}\label{eq:kish}
\lambda^{\mathrm{K}}_{i,t}=\frac{1}{1+(n_{i,t}-1)\,\hat\rho},
\end{equation}
the inverse design effect of a cluster of $n_{i,t}$ sites with intraclass correlation $\hat\rho$. In the ERK analyses $\hat\rho$ is computed separately for each reference model and horizon from training-field validation residuals only: Pearson correlations between the residual series of each pair of sites over their common eligible origins (pairs with fewer than three finite, non-constant matched values are omitted), then the median over pairs within each training field, the median over training fields, and clipping to $[0,1]$. The pixel simulation uses the same estimator on standardized validation residuals, with episodes in place of fields. This statistic was selected on synthetic data from three candidates under a rule committed before the evaluation; all three reached full power, the deciding comparison was narrow, and one tie-resolution detail of the rule was fixed only in code committed after the evaluation (\S\ref{sec:pooling}). It is the statistic registered for the confirmatory test M5 (\S\ref{sec:real}).

\paragraph{Contrast tempering (exploratory).} Kish tempering measures how strongly residuals are correlated across sites. Amplification, however, depends on how strongly the sites' contributions to the score \emph{difference between candidate blocks} are correlated. For a Gaussian score with predicted mean $\mu_{c,t}(a)$ and site variance $v_c$, replacing the observed block by a design draw $a^{b}$ lowers the site log score, to first order in the mean and when the per-site floor is inactive, by
\[
z^{b}_{c,t}=\frac{r_{c,t}\,D^{b}_{c,t}}{v_c},\qquad
r_{c,t}=Y_{c,t+h}-\mu_{c,t}(A^{F}_{t}),\qquad
D^{b}_{c,t}=\mu_{c,t}(A^{F}_{t})-\mu_{c,t}(a^{b}),
\]
where $D^{b}_{c,t}$ is the \emph{action contrast}. The contrast-tempering factor is the ratio of the mean square that the pooled score implicitly assumes for the sum of these contributions (independent sites) to their actual mean square, estimated on validation data with $B_c$ design draws $a^{b}\sim Q_t$ per origin:
\begin{equation}\label{eq:contrast}
\lambda^{\mathrm{c}}=\frac{\sum_{t,b}\sum_{c}\bigl(z^{b}_{c,t}\bigr)^{2}}{\sum_{t,b}\bigl(\sum_{c}z^{b}_{c,t}\bigr)^{2}},
\qquad\text{clipped to }[1/n,\,1].
\end{equation}
It is computed once per model and setting from validation episodes only, and is written and hashed before any test e-value is computed. The ratio is defined when its denominator is positive. If the contributions have zero cross moments across sites, its expected numerator and denominator agree; the ratio itself need not have expectation one, but it tends to one when the two sums, suitably normalized, satisfy a law of large numbers with a common positive limit. If the contributions are identical across a common number $n$ of sites, it equals $1/n$. The two factors can disagree in either direction. When a model misplaces a moving object, its residuals take both signs across pixels and their pairwise correlation averages to nearly zero, so $\lambda^{\mathrm{K}}\approx1$; weighted by the contrast, which has the same spatial pattern as the error, the same residuals add coherently, and $\lambda^{\mathrm{c}}\ll1$. Conversely, a frame-wide offset makes residuals strongly correlated, so $\lambda^{\mathrm{K}}$ is small, but the offset is orthogonal to a contrast that moves intensity without changing its total, so $\lambda^{\mathrm{c}}\approx1$. Contrast tempering was declared in the simulation protocol, before the pixel experiment was run, as an exploratory generalization. In the registered test on a published world model it was a secondary statistic and did not repair pooling amplification (\S\ref{sec:published}); we report it but do not recommend it.

\paragraph{When Kish tempering suffices.} If the action contrast is the same at every site and the site variances are equal---as for an additive model whose action branch sees only the unit's shared actions, the form of the ERK reference models---the contrast factor \eqref{eq:contrast}, where defined, satisfies the weighted cross-product identity of Proposition~\ref{prop:kish}; for a common site count it has the form of a Kish factor whose correlation is a contrast-weighted ratio of uncentred residual cross-products, which is not in general the median-of-Pearson estimate in \eqref{eq:kish}.

\subsection{Statistics that need no estimated factor}\label{sec:method:robust}

The factors of \S\ref{sec:method:tempering} are estimated from second moments of validation residuals. Three further statistics avoid that estimate or replace it by a direct search; all keep the design law, the bags and the aggregation, and all are admissible.

\paragraph{Tempering mixture.} For a fixed grid $\lambda_g=2^{-g}$, $g=0,\dots,G-1$, let $E(\lambda_g)$ be the e-value of a whole decision unit computed with the constant factor $\lambda_g$ at every origin, and
\begin{equation}\label{eq:mixture}
E^{\mathrm{M}}=\frac{1}{G}\sum_{g=0}^{G-1}E(\lambda_g).
\end{equation}
Each $E(\lambda_g)$ is an e-value because $\lambda_g$ is fixed, and so is their average; pathwise, $\log E^{\mathrm{M}}\ge\max_g\log E(\lambda_g)-\log G$. The average is taken over whole units, never over per-origin factors. The grid extends to or below $1/n$ at the largest site count of the application: $G=13$ in the pixel world and $G=15$ for the $64\times64\times3$ sites of \S\ref{sec:published}, where the 13-value version is also reported for continuity. Learning a learning rate over an exponentially spaced grid, or weighting a grid of learning rates by their performance, is standard in online learning~\citep{koolen2014learning,vanerven2016metagrad}; Proposition~\ref{prop:mixture} gives the guarantee we use.

\paragraph{Validation-tuned tempering.} $\lambda^{\mathrm{V}}$ is the grid value with the largest median validation $\log E$, computed per model with the same validation episodes and design draws as the contrast factor, ties going to the larger value; the test statistic is \eqref{eq:tempered} with $\lambda=\lambda^{\mathrm{V}}$. It depends on validation data only.

\paragraph{Validation-weighted mixture.} Validation tuning stakes everything on one factor chosen on validation data; the mixture spreads its stake evenly. A weighted mixture does both. With $g_V$ the grid index of $\lambda_V$ and a weight $w$ fixed in advance,
\begin{equation}\label{eq:weighted}E^{W}(w) = w\,E(\lambda_{g_V}) + \frac{1-w}{G-1}\sum_{g\ne g_V} E(\lambda_g).\end{equation}
Because $g_V$ depends on validation data only, $E^W(w)$ is an average of e-values with weights fixed before any test unit is seen, so it is an e-value. Pathwise, $\log E^W(w) \ge \log E(\lambda_{g_V}) - \log(1/w)$, and $\log E^W(w) \ge \log E(\lambda_g) - \log\{(G-1)/(1-w)\}$ for every other $g$. With $G = 15$ and $w = 3/4$ the losses are at most $\log(4/3) \approx 0.29$ against $\lambda_V$ and $\log 56 \approx 4.03$ against any other grid member, compared with $\log 15 \approx 2.71$ against every member for the uniform mixture, which is the case $w = 1/G$. Weighting a grid by prior information is standard; we claim no priority. $W(3/4)$ is the primary statistic of \S\ref{sec:p4}; $W(1/4)$ and $W(1/2)$ are reported as secondary.

\paragraph{Site-count-adaptive factors.} When units or origins differ in their number of sites, one factor $\lambda$ cannot suit all of them. A design-effect form lets the factor follow the site count: $\lambda_{i,t}(\rho) = 1/\{1 + (n_{i,t} - 1)\rho\}$, with $n_{i,t}$ the number of sites of unit $i$ at origin $t$ and $\rho$ on the grid $2^{-g}$. Validation tuning, the mixture and the weighted mixture over this grid are written V$_\rho$, M$_\rho$ and W$_\rho(w)$. They are e-values for the same reason as their counterparts. We use them in \S\ref{sec:erkrho}.

\paragraph{Online rules.} A factor can also be chosen within a unit, from that unit's earlier origins. At origin $t$, the follow-the-leader rule uses the grid factor with the largest cumulative log e-value over the origins before $t$; at the first origin it uses the smallest grid factor, and ties go to the smaller factor. The choice depends only on earlier origins, so each factor is fixed before its own origin's label is used, and the product over origins is an e-value under the conditions of Proposition~\ref{prop:P1}. O15 uses the 15 grid values, O57 a finer grid of 57 values $2^{-j/4}$, $j = 0, \dots, 56$, and O57h uses half of O57's choice at every origin. The uniform mixture is itself an online rule (Proposition~\ref{prop:overbet}(e)): it weights the grid factors at each origin by the wealth each has earned so far.

\paragraph{Rank e-values.} At each origin, let $k$ be the position of the observed block's score among the $B+1$ candidate scores, from the highest, with tied positions averaged (identical blocks receive one score). With $g(k)\propto(k/(B+1))^{\kappa-1}$ normalized to mean one over the $B+1$ positions, the factor is $g(k)$, averaged over tied positions. Under the exchangeable null a rank with uniform random tie-breaking is uniform, and averaging $g$ over the observed tie block preserves mean one; positive rescaling preserves ranks and tie blocks, so the factor is the same for the pooled and both tempered scores. We use $\kappa=1/\ln(B+1)$ (R) and $\kappa=1/2$ (R$_{1/2}$). Rank calibrators of this kind are p-to-e calibrators~\citep{vovk2021evalues}. For untied scores they pay an expected price per factor under the null (Proposition~\ref{prop:rank}); averaging over tied positions keeps validity but lowers that price, to zero when all candidates tie.

\subsection{A betting reading}\label{sec:method:betting}

In the exact form, $e_{i,t}=Q^{T}_{i,t}(A^{F}_{i,t})/Q_{i,t}(A^{F}_{i,t})$ with $Q^{T}_{i,t}(a)\propto Q_{i,t}(a)\,e^{T_{i,t}(a)}$. The e-value is the likelihood ratio of an alternative design law that up-weights blocks the model scores highly---equivalently, the wealth multiplier of a bet against the null placed according to the model~\citep{shafer2021testing,ramdas2023game,grunwald2024safe}. Tempering replaces this alternative by
\[
Q^{\lambda S}_{i,t}\;\propto\;Q_{i,t}^{\,1-\lambda}\,\bigl(Q^{S}_{i,t}\bigr)^{\lambda},
\]
a geometric interpolation between the null law and the pooled bet. Fractional Kelly betting~\citep{maclean1992growth,maclean2010longterm}, advocated when one is not confident about the alternative~\citep{ramdas2023game}, instead mixes arithmetically, $Q_{\kappa}=(1-\kappa)\,Q+\kappa\,Q^{S}$, which gives $e=1-\kappa+\kappa\,e^{S}$; we include it, with $\kappa=1/2$ applied per origin, as a reference arm in \S\ref{sec:pooling}. Here $\kappa$ denotes the fractional-Kelly mixing weight. It is unrelated to the rank-calibrator parameter in \S{}3.5 and is called $f$ in \S{}13. Both shrink the alternative toward the null as a guard against misspecification, and both keep the pooled bet's ranking of blocks. They interpolate along different paths: tempering along the exponential family $Q^{\lambda S}$, fractional Kelly along the mixture family $Q_{\kappa}$. Geometric shrinkage rescales the score: when per-site score differences are bounded, a factor $\lambda$ of order $1/n$ removes the linear growth of $|\log e|$ in the number of sites; arithmetic mixing bounds each factor below by $1-\kappa$. When the score takes only two values on the support of $Q$ (for example, a single binary action) the two families coincide, each $\lambda$ matching an origin-specific $\kappa$. With three or more distinct score values they differ at every interior parameter, $0<\lambda<1$, although they share their endpoints; the two are different members of one family of defences.

\subsection{Statistics used in this paper}\label{sec:method:table}

Table~\ref{tab:statistics} lists the principal statistics, with its label in the registered and protocol documents and its evidential status. Under the conditions of Proposition~\ref{prop:P1} the e-value statistics give valid tests. Section 13.6 gives the construction and validity conditions for PRseq. The pooled and tempered scores are admissible statistics in \eqref{eq:e-exact}--\eqref{eq:e-mc}; the field-mean and oracle variants of \S\ref{sec:pooling} are too, on their defined domains, when their candidate scores use only the unit's own history, targets and candidate block; the site average averages, over a nonempty site set fixed in the null conditioning, the per-site e-values that these equations give; and fractional Kelly averages an e-value with the constant one. Averages of e-values are again e-values. The full definitions of the variants are in the frozen protocols and are not restated here.

\begingroup\small
\setlength{\tabcolsep}{3pt}\begin{longtable}{@{}p{2.1cm} p{1.6cm} p{1.4cm} p{7.8cm}@{}}
\caption{Statistics used in this paper. ``Label'' is the name in the registered and protocol documents; the paper uses the names in the first column.}
\label{tab:statistics}\\
\toprule
Name & Label & Definition & Status \\
\midrule
Pooled & C0 & \eqref{eq:pooled} & Registered statistic of M4 (zxar3); secondary in ctfxv, j75g2 and 4wj7k. \\
Kish-tempered & C1 & \eqref{eq:tempered}, \eqref{eq:kish} & Selected on synthetic data (\S\ref{sec:pooling}); registered statistic of M5 (qezbx); secondary in ctfxv and, for Gaussian scores, in j75g2. \\
Field-mean & C2 & \S\ref{sec:pooling} & Candidate in the selection; not selected. \\
Site-average & C3 & \S\ref{sec:pooling} & Candidate in the selection; not selected. \\
Contrast-tempered & C1c & \eqref{eq:tempered}, \eqref{eq:contrast} & Exploratory; declared in the simulation protocol before the run; secondary in ctfxv and, for Gaussian scores, in j75g2. It did not repair pooling amplification on DIAMOND or on IRIS's decoded pixels. \\
Fractional Kelly & C4 & \S\ref{sec:method:betting} & Reference arm; simulation only. \\
Oracle variants & ORACLE\_* & \S\ref{sec:pooling} & True model; simulation only. \\
Tempering mixture & M (M13) & \eqref{eq:mixture} & Exploratory in the baseline study (\S\ref{sec:pooling:baselines}); primary statistic of ctfxv (\S\ref{sec:published}); secondary in j75g2 (\S\ref{sec:p4}) and 4wj7k (\S\ref{sec:p5}). \\
Validation-tuned & V & \S\ref{sec:method:robust} & Exploratory in the baseline study; secondary in ctfxv, j75g2 and 4wj7k. Selected by the regime map's rule (\S\ref{sec:regime}) and set aside after it collapsed on the ERK test fields (\S\ref{sec:erkxm}); with a site-count-adaptive factor it did not collapse there but lost most of the evidence for some models (\S\ref{sec:erkrho}). \\
Validation-weighted mixture & $W(w)$, $w\in\{\frac14,\frac12,\frac34\}$ & \eqref{eq:weighted} & Exploratory in the regime map (\S\ref{sec:regime}) and on the ERK test fields (\S\ref{sec:erkxm}); W(3/4) primary statistic of j75g2 (\S\ref{sec:p4}) and 4wj7k (\S\ref{sec:p5}); W(1/4), W(1/2) secondary. W(3/4) is primary for the 8mg9t horizon confirmation. \\
Site-count-adaptive variants & V$_\rho$, M$_\rho$, W$_\rho$(3/4) & \S\ref{sec:method:robust} & Exploratory, on the ERK test fields only (\S\ref{sec:erkrho}). \\
Online rules & O15, O57, O57h & \S\ref{sec:method:robust} & Exploratory on saved units (\S\ref{sec:overbet}); O57 and O57h secondary in 4wj7k (\S\ref{sec:p5}). \\
Rank e-values & R, R$_{1/2}$ & \S\ref{sec:method:robust} & Exploratory in the baseline study; secondary in ctfxv, j75g2 and 4wj7k. \\
T & T & \S\ref{sec:v5-temperature} & Registered secondary statistic of 8mg9t. Also fitted or recomputed in exploratory comparisons. \\
TS & TS\_C & \S\ref{sec:v5-bounded} & Exploratory (E1 to E7); calibrates every evidence rate in \S\ref{sec:v5-evidence}--\ref{sec:v5-trajectories}. \\
SW & SW & \S\ref{sec:v5-bounded} & Registered secondary statistic of 8mg9t and focal statistic of its stake confirmation. \\
PRseq & PRseq & \S\ref{sec:v5-horizons} & Exploratory reanalysis of the fourth study; a p-value. \\
\bottomrule
\end{longtable}\endgroup

N, RM, SM and UW are registered secondary statistics of 8mg9t, defined in Appendix M.4.

\paragraph{Notation in the registered documents.} The registered M4 documents write $\lambda$ for the precision $1/(2\sigma^{2})$ of the Gaussian score; in this paper $\lambda$ denotes only the tempering factor. The labels C1--C5 of the sealed-challenge criteria (\S\ref{sec:synthetic}) are unrelated to the statistic labels in Table~\ref{tab:statistics}. The specification writes $G_t$ for the observed history and $\mathcal{G}$ for the spillover graph; we write $\mathcal{F}_t$ and $\mathcal{N}$. Internal labels of our studies (P1--P4, A1, A1-G, A2, ERK-XM) appear in repositories and registrations; the paper uses descriptive names, mapped in Appendix~\ref{app:exploratory}.

%% file: sections/04_theory.tex
\section{Theory}\label{sec:theory}

The statements below restate the obligations of the registered specification, whose labels are given in parentheses, with notation adapted and with the hypotheses that the frozen proofs use stated explicitly. The proofs are in Appendix~\ref{app:proofs}. They follow the theory documents that were written before the registration and frozen with it (Appendix~\ref{app:proofs} lists them with their hashes), and Appendix~\ref{app:proofs} marks each place where we have added a hypothesis or tightened an argument; none of these changes affects the registered method. Proposition~\ref{prop:kish} and Propositions~\ref{prop:growth}--\ref{prop:closedloop} are new in this paper and were not part of any registration. Every guarantee is conditional on the design law used for resampling being the law that generated the actions; \S\ref{sec:theory:approx} treats departures from it.

\subsection{Validity}\label{sec:theory:validity}

\begin{proposition}[Validity; P1(a)--(c)]\label{prop:P1}
Suppose Assumption~\ref{ass:design} holds at every origin, and that
\begin{enumerate}
\item[(i)] the units, origins and sites that enter each product are fixed by structural events (layout, design allocation) that do not depend on the tested blocks, by own-path eligibility events treated in Proposition~\ref{prop:elig}(b), or by the observation mask, which under the null in force does not respond to the tested blocks (\S\ref{sec:setup:nulls});
\item[(ii)] the predictor and every score parameter (floor, variance, tempering factor) are functions of the observation mask and of training information that is conditionally independent of the joint vector of tested blocks given $\mathcal{F}_t$ and the potential outcomes; the same finite score function is applied to every candidate block, without using the position of the observed block among the candidates, and when the exact form is used its normaliser is finite and positive;
\item[(iii)] Monte Carlo draws are generated from $Q_{i,t}$ at the realised past, with randomness independent of the potential outcomes, of the assignments, of the training and design information of (ii) and of any auxiliary randomness used to construct the score, and across units and origins.
\end{enumerate}
Then:
\begin{enumerate}
\item[(a)] Under $H_0^{G}(h)$, conditionally on $\mathcal{F}_t$, the potential outcomes and the targets at $t+h$ of all units, each exact or MC factor has mean one, and so does their product across units. Along each residue grid the running product is a non-negative martingale with initial value one, with respect to the filtration generated by the history at the grid origins, the frozen training and design information, and the Monte Carlo seeds already used.
\item[(b)] Under $H_0^{\mathrm{own}}(h;\mathcal{N})$, let $I$ be an independent set of $\mathcal{N}$ fixed before testing, and let the statistic of each $i\in I$ use, besides $\mathcal{F}_t$-measurable covariates and its own candidate block, only unit~$i$'s own targets. Then, conditionally on $\mathcal{F}_t$, the potential outcomes, the future blocks of the units outside $I$ and the targets of the units in $I$, each exact or MC factor of a unit in $I$ has mean one, and so does their product over $I$; along each residue grid the running product over $I$ is a martingale as in (a). The conditioning may not include the targets of units outside $I$: through spillover these can depend on the blocks in $I$.
\item[(c)] Prechosen arithmetic averages over residue classes, independent sets and models of the quantities in (a)--(b) are e-values at the fixed analysis time, so $\Pr(E\ge1/\alpha)\le\alpha$. A batch randomization p-value computed at one prespecified common origin, with a statistic that is a fixed function of the joint candidate block given the conditioning field of (a) (respectively (b)), resampling all units jointly under $H_0^{G}$ or the units of one independent set under $H_0^{\mathrm{own}}$, and counting ties as exceedances, is super-uniform.
\end{enumerate}
\end{proposition}

The argument is short. Under $H_0^{G}$ and no anticipation, the targets at $t+h$ are functions of the potential-outcome functions and of the actions up to $t$, so conditioning on them leaves the design law of the tested blocks unchanged. The denominator of \eqref{eq:e-exact} is then the conditional expectation of its numerator; in the MC form, the observed block and its draws are exchangeable and the observed share of the total score has mean $1/(B+1)$. Shared outcome noise is part of the potential outcomes and requires no independence. A factor at a grid origin $t_k$ is determined by the history at $t_{k+1}=t_k+h$ and the MC draws used at $t_k$, which gives the martingale; averages of e-values are e-values without independence~\citep{vovk2021evalues}. Under $H_0^{\mathrm{own}}$ the targets of units in an independent set are invariant to the set's own blocks but those of neighbours outside it need not be, which is why (b) conditions on, and scores with, only the set's own targets. Two limits are worth stating. Optional stopping is licensed only along a single residue grid (by Ville's inequality); the average over classes is a fixed-time e-value, not an e-process on a common clock. And nothing in the argument depends on the predictor's accuracy, on independence between sites, or on the noise distribution: these affect power only.

\begin{proposition}[Eligibility; P1(d), P1(e)]\label{prop:elig}
\begin{enumerate}
\item[(a)] Let the structural eligibility events not depend on the tested blocks, and run the structural and the classifier-based pipelines with the same statistics and the same Monte Carlo draws whenever their eligible sets coincide. If the event that any classifier decision differs from its structural event has probability at most $\delta$ under the law that generated the commands, the classifier-based test has size at most $\min(1,\alpha+\delta)$. The bound is unconditional over repetitions of the experiment and classifier; it is not a guarantee conditional on the selected set.
\item[(b)] Suppose Assumption~\ref{ass:design} holds at every origin. Let unit~$i$ be analysed only on an event $\mathcal{E}_i$ that is a fixed function of its own complete action path, chosen before the data and depending on no other unit's actions and on no outcome (for example, that the path avoids prespecified fixed templates). Suppose that every unit on which $\mathcal{E}_i$ holds follows $Q_i$ (the template list contains every fixed design), and that $\Pr(\mathcal{E}_i\mid A_{i,\le t})>0$ at every analysed origin. Then replacing $Q_{i,t}$ by $Q_{i,t}(\cdot\mid\mathcal{E}_i,A_{i,\le t})$ keeps Proposition~\ref{prop:P1} exact, conditionally on the eligibility pattern of all units. For a hidden-Markov design this conditional law is computed exactly by a forward--backward recursion on the hidden state augmented with template-match indicators.
\end{enumerate}
\end{proposition}

Part (b) is used for the CcaS random-walk lineages, whose resampling law is conditioned on not reproducing either duplicated fixed template; lineages that reproduce a template are excluded (Appendix~\ref{app:assignment}). Part (a) enters only the secondary own-channel analyses; with no independent record of the physical layout, the only bound we can establish is $\delta\le1$, and we display it as vacuous next to every such result.

\begin{proposition}[Distributional nulls do not compose; P1b]\label{prop:P1b}
Let $(A_{1,k},A_{2,k},U_k)$, $k=1,2,3$, be independent fair bits, drawn afresh at each of three origins; take horizon $h=1$ and design law $Q=(\tfrac12,\tfrac12)$. Consider the systems (i) $Y_{1,k}=U_k$, $Y_{2,k}=U_k\oplus A_{1,k}\oplus A_{2,k}$, and (ii) $Y_{i,k}=A_{i,k}\oplus U_k$, $i=1,2$. In both, for every unit $i$ and origin $k$, $A_{i,k}$ is independent of $Y_{i,k}$ given both units' earlier actions and outcomes and the other unit's actions, so each unit-wise conditional-independence null holds and each unit-wise e-process has mean one. Score each unit by the Gaussian density $\mathcal{N}(Y_{i,k};A_{i,k},\sigma^2)$ with $\sigma^2=1/(2\log 9)$, so that the matching bit is nine times as likely as the other and each exact factor equals $9/5$ or $1/5$. Then the product of the unit-wise e-values across both units and the three origins has mean $1.64^3\approx4.41$ and exact rejection probability $1/8$ at $\alpha=0.05$.
\end{proposition}

The value $1/8$ is the probability that all six factors equal $9/5$; among density ratios $r\ge1$ it holds exactly when $r\ge20^{1/6}/(2-20^{1/6})\approx4.675$, and the rejection probability is zero for $1\le r<4.675$. Fresh noise is part of the construction: if $U$ were drawn once and held fixed, earlier outcomes would reveal it, and after the first origin the unit-wise null would fail for unit 2 in system (i) and for both units in system (ii). Both systems violate the sharp nulls, so for our tests they are alternatives. The proposition explains why the product across units in Proposition~\ref{prop:P1}(a) needs a sharp null: unit-wise distributional statements say nothing about how one unit's action enters another unit's outcome.

\subsection{What the design permits}\label{sec:theory:limits}

\begin{proposition}[Design limits; P2(a)--(e), regrouped]\label{prop:P2}
\begin{enumerate}
\item[(a)] [P2(a), P2(c)] Pathwise,
\[
e^{\mathrm{ex}}_{i,t}\le \frac{1}{Q_{i,t}(A^{F}_{i,t})},
\qquad
e^{\mathrm{mc}}_{i,t}\le\frac{B+1}{1+K},
\]
where $K$ is the number of draws equal to the observed block. A zero-entropy design gives $e\equiv1$.
\item[(b)] [P2(b), P2(c)] Under the open-loop design of Assumption~\ref{ass:design}, the action-path law is the same under every alternative, so
\[
\Pr\bigl(E\ge1/\alpha\bigr)\;\le\;\Pr_{Q}\Bigl(\max_{\text{components}}\;\sum_{\text{factors}}-\log Q_{i,t}(A^{F}_{i,t})\;\ge\;\log(1/\alpha)\Bigr),
\]
the maximum running over residue classes and independent sets, and the probability over coherent full command paths (blocks of different residue classes share actions). Under this open-loop design, the command-path law suffices to compute the right-hand side for a fixed analysis schedule. With $m$ prespecified hypotheses the threshold $1/\alpha$ becomes $m/\alpha$; under Proposition~\ref{prop:elig}(b), $Q_{i,t}$ and $\Pr_Q$ are replaced by their conditioned versions. The MC form has an analogous ceiling, with $\log(B+1)-\log(1+K)$ in place of $-\log Q$ and $K\sim\mathrm{Binomial}(B,Q_{i,t}(A^{F}_{i,t}))$; it can exceed the exact-form ceiling.
\item[(c)] [P2(d)] With $\mathcal{T}$ the frozen training information on which the score depends, $\mathbb{E}\bigl[\log e^{\mathrm{ex}}_{i,t}\mid\mathcal{F}_t,\mathcal{T}\bigr]\le I\bigl(Y_{\cdot,t+h};A^{F}_{i,t}\mid\mathcal{F}_t,\mathcal{T}\bigr)\le H_{Q}\bigl(A^{F}_{i,t}\mid\mathcal{F}_t\bigr)$.
\item[(d)] [P2(e)] For the batch test, $p\ge Q^{\mathrm{joint}}(A^{F}_{\mathrm{obs}})$ in the exact form and $p\ge(1+K)/(B+1)$ in the add-one MC form.
\end{enumerate}
\end{proposition}

Part (a) is one line---the observed term appears in its own denominator---and we do not present it as a deep result. Its value is practical. For a known open-loop law and fixed analysis schedule, the bounds can rule out reaching the rejection threshold before outcomes are opened. A point-mass conditional law gives a neutral factor. Because averaging models adds no design entropy, the same command-only bound applies to the model average. For the ERK analyses (M4, X and M5; Figure~\ref{fig:realchain}) we report the exact-form bound $\log E_{\mathrm{upper}}$ as a command-only comparator. Those analyses use Monte Carlo e-values, which the exact-form bound does not bound pathwise (part (b)), so the ratio $\log E/\log E_{\mathrm{upper}}$ is descriptive: it could in principle exceed one, and it is not a power percentage. Part (c) is a growth statement only.

\subsection{Approximate design laws}\label{sec:theory:approx}

\begin{proposition}[Approximate law; P3]\label{prop:P3}
Suppose the test resamples from a working law $\hat Q$ that is fixed independently of the tested blocks, with $\hat Q_{i,t}$ depending only on $A_{i,\le t}$ and the stratum, while the actions follow $Q$ (Assumption~\ref{ass:design}). Let $\Delta$ be the total-variation distance between the laws, under $Q$ and under $\hat Q$, of the complete command path from the first tested origin on, given the history before it and the frozen training and working-law information; under Assumption~\ref{ass:design} these laws do not involve the potential outcomes. Then, under the null,
\[
\Pr(E\ge1/\alpha)\;\le\;\alpha+\mathbb{E}[\Delta].
\]
If moreover $0\le E\le M$, then $\mathbb{E}[E]\le1+M\,\mathbb{E}[\Delta]$.
\end{proposition}

The event bound needs neither a coupling construction nor a bound on $E$. The moment bound, with $M=(B+1)^{m}$ for an MC product of $m$ factors, is weaker whenever rejection is possible ($M\ge1/\alpha$). By a telescoping argument over time steps, $\Delta$ is at most the expected sum, along the true command path, of the one-step total-variation distances between the true and working kernels, and for product kernels each one-step distance is at most the sum of the unit distances. It is unknown for the working laws used here, and those laws were not fitted independently of the released commands, so the proposition does not license our real-data tests; we rely instead on calibration with the real command records (\S\ref{sec:synthetic}). The closest published result, Theorem~4 of \citet{berrett2020conditional}, bounds the type~I error of the conditional randomization test, conditionally on $Y$ and $Z$, by $\alpha$ plus the total variation distance between the true and fitted conditional laws of $X$ given $Z$ for the whole sample, for independent and identically distributed data with the statistic and fitted law chosen independently of $X$ and $Y$; it covers both the randomization and the permutation forms, and it does not directly cover this longitudinal product. The resampler of \S\ref{sec:trap} that ignores part of the design state is far from the design law in total variation, and nothing in Proposition~\ref{prop:P3} protects it.

\subsection{Why replacement diagnostics fail}\label{sec:theory:trap}

\begin{proposition}[Failure of replacement diagnostics; P4]\label{prop:P4}
Let the units be independent and identically distributed copies of the following experiment, observed at one fixed clock, with all its parameters fixed as the number of units grows. Each unit has a stratum $\theta_i\in\{0,1\}$ known to the design, with $\Pr(\theta_i=1)\in(0,1)$. Let $\theta_i$ enter only an outcome recorded before the predictor's window, let the outcomes inside the window be constant, and let the focal outcome be $\theta_i$ plus independent centred noise of finite variance, for every action assignment, so that $H_0^{G}$ holds. Let the actions follow the Markov chain $\Pr(A_s=1\mid A_{s-1}=a)=\rho a+(1-\rho)p_{\theta_i}$, started in its stationary law, with $0\le\rho<1$, $p_0\ne p_1$ and $p_0,p_1\in(0,1)$. A predictor sees a window $W$ of finitely many recent commands, the recent outcomes and the clock, together with the future block $F$ of a fixed length $h\ge1$, but neither $\theta_i$ nor the earlier outcome; its ideal form is $\mu=\mathbb{E}[Y\mid W,F]$. Under squared-error loss, the population risk gaps of (i) retraining without actions ($\mu_0=\mathbb{E}[Y\mid W]$, or $\mathbb{E}[Y]$ if the window's commands are removed as well), (ii) replacing $F$ by zeros and (iii) replacing $F$ by the block of an independent unit are strictly positive. For these ideal predictors, as the number of units grows the empirical gaps converge to these constants while unit (for (iii), pair) bootstrap intervals shrink, so each diagnostic reports an effect with probability tending to one. Under the same construction, Proposition~\ref{prop:P1} bounds by $\alpha$ the size of the randomization test that resamples from $Q(\cdot\mid A_{i,t},\theta_i)$.
\end{proposition}

The window must not reveal the stratum: if the recent outcomes did (for instance, if they equalled $\theta_i$), the future block would add nothing and all three gaps would vanish. The proposition concerns ideal predictors, as the frozen proof does; fitted predictors inherit it only under further convergence conditions, which we do not pursue. The construction is the stand-in effect of \S\ref{sec:setup:standin} in its simplest form: the future block reveals $\theta_i$ beyond what the window does, so a good predictor uses it, and a replacement that breaks the dependence costs predictive accuracy without any action effect. Design draws preserve the dependence, which is why the randomization test is unaffected. The proposition covers replacement by another unit's block, not by a block taken from another time of the same unit. The development simulations confirm the construction numerically (\S\ref{sec:synthetic}).

\subsection{Pooling and tempering}\label{sec:theory:tempering}

Tempering needs no separate validity result: the tempered score \eqref{eq:tempered} is admissible (\S\ref{sec:method:tempering}), so Proposition~\ref{prop:P1} applies to it. The bound \eqref{eq:loge-bound} of \S\ref{sec:method:free} describes how the pooled score loses power. The remaining question is when the registered Kish factor approximates the contrast factor.

\begin{proposition}[Contrast tempering under a shared contrast; new]\label{prop:kish}
If the action contrast is common to the sites, $D^{b}_{c,t}=D^{b}_{t}$, the site variances are equal, $v_c=v>0$, the same sites enter every candidate at each origin, and the numerator and denominator of \eqref{eq:contrast} are positive, then the unclipped contrast factor equals $1/(1+X)$ with
\[
X=\frac{\sum_{t}w_t\sum_{c\ne c'}r_{c,t}\,r_{c',t}}{\sum_{t}w_t\sum_{c}r_{c,t}^{2}},
\qquad w_t=\sum_{b}\bigl(D^{b}_{t}\bigr)^{2}.
\]
For a common site count $n\ge2$ this is $\lambda^{\mathrm{c}}=1/[1+(n-1)\bar\rho_{D}]$ with $\bar\rho_{D}=X/(n-1)\in(-1/(n-1),1]$.
\end{proposition}

\begin{proof}
Under the hypotheses, $z^{b}_{c,t}=r_{c,t}D^{b}_{t}/v$, so $\sum_c(z^{b}_{c,t})^2=(D^{b}_{t}/v)^2\sum_c r_{c,t}^2$ and $(\sum_c z^{b}_{c,t})^2=(D^{b}_{t}/v)^2\bigl[\sum_c r_{c,t}^2+\sum_{c\ne c'}r_{c,t}r_{c',t}\bigr]$. Summing over $b$ gives the weight $w_t/v^2$ for each origin; dividing the two sums over $t$ gives the result.
\end{proof}

\begin{remark}
The identity $\lambda^{\mathrm{c}}=1/(1+X)$ holds for any site counts. Its reading as a Kish factor with intraclass correlation $\bar\rho_D$ requires a common site count. When the count $n_t$ varies across origins, $X$ is a weighted mean of $(n_t-1)\rho_t$, so the single constant $\lambda^{\mathrm{c}}$ corresponds to a Kish factor at an effective site count and need not equal $1/[1+(n_t-1)\hat\rho]$ at every origin; with at least two distinct counts, equality at every origin requires $\hat\rho=0$ and $\lambda^{\mathrm{c}}=1$. For a common $n$, clipping $\lambda^{\mathrm{c}}$ to $[1/n,1]$ is the same as clipping $\bar\rho_D$ to $[0,1]$.
\end{remark}

When the contrast is shared, contrast tempering is therefore Kish tempering with a particular estimate of the intraclass correlation: a pooled, contrast-weighted mean of uncentred within-origin residual cross-products, instead of the median of pairwise Pearson correlations used in \eqref{eq:kish}. The two estimate the same correlation only under conditions on the contrast weights as well as on the residuals: for instance, centred, equally variable residuals with an exchangeable and stable correlation, and contrast weights independent of the residuals. Without the condition on the weights, the contrast-weighted correlation can differ from the ordinary one even when the residuals are uncorrelated. The contrast hypothesis holds for additive models whose action branch sees only the unit's shared inputs; the ERK reference models have this form, so agreement between the two factors on ERK is partly structural. Because ERK site counts vary across fields and origins, the meaningful comparison there is between $\hat\rho$ and the pooled correlation implied by $\lambda^{\mathrm{c}}$, not between $\lambda^{\mathrm{K}}_{i,t}$ and a single constant. The ERK training diagnostic (\S\ref{sec:real:diag}) finds the implied correlation lower than both the median and the mean of the pairwise correlations, so Kish tempering is somewhat conservative there; it is a description of how these correlation summaries differ, not an independent test of the Kish rule. When the contrast varies across sites, as when an action moves an object across pixels, the hypothesis fails and the two factors can differ in either direction (\S\ref{sec:method:tempering}, \S\ref{sec:pooling}).

\paragraph{When tempering helps, and by how much.} Write $\Delta S_{i,t}(a)=S_{i,t}(A^{F}_{i,t})-S_{i,t}(a)$ for the observed block's advantage over a candidate $a$, so that the exact factor with $T=\lambda S$ is $\log e^{\mathrm{ex}}_{i,t}(\lambda)=-\log\mathbb{E}_{a\sim Q_{i,t}}e^{-\lambda\,\Delta S_{i,t}(a)}$; the MC factor obeys the same identity with $Q_{i,t}$ replaced by the empirical law $\hat Q_{i,t}$ of its observed-inclusive bag, which is not an assignment law (its validity still comes from exchangeability under $Q_{i,t}$). Sums over origins below refer to one product component, one residue class of one unit, unless an averaging formula is given. A factor $\lambda>1$ appears only in the optimization and is compatible with validity; the tempering procedure of \S\ref{sec:method:tempering} uses $\lambda\le1$.

\begin{proposition}[A conditional growth bound; new]\label{prop:growth}
Fix the target, the score function and the observed block, and suppose $\mathbb{E}_{Q_{i,t}}|\Delta S_{i,t}|<\infty$ and $\mathbb{E}_{Q_{i,t}}e^{-\eta\,\Delta S_{i,t}}<\infty$ for some $\eta>0$. On the domain where this exponential moment is finite, $\lambda\mapsto\log e^{\mathrm{ex}}_{i,t}(\lambda)$ is concave, with right derivative $\mu_{i,t}=\mathbb{E}_{Q_{i,t}}\Delta S_{i,t}$ at zero. If a finite $\nu_{i,t}\ge0$ satisfies $\log\mathbb{E}_{Q_{i,t}}e^{-\lambda(\Delta S_{i,t}-\mu_{i,t})}\le\lambda^{2}\nu_{i,t}/2$ for every $\lambda\in[0,\Lambda]$, then on that interval
\[
\log e^{\mathrm{ex}}_{i,t}(\lambda)\ge\lambda\mu_{i,t}-\tfrac12\lambda^{2}\nu_{i,t},
\qquad
\log E(\lambda)\ge f(\lambda):=\lambda\mathcal{M}-\tfrac12\lambda^{2}\mathcal{V},
\]
with $\mathcal{M}=\sum_t\mu_{i,t}$ and $\mathcal{V}=\sum_t\nu_{i,t}$ over one component. If $\mathcal{M}>0$ and $\mathcal{V}>0$, the lower bound is maximized at $\lambda_{\mathrm{quad}}=\min(\mathcal{M}/\mathcal{V},\Lambda)$, with value $\mathcal{M}^{2}/(2\mathcal{V})$ when $\mathcal{M}/\mathcal{V}\le\Lambda$. For bounded score differences, Hoeffding's lemma gives $\nu_{i,t}=(\operatorname{ess\,sup}\Delta S_{i,t}-\operatorname{ess\,inf}\Delta S_{i,t})^{2}/4$ and $\Lambda=\infty$. The same statements hold for the MC factor with $\hat Q_{i,t}$ in place of $Q_{i,t}$. Under the null, for temperatures within a common moment domain and with the stated integrability, $\mathbb{E}[\mu_{i,t}\mid\text{conditioning field}]=0$, so $-\tfrac12\lambda^{2}\mathbb{E}[\nu_{i,t}]\le\mathbb{E}[\log e_{i,t}(\lambda)]\le0$.
\end{proposition}

The null bound is an expected-loss bound. On a single path the first-order term $\lambda\mu_{i,t}$ remains and may be negative, and a realized $\mathcal{M}>0$ is a pathwise, not an expected, growth condition. The proposition separates what the factor needs from what Kish and contrast tempering estimate: $\mu$ and $\nu$ describe score differences between candidate blocks, not correlations between sites.

\begin{proposition}[Precision rescaling; new]\label{prop:blind}
Let the site scores be unfloored Gaussian log densities, $\ell_{i,c,t}(a)=\log\phi(Y_{i,c,t+h};\mu_{c,t}(a),v_c)$, with the scored sites, predicted means, data, validation residuals and contrasts held fixed, and multiply every precision by $\gamma>0$ ($v_c\mapsto v_c/\gamma$). Then:
\begin{enumerate}
\item[(a)] $S^{(\gamma)}_{i,t}(a)=\gamma\,S^{(1)}_{i,t}(a)+C_{i,t}(\gamma)$ with $C_{i,t}(\gamma)$ independent of $a$, so $\log E^{(\gamma)}(\lambda)=\log E^{(1)}(\gamma\lambda)$: on a log-$\lambda$ axis the tempered curve shifts by $-\log\gamma$, its supremum is unchanged when the allowed temperature domain is rescaled along with the curve. If the scaled proxies satisfy the moment bound over the same interval $[0,\Lambda]$, the maximiser of the quadratic lower bound is $\min\{\mathcal{M}/(\gamma\mathcal{V}),\Lambda\}$, which equals $\lambda_{\mathrm{quad}}/\gamma$ if the endpoint is also scaled to $\Lambda/\gamma$ or if neither maximiser is capped;
\item[(b)] the contrast factor \eqref{eq:contrast}, where defined, does not depend on $\gamma$, nor on a rescaling $D\mapsto\kappa D$ of the contrasts, and the same holds for any factor that uses standardized residuals only through correlation coefficients;
\item[(c)] hence, by \eqref{eq:loge-bound}, for the exact factor at an origin with a competitor of design probability at least $q$ and margin $m>0$ at $\gamma=1$, $\log e^{\mathrm{ex},(\gamma)}_{i,t}(\lambda^{\mathrm{c}})\le\log(1/q)-\gamma\lambda^{\mathrm{c}}m$; for a fixed observed-inclusive MC bag in which the competitor occurs $k$ times, $\log(1/q)$ is replaced by $\log((B+1)/k)$.
\end{enumerate}
\end{proposition}

The proposition concerns factors that are invariant under precision rescaling; it does not exclude an estimated factor that rescales correctly. With the per-site floor used in our applications the identity in (a) holds only where no floor is active at either scale; Appendix~\ref{app:proofs:blind} gives the error bound for floored scores and shows that flooring can also stop the collapse in (c).

\begin{proposition}[Tempering mixture; new]\label{prop:mixture}
Let $\lambda_g=\lambda_0 2^{-g}$, $g=0,\dots,G-1$, and for one product component let $E^{\mathrm{M}}=G^{-1}\sum_gE(\lambda_g)$. Then:
\begin{enumerate}
\item[(a)] under the conditions of Proposition~\ref{prop:P1}, $E^{\mathrm{M}}$ is an e-value;
\item[(b)] pathwise, $\log E^{\mathrm{M}}\ge\max_g\log E(\lambda_g)-\log G$;
\item[(c)] if the bound of Proposition~\ref{prop:growth} holds at the grid points with $\mathcal{M}>0$ and $\mathcal{V}>0$, then $\log E^{\mathrm{M}}\ge f(\lambda_{\mathrm{quad}})-\log G$ when $\lambda_{\mathrm{quad}}$ is a grid point, $\log E^{\mathrm{M}}\ge\tfrac89f(\lambda_{\mathrm{quad}})-\log G$ when it lies strictly between two grid points, and $\log E^{\mathrm{M}}\ge\lambda_0\mathcal{M}/2-\log G$ when $\mathcal{M}/\mathcal{V}>\lambda_0$;
\item[(d)] with $R$ residue classes averaged as in \S\ref{sec:method:aggregation}, the mixture of the class average is the average of the class mixtures, and if $\log E^{\mathrm{M}}_{r}\ge b_r-\log G$ for each class, then $\log E^{\mathrm{M}}\ge\log\sum_r e^{b_r}-\log R-\log G$.
\end{enumerate}
\end{proposition}

In the setting of Proposition~\ref{prop:blind}, the guarantee in (c) holds uniformly over the finite range of precision multipliers $\gamma$ that keeps the maximiser of the scaled bound, $\min\{\mathcal{M}/(\gamma\mathcal{V}),\Lambda\}$, within the grid (and the comparison points within the moment domain); outside that range no such claim is made, and the bound need not be positive. Pathwise, $\log E^{\mathrm{M}}\ge\log E(\lambda_{G-1})-\log G$. For one product component under the valid null, if $\lambda_{G-1}$ lies in a common deterministic moment domain of Proposition~\ref{prop:growth} and the required integrability and $\mathbb{E}[\mathcal{V}]<\infty$ hold, the expected log loss is at most $\log G+\tfrac12\lambda_{G-1}^{2}\,\mathbb{E}[\mathcal{V}]$; for a residue-class average, apply Jensen with the class-averaged proxy, under these conditions for every class.

\setcounter{proposition}{12}
\begin{proposition}[Over-betting; new]\label{prop:overbet} Fix one product component of a unit, with pooled scores $S_t(b)$, $b = 0, \dots, B$, at its origins $t$, and let
\[L(\lambda) = \sum_t \Big\{ \lambda S_t(0) - \log \frac{1}{B+1} \sum_{b=0}^{B} e^{\lambda S_t(b)} \Big\}, \qquad \lambda \ge 0,\]
so that $E(\lambda) = e^{L(\lambda)}$. Then, pathwise:
\begin{enumerate}

\item[(a)] $L$ is concave and $L(0) = 0$;

\item[(b)] $L(\lambda) \ge (\lambda/\lambda')\,L(\lambda')$ for $0 \le \lambda \le \lambda'$;

\item[(c)] if $L(\lambda) < 0$ for some $\lambda$, then $L(\lambda'') \le (\lambda''/\lambda)\,L(\lambda) < 0$ for every $\lambda'' \ge \lambda$; and if $L(\lambda') \ge 0$, then $L(\lambda) \ge 0$ for every $\lambda \le \lambda'$;

\item[(d)] for the mixture of Proposition~\ref{prop:mixture} with $\lambda_g = \lambda_0 2^{-g}$, and any $\lambda^\ast \in [\lambda_{G-1}, \lambda_0]$ with $L(\lambda^\ast) \ge 0$: $\log E^{\mathrm M} \ge \tfrac12 L(\lambda^\ast) - \log G$;

\item[(e)] $E^{\mathrm M} = \prod_t \sum_g w_{g,t}\, e_t(\lambda_g)$, where $e_t(\lambda)$ is the factor of $E(\lambda)$ at origin $t$ and $w_{g,t} \propto \prod_{s<t} e_s(\lambda_g)$, with $w_{g,1} = 1/G$.

\end{enumerate}
\end{proposition}
\setcounter{proposition}{10}

Part (c) says that the factors at which a tempered statistic loses form an upper interval: a negative log e-value means that the factor exceeds the break-even factor, and beyond it the loss grows at least in proportion to the factor. A factor below a profitable one never loses, and by (b) it keeps at least the share $\lambda/\lambda'$ of the log e-value. Part (d) needs no moment bound, unlike Proposition~\ref{prop:mixture}(c); it gives nothing when no factor in the range is profitable. Part (e) is the standard identity between a mixture of products and a product of wealth-weighted averages; we claim no priority for it. The proposition is stated for one product component; for a residue-class average, (a) can fail, and Proposition~\ref{prop:mixture}(d) applies instead.

\begin{proposition}[Rank calibrators; new]\label{prop:rank}
Let $g$ be positive and nonincreasing on the $N_{\mathrm{cand}}=B+1$ positions, with mean one; put $c_g=N_{\mathrm{cand}}^{-1}\sum_k\log g(k)$ and $r_g=\log g(1)-\log g(N_{\mathrm{cand}})$. For $e^{\mathrm{R}}_{i,t}=g(R_{i,t})$, with $R_{i,t}$ the unique rank of the observed score or a rank drawn by uniform random tie-breaking, let $\tau_{i,t}$ be the total-variation distance of its conditional law from the uniform one. Then $\mathbb{E}[\log e^{\mathrm{R}}_{i,t}\mid\mathcal{F}_t]\le c_g+\tau_{i,t}r_g$, and over $T$ factors $\mathbb{E}[\log E^{\mathrm{R}}]\le Tc_g+r_g\sum_t\mathbb{E}[\tau_{i,t}]$. If $g$ is not constant, $c_g<0$ and the bound is negative whenever the average $\mathbb{E}[\tau_{i,t}]$ is below $-c_g/r_g$.
\end{proposition}

For the normalized power family of \S\ref{sec:method:robust} with $N_{\mathrm{cand}}=128$, $c_g=-0.414$, $r_g=3.852$ and the threshold is $0.1075$ for $\kappa=1/\ln128$ (with $\log g(1)=2.665$), and $-0.142$, $2.426$ and $0.0584$ for $\kappa=1/2$. The statistic we use averages $g$ over tied positions. That keeps the factor's mean at one, but its null baseline is then a bag-dependent constant between $c_g$ and zero, equal to zero when all candidates tie (Appendix~\ref{app:proofs:rank}); the untied price is therefore not the price of the implemented statistic when ties are frequent, as in the Atari games of \S\ref{sec:published}.

\subsection{Designs with delayed feedback}\label{sec:theory:feedback}

\begin{proposition}[Delayed feedback; new]\label{prop:closedloop}
\begin{enumerate}
\item[(a)] Under Assumption~\ref{ass:design-cl} and hypotheses (i)--(iii) of Proposition~\ref{prop:P1}, with (iii) drawing from $Q_{i,t}$, the conclusions (a)--(c) of Proposition~\ref{prop:P1} hold under $H_0^{G}$, and under $H_0^{\mathrm{own}}$ when the product form holds for the tested set conditionally on the outside blocks.
\item[(b)] A sufficient condition for Assumption~\ref{ass:design-cl}, before any eligibility conditioning, is that actions are generated as $A_{i,s}=\pi_{i,s}(H_{\cdot,s-L},A_{i,<s},U_{i,s})$ with $L\ge h$, where the policies and their parameters are fixed given the permitted information, $H_{\cdot,s-L}$ is the history of all units up to time $s-L$ that the design records (outcomes, logged states and actions), no policy reads another unit's actions inside the block, and the innovations $U_{i,s}$ are mutually independent and independent of $\mathcal{F}_t$ and of all potential outcomes.
\end{enumerate}
\end{proposition}

Restricting only the tested units' policies is not enough under spillovers: an outside unit that reacts to a tested unit's action inside the block can carry that action into the tested target. A known non-product joint design could support a single jointly normalized factor but not a product of unit-wise factors. Eligibility conditioning must be shown separately to preserve the asserted law. The condition $L\ge h$ is sufficient, not necessary. In the trap experiment of \S\ref{sec:trap}, each episode is tested as one unit, and its frozen episode-local policy draws each action from a law that depends only on that episode's own state at least $h=4$ steps before the action, with fresh randomization; treating the simulator's pseudorandom draws as independent innovations, the design-law analysis therefore falls under (b), with no eligibility conditioning and a target mask that does not depend on the actions under the null. This does not cover the deliberately marginal resampler of the negative control (\S\ref{sec:trap:nc}).

%% file: sections/05_synthetic_validation.tex
\section{Synthetic validation}\label{sec:synthetic}

Before any real outcome was reopened, the registered pooled statistic passed through four layers of synthetic checks, each fixed before it was run: development simulations with known truth (M2a), calibration on the real command records with synthetic outcomes (M2b), a power pilot and an end-to-end rehearsal (M2c, M2e), and a sealed challenge built in a separate session after the method was frozen (M3). We report what each layer tested and, equally, what it could not have detected. A fifth check, a single-field rehearsal for M5, is described at the end. Full tables are in Appendix~\ref{app:disclosures}.

\subsection{Development simulations}\label{sec:synthetic:dev}

The development grid crossed five design laws (renewal, bounded walk, Markov, clock-locked and shuffle) with linear, saturating and delayed outcome dynamics, for each of the four reference models, with 256 replicates per cell. It contained four null families---the stand-in construction of Proposition~\ref{prop:P4}, sharp-null mutual spillover (an own-channel null), shared noise under the global null, and zero-entropy designs---and two alternatives built from the masked effects of Proposition~\ref{prop:P1b}.

\paragraph{Validity.} Across all designs, dynamics and models, the exact and MC e-value tests never rejected in the stand-in family (\MtwoaDevProxyTrapPooledPerModelAcrossDesignsDynamicsTestExactRejectionsMax{} for the exact form and \MtwoaDevProxyTrapPooledPerModelAcrossDesignsDynamicsTestMCRejectionsMax{} for the MC form, in every cell), and rejected at most \MtwoaDevSharedNoisePooledPerModelAcrossDesignsDynamicsTestExactRejectionsMax{} under shared noise and \MtwoaDevSharpSpilloverPooledPerModelAcrossDesignsDynamicsTestMCRejectionsMax{} under sharp-null spillover. The batch randomization p-value at a single origin is a p-value test at level $0.05$ with $B=127$, so its attainable nominal rate is $6/128$, or $12$ of $256$ replicates. Across the \MtwoaCRTNullCells{} null cells of the grid (\MtwoaCRTNullCellsPerFamily{} in each null family) it rejected in at most \MtwoaCRTMaxCount{} of $256$ replicates, a count reached in \MtwoaCRTCellsGeTwenty{} cell, under shared noise; \MtwoaCRTCellsGeThirteen{} cells exceeded $12$. With this many cells, some counts above $12$ are expected from binomial variation alone. In the largest stand-in setting, with $512$ units, the four models rejected at rates from \MtwoaBatchRatePerModelMinPercent{} to \MtwoaBatchRatePerModelMaxPercent{}, none significantly above the nominal rate (smallest one-sided binomial tail probability \MtwoaBatchTailPMinAcrossModels{}). Validity means a rejection rate at most $\alpha$, not zero rejections; the e-value tests reject far less often than $\alpha$ because Markov's inequality is conservative.

\paragraph{The stand-in construction.} In the stand-in family the replacement diagnostics reported an action effect in up to \MtwoaDevProxyTrapPooledPerModelAcrossDesignsDynamicsBzeroRejectionsMax{} (zero block), \MtwoaDevProxyTrapPooledPerModelAcrossDesignsDynamicsBswapRejectionsMax{} (another unit's block) and \MtwoaDevProxyTrapPooledPerModelAcrossDesignsDynamicsBablRejectionsMax{} (retraining without actions) replicates, although no action had any effect. In the largest setting of the construction every model's zero-block and swap diagnostics reported an effect in every replicate (minima across models \MtwoaDevProxyTrapMarkovLinearUnitsFiveTwelvePooledPerModelRangeBzeroRejectionsMin{} for the zero block and \MtwoaDevProxyTrapMarkovLinearUnitsFiveTwelvePooledPerModelRangeBswapRejectionsMin{} for the swap), while the e-value tests never rejected (\MtwoaHeadlineExactTestRejectionsMaxAcrossModels{}; Figure~\ref{fig:trap}B). The diagnostics fail just as badly in zero-entropy designs, where the actions follow a deterministic schedule and the randomization test returns $e\equiv1$ by construction: there they reported effects in up to \MtwoaDevZeroEntropyPooledPerModelAcrossDesignsDynamicsBswapRejectionsMax{} replicates. A randomization test that resamples from the marginal frequency of each action instead of the design law (B-marg) reported effects in up to \MtwoaDevProxyTrapPooledPerModelAcrossDesignsDynamicsBmargRejectionsMax{} stand-in replicates, the same failure that \S\ref{sec:trap} reproduces in the pixel world.

\paragraph{Power where the design allows little.} The masked alternatives were built to be hard: their effects are invisible to unit-wise marginal comparisons. The e-value tests rejected in at most \MtwoaDevMaskedXorPooledPerModelAcrossDesignsDynamicsTestExactRejectionsMax{} replicates for masked XOR effects and \MtwoaDevMaskedSharedPooledPerModelAcrossDesignsDynamicsTestExactRejectionsMax{} for masked shared effects. We report this as a limitation, not a failure of validity; the sealed challenge reproduces it (\S\ref{sec:synthetic:m3}).

\subsection{Calibration with the real command records}\label{sec:synthetic:calib}

Proposition~\ref{prop:P3} does not license tests under working laws whose distance from the true law is unknown, so the working laws were calibrated on the released command records themselves. Each replicate drew a fresh random subset of real command paths and fresh synthetic outcomes from a model satisfying the sharp null, so that the estimated size averages over whatever law generated the released commands while the test resamples from the working law. The pre-committed criterion required the one-sided 95\% Clopper--Pearson upper bound on size to be at most $0.06$ in every primary cell. All \MtwobPrimaryAllDatasetsPooledPerModelCellsPassingCPNinetyFive{} of \MtwobPrimaryAllDatasetsPooledPerModelCellsTotal{} primary cells passed; the largest point false-positive rate was \MtwobPrimaryAllDatasetsPooledPerModelCellsPointFalsePositiveRateMax{} and the largest upper bound \MtwobPrimaryAllDatasetsPooledPerModelCellsCPNinetyFiveUpperMax. One generator placed the stand-in construction on the real command records, with a slowly decaying response to past commands and strongly autocorrelated noise; there the test rejected in at most \MtwobProxyTrapSTwoPrimaryAllDatasetsPooledPerModelAcrossHorizonsTestRateMax{} of replicates, whereas retraining without actions and zero-block replacement reported effects in up to \MtwobProxyTrapSTwoPrimaryAllDatasetsPooledPerModelAcrossHorizonsBablRateMax{} and \MtwobProxyTrapSTwoPrimaryAllDatasetsPooledPerModelAcrossHorizonsBzeroRateMax{} of them.

\subsection{Power pilot and end-to-end rehearsal}\label{sec:synthetic:pilot}

On the full M4 grid of origins, all \MtwocSizeFullGridPooledFourModelAvgThresholdTwentyCellsPassingCPNinetyFive{} of \MtwocSizeFullGridPooledFourModelAvgThresholdTwentyCellsTotal{} null cells passed the registered validity gate (largest upper bound \MtwocSizeFullGridPooledFourModelAvgThresholdTwentyCPNinetyFiveUpperMax), and the four-model average had minimum power across cells of \MtwocPowerPooledFourModelAvgSignalTenPercentThresholdHundredAcrossCellsMin{} at $10\%$ signal and \MtwocPowerPooledFourModelAvgSignalThirtyPercentThresholdHundredAcrossCellsMin{} at $30\%$. At $2\%$ signal, power for ERK at thirty minutes ranged from \MtwocPowerErkHthirtyPooledFourModelAvgSignalTwoPercentThresholdHundredAcrossFamiliesMin{} to \MtwocPowerErkHthirtyPooledFourModelAvgSignalTwoPercentThresholdHundredAcrossFamiliesMax, which the specification recorded as a limitation. An end-to-end rehearsal of the frozen pipeline on a simple synthetic generator rejected strongly ($\log E=\MtwoeRehearsalLogEHtenDisp$ at ten minutes).

Neither check could have revealed pooling amplification. The pilot gave each synthetic ERK field a single outcome trajectory, not hundreds of cells sharing one light program, and the rehearsal's generator was simple enough for the fitted models to be nearly correctly specified. The pooled statistic therefore entered M4 without ever having been scored on many sites that share an action sequence under a misspecified model---precisely the setting in which it fails (\S\ref{sec:pooling}). We regard this as the main gap in our pre-registration validation.

\subsection{Sealed challenge}\label{sec:synthetic:m3}

The sealed challenge was generated in a separate session after the method was frozen and registered, from generators the method author never saw, and scored with the frozen implementation; the analysis functions used were verified identical to those of M4. It comprised \MthreeFamilyCount{} families: new designs and dynamics, an omitted-history stand-in construction, spillover on a given graph, shared autoregressive outcome noise, masked XOR effects, a known approximate-law error and zero-entropy designs. Its frozen criteria were a size bound, power where the design permits it, and exact neutrality under zero entropy.

The pooled statistic registered for M4 rejected in \MthreeSizeRegisteredStatisticPooledAllFamiliesRejections{} null replicates (pooled upper bound \MthreeSizeRegisteredStatisticPooledAllFamiliesCPNinetyFiveUpper; largest per-family upper bound \MthreeSizeRegisteredStatisticPooledAcrossFamiliesCPNinetyFiveUpperMax). Zero-entropy cells returned $e$ and $E$ equal to one exactly. Of the \MthreePowerEligibleCellsTotal{} alternative cells whose exact design ceiling was at least $0.95$, \MthreePowerEligibleCellsMeetingEighty{} reached the required power of $0.80$. The exception was the masked XOR family, with \MthreePowerRegisteredStatisticPooledFZeroNineCellcZeroTwoFiveHOneExactCeilingEligibleRejections{} rejections (\MthreePowerRegisteredStatisticPooledFZeroNineCellcZeroTwoFiveHOneExactCeilingEligibleRate) against a ceiling of \MthreePowerRegisteredStatisticPooledFZeroNineCellcZeroTwoFiveHOneExactCeilingEligibleCeiling. Its effects are sharp but masked marginally, as in the masked alternatives of the development grid (\S\ref{sec:synthetic:dev}), which were also the hard case there. As in M2, the replacement diagnostics reported effects at high rates in null families (Figure~\ref{fig:m3diag}). Two disclosures apply. The challenge's designer had seen the source of the reference models and one correction to the challenge design. And none of its families was built around units with many sites sharing an action sequence (Appendix~\ref{app:disclosures} lists them), so, like the pilot, it was not designed to reveal pooling amplification.

\subsection{A single-field rehearsal for M5}\label{sec:synthetic:p3}

Before the M5 plan was finalized, the Kish-tempered statistic was rehearsed on synthetic outcomes attached to one real training field's command path, with the linear-Gaussian model only and the rejection threshold $E\ge40$ of the draft plan, which then had two primary horizons. With the simple synthetic response the statistic rejected in every replicate at both horizons (for example \PthreeRejectionsPerModelLGSoneHtenSignalTenPercentSharedCellsFiveHundredSingleField{} at ten minutes with field-wide shared noise). With the harder generator and field-wide shared noise it rejected in \PthreeRejectionsPerModelLGMtwoeHtenSignalTenPercentSharedCellsFiveHundredSingleField{} replicates at ten minutes but only \PthreeRejectionsPerModelLGMtwoeHthirtySignalTenPercentSharedCellsFiveHundredSingleField{} at thirty minutes. The registered plan therefore made ten minutes the only primary horizon, with threshold $E\ge20$; because $E\ge40$ implies $E\ge20$, these counts are lower bounds on the rehearsal's rejections at the registered threshold.

%% file: sections/06_pooling_amplification.tex
\section{Pooling amplification and its repair}\label{sec:pooling}

The registered M4 analysis failed on its own positive controls (\S\ref{sec:real:m4}). This section reports how we diagnosed the failure and selected a repair, on synthetic outcomes attached to the ERK command records, and a pre-declared demonstration of the same failure in a pixel world model. Everything here is exploratory. The diagnosis and the selection used no test outcome, and the repaired statistic was then confirmed on a field that no one had opened (\S\ref{sec:real:m5}). A later exploratory study (\S\ref{sec:pooling:baselines}) compared the repairs with statistics that estimate no factor, and its outcome fixed the primary statistic of a registered test on a published world model (\S\ref{sec:published}).

\subsection{Diagnosis on ERK-like synthetic data}\label{sec:pooling:diag}

\begin{figure}[t]
\centering
\includegraphics[width=\linewidth]{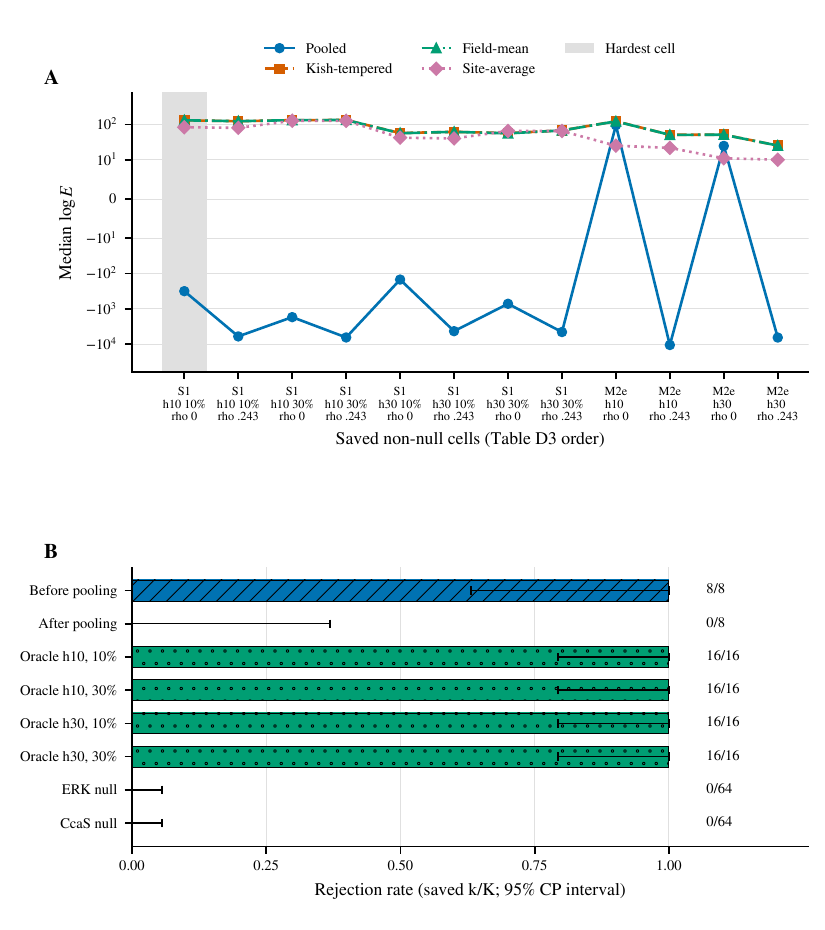}
\caption{Diagnosis and repair on ERK-like synthetic data. (A) Median $\log E$ of the pooled score and of the three repair candidates in the twelve alternative cells of the selection grid; the shaded cell is the one the selection script designated hardest (\S\ref{sec:pooling:fix}). The Kish-tempered and field-mean curves nearly coincide. The pooled score fails everywhere except without shared noise under the simple generator. (B) Rejection counts before and after the field-pooling step, for the oracle predictor in four cells, and for the pipeline under the null on ERK and CcaS commands. Error bars: $95\%$ Clopper--Pearson intervals. Symmetric-log axis in A.}
\label{fig:biosynth}
\end{figure}

We attached synthetic outcomes to the command records of the M4 test fields, with five hundred cells per field, and fitted the linear-Gaussian reference model exactly as in M4. Three findings located the failure.

First, the pooled score collapsed without any shared noise. With independent cell noise, its median $\log E$ at ten minutes was \PartCLogEPerModelLGCzeroHtenSignalTenPercentRhoZeroCellsFiveHundred{} at $10\%$ signal and \PartCLogEPerModelLGCzeroHtenSignalThirtyPercentRhoZeroCellsFiveHundred{} at $30\%$: a stronger effect made the evidence \emph{more} negative, as \eqref{eq:loge-bound} predicts when a misspecified model's error along the contrast grows with the effect it misrepresents. These medians come from the diagnostic runs, with thirty-two replicates per cell; the selection grid of \S\ref{sec:pooling:fix}, with sixteen replicates per cell, gives similar values (Table~\ref{tab:d3-fix}).

Second, the oracle predictor rejected in every replicate (\DebugOracleRejectionsPerModelOracleHtenSignalTenPercentRhoZeroCellsFiveHundred{} at ten minutes and \DebugOracleRejectionsPerModelOracleHthirtySignalThirtyPercentRhoZeroCellsFiveHundred{} at thirty), and the pipeline remained valid under the null (\DebugNullRejectionsPooledErk{} rejections with mean $e$ \DebugNullMeanFactorEErk{} on ERK commands, \DebugNullRejectionsPooledCcas{} with mean $e$ \DebugNullMeanFactorECcas{} on CcaS commands). The oracle's $\log E$ of \DebugOracleTieBoundHtenDisp{} at ten minutes is a ceiling set by ties among resampled blocks---the generator responds only to the last action of the block---and not the design bound of Proposition~\ref{prop:P2}.

Third, changing the pipeline one component at a time showed that a single step caused the collapse: summing the cells' log scores and then normalizing once over the field's design (rejections \DebugComponentSwapRejectionsPerModelLGBefore{} before that step, \DebugComponentSwapRejectionsPerModelLGAfterFieldPooling{} after it; Figure~\ref{fig:biosynth}B). Adding field-wide shared noise to the simple rehearsal generator of \S\ref{sec:synthetic:pilot} also collapsed the pooled score (median $\log E$ \FixCzeroMedianLogEPerModelLGMtwoeHtenSharedCellsFiveHundred{} at ten minutes in the selection grid of \S\ref{sec:pooling:fix}): noise shared by all cells moves them in the direction of the contrast, so it amplifies in the same way as model error.

\subsection{Selecting a repair}\label{sec:pooling:fix}

Before evaluating any candidate, we committed a protocol listing three repairs and a selection rule. The candidates were the Kish-tempered score (\S\ref{sec:method:tempering}), a field-mean score that scores the field-average outcome under a Gaussian whose variance accounts for the within-field correlation, and a site-average that averages per-cell e-values; each kept the design law, the Monte Carlo bags and the aggregation of M4. The grid crossed two generators (the first at two signal levels), two levels of shared noise and both horizons, giving twelve alternative cells with sixteen replicates each, and two null cells with thirty-two replicates each. The rule was: (1) exclude a candidate with two or more null rejections in any null cell; (2) choose the highest minimum rejection rate across alternative cells; (3) break ties by the highest median $\log E$ in the hardest cell, the one with the lowest pooled rejection rate; (4) break remaining ties by simplicity, Kish first.

All three candidates rejected in every replicate of every alternative cell (minimum rates \FixSelectionPooledPerModelLGCOneMinNonNullRejectionRate{} for Kish tempering, \FixSelectionPooledPerModelLGCTwoMinNonNullRejectionRate{} for the field mean and \FixSelectionPooledPerModelLGCThreeMinNonNullRejectionRate{} for the site average) and in no null replicate (Kish tempering: \FixSelectionPooledPerModelLGCOneNullRhoZeroHtenRejections{} and \FixSelectionPooledPerModelLGCOneNullRhoSharedHtenRejections{} in the two null cells; the other candidates likewise, Appendix~\ref{app:disclosures}). Step (2) therefore tied, and the selection fell to step (3). Because every cell tied on the pooled rejection rate, the ``hardest cell'' was not defined by the rule; the selection script took the first cell in grid order. That script was first committed together with the selection result, after the evaluation, so we cannot show that this detail was fixed in advance. In that cell the median $\log E$ was \FixSelectionPooledPerModelLGCOneHardestCellMedianLogETab{} for Kish tempering, \FixSelectionPooledPerModelLGCTwoHardestCellMedianLogETab{} for the field mean and \FixSelectionPooledPerModelLGCThreeHardestCellMedianLogETab{} for the site average. Across the twelve possible designations of the hardest cell, the rule would have selected Kish tempering in \FixCellwisePooledPerModelLGCOneMedianWinCells{} of them, the field mean in \FixCellwisePooledPerModelLGCTwoMedianWinCells{} and the site average in \FixCellwisePooledPerModelLGCThreeMedianWinCells; the Kish-minus-field-mean difference in median $\log E$ ranged from \FixCellwisePooledPerModelLGCOneMinusCTwoMin{} to \FixCellwisePooledPerModelLGCOneMinusCTwoMax. On this evidence Kish tempering and the field-mean score perform alike, both well ahead of the site average (Figure~\ref{fig:biosynth}A). Kish tempering was registered for the confirmatory test; its registration describes the selection as following the pre-committed rule, and we correct that description here (Appendix~\ref{app:disclosures}).

\subsection{A pixel world model}\label{sec:pooling:pixel}

To show the failure where machine-learning practice meets it, we ran a demonstration under a protocol frozen before any learned model was trained, with dated predictions committed before the main run. A Gaussian blob moves on an $R\times R$ frame; each action pushes it in one of four directions or leaves it, and a delayed restoring force adds memory. Frames carry independent pixel noise and one of three kinds of frame-level shared noise: none, an additive offset, or a multiplicative gain on the blob. Actions are drawn uniformly at random, so the design law is known exactly. Two learned models predict the frame four steps ahead from the last two frames and the action window, a convolutional network with action conditioning and a per-pixel ridge regression, and are compared with oracles that know the true dynamics. Each episode is one unit and its $n=R^2$ pixels are its sites, from $64$ to $4096$; the effect size was calibrated on the oracle alone. Every statistic uses Monte Carlo e-values with $B=127$ and the rejection rule $E\ge20$, and reports the average of the two models.

\paragraph{The pooled score collapses as pixels are added, and the cause is misspecification.} Figure~\ref{fig:pooling} shows the main grid. With no shared noise and the larger effect, the median $\log E$ of the pooled score fell from \PxTTwoCzeroTwoModelAvgNoneDHiPixelsSixtyFourTrainSixtyFourGridGOneMedianLogEDisp{} at $64$ pixels to \PxTTwoCzeroTwoModelAvgNoneDHiPixelsFortyNinetySixTrainSixtyFourGridGOneMedianLogEDisp{} at $4096$, while the oracle's rose from \PxTTwoOracleCzeroOracleNoneDHiPixelsSixtyFourTrainSixtyFourGridGOneMedianLogEDisp{} to \PxTTwoOracleCzeroOracleNoneDHiPixelsFortyNinetySixTrainSixtyFourGridGOneMedianLogEDisp. The pattern was the same under offset and gain noise (pooled score at $4096$ pixels \PxTTwoCzeroTwoModelAvgOffsetDHiPixelsFortyNinetySixTrainSixtyFourGridGOneMedianLogEDisp{} and \PxTTwoCzeroTwoModelAvgGainDHiPixelsFortyNinetySixTrainSixtyFourGridGOneMedianLogEDisp; oracle \PxTTwoOracleCzeroOracleOffsetDHiPixelsFortyNinetySixTrainSixtyFourGridGOneMedianLogEDisp{} and \PxTTwoOracleCzeroOracleGainDHiPixelsFortyNinetySixTrainSixtyFourGridGOneMedianLogEDisp), and from $256$ pixels on the pooled score never rejected under any noise (\PxCzeroRejectRateTwoModelAvgDHiMaxPixelsGeTwoFiftySix{}). Because the oracle does not collapse under either form of shared noise, the collapse of the fitted models is due to their misspecification. Our protocol had predicted that the gain noise would make even the oracle collapse; it did not, because a multiplicative gain rescales the oracle's evidence without changing its sign.

\paragraph{Kish tempering does not repair it here.} The pixel residuals of a misplaced blob take both signs, so their median pairwise correlation is nearly zero without shared noise or with gain noise (at most \PxRhoHatMaxNoneGain{} across cells), the Kish factor is close to one, and Kish tempering behaves like the pooled score: it rejected in \PxTTwoConeTwoModelAvgNoneDHiPixelsSixtyFourTrainSixtyFourGridGOneRejectRateDisp{} of episodes at $64$ pixels and in none from $256$ on. Under offset noise the residual correlation rose to between \PxRhoHatOffsetMin{} and \PxRhoHatOffsetMax{}, and Kish tempering did reject, but for the wrong reason: the offset raises the residual correlation while being orthogonal to the action contrast, and at $4096$ pixels the resulting over-tempering cut its rejection rate to \PxTTwoConeTwoModelAvgOffsetDHiPixelsFortyNinetySixTrainSixtyFourGridGOneRejectRateDisp. Kish tempering, the statistic registered for M5, therefore does not repair pooling amplification in the pixel world. The fractional-Kelly reference arm of \S\ref{sec:method:betting}, which bounds each factor below but keeps the pooled bet, rejected in \PxCfourRejectRateTwoModelAvgDHiMin{} to \PxCfourRejectRateTwoModelAvgDHiMax{} of episodes at the larger effect (Appendix~\ref{app:pixel-grid}).

\paragraph{Contrast tempering restores power, not information.} Contrast tempering, declared in the protocol as an exploratory generalization, rejected in \PxConeCRejectRateTwoModelAvgDHiMin{} to \PxConeCRejectRateTwoModelAvgDHiMax{} of episodes across all noise conditions and frame sizes at the larger effect, against \PxTTwoOracleJointOracleNoneDHiPixelsSixtyFourTrainSixtyFourGridGOneRejectRateDisp{} for the oracle with the exact joint likelihood. Its median $\log E$, however, stayed between \PxConeCMedianLogETwoModelAvgDHiMin{} and \PxConeCMedianLogETwoModelAvgDHiMax{} and did not grow with the number of pixels, far below the oracle: it removes the amplification but does not convert the additional pixels into additional evidence. At the smaller effect every learned statistic was weak (contrast tempering at most \PxConeCRejectRateTwoModelAvgDLoMax{}). Section~\ref{sec:pooling:baselines} shows that statistics estimating no factor matched these rejection rates, and \S\ref{sec:published} that contrast tempering did not carry over to a published world model. \begin{figure}[t]
\centering
\includegraphics[width=\linewidth]{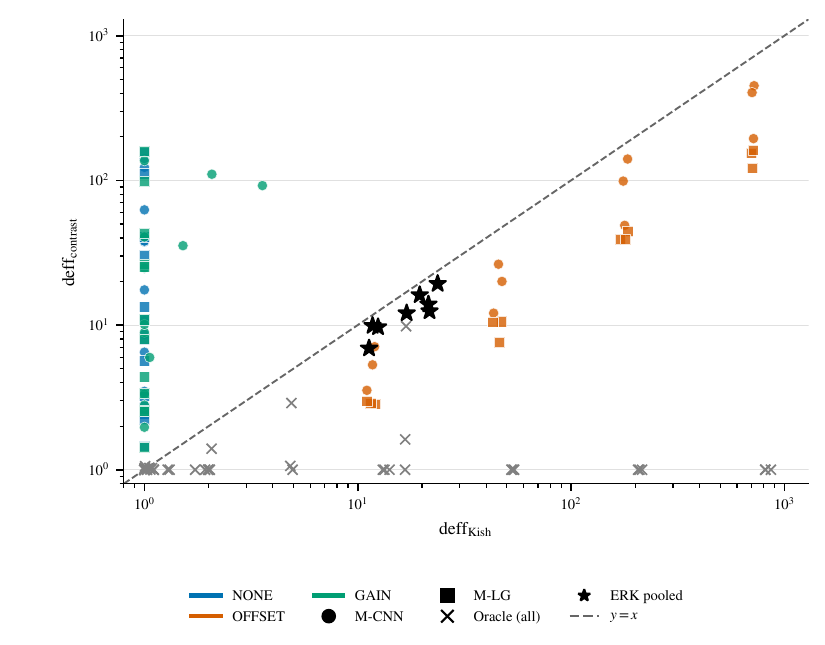}
\caption{Two design effects. The Kish design effect from the median pairwise residual correlation ($x$) against the design effect along the action contrast ($y$), for every learned-model cell of the pixel main grid (colour: shared noise; shape: model), every oracle cell (grey crosses), and the pooled ERK training diagnostic (stars; four models and two horizons, at the diagnostic's cell counts). Without shared noise or with gain noise, residual correlations are near zero and the Kish factor misses the correlation along the contrast (points above the diagonal); a frame-wide offset correlates the residuals but is orthogonal to the contrast (points below). The ERK points lie slightly below the diagonal, so Kish tempering is somewhat conservative there. Oracle residuals are independent pixel noise, so their contrast design effect is close to one.}
\label{fig:deff}
\end{figure}

The mechanism of \S\ref{sec:method:tempering} was visible directly (Figure~\ref{fig:deff}). For the learned models in the main grid, the contrast design effect exceeded the Kish design effect in every row without shared noise or with gain noise (\PxDeffContrastExceedsKishRowsNoneGain{}) and fell below it in every offset row (\PxDeffContrastBelowKishRowsOffset{}), as predicted before the main run. For the oracles, whose residuals are independent pixel noise, both design effects are close to one and their ordering carries no information (Appendix~\ref{app:pixel-grid}).

\paragraph{Scope.} Every statistic was valid in every null cell. The demonstration uses one simulator family, two model classes and one training run per cell; compute limits reduced the episode length and the number of test episodes, so the intervals for alternative cells are wide (Appendix~\ref{app:pixel-grid}). Contrast tempering has only this simulation evidence; on a published world model it did not repair pooling amplification (\S\ref{sec:published}), and we do not recommend it.

\subsection{Statistics that estimate no factor}\label{sec:pooling:baselines}

Both repairs above estimate a tempering factor from validation residuals. We asked whether statistics that need no such estimate do as well: the tempering mixture M over a fixed grid of thirteen factors, validation-tuned tempering V, and the rank e-values R and R$_{1/2}$ (\S\ref{sec:method:robust}). Before any of them was computed we committed a protocol with dated predictions and mechanical decision rules. We then recomputed every existing statistic from the saved models, episodes and seeds of the pixel demonstration and of the selection grid of \S\ref{sec:pooling:fix}, on a different machine, and computed the new statistics on the same draws. The recomputation reproduced every rejection decision of the original runs. Its reproduction criterion, as first written, used a relative tolerance that is ill-posed for values near zero; we amended it before any new statistic was computed (Appendix~\ref{app:baselines}). This study is exploratory.

\begin{table}[t]
\centering
\small
\caption{Exploratory baseline study. Rejections in the pixel world (twelve cells of sixteen episodes at each effect size, two-model average, $E\ge20$) and the number of ERK-like alternative cells (of twelve) in which every one of sixteen replicates was rejected ($E\ge100$). A dash: not defined in that setting.}
\label{tab:baselines}
\begin{tabular}{@{}l r r r@{}}
\toprule
Statistic & Pixel, larger effect & Pixel, smaller effect & ERK-like, cells at 16/16 \\
\midrule
Contrast-tempered (C1c) & \PoneDhiCOnecRejections/192 & \PoneDloCOnecRejections/192 & -- \\
Kish-tempered (C1) & \PoneDhiKishRejections/192 & \PoneDloKishRejections/192 & \PoneFixKishFullCells/12 \\
Tempering mixture (M) & \PoneDhiMRejections/192 & \PoneDloMRejections/192 & \PoneFixMFullCells/12 \\
Validation-tuned (V) & \PoneDhiVRejections/192 & \PoneDloVRejections/192 & -- \\
Rank e-value (R) & \PoneDhiRRejections/192 & \PoneDloRRejections/192 & \PoneFixRFullCells/12 \\
Rank e-value (R$_{1/2}$) & \PoneDhiRhalfRejections/192 & \PoneDloRhalfRejections/192 & \PoneFixRhalfFullCells/12 \\
\bottomrule
\end{tabular}
\end{table}

Table~\ref{tab:baselines} gives the results. At the larger effect the mixture and validation tuning matched contrast tempering, and the mixture also rejected where Kish tempering had failed. The pre-registered rule therefore returned outcome~\PoneOutcome: a statistic that estimates no factor matched contrast tempering in the pixel world and Kish tempering on ERK-like data. Validation tuning matched contrast tempering, so the contrast formula added nothing over a temperature chosen on validation data. At the smaller effect the mixture was weaker than contrast tempering and validation tuning, though not than Kish tempering. Averaging can reduce log evidence relative to the best grid member by up to $\log 13\approx2.56$, which can matter near a rejection threshold; the bound alone does not determine these observed differences. The rank e-value R failed, against our dated prediction. On the ERK-like grid its largest count in any cell was \PoneFixRMaxCellRejections{} of sixteen replicates, consistent with the expected price per factor that a rank calibrator pays when evidence is spread over thousands of weakly informative factors (Proposition~\ref{prop:rank}, stated for untied scores); R$_{1/2}$, a secondary arm, did well in the pixel world but likewise failed on ERK-like data (largest count \PoneFixRhalfMaxCellRejections{} of sixteen). No new statistic rejected more than \PoneNullMaxRejectionsNewArms{} of the episodes of any null cell. On the strength of this study the mixture became the primary statistic of the registered test in \S\ref{sec:published}.

\subsection{A regime map (exploratory)}\label{sec:regime}

Section~\ref{sec:pooling:baselines} compared the statistics in two settings. To see where each one works, we ran a simulator in which the relevant quantities can be set directly, under a protocol committed before any statistic was computed (Appendix~\ref{app:exploratory}). A unit has 200 origins, and recorded actions are drawn uniformly from four. The executed action equals the recorded one with probability $p$ and is otherwise an independent draw. Each of $n$ sites responds to the executed action through a fixed pattern, plus shared noise of variance $\rho$ that moves the sites the way an action would, plus independent noise. The model predicts the response to the recorded action, with an error along the action contrast that is a fraction $\delta$ of the effect and is shared by all sites; it assumes that the recorded action was executed. The grid crossed $n \in \{16, 256, 4096\}$, $\rho \in \{0, 0.05\}$, $\delta \in \{0, 0.25, 0.5\}$, $p \in \{1, 0.5, 0.2\}$ and three effect sizes calibrated on the oracle alone, with 32 test units in each of the 162 alternative cells and 64 in each of the 18 null cells.

The pooled score behaved as \S\ref{sec:method:free} predicts. With $\delta = 0$ and $p \le 0.5$ it collapsed at 4096 sites in all 12 such cells: when the recorded action is not always executed, a model that assumes it was is misspecified along the contrast even if its response pattern is exact. Kish tempering did no better than the pooled score, with mean rejection rates of 0.222 and 0.221 over the alternative cells; contrast tempering reached 0.496. Validation tuning had the highest mean rate, 0.635, followed by the weighted mixtures (0.625, 0.609 and 0.588 for $w = 3/4, 1/2, 1/4$) and the uniform mixture (0.560). In the 98 cells of the simulator and of the saved pixel runs in which some candidate rejected in at least 80%

The protocol's mechanical rule chose the primary statistic of the next registered test among M, W(1/4), W(1/2), W(3/4) and V. It selected V, which tied with W(3/4) within the rule's tolerance and won on mean rate. The simulator could not test V on ERK-like data, because the saved ERK-like runs lack the validation residuals V needs. Section~\ref{sec:erkxm} tests V on the real ERK test fields, where it collapsed. For that reason the next registered test used W(3/4) instead of V, a departure from the rule that was recorded in the registered protocol before any of that test's data existed (\S\ref{sec:p4:design}). A later exploratory analysis showed that the collapse on the ERK fields was specific to tuning one factor for fields with different numbers of cells: with a site-count-adaptive factor, validation tuning no longer collapsed, though it still lost most of the evidence for some models (\S\ref{sec:erkrho}). The choice of W(3/4) was registered before that analysis.

%% file: sections/07_standin_trap.tex
\section{The stand-in trap}\label{sec:trap}

Section~\ref{sec:setup:standin} and Proposition~\ref{prop:P4} describe how a predictor can make future actions look effective when they are not. The development simulations and the calibration on real command records showed the trap on biological designs (\S\ref{sec:synthetic}). Here we build it into the pixel world, where every quantity is known, and test what the randomization test needs in order to avoid it.

\subsection{A pixel world with a stand-in}\label{sec:trap:pixel}

\begin{figure}[t]
\centering
\includegraphics[width=\linewidth]{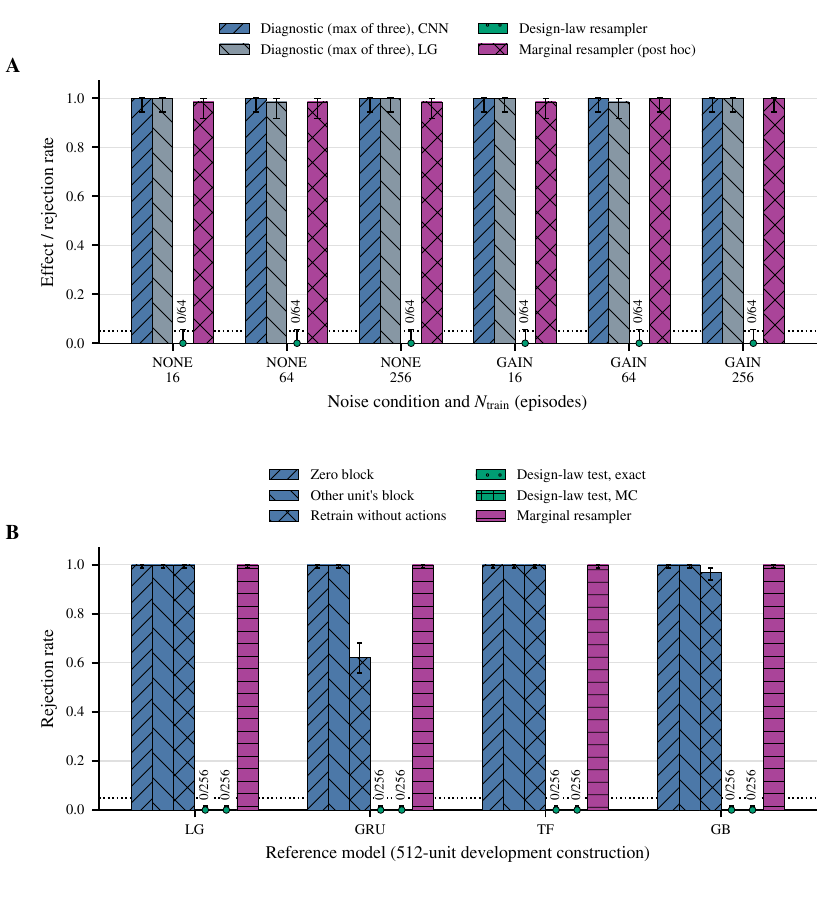}
\caption{The stand-in trap. (A) Pixel world whose logging policy depends on the blob's position six steps earlier, which the models do not see; no action has any effect. For each cell (shared noise and number of training episodes; $64$ test episodes), the highest effect-report rate among the three replacement diagnostics for each learned model, and the contrast-tempered randomization test with the design-law resampler, which conditions on the full logged state, and with a resampler that draws actions from their marginal frequencies (post hoc, \S\ref{sec:trap:nc}). (B) Development stand-in construction with $512$ units ($256$ replicates), for each reference model: the three replacement diagnostics, the exact and Monte Carlo e-value tests with the design law, and a randomization test with a marginal resampler. Error bars: $95\%$ Clopper--Pearson intervals. Dotted line: $\alpha=0.05$.}
\label{fig:trap}
\end{figure}

The logging policy now depends on state the model cannot see: with probability $0.6$ it chooses the action that points from the blob's position six steps earlier toward the centre of the frame, and each other action with probability $0.1$. The model sees only the last two frames, so the future actions carry information about an earlier position that the restoring force still acts on. The actions themselves have no effect. We used $256$-pixel frames, training sets of $16$, $64$ and $256$ episodes, and either no shared noise or gain noise. The randomization test resamples each future action from the logging policy given the full logged state, which the design records.

The replacement diagnostics---retraining without actions, setting the actions to zero, and shuffling the action windows across origins within an episode---reported an action effect in most test episodes (Figure~\ref{fig:trap}A): for the convolutional model in \PxcheckTrapDiagRatePerModelCNNGainAcrossTrainingSizesHfourPixelsTwoFiftySixMin{} to \PxcheckTrapDiagRatePerModelCNNNoneAcrossTrainingSizesHfourPixelsTwoFiftySixMax{} of them, and for the ridge model in \PxcheckTrapDiagRatePerModelLGNoneAcrossTrainingSizesHfourPixelsTwoFiftySixMin{} to \PxcheckTrapDiagRatePerModelLGGainAcrossTrainingSizesHfourPixelsTwoFiftySixMax, across diagnostics, training sizes and noise conditions. With the design-law resampler the randomization test never rejected: the Kish-tempered statistic rejected in \PxTrapConeTwoModelAvgRejectRateMaxGTwo{} of episodes in every cell, and the contrast-tempered statistic plotted in Figure~\ref{fig:trap}A in none for the two-model average (\S\ref{sec:trap:nc}). The experiment does not show the growth of the diagnostics' confidence with data that Proposition~\ref{prop:P4} describes: with the smallest training set, the convolutional model's diagnostics already reported effects in at least \PxTrapDiagRateCNNTrainSixteenMin{} of episodes, leaving no room to grow. That property rests on the proposition and on the development simulations (\S\ref{sec:synthetic:dev}).

\subsection{The resampler must condition on the design state}\label{sec:trap:nc}

The protocol included a negative control: the Kish-tempered statistic with a deliberately wrong resampler that draws each action from its marginal frequency in the training episodes, ignoring the logged state. It was expected to reject often. It never rejected (\PxTrapNcFrozenConeRejectionsMaxGTwo{} episodes in every cell), and the pre-declared criterion for the negative control was not met. The result is uninformative rather than reassuring: without shared noise or with gain noise, Kish tempering coincides with the pooled score at this frame size, and the pooled score has no power there (\S\ref{sec:pooling:pixel}). A statistic whose $\log E$ is driven far below zero by amplification cannot reveal that a resampler is invalid.

After seeing this result we added one post hoc analysis, declared in a dated note committed before any of its e-values were computed: the same wrong resampler with the contrast-tempered statistic, which does have power, on the sealed episodes, frozen models and frozen tempering factors of the six trap cells. The wrong resampler rejected in \PosthocNcWrongRateTwoModelAvgAcrossNoiseAcrossTrainingSizesHfourPixelsTwoFiftySixMin{} to \PosthocNcWrongRateTwoModelAvgAcrossNoiseAcrossTrainingSizesHfourPixelsTwoFiftySixMax{} of episodes for the two-model average (\PosthocNcWrongRatePooledPerModelRangeAcrossNoiseTrainingSizesMin{} to \PosthocNcWrongRatePooledPerModelRangeAcrossNoiseTrainingSizesMax{} for the individual models). With the correct resampler, whose results were read from the sealed records rather than recomputed, the same statistic never rejected for the two-model average (\PosthocNcProperRejectionsTwoModelAvgAcrossNoiseAcrossTrainingSizesHfourPixelsTwoFiftySixMax{} episodes in every cell); one single-model cell had \PosthocNcProperRejectionsPerModelLGGainTrainSixteenHfourPixelsTwoFiftySix. The wrong resampler rejected more often in \PosthocNcWrongExceedsCorrectComparisonsPooled{} comparisons. This analysis is post hoc and exploratory; it does not replace the frozen negative control, whose criterion remains unmet.

Two lessons follow. First, the conditioning in Assumption~\ref{ass:design} is not a technicality: a resampler that ignores part of the state the design used is far from the design law in total variation, nothing in Proposition~\ref{prop:P3} protects it, and it reproduces the false effect that the replacement diagnostics report. Second, a negative control needs a statistic with power; pooling amplification can hide invalidity as easily as it hides an effect.

\subsection{Summary across settings}\label{sec:trap:summary}

The same three-way pattern appears in every setting we examined. Replacement diagnostics report effects at high rates when there is none: in the development simulations, on real command records with synthetic outcomes, in the sealed challenge, and in the pixel world. A randomization test that resamples from the design law, conditioning on the full state the design used, keeps its error rate. A randomization test that resamples from a simpler law loses that guarantee and reproduces the diagnostics' error. On the real ERK data the diagnostics reported an effect that is biologically established, so we do not count those reports as instances of the trap (\S\ref{sec:real:m4}).

%% file: sections/08_real_data.tex
\section{Real data}\label{sec:real}

The real-data evidence has three stages, each with a different evidential status. M4 is the registered analysis of two previously analyzed datasets. X is a single exploratory application of the repaired statistic to the M4 test fields. M5 is a registered confirmatory test of the repaired statistic on a field that no one involved had opened. Appendix~\ref{app:osf} gives registration identifiers, hashes and the time-ordering of every data access; Appendix~\ref{app:disclosures} lists every deviation and disclosure.

\subsection{Data, units and design laws}\label{sec:real:data}

\paragraph{ERK.} In the ERK experiment~\citep{naleczjawecki2023erk}, with data released as \citet{naleczjawecki2022erkdata}, each microscope field receives one light program, a sequence of pulses whose inter-pulse intervals were randomized by the acquisition protocol, and the field's hundreds of cells are recorded under that common program. The field is therefore the assignment unit and its cells are sites; the spillover graph between fields is empty, so the primary null is $H_0^{G}$. The working design law, V3, shuffles the inter-pulse intervals uniformly within interval groups, conditionally on the elapsed interval at the origin, and preserves the protocol's rest segment; two variants, V1 and V2, serve as sensitivity laws. Fields run under a binary policy are deterministic and contribute $e\equiv1$. The reference models were trained on four minimum-gap fields from the first replicate and tested on four minimum-gap fields from the second replicate and five ordinary-interval fields.

\paragraph{CcaS/CcaR.} In the CcaS/CcaR experiment~\citep{lugagne2024ccas}, with data and trained models released as \citet{lugagne2023ccasdata}, each cell lineage receives its own light program from a random walk, bounded in the released acquisition code, so each lineage is a unit with a single site. The working law, D1, is the generator in the released acquisition code delayed by one step, consistent with a shift at data export; the delay was found after the first design-law checks failed (Appendix~\ref{app:assignment}). Every CcaS rejection below is conditional on D1. Lineages that follow one of two duplicated fixed templates have zero design entropy and contribute $e\equiv1$; the remaining lineages are tested under the design law conditioned on avoiding both templates (Proposition~\ref{prop:elig}(b)), which keeps the test exact. In addition to the four reference models, we test the published model of \citet{lugagne2024ccas} with its released preprocessing, after replacing its neighbour-exposure input by the lagged, admissible version (\S\ref{sec:setup:nulls}).

\paragraph{Models and prior use.} The four reference models (linear-Gaussian, GRU, small Transformer and gradient boosting) are additive: an own-history branch plus an action branch fitted to its residual. They were trained on training-unit outcomes only and frozen by hash before any test outcome was opened; their Gaussian variances come from held-out validation data. Both datasets were analyzed earlier in our BioWM-Eval study (Appendix~\ref{app:biowm-eval}); M4 is therefore a re-analysis, not a blind test, although every M4 setting was fixed on synthetic data and command records before the outcomes were reopened.

\subsection{M4: the registered analysis}\label{sec:real:m4}

M4 tests five prespecified hypotheses with the pooled statistic (Table~\ref{tab:m4}): ERK at horizons of ten and thirty minutes, registered as positive controls because optogenetic stimulation is known to drive ERK activity; CcaS at one and six five-minute steps with the four-model average; and CcaS at twelve steps with the published model. A slot is rejected iff $E\ge100$, the e-Bonferroni rule for five hypotheses at family level $0.05$.

\begin{table}[t]
\centering
\small
\caption{M4 primary results (pooled statistic, registered analysis). Natural-log e-values; reject iff $E\ge100$.}
\label{tab:m4}
\begin{tabular}{@{}l l l r l@{}}
\toprule
Slot & Data & Horizon & $\log E$ & Decision \\
\midrule
1 & ERK (positive control) & 10 min & \MfourLogESlotOneTab & not rejected \\
2 & ERK (positive control) & 30 min & \MfourLogESlotTwoTab & not rejected \\
3 & CcaS, four-model average & 1 step & \MfourLogESlotThreeTab & not rejected \\
4 & CcaS, four-model average & 6 steps & \MfourLogESlotFourTab & rejected \\
5 & CcaS, published model & 12 steps & \MfourLogESlotFiveTab & rejected \\
\bottomrule
\end{tabular}
\end{table}

Neither positive control was rejected, and the failure was not marginal: every reference model gave a strongly negative $\log E$ for ERK (from \MfourLogEPerModelSummaryErkMinDisp{} to \MfourLogEPerModelSummaryErkMaxDisp), and so did both sensitivity laws. By the interpretation registered with M4, a strongly negative $\log E$ for a model means that its predictions under the observed actions scored worse than its predictions averaged over the design; it is not evidence against an effect. A dated note written after we had read the M4 report fixes further interpretation rules; it was written post hoc, and its rules only restrict what may be claimed (Appendix~\ref{app:disclosures}). Section~\ref{sec:pooling} shows that the failure is pooling amplification: the ERK score sums hundreds of cells that share one action sequence.

Both CcaS hypotheses at the longer horizons were rejected. Restricting the published model to lineages whose lagged exposure could be verified from commands gives $\log E=\MfourLugagneVerifiedLogEDisp$. A second sensitivity analysis that uses the raw neighbour exposure with singleton averaging gives \MfourLugagneRawNSingletonLogEDisp; it changes both the exposure alignment and the aggregation, is not a valid test, and is reported descriptively only. The secondary own-channel analysis at six steps gives $\log E$ from \MfourOwnLogEPerModelSummaryCcasHsixMinDisp{} to \MfourOwnLogEPerModelSummaryCcasHsixMaxDisp{} across models, but its eligibility bound is vacuous (Proposition~\ref{prop:elig}(a)), so it is descriptive.

Two secondary results bear on the diagnosis. The batch randomization p-value at a single prespecified origin gave per-model values from \MfourBatchPPerModelSummaryErkHthirtyMinFrac{} to \MfourBatchPPerModelSummaryErkHthirtyMaxFrac{} for ERK at thirty minutes and \MfourBatchPPerModelSummaryErkHtenMinFrac{} for every model at ten minutes (unadjusted, secondary). A rank-based p-value depends only on where the observed score ranks among the design draws, not on the size of the score differences, so it does not suffer from the over-aggressive betting that drives pooling amplification. The contrast with the e-value product is consistent with the diagnosis; but the registered interface covers such p-values only at one common origin, whereas e-values combine evidence across all origins. Second, the replacement diagnostics reported an ERK action effect, with risk differences from \MfourRiskDifferencePooledSummaryErkMin{} to \MfourRiskDifferencePooledSummaryErkMax. The ERK effect is biologically real, so these are most likely true positives and we do not use them as examples of the stand-in trap. For CcaS at one step the same diagnostics reported effects of magnitude \MfourRiskMagnitudePooledSummaryCcasHoneMin{} to \MfourRiskMagnitudePooledSummaryCcasHoneMaxSci{} while the randomization test did not reject; the truth there is unknown, so this is descriptive only.

\subsection{X: an exploratory check on the M4 test fields}\label{sec:real:x}

After diagnosing the failure (\S\ref{sec:pooling}), we selected the Kish-tempered statistic on synthetic data and applied it once, with every parameter fixed on training data, to the nine ERK test fields of M4. We call this analysis X. It is exploratory: the fields had been opened in M4, and the analysis was run because the pooled statistic had failed there. No test outcome was used to choose the statistic or its parameters. The four-model $\log E$ was \XLogEHtenDisp{} at ten minutes and \XLogEHthirtyDisp{} at thirty minutes, against \MfourLogESlotOneDisp{} and \MfourLogESlotTwoDisp{} for the pooled statistic on the same data; the individual models ranged from \XLogEPerModelSummaryAcrossHorizonsMinDisp{} to \XLogEPerModelSummaryAcrossHorizonsMaxDisp. The exact-form design bound of Proposition~\ref{prop:P2} for these fields is \XExactDesignBoundHtenDisp{} at both horizons, and X reached \XLogRatioToExactBoundPooledHten{} and \XLogRatioToExactBoundPooledHthirty{} of it on the log scale. Because X used Monte Carlo e-values, this is a descriptive comparison rather than a ceiling; the corresponding Monte Carlo bound of Proposition~\ref{prop:P2}(a), computed in the M5 registration package from the frozen Monte Carlo bags, is \XMCDesignBoundHtenDisp{} at ten minutes and \XMCDesignBoundHthirtyDisp{} at thirty. The ratios are not power estimates. These values are reported descriptively: X was prompted by outcomes already opened in M4, and no formal error guarantee is claimed for it.

\subsection{M5: a confirmatory test on an unopened field}\label{sec:real:m5}

\begin{figure}[t]
\centering
\includegraphics[width=\linewidth]{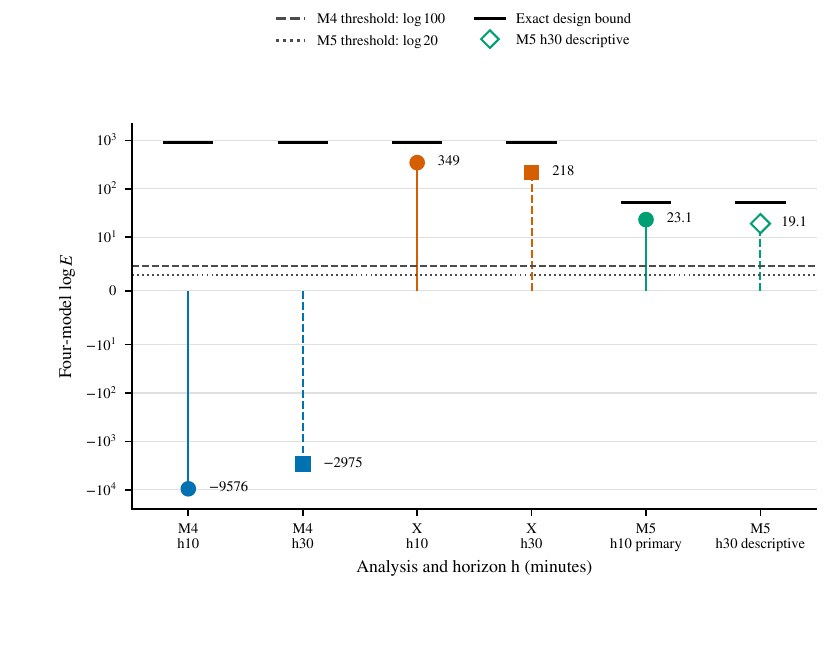}
\caption{The ERK evidence chain. Four-model $\log E$ of the registered pooled statistic in M4 (nine test fields), and of the Kish-tempered statistic in the exploratory analysis X (the same fields) and in the confirmatory test M5 (one unopened field; thirty minutes descriptive only). Horizontal bars mark the exact-form design bound of each analysis (Proposition~\ref{prop:P2}), a command-only comparator for these Monte Carlo e-values; dashed and dotted lines mark the rejection thresholds of M4 ($E\ge100$) and M5 ($E\ge20$). Symmetric-log axis.}
\label{fig:realchain}
\end{figure}

\paragraph{Design.} M5 tests the repaired statistic on one ERK field, a minimum-gap field with thirty-minute minimum intervals from the first replicate. It had been excluded from M4 by the scope of the registered sample (minimum gaps of ten to twenty-five minutes, two replicates), not for data quality. None of our projects had opened it, and before registration the first author recorded a statement that he had never opened, previewed or plotted its outcomes (Appendix~\ref{app:osf}). The analysis plan was committed before X was run and was not changed afterwards (its committed and registered bytes are identical). It was registered before the data were obtained, in a new registration linked to the M4 registration, together with a disclosure that the X results were available when the plan was finalized. The only primary hypothesis is $H_0^{G}$ at ten minutes, rejected iff $E\ge20$; thirty minutes is computed and reported descriptively only. The registered plan chose this horizon before any M5 data were obtained, because a single-field rehearsal had shown weak power at thirty minutes under shared noise (\S\ref{sec:synthetic}). The statistic is the Kish-tempered score with all parameters frozen on the training fields, the working law V3, and Monte Carlo e-values with $B=127$.

\paragraph{Command-law gate.} Before any outcome statistic was computed, a registered gate checked the reconstructed command series alone: that commands were shared by all cells, that every non-rest inter-pulse interval was at least thirty minutes (the minimum was \MfiveMinimumNonRestMinutes{} minutes), that the frozen parser found the rest segment (at interval rank \MfiveRestSegmentIndex), and that the exchangeability diagnostics used for the M4 fields did not flag the interval sequence ($p=\MfiveGateDiagnosticPOne$, $\MfiveGateDiagnosticPTwo$ and $\MfiveGateDiagnosticPThree$, against the inherited per-statistic threshold $0.01/39$). The gate passed.

\paragraph{Result.} The primary hypothesis was rejected (Figure~\ref{fig:realchain}): the four-model $\log E$ at ten minutes was \MfiveLogEHtenDisp, that is $E\approx\MfiveEHtenScientific$. All four models agreed (\MfiveLogEPerModelLGHtenDisp, \MfiveLogEPerModelGRUHtenDisp, \MfiveLogEPerModelTFHtenDisp{} and \MfiveLogEPerModelGBHtenDisp). The field had \MfiveHtenCells{} cells and \MfiveHtenOrigins{} origins; \MfiveHtenValidCellOrigins{} cell--origin pairs contributed, \MfiveHtenExcludedCellOrigins{} were excluded by the observation mask, and \MfiveHtenZeroEntropyFactors{} factors had zero design entropy. The exact-form design bound is \MfiveExactDesignBoundHtenDisp, and the statistic reached \MfiveLogRatioToExactBoundPooledHten{} of it on the log scale, a descriptive comparison as for X. At thirty minutes the descriptive $\log E$ was \MfiveLogEHthirtyDisp{} (models from \MfiveLogEPerModelSummaryHthirtyMinDisp{} to \MfiveLogEPerModelSummaryHthirtyMaxDisp); it carries no confirmatory decision.

\paragraph{What M5 shows.} Conditional on the registered design, observation-process and frozen-training assumptions, with the working law V3, the rejection is evidence against the sharp null $H_0^{G}$ on a new field: M5 detects the known ERK action effect with the repaired statistic. It does not estimate the size of the effect, it says nothing about power for CcaS or other systems, and it does not validate the forecast gains reported in BioWM-Eval, which answer a different question with a with-versus-without-actions comparison of the kind \S\ref{sec:trap} shows can be inflated. The evidence is narrow: one field, one dataset, one horizon. The field's thirty-minute intervals lie outside the training range of ten to twenty-five minutes, which may reduce power; the exact acquisition law for this field is unverified, so validity is conditional on V3; and the gate's threshold, inherited from the multi-field M1 analysis, is lenient for a single field.

\subsection{Kish versus contrast tempering on the ERK training fields}\label{sec:real:diag}

To ask whether the registered Kish factor is a reasonable choice on ERK, we compared it with the contrast factor of \S\ref{sec:method:tempering} on the training fields' validation cells. This diagnostic is exploratory and descriptive: it used training outcomes only, computed no e-value, and touched no test field; its reading rule was fixed before the data were accessed. Expressing the contrast factor as an implied correlation $\rho_c$, the pooled ratio $\rho_c/\hat\rho$ ranged from \ErkDiagRhoRatioPooledHtenMin{} to \ErkDiagRhoRatioPooledHtenMax{} across models at ten minutes and from \ErkDiagRhoRatioPooledHthirtyMin{} to \ErkDiagRhoRatioPooledHthirtyMax{} at thirty minutes. All values fall inside the pre-declared band from $0.5$ to $2$, so by the pre-declared reading Kish tempering is approximately adequate on ERK; all are below one, so Kish tempering over-tempers somewhat and is conservative there. Individual fields gave ratios from \ErkDiagRhoRatioSingleFieldMin{} to \ErkDiagRhoRatioSingleFieldMax{}; Appendix~\ref{app:erk-field} lists every field.

Two qualifications follow from Proposition~\ref{prop:kish}. Because the action branch of the ERK reference models predicts the same response for every cell of a field, the contrast is shared across cells and the contrast factor has the form of Proposition~\ref{prop:kish}: a Kish factor with a contrast-weighted residual correlation, at an effective cell count since counts vary across origins. The agreement is therefore partly structural, and the diagnostic is not an independent validation of the Kish rule. And the gap is not only a matter of median versus mean: against the mean pairwise residual correlation the ratio at ten minutes was still between \ErkDiagPooledPerModelGRUHtenRhoContrastOverRhoMean{} and \ErkDiagPooledPerModelTFHtenRhoContrastOverRhoMean, so contrast weighting itself lowers the effective correlation; we did not decompose the remainder further. The diagnostic used the validation cells of each training field, fewer than the cells of a test field, so comparisons at test-field size assume that correlations do not change with the number of cells. We did not compute contrast-tempered e-values on the M5 field, and we make no claim about what they would have been.

\subsection{Validation tuning on the ERK test fields (exploratory)}\label{sec:erkxm}

Validation tuning and the mixtures had no ERK entry in \S\ref{sec:pooling:baselines}, because the saved ERK-like runs lack the validation quantities they need. We therefore computed them on the nine ERK test fields of M4, as a second exploratory analysis after X, under a protocol committed before the fields were reopened (Appendix~\ref{app:exploratory}). Everything was as in X---the models, the design law V3 and X's frozen Monte Carlo bags---except that each model's factor for V was chosen on the validation cells of the four training fields, with 31 design draws per origin. A reproduction gate first recomputed the four-model log $E$ of the pooled statistic in M4 and of the Kish-tempered statistic in X; both matched.

Validation tuning collapsed (Table~\ref{tab:erkxm}). Its four-model log $E$ was 210 at ten minutes, with two of the four models strongly negative ($-1420$ and $-979$), and $-101$ at thirty minutes, with all four models negative. The uniform mixture gave 347 and 218, W(3/4) 345 and 217, and Kish tempering 349 and 218 as in X; the mixture was within $\log 15$ of the best fixed grid member at both horizons, as Proposition~\ref{prop:mixture} requires. The factor chosen on the training fields was 1/32 to 1/4, whereas the per-origin Kish factor on the test fields was about 0.008 to 0.04. One reading is that a factor tuned on the training fields' validation cells, which are fewer than the cells of a test field, is too large for the test fields. Section~\ref{sec:erkrho} tests this reading. The analysis is exploratory for the same reasons as X: the fields had been opened, and no error guarantee is claimed for it.

These fields cannot serve a confirmatory test of the mixture, and no other field of this dataset can either: an inventory of the released ERK data, using archive metadata and our own records only, found no randomized field that we could certify as never opened by any of our projects (Appendix~\ref{app:exploratory}). A registered biological test of the mixture therefore needs new data.

\input{appendix/tables/V3_erkxm.tex}

\subsection{A site-count-adaptive factor on the ERK test fields (exploratory)}\label{sec:erkrho}

If \S\ref{sec:erkxm}'s reading is right, a factor that follows the number of cells should not collapse. We repeated the check of \S\ref{sec:erkxm} with the site-count-adaptive statistics of \S\ref{sec:method:robust}, under a protocol committed before the fields were reopened. The analysis was gated on reproducing the values of \S\ref{sec:erkxm} for C0, C1, V and M at both horizons. A first attempt failed this gate because of a difference in preprocessing and was discarded; after the preprocessing was matched, all eight comparisons passed (Appendix~\ref{app:a3e}).

The collapse disappeared (Table~\ref{tab:rho}). The four-model log $E$ of V$_\rho$ was 350.5 at ten minutes and 206.1 at thirty, against 210 and $-101$ for V. But V$_\rho$ was still not safe for every model: at ten minutes its log $E$ was 36.2 for the linear-Gaussian model, 176.9 for the GRU, 300.7 for the Transformer and 351.9 for gradient boosting, while the adaptive mixture M$_\rho$ gave 293.7, 321.3, 298.0 and 349.2. The four-model value hides this, because it is dominated by the best model. The mixtures and Kish tempering were within a few units of each other at both horizons. So the statement of \S\ref{sec:erkxm} needs a qualification: on these fields validation tuning collapsed because one factor was tuned for fields of different sizes; tuned in a form that follows the site count, it did not collapse, but it still lost most of the evidence for one model, and the mixtures did not. The analysis is exploratory for the same reasons as \S\ref{sec:erkxm}.
\input{appendix/tables/V4_rho.tex}

%% file: sections/08b_published_model.tex
\section{A published world model}\label{sec:published}

Every model in the preceding sections was built by us. If pooling amplification is a property of pooled scores and not of our models, it should appear on a world model trained by others, and the statistic chosen in \S\ref{sec:pooling:baselines} should repair it there. We tested both claims in a registered confirmatory study on DIAMOND~\citep{alonso2024diamond}, a diffusion world model for Atari games whose authors released trained checkpoints. We neither trained nor modified the models.

\subsection{Design}\label{sec:published:design}

\paragraph{Models and environment.} We used the released checkpoints for Boxing, Pong and Breakout at a fixed revision, with their hashes recorded before registration (Appendix~\ref{app:published}). The Arcade Learning Environment~\citep{bellemare2013arcade}, an interface to the Stella Atari 2600 emulator, was configured exactly as for DIAMOND's training: four frames per action, the maximum of the last two frames, $64\times64$ RGB frames, no sticky actions and the minimal action sets of $18$, $6$ and $4$ actions. The model predicts the next frame from the last four frames and actions with a three-step sampler; at each origin one noise image was drawn and shared by every candidate action, so the prediction is a fixed function of the candidate. The candidate block is the next action ($h=1$), and the sites are the $n=64\cdot64\cdot3=12{,}288$ pixel channels. The score is the Gaussian log density of the observed next frame around the predicted one, with per-site variances estimated on validation units and floored at the square of one intensity step.

\paragraph{Two conditions.} In the ACTIVE condition the emulator executed the recorded action, drawn uniformly at random; the games include action-responsive motion of the player, paddle or ball, and the rejection counts assess detection in this setting, without asserting that the scored frame differs at every eligible origin. In the NULL condition the recorded labels were drawn the same way but the emulator executed an independent random sequence, so the labels have no effect on any frame and the null holds exactly. A unit is one segment of up to $500$ agent steps from a reset. Each game and condition had $16$ validation units; there were $32$ ACTIVE and $64$ NULL test units per game.

\paragraph{Statistics and criteria.} The primary statistic was the tempering mixture over fifteen factors, the grid extending below $1/n$ (\S\ref{sec:method:robust}). The pooled score, Kish and contrast tempering, validation-tuned tempering, the thirteen-value mixture and the two rank e-values were secondary. Every statistic used the same Monte Carlo bags ($B=127$), rejected at $E\ge20$, and took its factors from the validation units of its own game and condition. The registered criteria, per game, were: \emph{validity}, each statistic rejecting in at most $2$ of the $64$ NULL units; \emph{collapse}, a median pooled $\log E$ below zero in the ACTIVE units; and \emph{repair}, the mixture rejecting in at least $24$ of the $32$ ACTIVE units. Overall confirmation required validity in all three games and both collapse and repair in at least two. Five dated predictions were registered with the criteria.

\paragraph{Registration and seeds.} The protocol, the identity of the frozen analysis code and a note on its dates were registered on OSF (ctfxv) before any confirmatory unit existed. The confirmatory seeds were derived from the first beacon round scheduled at least two minutes after the start of the minute in which the registration was recorded, so no unit could have been generated earlier. A pilot on separate seeds had run the full pipeline under NULL only; under ACTIVE it computed predictions and validation quantities but no e-value. Two integrity checks of the frozen code stopped the confirmatory run before any ACTIVE test unit was generated: a field-name collision in the access log, and a storage tolerance tighter than float32 rounding. Each was fixed by a dated deviation published on OSF before the run resumed with the same seeds, and every regenerated unit was identical, array by array, to its earlier copy (Appendix~\ref{app:published}).

\subsection{Results}\label{sec:published:results}

\begin{table}[t]
\centering
\small
\caption{Registered test on DIAMOND. Rejections ($E\ge20$) in the ACTIVE ($/32$) and NULL ($/64$) test units of each game, and the median $\log E$ of the ACTIVE units. The mixture over fifteen factors (M) is the primary statistic.}
\label{tab:published}
\input{appendix/tables/P2_results.tex}
\end{table}

Table~\ref{tab:published} gives the results. Pooling amplification appeared on the published models: although actions visibly move the player, paddle or ball, the pooled score rejected in \PtwoCZeroActiveRejectionsTotal{} of the $96$ ACTIVE units, with median $\log E$ between \PtwoCZeroActiveMedianLogEMinDisp{} and \PtwoCZeroActiveMedianLogEMaxDisp{} across games. Tempering by an estimated factor did not prevent it: Kish tempering rejected in \PtwoKishActiveRejectionsTotal{} and contrast tempering in \PtwoContrastActiveRejectionsTotal{} of the $96$ units, with medians from \PtwoKishActiveMedianLogEMinDisp{} to \PtwoKishActiveMedianLogEMaxDisp{} and from \PtwoContrastActiveMedianLogEMinDisp{} to \PtwoContrastActiveMedianLogEMaxDisp. The mixture rejected in \PtwoMActiveRejectionsTotal{} of the $96$ units, with medians from \PtwoMActiveMedianLogEMinDisp{} to \PtwoMActiveMedianLogEMaxDisp; validation tuning, the thirteen-value mixture and both rank e-values did as well. Under the exact null the eight statistics made \PtwoNullRejectionsAllStatisticsTotal{} rejections in $8\times192$ tests, and the mixture's median $\log E$ stayed between \PtwoMNullMedianLogEMinDisp{} and \PtwoMNullMedianLogEMaxDisp, comparable in magnitude to $\log 15\approx2.71$, the largest log-e loss relative to the best member of the fifteen-factor grid (Proposition~\ref{prop:mixture}), although the proposition does not assert a null loss of that size.

All three criteria held in all three games, so the study confirmed its claim. Of the five predictions, \PtwoPredictionsMet{} held. The one that failed was ours about ranks: we predicted that the rank e-value would reject less often than the mixture in at least two games, and it matched the mixture in all three. The replacement diagnostic reported an action effect in \PtwoReplacementEffectsActiveTotal{} of the $96$ ACTIVE units and in \PtwoReplacementEffectsNullTotal{} of the $192$ NULL units; because the NULL labels are independent of everything the model sees, this design produces no stand-in effect, and the trap of \S\ref{sec:trap} is not tested here.

The estimated factors were an order of magnitude or more above the factor selected on validation data. Across games, $\lambda^{\mathrm{K}}$ ranged from \PtwoLambdaKishActiveMin{} to \PtwoLambdaKishActiveMax{} and $\lambda^{\mathrm{c}}$ from \PtwoLambdaContrastActiveMin{} to \PtwoLambdaContrastActiveMax, whereas $\lambda^{\mathrm{V}}$ was \PtwoLambdaVActiveMin{} to \PtwoLambdaVActiveMax; the contrast factor exceeded it by a factor of \PtwoContrastOverVActiveMin{} to \PtwoContrastOverVActiveMax. With predictions, residuals and contrasts held fixed, both estimated factors are invariant under a common multiplier of all site precisions, whereas for unfloored Gaussian scores the tempered log-e curve shifts by the inverse multiplier along the temperature axis (Proposition~\ref{prop:blind}). This is consistent with a possible mechanism for the failure; it is a descriptive, post hoc reading and not a test (Appendix~\ref{app:published} gives the saved score bags it rests on).

\subsection{What this shows}\label{sec:published:scope}

On world models that we did not build, the pooled design-based test collapsed although actions mattered, tempering by factors estimated from residuals did not prevent the collapse, and a mixture fixed in advance restored power while every statistic kept its error rate under an exact null. The study does not show that DIAMOND predicts the consequences of actions correctly: a rejection is evidence that the recorded actions affect the frames, read through the model's predictions, not a validation of those predictions. The action effects in these games are strong, so the study says little about power for weak effects, where the mixture was weaker than contrast tempering in the pixel world (\S\ref{sec:pooling:baselines}). It covers one model family, three games and a horizon of one step. The weighted mixture was defined after this study. Computed afterwards from the saved scores of the registered run, as part of the over-betting study (\S\ref{sec:overbet}, Appendix~\ref{app:exploratory}), W(3/4) rejected in all 96 ACTIVE units and in 0 of the 192 NULL units. This computation is post hoc and is not part of the registered test. Section~\ref{sec:p4} extends the test to two further model families.

\subsection{Score rule and sampling noise (exploratory)}\label{sec:score-sampling}

Two objections to \S\ref{sec:published:results} can be checked on the saved data: that the collapse is an artifact of the per-pixel Gaussian score, and that a single sampled prediction is too noisy. Both studies below are exploratory, under protocols committed before they were run (Appendix~\ref{app:exploratory}).

\paragraph{Score rule.} We rescored every saved DIAMOND unit under ten alternative score rules, with predictions, observations and bags unchanged: two higher variance floors, $(8/255)^2$ and $(32/255)^2$; one common variance for all sites; Student-t ($\nu = 3$) and Laplace densities; averages over blocks of 2\(\times\)2, 4\(\times\)4 and 8\(\times\)8 pixels; a random projection to 256 dimensions; and a joint Gaussian whose covariance is a rank-16 principal-component model of the validation residuals plus a diagonal. A reproduction gate first recovered every rejection decision of the registered run exactly. Under every rule, no statistic rejected in any of the 64 NULL units of any game, and the mixture rejected in all 32 ACTIVE units of every game. The pooled score collapsed under every rule in at least two of the three games, including under the joint Gaussian (median log $E$ $-366{,}580$, $-66{,}379$ and $-4324$). Coarser scores weakened the collapse: with 8\(\times\)8 blocks, 192 sites instead of 12,288, the median pooled log $E$ was $-6656$ in Boxing, $-77$ in Pong and 67 in Breakout, against $-222{,}582$, $-21{,}805$ and $-5408$ per pixel. Kish tempering reached 24 of 32 units in 1 of the 27 game--rule pairs in which it was defined. Contrast tempering depended on the rule: it reached 24 of 32 units in Pong or Breakout under the higher floors and the coarser blocks, and in Boxing only under the random projection.

\paragraph{Sampling noise.} DIAMOND samples its prediction, and the registered test used one sample per origin. We regenerated all 288 units of the three games with predictions averaged over four samples, after a reproduction gate had recovered a registered unit bit for bit. The collapse was unchanged: the median pooled log $E$ of the ACTIVE units was $-220{,}241$, $-22{,}354$ and $-5553$, against $-222{,}582$, $-21{,}805$ and $-5408$ with one sample. Kish and contrast tempering again rejected in 0 of 32 units in every game; validation tuning, the mixture and both rank e-values rejected in all 32; under the null every statistic rejected in 0 of 32 units. The factor chosen on validation data was 0.002 to 0.004, against Kish factors of 0.018 to 0.32.

%% file: sections/08c_two_models.tex
\section{Two further published world models}\label{sec:p4}

Section~\ref{sec:published} covers one model family. We registered a second confirmatory test on two families that differ from DIAMOND, and from each other, in how they predict and how they can be scored: IRIS (\citealp{micheli2023transformers}), which compresses each frame into 16 discrete tokens and predicts the next frame's tokens with an autoregressive Transformer, and TD-MPC2 (\citealp{hansen2024tdmpc2}), which predicts in a learned latent space and has no decoder. Both were trained and released by their authors; we neither trained nor modified them.

\subsection{Design}\label{sec:p4:design}

\paragraph{Models and environments.} For IRIS we used the released checkpoints of all 26 games of its benchmark, none excluded, with IRIS's own preprocessing: four frames per action, the maximum of the last two frames, 64 \(\times\) 64 RGB frames, no sticky actions and minimal action sets. For TD-MPC2 we used released single-task checkpoints for three DeepMind Control Suite tasks (\citealp{tassa2018control})---cartpole-balance, reacher-easy and walker-walk---and three MyoSuite tasks (\citealp{caggiano2022myosuite})---myo-pose, myo-reach and myo-obj-hold. Code revisions and checkpoint hashes, and one wrapper patch that changes no dynamics, were frozen before registration (Appendix~\ref{app:p4}).

\paragraph{Scores.} For IRIS, the primary score is the model's own log-likelihood of the 16 tokens of the observed next frame, computed by teacher forcing. It is a joint likelihood over the tokens, not a sum of independent site scores. As a secondary score in Boxing, Pong and Breakout, we decoded a sampled next frame and scored it with the per-pixel Gaussian of \S\ref{sec:published}, over 12,288 sites. For TD-MPC2, which has no decoder, the score is a Gaussian on the 512 coordinates of the predicted next latent state, centred on the encoding of the observed next observation, with per-coordinate variances from validation units. The candidate block is the next action ($h = 1$) throughout.

\paragraph{Conditions.} Recorded labels were drawn uniformly at random. In an ACTIVE condition with execution probability $p$, the executed action was the recorded label with probability $p$ and otherwise an independent uniform draw; $p = 1$ is full execution. In the NULL condition the executed actions were drawn independently of the labels, so the null holds exactly. Every game and task had $p = 1$ and NULL; Boxing, Pong, Breakout and the six control tasks also had $p = 0.5$, 0.2 and 0.05. Partial execution makes the action effect intermittent: the model assumes the recorded action was executed, so it is misspecified along the action contrast even where its dynamics are right. Each game or task and condition had 16 validation units and 32 test units, 64 under NULL. A unit is one segment of up to 500 steps from a reset, and of up to 100 or 75 steps for the MyoSuite tasks.

\paragraph{Primary statistic.} The primary statistic was the validation-weighted mixture W(3/4) of \S\ref{sec:method:robust}. The regime map's rule had selected V (\S\ref{sec:regime}), but V collapsed on the ERK test fields (\S\ref{sec:erkxm}). Before any data of this test existed, we therefore registered W(3/4), which keeps most of V's power when V's factor is right and the mixture's protection when it is wrong. The later analysis of \S\ref{sec:erkrho} showed that V's collapse on the ERK fields can be removed by a site-count-adaptive factor. The registered choice predates it; with that analysis in hand the case against V is weaker than it looked, though V$_\rho$ still lost most of the evidence for one model. The pooled score, Kish and contrast tempering (Gaussian scores only), V, M, M13, W(1/4), W(1/2) and both rank e-values were secondary. All statistics used the same bags ($B = 127$), took their factors from the validation units of their own game or task and condition, and rejected at $E \ge 20$.

\paragraph{Criteria and predictions.} For each primary arm---IRIS tokens and TD-MPC2---the registered criteria per game or task were validity, every statistic rejecting in at most 2 of the 64 NULL units, and power, W(3/4) rejecting in at least 24 of the 32 units at $p = 1$. An arm was confirmed if validity held in every game or task and power held in at least 18 of the 26 games, or in at least 4 of the 6 tasks. Eight dated predictions were registered with the criteria.

\paragraph{Registration, seeds and deviations.} The protocol and the identity of the frozen code were registered on OSF (j75g2) on 9 November 2026. As in \S\ref{sec:published}, the seeds were derived from the first beacon round scheduled at least two minutes after the start of the minute in which the registration was recorded. A pilot on separate seeds had run the full pipeline under NULL only; under ACTIVE it computed predictions and validation quantities but no e-value. The confirmatory run generated all 5392 units once. Its last step, writing the decision file, failed on a data-type error in the JSON writer. A dated deviation changed only that conversion and wrote the decision from the stored, hashed results, and an independent script that does not use the frozen code reproduced every decision value. An earlier attempt had stopped before generating anything, because of an error in our run instruction, not in the protocol (Appendix~\ref{app:p4}).

\subsection{Results}\label{sec:p4:results}

Table~\ref{tab:p4} summarizes the results; Appendix~\ref{app:p4} gives every game, task and condition.

\paragraph{Validity.} In the two primary arms the statistics made 4 rejections in 19,200 NULL tests, none of them by the primary statistic. No statistic rejected in more than 1 of the 64 NULL units of any game or task, so validity held everywhere. The secondary pixel score made 0 rejections in 2112 NULL tests.

\paragraph{Power.} W(3/4) rejected in at least 24 of 32 units at $p = 1$ in 24 of the 26 IRIS games and in all six TD-MPC2 tasks, so both arms were confirmed. It missed in BankHeist (16 of 32) and UpNDown (8 of 32), where no statistic reached the bar; validation tuning had 19 and 9.

\paragraph{The model's own likelihood.} We had predicted, with probability 0.5, that the pooled token score would collapse at $p = 1$ in at least two of Boxing, Pong and Breakout. It collapsed in none of the three (median log $E$ 144, 428 and 121). It collapsed in all three once execution was partial (at $p = 0.5$: $-3258$, $-2615$ and $-617$). The protocol had fixed this outcome as meaning that amplification needs misspecification along the action contrast, and that a proper likelihood does not prevent it once the action model is wrong. Across all 26 games, which the prediction did not cover, the pooled token score had a negative median log $E$ at $p = 1$ in 14 and reached 24 of 32 units in 10. The factor chosen on validation data for tokens was between 1/16 and 1 across the 26 games, with median 1/4; in the three games scored both ways it was 1/4 to 1/2 for tokens and 1/1024 to 1/256 for decoded pixels. The model's own likelihood needs far less tempering than an independent pixel score, but the factor chosen was below 1 in 25 of the 26 games.

\paragraph{Pixels and latents.} On the decoded pixels, the pattern of \S\ref{sec:published} recurred on a second model family: the pooled score and Kish tempering rejected in 0 of 96 units, contrast tempering in 20, and V, M and W(3/4) in 96. On TD-MPC2's latent score, the pooled score reached the power bar in 1 of 6 tasks, Kish tempering in 2 and contrast tempering in 4. In reacher-easy and walker-walk the Kish factor was 1 in the NULL and partial-execution conditions---the residuals showed no correlation to correct for---while the factor chosen on validation data was between $6 \times 10^{-5}$ and $8 \times 10^{-3}$.

\paragraph{Partial execution.} As execution became rarer, the pooled score failed first (Figure~\ref{fig:p4-partial}). At $p = 0.5$ it rejected in 0 of 32 units in all nine sets, while W(3/4) reached 24 of 32 in seven. At $p = 0.2$ W(3/4) reached the bar in four sets, and at $p = 0.05$ no statistic did in any set. W(3/4) rejected in more units than the pooled score at $p = 0.2$ in eight of the nine sets.

Of the eight dated predictions, seven held; the one that failed is the prediction about the token likelihood above. The replacement diagnostic reported an effect in 1 to 7 of the 64 NULL units of each game or task, and in most units at $p = 1$; as in \S\ref{sec:published}, the NULL labels are independent of everything the model sees, so this design produces no stand-in effect.

\input{appendix/tables/V3_p4.tex}
\input{figures/v3/figures.tex}
\subsection{What this shows}\label{sec:p4:meaning}

On two further model families and three kinds of score, the registered mixture kept its error rate under exact nulls and, at full execution, reached the power bar wherever any statistic did. Pooling amplification was not confined to Gaussian pixel scores: it appeared on a latent score and, in more than half of the games, on a model's own token likelihood, although there much less tempering was needed. Factors estimated from residuals failed again; on two control tasks the Kish factor saw nothing to correct. The study does not validate the counterfactual predictions of these models: a rejection says that the recorded actions affect the observations, read through the model. It covers simulators with random actions and a horizon of one step. At $p=.05$ none of the statistics evaluated in this study reached 24 of 32 in any set. Section 13 reports a subsequent registered comparison with bounded stakes.

%% file: figures/v3/figures.tex
\begin{figure}[t]
\centering\includegraphics[width=\linewidth]{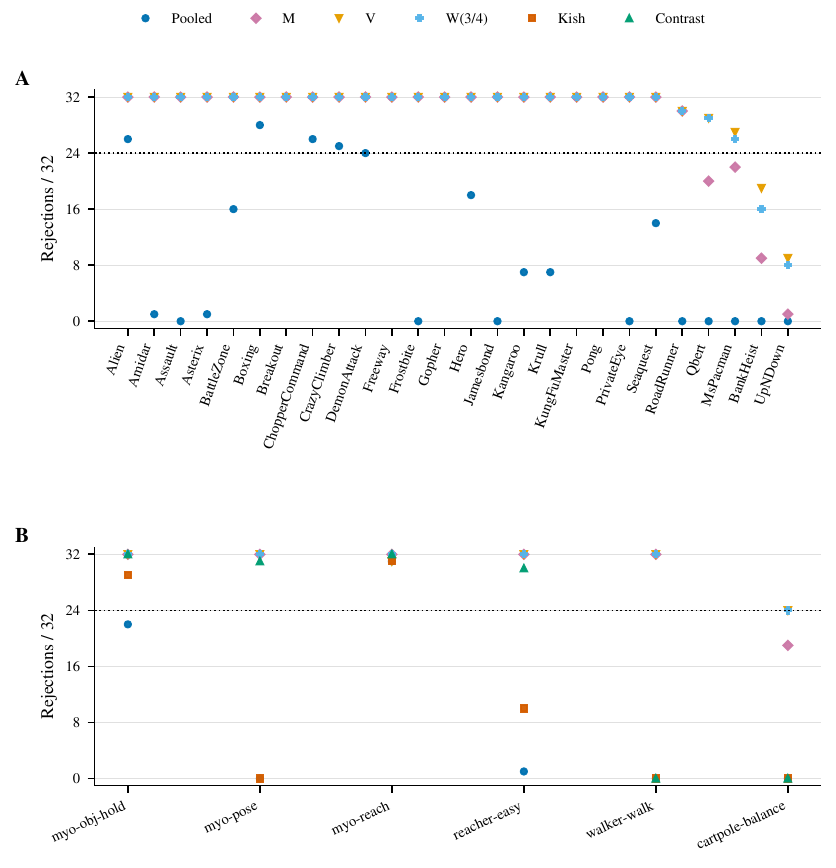}
\caption{Registered test on IRIS and TD-MPC2: rejections out of 32 test units at full execution. The line marks the registered power bar.}\label{fig:p4-full}
\end{figure}
\begin{figure}[t]
\centering\includegraphics[width=\linewidth]{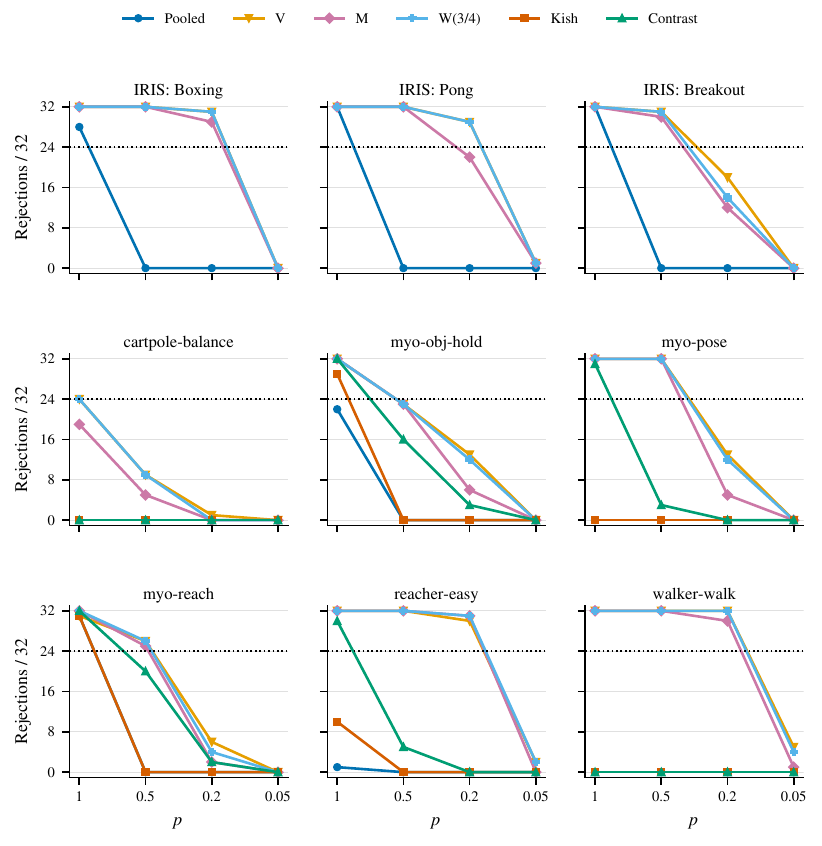}
\caption{Partial execution. Rejections out of 32 test units as the recorded action is executed with probability $p$. The pooled score fails first; at $p = 0.05$ no statistic reaches the power bar.}\label{fig:p4-partial}
\end{figure}

%% file: sections/11_robot_data.tex
\section{Real robot data with a published model}\label{sec:p5}

Sections~\ref{sec:published} and~\ref{sec:p4} used simulators, where the null can be imposed and the data regenerated. We registered a third confirmatory test on recorded robot video, which cannot be replayed, with a published model that we did not train.

\subsection{Design}\label{sec:p5:design}

\paragraph{Data and model.} BAIR robot pushing (\citealp{ebert2017self}) records a robot arm pushing objects: 30 frames per trajectory at 64 \(\times\) 64 pixels, with a four-component command at every frame. iVideoGPT (\citealp{wu2024ivideogpt}) compresses a frame into tokens and predicts the next frame's 16 tokens from one context frame and the command. We used its released action-conditioned checkpoint for these data, unchanged.

\paragraph{Units and score.} The study set is the official test trajectories 16--255. Their images had not been opened by us before registration; trajectories 0--15 had been read in a feasibility check and served the pilot only. The recorded commands change every two frames. We kept the 239 trajectories in which this holds exactly and excluded one. A unit is one trajectory with 14 origins, the frames at which a new command starts, without the first. The candidate block is the command at the origin ($h = 1$), and the score is the model's own log-likelihood of the 16 tokens of the observed next frame, by teacher forcing, as for IRIS in \S\ref{sec:p4}. A public beacon split the units at random into 80 validation and 159 test units.

\paragraph{Two arms.} In the NULL arm the label of each origin is our own draw, so the null holds by construction and the test is exact. In the recorded-command arm the robot is a known positive control. The rejection counts describe sensitivity under the two working laws and the assumed command/frame alignment.

\paragraph{Two design laws.} A design-based test needs the law from which the recorded commands were drawn. The software and configuration of the 2017 collection could not be established, so we worked from the command records: successive commands are uncorrelated (within \(\pm\)0.011) and show no dependence on the arm's position near the edge of the workspace. Under the primary law, L-emp, the candidates of a unit are recorded commands of training trajectories, drawn without replacement from a fixed pool of 14,294. Conditional on the realised pool, these candidates follow the pool's empirical law, with the remaining entries changing as draws are used. Even if every command were an independent draw from one scene-independent population, a fresh recorded command need not have that conditional law, so L-emp is a working law, not an exact design law. Under the sensitivity law, L-par, candidates are fresh draws from a law fitted to the records. Before the code was frozen, the commands of the study set and of the pool were compared from records alone; the four tests gave $p$ = 0.20, 0.34, 0.16 and 0.40. Validity for the recorded-command arm requires the conditional exchangeability and timing conditions in Appendix L.3. The constructed label-null test uses the assignments we controlled.

\paragraph{Statistics, criteria and predictions.} The primary statistic was again W(3/4). The pooled score, V, M, W(1/4), W(1/2), both rank e-values and two online rules (\S\ref{sec:method:robust}) were secondary. In the NULL arm, V and W(3/4) were also computed at the factor chosen by the ACTIVE arm: a factor chosen on null data sits near zero, where these statistics hardly move and their null check says little. All statistics used bags of 128 and rejected at $E \ge 20$. The registered criteria were validity, each of the twelve NULL statistics rejecting in at most 4 of the 159 test units under each law, and power, W(3/4) rejecting in at least 120 of the 159 under L-emp. Six dated predictions were registered with them.

\paragraph{Registration and run.} The protocol and the frozen code were registered on OSF (4wj7k) on 10 November 2026, after a pilot that ran the constructed null only, on 48 trajectories outside the study set, and a rehearsal of the complete confirmatory path on synthetic scores. The seeds came from the first beacon round scheduled at least two minutes after the start of the minute in which the registration was recorded. The run was generated and analysed once, with no deviation. An independent script, written after the results were committed, reproduced every unit e-value and decision count.

\subsection{Results}\label{sec:p5:results}

\paragraph{Validity.} Under the constructed null, no statistic rejected in more than 2 of the 159 test units under L-emp, or in more than 1 under L-par (Table~\ref{tab:p5}). The primary statistic rejected in none, also at the factor of the ACTIVE arm.

\paragraph{Power.} With the recorded commands, W(3/4) rejected in 155 of the 159 test units under L-emp, against the required 120, and in 156 under L-par. The registered count criterion was met under L-emp; as a statement about the recorded commands it is a sensitivity result under the working assumptions of \S\ref{sec:p5:design}. The two laws agreed in 158 of the 159 units.

\paragraph{The pooled score did not collapse.} It rejected in 143 units, with a median log $E$ of 13.9 against 15.3 for W(3/4). We had given a collapse 40\%. The validation factor was 1/2 under both laws. As in Boxing, Pong and Breakout at full execution (\S\ref{sec:p4:results}), a model's own joint likelihood over 16 tokens, applied to commands that were executed, over-bets only mildly.

\paragraph{What tempering still buys.} The pooled score failed in 16 units, and in 13 of them its log $E$ was negative, down to $-134.5$. In these units the loss was concentrated: the single worst origin accounted for the median unit's whole net loss (exploratory). The smallest log $E$ was $-52.5$ for V and $-1.8$ for W(3/4), whose loss against the best grid member is bounded. Each unit's best factor on a fine grid was below 1 in 150 of the 159 units (median 0.59), and the curve is flat below its maximum and steep above it (Figure~\ref{fig:p5}b), as Proposition~\ref{prop:overbet} implies: at a factor of 4 the median log $E$ was $-41$. Used on fewer origins, the statistics separate further: with the first four origins of each unit, W(3/4) rejected in 122 units and the pooled score in 104, and the pooled count fell from 130 to 127 between nine and eleven origins while the mixture's never fell (Figure~\ref{fig:p5}a). These descriptions were planned in the protocol and are not part of its criteria.

\paragraph{Other statistics.} M rejected in 154 units, V in 155, the rank e-values in 153 and 154, and the online rules in 149 and 153. The replacement diagnostic was positive in 151 of the 159 units with recorded commands and in 4 under the null.

\paragraph{Predictions.} Five of the six held: validity (95\%), power (65\%), the pooled score rejecting in fewer units than W(3/4) (65\%), a validation factor below 1 (75\%) and agreement of the two laws within 15 units (80\%). The prediction that the pooled score's median log $E$ would be negative (40\%) failed.

\input{appendix/tables/V4_p5.tex}
\input{figures/v4/fig_p5_float.tex}
\subsection{What this test shows}\label{sec:p5:meaning}

On recorded video, the constructed label-null checks used a known assignment law. With the recorded commands, the registered mixture rejected in 155 of 159 trajectories under L-emp, a sensitivity result under the working law and timing assumptions. The test does not show a collapse of the pooled score on real data. It places these data in the regime in which the pooled score mostly works, and there the mixture cost nothing in rejections and bounded the loss in the units where the pooled score failed.

Three limits apply to the ACTIVE arm. The design law is reconstructed from the records, not documented. The pool's commands were part of the model's training data and the recorded commands of the study units were not; L-par, whose candidates the model has never seen, gave the same result. And the study units belong to the official test set, which the model's authors may have used to choose the checkpoint. None of these affects the NULL arm.

%% file: figures/v4/fig_p5_float.tex
\begin{figure}[htbp]
\centering\includegraphics[width=\linewidth]{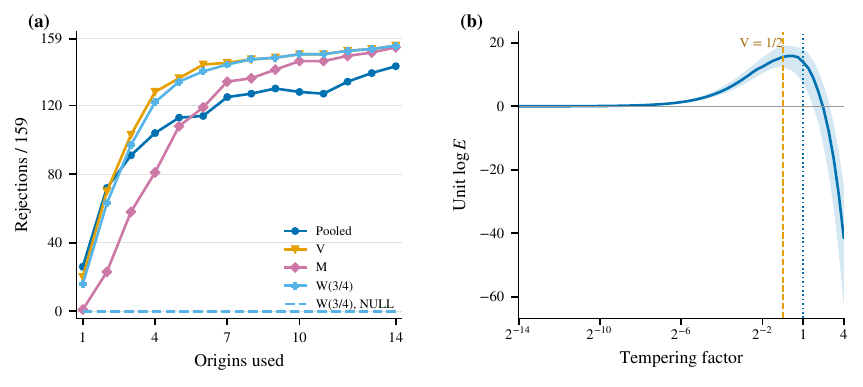}
\caption{Real robot data. (a) Rejections as origins are added. The pooled count is not monotone; the mixture's is. (b) The unit log e-value as a function of the tempering factor: flat below its maximum, steep above it. Descriptive; planned in the registered protocol.}\label{fig:p5}
\end{figure}

%% file: sections/12_overbetting.tex
\section{One account of every collapse (exploratory)}\label{sec:overbet}

Sections~\ref{sec:pooling} to~\ref{sec:p5} report collapses of the pooled score, of Kish and contrast tempering, and of validation tuning. Proposition~\ref{prop:overbet} says what they must have in common: a tempered one-step statistic can have a negative log e-value only if its factor exceeds the break-even factor of that product. We checked this, and how far the failing factors were from the best ones, on the saved scores of the DIAMOND study and of the IRIS and TD-MPC2 study, under a protocol committed before the computation (Appendix~\ref{app:exploratory}). The analysis is exploratory: the units had been analysed before.

\paragraph{The asymmetry holds in every unit.} In all 4512 units (288 DIAMOND, 288 IRIS decoded-pixel, 2784 IRIS token and 1152 TD-MPC2 test units, including NULL and every stored execution condition), the unit log e-value was concave in the factor on a fine grid, the proportional bound of Proposition~\ref{prop:overbet}(b) held, and every negative value lay above the break-even factor. The mixture's bound of Proposition~\ref{prop:mixture}(b) held in all 4512 units, and the half bound of Proposition~\ref{prop:overbet}(d) in all 3114 units to which it applies.

\paragraph{How far off the failing factors were.} In the twelve cases at full execution in which Kish or contrast tempering rejected in fewer than 24 of the 32 units of a game or task while W(3/4) rejected in at least 24, all on IRIS's decoded pixels or on TD-MPC2, the estimated factor exceeded the unit's best factor by a median, over the units of the case, of at least 6.26 times. On DIAMOND the best factor was between 0.0016 and 0.0046 across the three games and the break-even factor between 0.0046 and 0.031, while the Kish and contrast factors were 11 to 82 times the best factor in the median unit (Table~\ref{tab:overbet} gives the medians of each factor). In these comparisons the fitted factors could lie above break-even. Using one tenth of a profitable comparator's temperature would retain at least one tenth of its log e-value.

\paragraph{Online rules.} Choosing the factor within the unit from earlier origins removes the need for validation data. Under the null these rules made at most 1 rejection in the 64 units of any game or task. With actions executed they were weaker than the mixtures (Table~\ref{tab:overbet}): the follow-the-leader rules reached the power bar or rejected in fewer units than W(3/4) on every score, and halving the leader's factor (O57h) did better on three of the four scores and no worse on the fourth, as the asymmetry suggests. In the robot-data study, where two of them were registered as secondary statistics, they rejected in 149 and 153 of 159 units against 155 for W(3/4) (\S\ref{sec:p5:results}). A follow-the-leader rule stakes everything on one factor at each origin; the mixture, which is the wealth-weighted rule of Proposition~\ref{prop:overbet}(e), does not, and the identity was confirmed numerically (largest difference $5 \times 10^{-12}$).

\paragraph{What this explains.} The same asymmetry accounts for the three kinds of failure in this paper. The pooled score uses the factor 1, far above break-even when thousands of sites share a model error (\S\ref{sec:pooling}, \S\ref{sec:published}). Kish and contrast factors estimate a design effect from residuals and came out too large by one to two orders of magnitude on published models (\S\ref{sec:published}, \S\ref{sec:p4}). A factor tuned on validation units of a different size transfers badly in the same direction (\S\ref{sec:erkxm}, \S\ref{sec:erkrho}). It also explains the shape of the result on robot video, where the pooled score's factor of 1 was above the best factor in 150 of 159 units, but by less than a factor of two in the median unit, so the loss was small in most units and large in a few (\S\ref{sec:p5:results}). For practice it gives a rule that needs no model of the dependence between sites: when unsure, err toward the smaller factor, or do not choose and mix.

One prediction of the study's protocol was wrong as stated. The protocol claimed a half bound for the mixture without the condition that some factor in the range be profitable; the bound failed in 82 units, all of them units in which no grid factor was profitable. A dated amendment, written after the result, replaced the claim by the two bounds reported above; we record the prediction as contradicted.
\input{appendix/tables/V4_overbet.tex}

%% file: appendix/tables/V4_overbet.tex
\begin{table}[htbp]\centering\small
\caption{Over-betting study (exploratory). Top: DIAMOND, per game, the unit's best factor, break-even factor, Kish factor and contrast factor (medians over ACTIVE units). Bottom: units rejected (DIAMOND, of 96; IRIS pixels, of 96) or games and tasks at or above the power bar (IRIS tokens, of 26; TD-MPC2, of 6), with actions executed.}\label{tab:overbet}
\setlength{\tabcolsep}{4pt}\begin{tabular}{@{}lrrrr@{}}
\toprule
Game & Best factor & Break-even & Kish & Contrast \\
\midrule
Boxing & 0.0016424 & 0.0046453 & 0.01812 & 0.024074 \\
Pong & 0.0046453 & 0.026278 & 0.048787 & 0.077295 \\
Breakout & 0.0039062 & 0.03125 & 0.32019 & 0.05298 \\
\bottomrule\end{tabular}\par\medskip\begin{tabular}{@{}lrrrr@{}}\toprule
Rule & DIAMOND & IRIS tokens & IRIS pixels & TD-MPC2\\\midrule
O15 & 87 & 22 & 76 & 4 \\
O57 & 88 & 22 & 79 & 5 \\
O57h & 90 & 23 & 87 & 5 \\
W(3/4) & 96 & 24 & 96 & 6 \\
\bottomrule\end{tabular}
\end{table}

%% file: sections/13_partial_execution.tex
\section{Partial execution and bounded stakes}\label{sec:v5-stakes}

\setcounter{proposition}{13}

Sections 9–11 tested released scorers at one step, including partial execution in Section 10 and a working-law analysis of recorded robot commands in Section 11. We now study bounded stakes under partial execution and at longer horizons. This section adds a bounded stake to the statistics of \S\ref{sec:method}, derives a finite-bag oracle weight, and reports a fourth registered study and two exploratory analyses of its stored scores. The studies of this section and of \S\ref{sec:v5-evidence} and \S\ref{sec:v5-trajectories} share one design: ACTIVE units, in which the labels are the recorded actions, and NULL units built so that the null holds exactly. The e-value statistics reject at $E \ge 20$ and the sequential rank-sum test PRseq at $p \le 0.05$.

\subsection{Temperatures fitted by log-loss}\label{sec:v5-temperature}

Test bags here contain the observed candidate and $B=127$ decoys. In the fourth study and E1's primary comparison the parameters are fitted on independent validation contrast bags with 31 decoys. For T and TS the inverse temperature $\lambda$ ranges from $2^{-14}$ to 4. This is a wider fitting range than the original tempering methods in \S{}3.4. Write $e_t(\lambda)$ for the Monte Carlo factor \eqref{eq:e-mc} of the tempered score $\lambda S_t$ at origin $t$. \emph{T} is ordinary temperature scaling \citep{guo2017calibration}: it maximises the validation log-likelihood of the recorded action among the candidates, $\sum_t \log e_t(\lambda)$ up to a constant. V and W(3/4) are as in \S\ref{sec:method:robust}. Validation-dependent e-value parameters were fixed before test scoring. In the fourth study, NULL units used the independent full-execution configuration for the same task and horizon. The e-value statistics have the validity stated in Proposition~\ref{prop:P1}. Section 13.6 gives the conditions for the PRseq p-value.

\subsection{Bounded stakes}\label{sec:v5-bounded}

For a one-step product, or a single residue-class product at a longer horizon, choose a stake $f\in[0,1]$.
\begin{equation}\label{eq:stake} E(\lambda, f) = \prod_t \{\, 1 - f + f\, e_t(\lambda) \,\}.\end{equation}
The factor $1-f+fe_t(\lambda)$ is the fractional-Kelly factor of \S{}3.6, with $f$ in place of the mixing weight $\kappa$. For fixed admissible $\lambda$ and $f$, each factor has null conditional mean one under Proposition 1. At longer horizons multiply within each residue class and average the class products as in Appendix M.4. Read from the candidates, $1 - f + f e_t$ is $B+1$ times the probability of the recorded action under a softmax over the candidates' scores mixed with the uniform distribution over the candidates.

\begin{proposition}[Bounded stake]\label{prop:stake}
For finite scores, pathwise, with $T$ origins and $x_t = e_t(\lambda) - 1$: (a) $\log E(\lambda, f) \ge T \log(1-f)$ for $0 < f < 1$; (b) $\log E(\lambda, f) \ge f \sum_t x_t - f^2 \sum_t x_t^2$ for $0\le f\le1/2$.
\end{proposition}

Each step therefore lowers the log wealth by at most $-\log(1-f)$. For a fixed sequence whose mean e-value exceeds one, every sufficiently small positive stake has positive log growth; a given finite grid of stakes need not contain one. The unbounded product grows only when the mean log e-value is positive. If a fraction $\pi$ of the origins has e-value $e_+ > 1$ and the rest has $\epsilon > 0$, the two conditions are $\pi e_+ + (1-\pi)\epsilon > 1$ and $\pi \log e_+ + (1-\pi) \log \epsilon > 0$; for every fixed $0 < \pi < 1$ with $\pi e_+ > 1$, a small enough $\epsilon$ satisfies the first and fails the second. The factor $1 - f + f e$ is the usual bet of a fraction of one's wealth \citep{kelly1956, maclean1992growth, waudbysmith2024betting}.

We use the stakes $f_k = 2^{-k}$, $k = 0, \dots, 6$, with the 15 temperatures $\lambda_g=2^{-g}$, $g=0,\ldots,14$. \emph{SW} puts weight 3/4 on the pair with the largest median validation log e-value and spreads 1/4 evenly over the other 104 pairs. \emph{TS} fits $(\lambda, f)$ jointly, over $\lambda \in [2^{-14}, 4]$ and $f \in [0, 1]$, by maximising the validation log-likelihood $\sum_t \log\{1 - f + f e_t(\lambda)\}$: it is temperature scaling with uniform smoothing, the objective of T with one more parameter.

\subsection{The one-step finite-bag oracle}\label{sec:v5-estimates}

Fix one origin and condition on its history and admissibly frozen information. In a randomly ordered, observed-inclusive bag of $K=B+1$ slots, the recorded action and the $B$ decoys are independent draws from $Q_t$ before the outcome is observed. The recorded slot $J$ is uniform before observing the outcome. An independent Bernoulli($p$) coin selects execution of the recorded action. Otherwise an independent draw from $Q_t$ is executed, without using the bag. The fresh draw may equal a bag value. Assume the outcome laws have densities with respect to a common measure, and write $L_j$ for the density of $y$ under execution of slot $j$, $\bar L=K^{-1}\sum_j L_j$, and $m=\int L(y\mid h,a)Q_t(da)$. Assume these densities are finite on the relevant support.

\begin{proposition}[Stake and execution rate]\label{prop:v5-stake-oracle}
Given the bag and $y$, the recorded-slot probabilities are $\pi_j=[(1-p)m+pL_j]/[K((1-p)m+p\bar L)]$ wherever the denominator is positive. Against the conditional uniform-slot null, the log-optimal factor among nonnegative factors with slot average at most one is $K\pi_J$. For $\bar L>0$ it is $1-w+w e_J^\ast$, where $e_j^\ast=L_j/\bar L$ and $w=p\bar L/[(1-p)m+p\bar L]$. When $\bar L=0$ on the alternative support the factor is one.
\end{proposition}

The proof is Bayes' rule and Gibbs' inequality (Appendix~\ref{app:proofs:v5-execution}). The calculation is one-step. At longer horizons the design mixes different execution patterns inside the block. The weight $w$ is the posterior probability, given the bag and the observation, that the execution coin selected the recorded slot. It equals $p$ when $\bar L = m$. For almost every observation with $0<m<\infty$, the independent decoy average converges to $m$ as $B$ grows, and the one recorded slot has vanishing weight, so $w\to p$. For $0<p<1$ and $m>0$, the posterior odds $w/(1-w)$ equal the prior odds $p/(1-p)$ multiplied by $\bar L/m$. The proposition concerns the factor computed from the system's true densities. A model-supplied score with a fitted temperature does not give $e^\ast$, and nothing here identifies the stake returned by a fit with $p$; \S\ref{sec:v5-fitted} reports what the fits returned. This is the algebra of label noise and of uniform smoothing, not a new result.

\subsection{A fourth registered study}\label{sec:v5-registered}

The fourth registered study returned to the six TD-MPC2 tasks of \S\ref{sec:p4} at horizons of 1 to 8 steps and under partial execution: the simulator executed the recorded label with probability $p$ and a fresh random action otherwise, and the model always received the label. Its criteria and predictions were written after an exploratory run of the same design on separate seeds. The registration evaluated the same six tasks and checkpoints on new seeds, with a revised aggregate full-execution criterion and a gain margin set after the exploratory run (Appendix M). Over 576 partial-execution test units at one step, SW rejected in 376, $R_{1/2}$ in 286, T in 248 and W(3/4) in 243; each difference exceeded the registered margin of 58 units, and SW rejected in 93 units in which $R_{1/2}$ did not, against 3 the other way (Figure~\ref{fig:v5-horizon1}). The fourth S-GAIN comparator, R, rejected in 228 units. All four paired sign tests passed. At $p = 0.05$, 56 of SW's 63 rejections came from two tasks with long units and strong signals. Under full execution SW rejected in 643 of 768 unit--horizon pairs and W(3/4) in 633, with a largest per-set shortfall of one unit.

The protocol and code were registered with OSF (8mg9t, 12 November 2026 UTC; Appendix~\ref{app:p6}).

\input{figures/v5/partial_execution.tex}

\subsection{Matched comparisons (E1, exploratory)}\label{sec:v5-comparators}

Relative to T, the registered SW rule allows a stake, searches a discrete temperature grid, selects by median validation log e-value, and forms a weighted mixture over the candidate pairs. On the same units, TS, which is T's objective with a stake added, rejected in 381 of the 576. For SW minus TS the mean difference was \ensuremath{-}0.87 percentage points (task-reweighting interval \ensuremath{-}1.9 to 0.2 points). Holding the grid, objective and tie rules fixed and adding only the stake raised a temperature rule from 242 to 378 and a rank rule from 284 to 348; a staked rank mixture rejected in 351 against 286 for $R_{1/2}$. An online stake that uses only earlier origins rejected in 311. A second set of 576 units, from the exploratory run, gave TS 378, SW 364 and T 255. In the registered TD-MPC2 units at full execution and one step, TS and T each rejected in 179 of 192 and SW in 183. In the exploratory IRIS-token comparison, SW, TS and T rejected in 178, 177 and 174 of 288 partial-execution units and in 780, 783 and 783 of 832 full-execution units, respectively. The matched one-step TD-MPC2 comparisons show an observed gain from allowing a stake in those cells. The counts vary with the scorer, execution condition and horizon.

\subsection{Longer horizons}\label{sec:v5-horizons}

Nonconstant rank e-factors contribute negative expected log growth under the null. PRseq instead uses a fixed schedule of possible origins. Within each nonoverlapping residue class it sums the recorded action's ranks, breaking ties uniformly and inserting independent uniform ranks at absent scheduled origins. It computes the exact independent-uniform convolution tail for each class and combines the tails by Bonferroni, assigning p=1 to an empty unit. This gives a super-uniform p-value under the per-origin null with conditionally exchangeable candidate bags, history-measurable or null-invariant eligibility, and independent auxiliary draws. At one step it rejected in 287 of the 576 units; at 2, 4 and 8 steps it rejected in 241, 201 and 148, against 271, 182 and 94 for TS, 266, 169 and 107 for SW and 191, 142 and 74 for T (Table~\ref{tab:v5-horizons}). Under the null, e-value statistics rejected in at most 4 of 384 units at any horizon and PRseq in 17 to 24, close to its level of 19.

\input{appendix/tables/V5_horizons.tex}

\subsection{What the stake estimated (E2, exploratory)}\label{sec:v5-fitted}

Proposition~\ref{prop:v5-stake-oracle} motivates checking whether a stake fitted with a well-calibrated score is close to $p$. In the one-step exploratory analysis, a cell was classified as strong if its full-execution TS\_C log e-values summed over test units, divided by their total eligible-origin count, reached 0.5 nats per origin. Among the strong partial-execution cells the fitted stake was within a factor of two of $p$ in 29/31, 14/15 and 13/15 cells in the three data sets, with median $f/p$ of 0.96, 0.97 and 1.09. In the two data sets with three partial-execution rates, 11 of 12 task series were non-increasing as $p$ fell. T's factor was lower than its full-execution value in all 61 strong cells. Five of eight weaker cells were within the band. The registered-score data set includes reacher-easy, where an earlier grid-stake fit suggested the question. These are descriptive comparisons of fitted values with the known execution probability.

%% file: figures/v5/partial_execution.tex
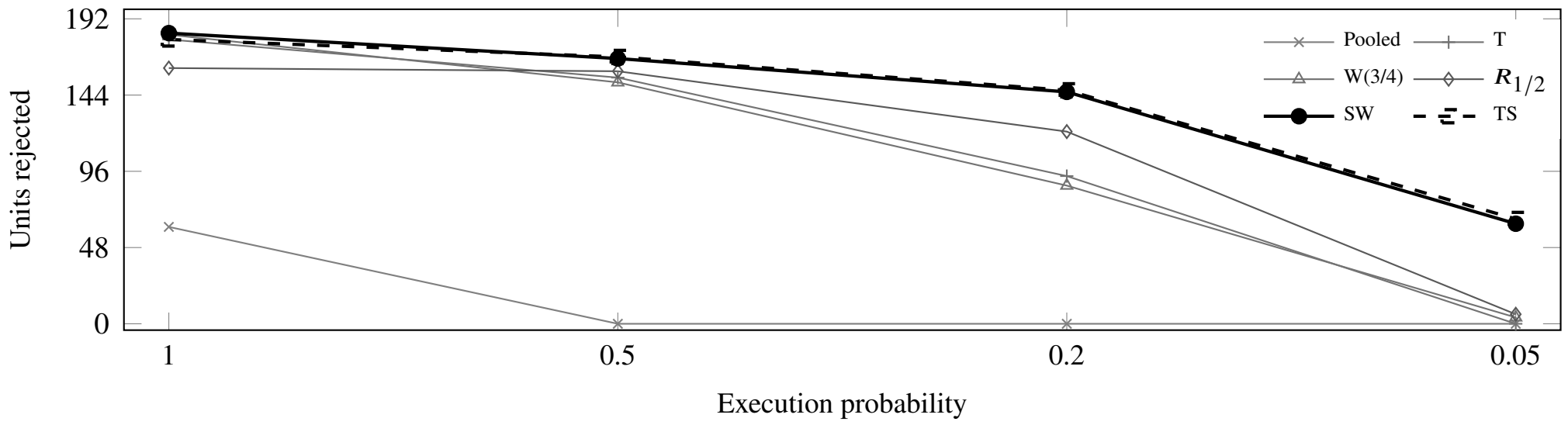
\begin{figure}[!htbp]\centering
\begin{tikzpicture}
\begin{axis}[width=.99\columnwidth,height=1.70in,scale only axis=false,
        xmin=-.1,xmax=3.1,ymin=-4,ymax=198,xtick={0,1,2,3},xticklabels={1,0.5,0.2,0.05},
        ytick={0,48,96,144,192},xlabel={Execution probability},ylabel={Units rejected},
        label style={font=\scriptsize},tick label style={font=\scriptsize},
        legend style={font=\tiny,at={(1,.97)},anchor=north east,legend columns=2,
        draw=none,fill=white,inner sep=1pt},legend cell align=left]
\addplot[gray,mark=x,mark size=1.6pt] coordinates {(0,61) (1,0) (2,0) (3,0)};
\addlegendentry{Pooled}
\addplot[black!55,mark=+,mark size=1.6pt] coordinates {(0,179) (1,155) (2,93) (3,0)};
\addlegendentry{T}
\addplot[black!55,mark=triangle,mark size=1.6pt] coordinates {(0,182) (1,152) (2,87) (3,4)};
\addlegendentry{W(3/4)}
\addplot[black!65,mark=diamond,mark size=1.6pt] coordinates {(0,161) (1,159) (2,121) (3,6)};
\addlegendentry{\ensuremath{R_{1/2}}}
\addplot[black,thick,mark=*,mark size=1.6pt] coordinates {(0,183) (1,167) (2,146) (3,63)};
\addlegendentry{SW}
\addplot[black,dashed,thick,mark=square,mark size=1.6pt] coordinates {(0,179) (1,168) (2,147) (3,66)};
\addlegendentry{TS}
\end{axis}
\end{tikzpicture}
\caption{Partial execution, one step, six TD-MPC2 tasks. Test units rejected (of 192) as the recorded action is executed less often. Solid: registered statistics. Dashed: temperature and stake fitted jointly by validation log-loss, computed afterwards on the same units (exploratory).}\label{fig:v5-horizon1}
\end{figure}

%% file: appendix/tables/V5_horizons.tex
\begin{table}[!htbp]
\caption{Partial execution at horizons of 1 to 8 steps: test units rejected (of 576: six tasks at execution probabilities 0.5, 0.2 and 0.05). Upper block: registered run. Lower block: computed afterwards on the same units (exploratory). PRseq uses padded rank sums within residue classes and Bonferroni across classes under the conditions of \S{}13.6.}\label{tab:v5-horizons}
\centering\small
\setlength{\tabcolsep}{2pt}
\begin{tabular*}{\columnwidth}{@{\extracolsep{\fill}}lrrrr@{}}
\toprule
Statistic & $h=1$ & $h=2$ & $h=4$ & $h=8$\\\midrule
Pooled & 0 & 0 & 4 & 15\\
T & 248 & 191 & 142 & 74\\
W(3/4) & 243 & 186 & 147 & 92\\
\ensuremath{R_{1/2}} & 286 & 232 & 144 & 97\\
SW & 376 & 266 & 169 & 107\\
\midrule
TS & 381 & 271 & 182 & 94\\
PRseq & 287 & 241 & 201 & 148\\
\bottomrule\end{tabular*}
\end{table}

%% file: sections/14_reading_evidence.tex
\section{Reading the evidence}\label{sec:v5-evidence}

\subsection{Evidence rate and references}\label{sec:v5-rate}

\textbf{Evidence rate.} For a scorer and a set of test units, let $G$ be the mean over planned test units of each unit's TS log e-value divided by its eligible-origin count. A realized zero-origin unit contributes zero and stays in the denominator. A structurally absent stratum is undefined. $G \le \log(B+1) \approx 4.85$. At one origin the TS factor is $B+1$ times the probability that a softmax mixed with the uniform distribution gives to the recorded label in the bag, so its logarithm is the improvement in log score over uniform guessing among the candidates. Under partial execution the recorded label can differ from the action actually applied, so $G$ concerns the recorded label. Proposition~\ref{prop:information} bounds the expected log factor of an origin by the information between the action and the next observation given the realised history and the frozen score information, provided the law of the recorded action given the history and that information is the design law; without that condition the bound contains an additional divergence of the assignment law. $G$ is a sample mean of these log factors, not a confidence bound on that information.

\textbf{What it compares.} Different scorers applied to the same units and the same draws can be compared on $G$, and on paired unit log e-values. A scorer need not be a world model: an inverse-dynamics model, trained to predict the action from the current and next observation \citep{pathak2017curiosity, baker2022vpt}, gives a statistic of the same form whose candidates are scored by its own likelihood. We use such models as references: for a world model with evidence rate $G$ and a reference's rate $G_{\mathrm{ref}}$ on the same units, $\rho = G / G_{\mathrm{ref}}$ (undefined when $G_{\mathrm{ref}} < 0.05$). A second kind of reference, used in E5 to E7, is a \emph{contrastive critic}: a network $s(o_t, o_{t+1} - o_t, a)$ trained by cross-entropy to pick the executed action among 127 draws from the design law of that step, which is the test's own task. No reference is an upper bound; $\rho$ is not a fraction of the available information, it depends on the reference's features, fit and calibration, and it can exceed one. It is different from the ratio of a scorer's rate at two levels of a trajectory intervention, used in \S\ref{sec:v5-trajectories}.

\textbf{What it does not say.} $G$ concerns a scorer's ability to pick out the recorded label at the observed histories under the design law. A high value can come from one easily decoded feature, so it does not measure the size of the system's causal response or a forward model's accuracy. It is not a measure of control quality, which also depends on the policy, the value and reward estimates and the planner \citep{lambert2020objective}, and it changes with the trajectories on which it is computed (\S\ref{sec:v5-trajectories}).

\begin{proposition}[Information bound]\label{prop:information}
Fix an origin $t$ and condition on a realization $h$ of the history $\mathcal F_t$ and on the frozen score information $\mathcal D$. Write $q=Q_t(h)$, $P$ for the conditional joint law of $(A_t,Y_{t+1})$, and $P_Y$ for the corresponding observation marginal. Take $Q_t(h)$ as the reference law. Given $(h,\mathcal D)$, draw the decoys independently from this law and independently of the recorded action and outcome. Apply the same finite score function to each candidate without using the recorded slot. The bound is $D_{\rm KL}(P_{A_t,Y_{t+1}\mid h,\mathcal D}\|Q_t(h)\otimes P_{Y_{t+1}\mid h,\mathcal D})$. This equals mutual information at the fixed conditioning values plus $D_{\rm KL}(P_{A_t\mid h,\mathcal D}\|Q_t(h))$.
\[\mathcal I_t(h)=D_{\rm KL}\big(P \,\big\|\, q\otimes P_Y\big)\in[0,\infty].\]
With this notation,
\[\mathbb E[\log e_t(\lambda) \mid \mathcal F_t = h, \mathcal D] \;\le\; \mathcal I_t(h).\]
\end{proposition}

The bound applies to the TS factor $1-f+f\,e_t(\lambda)$, which is also $B+1$ times a probability vector's entry at the recorded slot. A finite bag leaves the non-negative bag-marginal divergence of the proof, and a fitted scorer also pays its error relative to the best slot posterior. Proposition~\ref{prop:P2}(c) bounds the exact factor, whereas Proposition~\ref{prop:information} bounds the Monte Carlo factor formed from a finite bag.

See Appendix~\ref{app:proofs:v5-execution} for the proof and remark.

\subsection{Status of the analyses}\label{sec:v5-status}

E1–E8 were exploratory and unregistered. Their dated protocols and forecasts were finalized before the specified new analyses. Separate automated reviews assessed the protocols before execution. E1 reanalyzed stored scores and passed its validation checks after a documented tolerance correction. E2 independently checked tabulations of stored fits and unit statistics. E3–E7 had separately written recomputations using stored numerical results. E1 and E2 both use outputs from the fourth study and additional datasets. Five analyses generate new units with released TD-MPC2 models: weakening of the model's action input, with an inverse-dynamics reference, on six tasks (E3); five model sizes on 30 tasks under two action laws (E4); stronger references, and trajectories that mix the two laws (E5); trajectories whose share of uniform steps is varied in six levels (E6); and a repetition of E6 on fresh units with every fitted quantity frozen, with small forward models that we trained (E7). In these simulated positive-control settings the systems respond to actions. At partial execution the recorded label is applied with the stated probability; otherwise the simulator applies an independent action. Rejection counts describe detection with the specified score and evidence rates describe discrimination of the recorded label. E8 reports descriptive pointwise trajectory-bootstrap intervals for evidence rates and contrasts at P-selected steps in E6 and E7, including midpoint changes and retention ratios. These intervals quantify trajectory sampling conditional on the evaluated design, fitted models, configurations and tasks.
E8 reanalysed the stored E6 and E7 results with trajectory-bootstrap intervals. Its protocol was reviewed in the same way before execution, through four rounds, and a separately written verifier recomputed every reported value; one documented correction changed only how the verifier read missing-value codes.

\subsection{Does the evidence follow the model's use of the action? (E3)}\label{sec:v5-weakening}

In every ACTIVE arm the null is false, so rejections cannot rank models. E3 asks whether the evidence rate responds when a model's use of the action is degraded on purpose. We took the released single-task TD-MPC2 models of six tasks and multiplied the action input of the dynamics by a gain $\alpha$ before every prediction, for every candidate alike, on 32 fresh test units per task under uniformly random actions, and refitted TS for each variant, so the variants are compared as recalibrated scorers. Summed over the six tasks, the evidence rate was 15.9, 10.3, 8.7 and 7.5 nats per origin at $\alpha = 1$, 1/2, 1/4 and 1/10, and zero at $\alpha = 0$, where every candidate is scored alike by construction (Figure~\ref{fig:v5-models}a). Rejection counts hardly moved (184 to 189 of 192) until $\alpha = 0$; under partial execution at $p = 0.2$ they fell from 139 to 104. The decline was not uniform: in cartpole-balance the evidence rose as the gain shrank, from 0.016 to 0.072, and in one Myo task it was not monotone.

We also measured each variant's mean squared next-latent prediction error under the recorded action. With the action input removed, the mean squared error of the next-latent prediction under the recorded action changed by a factor of 0.97 in myo-pose and fell to 0.58 of its intact value in cartpole-balance; in the other four tasks it rose, by factors between 1.4 and 43. Prediction error can thus fail to register that a model has stopped using its action. Masking a random fraction of the action input's coordinates had a similar effect on the evidence rate (13.9, 8.5 and 5.7 at 3/4, 1/2 and 1/4 of the coordinates kept). The sign-flipped scorers had fitted TS evidence rates of zero at full execution in the six tasks.

\textbf{Inverse dynamics.} IDM-L had a higher full-execution evidence rate than the intact single-task scorer in each of the six tasks. IDM-L was trained for 20,000 updates on 230 units per task to predict the action from $(o_t, o_{t+1} - o_t)$. Summed over the six tasks, the rates were 27.1 against 15.9 nats per origin, with 181 against 139 rejections at $p = 0.2$. In cartpole-balance at $p = 0.2$ it rejected in 32 of 32 units where the world model rejected in none. Part of the gap has a simple source: in the three Myo tasks the post-action muscle-activation state gives the inverse-dynamics scorer an explicit actuator-state trace, and IDM-L was within displayed precision of the cap at full execution ($\log 128$). The three DMControl tasks have no such trace, and in walker-walk IDM still had 4.15 nats per origin against 1.70. A smaller inverse-dynamics variant was weaker than the matched world model under partial execution (Appendix~\ref{app:J}), so these statements concern this particular reference. A test of the system does not need a world model, and a world model's evidence rate is better read against a strong reference than on its own.

\input{figures/v5/model_evidence.tex}

\subsection{Thirty tasks, five model sizes, two action laws (E4, E5)}\label{sec:v5-sizes}

E4 extends this to the 30 tasks of the TD-MPC2 multitask suite \citep{hansen2024tdmpc2, tassa2018dmcontrol}. Each task was scored by the released multitask models of 1M, 5M, 19M, 48M and 317M parameters, by a released single-task model of 5M parameters and by IDM, on the same 32 test units. Each model's planner was also run for five episodes to measure its return $R$. Actions were drawn under two laws: uniform on the action box (U), and centred on the single-task model's policy prior with Gaussian noise of scale 0.3, clipped (P). Three tasks of E3 suggested the questions; the appendix also gives the 27 other tasks alone (Appendix~\ref{app:K}).

\textbf{Size.} The 317M model had a higher evidence rate than the 1M model in all 30 tasks under both laws; the full ordering over the five sizes was non-decreasing in 12 of the 30 tasks under U and 19 under P. Across the five sizes return and evidence rate were positively rank-correlated in 27 of 30 tasks under U and 28 under P (Kendall's $\tau_b$; median 0.6 and 0.7). The median ratio $\rho$ under U rose from 0.45 at 1M to 0.84 at 317M (Table~\ref{tab:v5-size}). These are associations within one model family; they do not show that the evidence rate measures control quality or that size improves dynamics.

\textbf{Return without evidence.} We call a cell high-return, low-evidence when a model's mean return is at least 90\% of the best of the six models in that task, that best is at least 100, and $\rho \le 1/4$. Under U, 7 of the 30 tasks had such a cell: acrobot-swingup, pendulum-swingup, pendulum-spin and the four cart-pole tasks. All seven tasks still met the criterion in each sensitivity analysis that omitted the same one of the five reset indices from all six models' returns. The single-task model accounted for six of them; in cartpole-balance it reached a return of 999 of a possible 1000 with an evidence rate of 0.013 nats per origin, 0.3\% of the reference's 3.88. With thresholds of 80\% and 1/2 the single-task model had 16 such cells, and with 95\% and 1/10 it had 3. None of the walker, cheetah, hopper, reacher, finger, fish or cup tasks had one. Return is a five-episode measurement of the whole planning agent.

\textbf{Policy-centred actions.} Under P only 2 tasks had a high-return, low-evidence cell, and in all six single-task cases of U the evidence rate rose; in cartpole-balance from 0.013 to 3.45 nats per origin. The U and P evaluations change the conditional action law, the trajectory histories and each scorer's validation-fitted configuration. The inverse-dynamics reference was also trained separately under each law. The single-task model also both scored P and supplied its centre.

\textbf{Single-task against multitask.} In paired sign tests on the 32 test units (300 comparisons, Holm-adjusted, level 0.01), the single-task model had less evidence than the 317M model in 23 tasks and more in 4 under U, although its return was at least as high in 23 of the 30 tasks; under P it had more in 17 and less in 8.

\textbf{A stronger reference (E5).} Under U, IDM had more evidence than every world model in all 30 tasks; under P it did not in six, the five walker tasks and reacher-three-easy (seen after the fact). We specified four additional reference designs and evaluated five variants (R1, R2, R3, R3c and R4) alongside the original IDM (R0) on fresh P units. The critic R4 had the highest reference rate in the five walker tasks, and the quadratic R3c score had the highest rate in reacher-three-easy (4.68 versus 3.60 for 317M). R0 itself exceeded 317M on the fresh reacher-three-easy units. These are descriptive comparisons of selected reference procedures on fresh units. On the fresh P units in each of the five walker tasks, the 317M rate was positive and R4's rate exceeded R0's. The 317M rate relative to R0 was therefore larger than its rate relative to R4. Appendix R identifies all six references in l-001.tsv and gives their fitting budgets.

\textbf{Mixing the two laws within a trajectory (E5).} In ten tasks an independent fair coin chose, at every step, whether the action was drawn from U or from P, so that at each step the choice between the two laws is independent of the state reached, although the P kernel itself depends on the state. With the critic as reference, the single-task model's evidence rate at U steps and at P steps was 0.67 and 0.24 nats per origin in acrobot-swingup, 0.94 and 0.64 in pendulum-swingup, 0.51 and 0.41 in cartpole-balance and 0.51 and 0.21 in cartpole-swingup: at most a quarter of the critic's rate under either law, and no higher at P steps than at U steps. The same held with one configuration shared by both kinds of step. These numbers describe the mixed trajectories; earlier coins affect later states, the realised states at the two kinds of step can differ, and each unit's rate is averaged over a random number of steps. In cartpole-balance the same model, tested with the same law P, had 3.45 nats per origin on trajectories driven by P throughout (E4) and 0.41 on the mixed trajectories. The all-P and mixed runs differ in trajectory populations and fitted scorer configurations, so the difference in $G$ does not isolate either change. Section 15 varies the mixture probability with the current P kernel and scorer configuration held fixed.

\textbf{Null calibration.} Under exact nulls in E4, over 56 panels of 1920 null units (seven scorers, two laws, four e-value statistics), the e-value statistics rejected once. In E5, over the 76 primary panels (46,336 null unit tests), they rejected once. PRseq rejected in 91 to 119 units per panel of 1920, around its level of 96; the smallest binomial tail probability over its 14 panels was 0.011. These counts are repeated tests on shared units and serve as an implementation check.

\input{appendix/tables/V5_size.tex}

%% file: figures/v5/model_evidence.tex
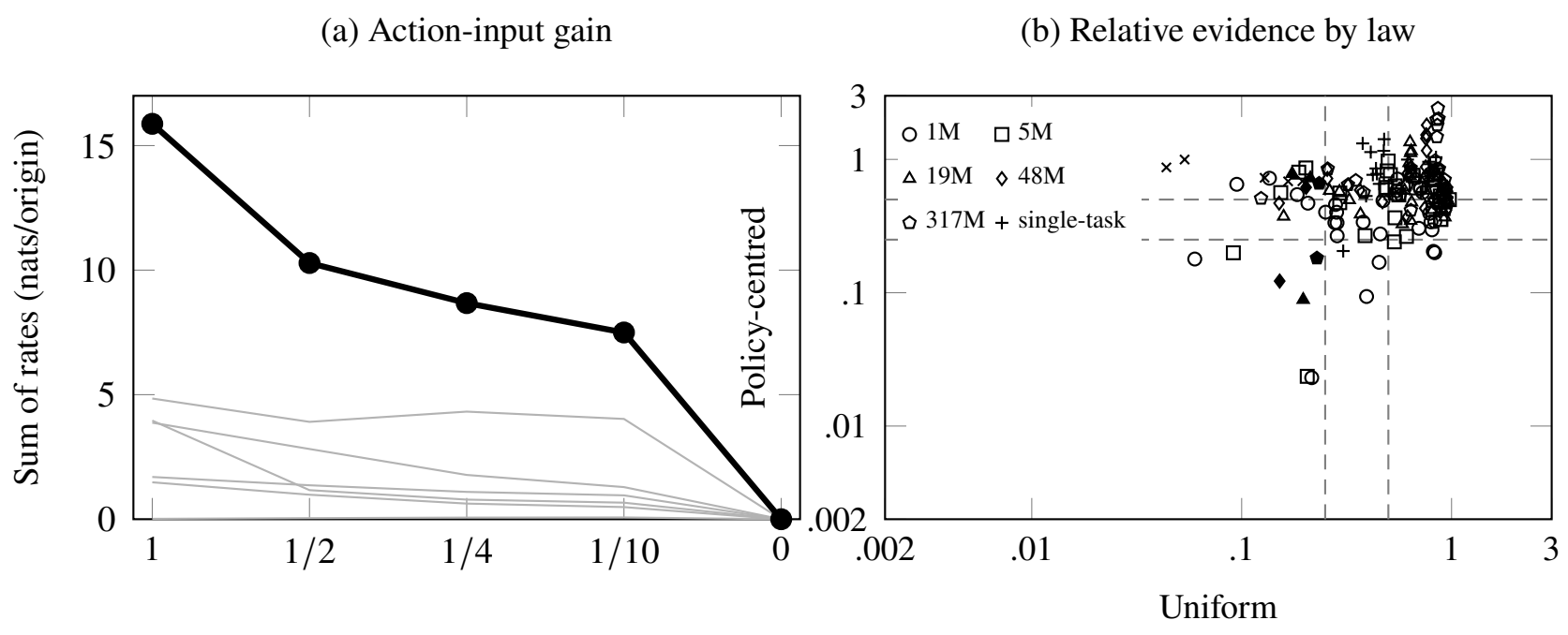
\begin{figure}[!htbp]\centering
\begin{tikzpicture}
\begin{groupplot}[group style={group size=2 by 1,horizontal sep=.6cm},
         width=.45\textwidth,height=1.80in,label style={font=\scriptsize},
         tick label style={font=\scriptsize},title style={font=\scriptsize,align=center}]
\nextgroupplot[title={(a) Action-input gain},xmin=-.12,xmax=4.12,ymin=0,ymax=17,
  xtick={0,1,2,3,4},xticklabels={1,$1/2$,$1/4$,$1/10$,0},ylabel={Sum of rates (nats/origin)}]
\addplot[gray!60,thin,mark=none] coordinates {(0,0.015564906443232473) (1,0.04388898730459757) (2,0.06531830610011412) (3,0.07161822923291523) (4,0.0)};
\addplot[gray!60,thin,mark=none] coordinates {(0,3.8751699968889106) (1,2.819306312463478) (2,1.7777667335619847) (3,1.2896676460375396) (4,0.0)};
\addplot[gray!60,thin,mark=none] coordinates {(0,1.4824225425015805) (1,0.9836452778674517) (2,0.627837943964102) (3,0.486963600884897) (4,0.0)};
\addplot[gray!60,thin,mark=none] coordinates {(0,4.845405034410069) (1,3.9090326535551947) (2,4.320014484117563) (3,4.02629563835441) (4,0.0)};
\addplot[gray!60,thin,mark=none] coordinates {(0,3.9596117686975623) (1,1.1670846388247185) (2,0.7877184535828989) (3,0.6638626253340189) (4,0.0)};
\addplot[gray!60,thin,mark=none] coordinates {(0,1.6960407247313039) (1,1.3631095422481871) (2,1.0971836465999085) (3,0.9581116713098079) (4,0.0)};
\addplot[black,very thick,mark=*,mark size=1.6pt] coordinates {(0,15.874214973672658773) (1,10.28606741226362757) (2,8.67583956792657122) (3,7.49651941115358863) (4,0.0)};
\nextgroupplot[title={(b) Relative evidence by law},xmode=log,ymode=log,
         xmin=.002,xmax=3,ymin=.002,ymax=3,xtick={.002,.01,.1,1,3},ytick={.002,.01,.1,1,3},
         xticklabels={.002,.01,.1,1,3},yticklabels={.002,.01,.1,1,3},xlabel={Uniform},ylabel={Policy-centred},
         legend style={font=\tiny,at={(.018,.975)},anchor=north west,draw=none,fill=white,inner sep=1pt,legend columns=2},
         legend cell align=left]
\draw[densely dashed,gray,thin] (axis cs:0.25,.002)--(axis cs:0.25,3);
\draw[densely dashed,gray,thin] (axis cs:.002,0.25)--(axis cs:3,0.25);
\draw[densely dashed,gray,thin] (axis cs:0.5,.002)--(axis cs:0.5,3);
\draw[densely dashed,gray,thin] (axis cs:.002,0.5)--(axis cs:3,0.5);
\addplot[only marks,black,mark=o,mark size=1.30pt] coordinates {(0.2851146662772547,0.26715530822254674) (0.280107605641082,0.4589351541781097) (0.2847635334202195,0.33469832689035656) (0.8312436292564667,0.19968874032452297) (0.5485929041493122,0.5826795115877168) (0.5530287574612647,0.7117238681516918) (0.05979236578283012,0.17894205296028481) (0.206836855074377,0.46954329210754236) (0.09478465516307927,0.6505091694442632) (0.13566578534560245,0.7222107551041049) (0.18386138692233797,0.5443272399262427) (0.25118189703533544,0.4020939274147408) (0.5418646023961377,0.5441283993311433) (0.3937221769188285,0.09367041249968938) (0.7111843219678818,0.6011398569856334) (0.7091874620917307,0.6086321264782234) (0.8322406616498774,0.3424115852428516) (0.47146690221464105,0.49056197065506885) (0.4575396619156184,0.27587435170769914) (0.28343080830386663,0.40149077836097424) (0.2780102652718982,0.3353753622755419) (0.7867060810483102,0.3941050554971402) (0.8192290110469905,0.20420104107980336) (0.798285234567375,0.3399971349795862) (0.8070118635042817,0.2962345079560245) (0.45178188709200057,0.16916676660119495) (0.7285129790017941,0.5647449300923191) (0.6990861529714512,0.30417411850639636) (0.37938438025518334,0.33850424260494744) (0.21571558336414937,0.023052302844366452)};
\addlegendentry{1M}
\addplot[only marks,black,mark=square,mark size=1.30pt] coordinates {(0.4857100921562023,0.5914686019050256) (0.4989271749058018,0.9723967864561915) (0.4905533225453813,0.615023568147701) (0.8903903140956516,0.35474840399396107) (0.6772639357444319,0.6108637644698183) (0.6555358637996543,0.7608896143390038) (0.09130032608978775,0.19933899723579468) (0.15356762542238284,0.5638428851046612) (0.20230276296988442,0.8628606605941499) (0.1871478854052743,0.802360897755477) (0.2935069287831068,0.5595383740960321) (0.2932813553953151,0.47351533434293347) (0.5609105210811045,0.541420710764963) (0.609665260306892,0.2643247135237535) (0.8233877393396857,0.7218539199985194) (0.8379908543622844,0.7008519516503219) (0.9764766965204253,0.4991938780149423) (0.5391389786203091,0.6340365988971022) (0.5379406168920168,0.36470871257328186) (0.4935279870108523,0.8200760392249737) (0.5112496566921907,0.7682780214773407) (0.8775498850254949,0.48615482196285326) (0.8863227036817529,0.35521184748638135) (0.8841910561290943,0.531664132114795) (0.8820046696866293,0.47657659440841227) (0.5338078594409498,0.24073869262956787) (0.8029838971500007,0.6184760799295548) (0.8157926185768934,0.41631284766975896) (0.387235192030357,0.2679593835021265) (0.20535108562360427,0.023589425611203155)};
\addlegendentry{5M}
\addplot[only marks,black,mark=triangle,mark size=1.30pt] coordinates {(0.6275474739368297,0.9417568353170166) (0.6343635253139768,1.3489375023288543) (0.6407384620727324,0.8782020748705749) (0.9299484723417529,0.36838404838215144) (0.6357891142453878,0.7486219281683759) (0.6445483631564772,0.8281061994746661) (0.15850919368073094,0.3733953053391323) (0.26226218724468187,0.5830297336931626) (0.2925347076254776,0.5736198381404647) (0.32581650512386673,0.49660047997504453) (0.5973886345400735,0.5842194985595781) (0.5824264764755118,0.3276509333798856) (0.7866716801110912,0.7277009335558186) (0.7342787618894749,0.7195216881310297) (0.9632704407189164,0.48931265925573536) (0.6453980923349878,0.7791931162518123) (0.6453125640046464,0.4767664862474005) (0.6380284762629839,1.1188203903317033) (0.6424675193792433,1.1572580535373884) (0.925639850076642,0.5993270594863661) (0.9157612451528364,0.4054116946425513) (0.8973145567632377,0.6011028282885699) (0.9196955561900522,0.5150092509693963) (0.6389935483713144,0.34715357548585246) (0.8247620506529273,0.7351819104516142) (0.8139944187307935,0.4184574845812571) (0.3683023130382138,0.38796624681550945)};
\addlegendentry{19M}
\addplot[only marks,black,mark=triangle*,mark size=1.30pt,forget plot] coordinates {(0.17409857426727854,0.7810184683743058) (0.21241515085985724,0.7238645342601687) (0.19645883807592426,0.08865211989664715)};
\addplot[only marks,black,mark=diamond,mark size=1.30pt] coordinates {(0.75602471526577,1.5265856555185926) (0.7613707607577127,1.8084030313030235) (0.7613804098018168,1.1641700134771824) (0.9446265230059001,0.4153230930449738) (0.6281341689722123,0.7401798102296168) (0.6173131662208668,0.7874042365075441) (0.15088335603510047,0.4697977295811504) (0.25630207869537036,0.7186343243177044) (0.2557417054785945,0.8424267582225891) (0.32149237803487096,0.6416510796831195) (0.3649610101867036,0.5889349313385589) (0.564598344359044,0.5992685123973506) (0.6169392319838926,0.3708334184068828) (0.808963607277222,0.7869496371187682) (0.7568795004209289,0.752692122824642) (0.955999276058413,0.5112662781109075) (0.7534855538693621,0.8783383540061981) (0.7626443195013007,0.5348265432840835) (0.755166031276702,1.4362210599517022) (0.7632353916220495,1.5138424564080417) (0.9534253012009468,0.6180098035535706) (0.952758433858225,0.48733477618144844) (0.9130635163482774,0.6680555852506654) (0.9568874938296482,0.5767742273916229) (0.744901166397097,0.43010850647048493) (0.8323868185714794,0.8296073506398604) (0.8199542236285146,0.5013012766849797) (0.46863580808173827,0.4813649784514586)};
\addlegendentry{48M}
\addplot[only marks,black,mark=diamond*,mark size=1.30pt,forget plot] coordinates {(0.20281919744641463,0.6170093337174593) (0.15172251988997046,0.12226935629473173)};
\addplot[only marks,black,mark=pentagon,mark size=1.30pt] coordinates {(0.8480738203245918,1.9859980112785094) (0.8610704013241244,2.416364350388059) (0.851860567544617,1.4729404137724875) (0.9375301467636711,0.4201183690468524) (0.6756679888043299,0.7175242397243465) (0.638599781552555,0.8595890479557731) (0.12371262825202155,0.5096810885005114) (0.34978886941032794,0.6944013090070892) (0.2571785289749889,0.847741882062796) (0.32111367848260164,0.6446206010732165) (0.38142593449915374,0.5640557118277219) (0.6122376813762359,0.6391870971661457) (0.6399454231977958,0.41521815729302125) (0.8563137880327696,0.8413786707606221) (0.8054641658417895,0.811941361712934) (0.9449058100720957,0.5056464622298159) (0.8357468033761922,0.9795157771187014) (0.8441212296952573,0.5905226623794204) (0.8556639849285534,1.8058891244644353) (0.8675212353469556,2.017439525367466) (0.9192485020857014,0.630465194304801) (0.9416959001428821,0.4697112768823361) (0.9277454908043192,0.7035700262325906) (0.9424715233863079,0.5988483005569573) (0.8297925872385014,0.48961229689602553) (0.871344493412885,0.8483663503502449) (0.8660573575650279,0.46383596971900304) (0.5707405790316701,0.5414290390699457)};
\addlegendentry{317M}
\addplot[only marks,black,mark=pentagon*,mark size=1.30pt,forget plot] coordinates {(0.2339546950077577,0.6606714134973501) (0.22830267913502014,0.18174499584420176)};
\addplot[only marks,black,mark=+,mark size=1.30pt] coordinates {(0.47744077368331433,1.4179195523461001) (0.3773139084186245,1.3181760361412047) (0.43651867966960134,0.8553691446701065) (0.9161945003107267,0.5394044244805049) (0.778558974787947,0.959488510871644) (0.6169292285415073,0.996575916242579) (0.4408317789135994,0.779431746043254) (0.581408491284984,0.552613421869312) (0.791211283378708,0.9211886374966097) (0.8451355880530641,0.8576900138687085) (0.9173022911657966,0.605959134542854) (0.42184806526983787,0.7663183205229687) (0.39174172182478356,0.5315255861729036) (0.41219501362973354,1.1349689932864144) (0.47524503175983207,1.1590784623159063) (0.7720269288702006,0.6779336332317423) (0.8270198910744185,0.5099758230154667) (0.7042085074924784,0.743647603192716) (0.7133695432497055,0.5702641292811271) (0.30471859520885203,0.20549137967120382) (0.8432746889445865,1.0385176933679041) (0.877485634329919,0.7119319504843757) (0.506464688637992,0.5520796703267941) (0.4525818535417671,0.6551920937371047)};
\addlegendentry{single-task}
\addplot[only marks,black,mark=x,mark size=1.30pt,forget plot] coordinates {(0.19481589080805617,0.6862337361105842) (0.16650801049109015,0.6846014274036561) (0.0033116683095223793,0.8133318507053187) (0.0534866951939092,0.9976432436189842) (0.043789532268988275,0.8706223245608062) (0.12865546280617768,0.7280380322443812)};
\end{groupplot}
\end{tikzpicture}
\caption{What the evidence rate responds to. (a) The heavy curve is the sum of the six task evidence rates; the light curves are the rates for individual tasks (exploratory); at \ensuremath{\alpha} = 0 the evidence is zero by construction. (b) The same model's evidence relative to an inverse-dynamics reference under uniform actions (x) and actions drawn near the single-task model's policy (y), 30 tasks. Filled: models with near-best return and at most a quarter of the reference's evidence under uniform actions. The vertical positions are computed relative to the original Gaussian IDM. On fresh P units in each of the five focal walker tasks, R4 had a higher rate than the original Gaussian IDM (R0).}\label{fig:v5-models}
\end{figure}

%% file: appendix/tables/V5_size.tex
\begin{table}[!htbp]
\caption{Thirty tasks (exploratory). Medians over tasks of the evidence rate G (nats per origin) and of \ensuremath{\rho} = G/\ensuremath{G_{\mathrm{ref}}} under uniform (U) and policy-centred (P) actions; median over tasks of planner return R, the mean of five episodes in each task; number of cells with high return and low evidence (return at least 90\% of the task's best, best at least 100, \ensuremath{\rho} \ensuremath{\le} 1/4) under U and P. IDM is the reference. \ensuremath{\rho} (P) is computed relative to the original Gaussian IDM. On fresh P units in each of the five focal walker tasks, R4 had a higher rate than the original Gaussian IDM (R0).}\label{tab:v5-size}
\centering\small
\setlength{\tabcolsep}{2pt}
\begin{tabular*}{\columnwidth}{@{\extracolsep{\fill}}lrrrrrrr@{}}
\toprule
Model & $G$ U & \ensuremath{\rho} U & $G$ P & \ensuremath{\rho} P & $R$ & U & P\\\midrule
mt30-1M & 1.84 & 0.45 & 1.26 & 0.37 & 94 & 0 & 0\\
mt30-5M & 2.19 & 0.54 & 1.74 & 0.55 & 201 & 0 & 0\\
mt30-19M & 2.63 & 0.64 & 1.88 & 0.59 & 560 & 3 & 1\\
mt30-48M & 3.10 & 0.76 & 2.18 & 0.63 & 786 & 2 & 1\\
mt30-317M & 3.46 & 0.84 & 2.17 & 0.65 & 813 & 2 & 1\\
single-task 5M & 2.07 & 0.48 & 2.45 & 0.75 & 882 & 6 & 1\\
IDM & 4.38 & -- & 3.73 & -- & -- & -- & --\\
\bottomrule\end{tabular*}
\end{table}

%% file: sections/15_trajectory_evidence.tex
\section{Evidence away from the policy's trajectories (E6, E7)}\label{sec:v5-trajectories}

\textbf{Design (E6).} We took the four pole tasks in which E4 and E5 raised the question (acrobot-swingup, pendulum-swingup, cartpole-balance, cartpole-swingup) and two locomotion tasks as controls (walker-walk, cheetah-run); all six were chosen after earlier results. Every trajectory has a level $q \in \{0, 0.1, 0.25, 0.5, 0.75, 1\}$: at each step an independent coin selects the uniform law U with probability $q$ and the policy-centred conditional kernel P otherwise, and the recorded action and its decoys are drawn from the selected law. The e-value is formed separately over the steps at which the coin selected P and over those at which it selected U; each is valid for the reason given for the mixed law of E5. At $q = 0$ a trajectory is driven by P throughout, as in E4. Three things are the same at every level: each scorer's configuration (one per task, fitted on validation units of all six levels together), the reference (one contrastive critic per task, trained on training units of all six levels), and the conditional kernel P of the tested action given the state. Because P is centred on the policy at the state reached, the marginal law of P actions can still differ between levels. Units at different levels are generated independently. At a fixed clock, the population contrast between levels for a P-selected step operates through the distribution of the pre-action history, because the current kernel and the fitted scorer are fixed. The equal-unit rate also weights steps by the realized number of P steps in each unit, and sample rates vary. The origin-weighted and fixed-clock-normalized summaries showed the same broad midpoint pattern in the four pole tasks (Appendix S and the m-series ancillary tables). We compute no test between levels.

\textbf{World models lose their evidence; the critic keeps most of it.} Figure~\ref{fig:v5-occupancy} and Table~\ref{tab:v5-occupancy} give the evidence rate at P steps; percentages are rounded to whole numbers. For the single-task model it was 1.42, 2.16, 2.93 and 2.45 nats per origin at $q = 0$ in the four pole tasks and 0.22, 0.53, 0.28 and 0.17 at $q = 0.5$: 16\%, 25\%, 9\% and 7\% of its value on the policy's own trajectories. For the 317M multitask model the same ratios were 6\%, 26\%, 33\% and 27\%. The critic kept 74\%, 73\%, 77\% and 80\%. Relative to the critic, the single-task model went from between 0.55 and 0.63 of its rate at $q = 0$ to between 0.05 and 0.19 at $q = 0.5$. In the two control tasks nothing of the kind happened: the ratios were 81\% and 115\% for the single-task model and 101\% and 124\% for the 317M model. At $q = 0.75$ the single-task model's rate was negative in the two cart-pole tasks (\ensuremath{-}0.45 and \ensuremath{-}0.33): the shared configuration over-bets there. With a configuration fitted separately at each level those two rates were 0.02 and 0.04, and the ratios at $q = 0.5$ were 15\%, 25\%, 10\% and 7\%, so the fall is not produced by the shared configuration. Some series are not monotone in $q$, and the critic's rate also varies.

\textbf{Forecasts.} Seven forecasts were dated before the run. Six held, among them that the single-task model would keep at most half of its evidence at $q = 0.5$ in at least three pole tasks (it did in four), that the critic would keep at least half (four), and that each control task would retain more than at least three of the four pole tasks (both controls exceeded all four in this sample). One failed: we expected the 317M model to fall to half or less in at most two pole tasks, and it did in all four. The fall is thus not particular to models trained on one task, and its extent for the 317M model was not anticipated.

\textbf{The same on fresh trajectories (E7).} E7 generated 2,592 new units. The frozen repeat used the same TD-MPC2 checkpoints, critic weights and E6 configurations on 1,152 new test units and 1,152 new NULL units. The remaining 288 units were validation units for configuring the forward-model arms. The ratios at $q = 0.5$ were 14\%, 25\%, 6\% and 9\% for the single-task model, 11\%, 24\%, 25\% and 29\% for the 317M model and 71\%, 72\%, 73\% and 81\% for the critic; in the control tasks they lay between 84\% and 137\%. Three forecasts about these ratios, dated after E6, held. This repeats the evaluation on new random draws of the same design. It is not a replication on other tasks, other models or another training run.

\textbf{Uncertainty (E8).} A further exploratory analysis, E8, resampled whole trajectories within each task and level cell of E7, with 10,000 bootstrap replicates, 95\% percentile intervals, and one index array shared by all scorers scored on the same units, and compared three ways of averaging the evidence. In all four pole tasks the interval for the L050-minus-L000 change lay below zero for ST, for the 317M model and for the critic, and the interval for ST's change minus the critic's change also lay below zero. ST kept 0.14 [0.10, 0.19], 0.25 [0.23, 0.27], 0.06 [\ensuremath{-}0.01, 0.13] and 0.09 [0.01, 0.18] of its evidence at L000 in acrobot-swingup, pendulum-swingup, cartpole-balance and cartpole-swingup; the 317M model kept 0.11 [0.07, 0.16], 0.24 [0.23, 0.26], 0.25 [0.15, 0.36] and 0.29 [0.23, 0.35]; the critic kept 0.71 [0.63, 0.80], 0.72 [0.70, 0.73], 0.73 [0.67, 0.79] and 0.81 [0.79, 0.84]. In the two control tasks the world models kept between 0.84 [0.66, 1.05] and 1.37 [1.24, 1.55]. ST's retention differed by less than 0.004 between the equal-unit, origin-weighted and fixed-clock summaries in every pole task. All five dated forecasts of E8 held. These intervals describe trajectory sampling conditional on the fitted models, configurations and tasks; they do not cover variation across training runs, checkpoints, critics or task choice (Appendix~\ref{app:v5-e8}).

\textbf{A second model class (E7).} In E7 we compared three budget-matched forward-model training procedures on common evaluation units. The arms used different training trajectories, with arm-specific preprocessing, checkpoint fits and calibration. For each task and arm, we used 29,000 optimization transitions from the stored E6 training trajectories at $q=0$ only (D-pol), at $q=1$ only (D-uni), or across all six levels (D-mix), with equal update budgets and common checkpoint-selection rows. D-mix had more evidence than D-pol at $q = 0.5$ in all four pole tasks (0.30 against 0.07, 2.26 against 1.82, 1.09 against \ensuremath{-}0.02, 1.31 against 0.91), as forecast. Two directional forecasts failed: the policy-only procedure met the half-baseline decline criterion in two of four tasks, and the mixed procedure retained at least half in two of four. Both forecasts required three. These comparisons describe the fitted procedures on these tasks. In cartpole-balance the shared configuration hid D-pol's evidence on its own trajectories (0.16 at $q = 0$); with level-specific configurations D-pol had 4.28 at $q = 0$ and 0.006 at $q = 0.5$, against 3.92 and 1.09 for D-mix. A third forecast, that D-pol would exceed D-uni at $q = 0$ in at least three tasks, also failed (two). Five of E7's eight forecasts held. D-all used 174,000 optimization transitions per task at the same update budget. Appendix~\ref{app:N} reports it as a sensitivity analysis.

\textbf{Seen after the fact.} At $q = 0.5$ D-mix had more evidence than both TD-MPC2 models in all four pole tasks (0.30, 2.26, 1.09, 1.31 against at most 0.18, 0.56, 0.58, 0.60). D-mix had higher midpoint evidence than both released TD-MPC2 scorers on these four tasks under the stated evaluation. Its different training data and observation-space score leave the source of the difference unresolved.

\textbf{What this does and does not say.} In these four tasks, world models that carried clear evidence about the action on trajectories generated around a policy carried little of it at the same kind of step on trajectories in which each step was uniform with probability one half, although a critic still found it. We do not identify which feature of the history is responsible; observations shifted with $q$ in the control tasks as well (Appendix~\ref{app:M}). We do not distinguish a change in how recoverable the action is from scorer-specific approximation error. We make no claim about planning or return. An evidence rate is a property of a scorer on a distribution of trajectories, and a rate measured near one policy should not be carried over to others.

\textbf{Null calibration.} In E6 the e-value statistics rejected twice in 23,040 null unit tests over the 24 primary panels of 960 units, and in E7 16 times in 53,760 over 56 panels; no panel triggered the audit. PRseq rejected in 41 to 52 units per panel of 960 in E6 and in 35 to 61 in E7, around its level of 48; the smallest binomial tail probabilities were 0.30 and 0.036.

\input{figures/v5/trajectory_evidence.tex}

\input{appendix/tables/V5_occupancy.tex}

%% file: figures/v5/trajectory_evidence.tex
\begin{figure}[!htbp]\centering
\resizebox{\textwidth}{!}{%
\begin{tikzpicture}
\begin{groupplot}[group style={group size=2 by 2,horizontal sep=.64cm,vertical sep=.84cm},
           width=.46\textwidth,height=1.46in,xmin=-.18,xmax=4.18,xtick={0,1,2,3,4},
           xticklabels={0,.1,.25,.5,.75},title style={font=\scriptsize},
           tick label style={font=\tiny},label style={font=\tiny},
           legend columns=3,legend style={draw=none,font=\tiny,inner sep=1pt}]
\nextgroupplot[title={Acrobot},ymin=0,legend to name=qlegend,xticklabels={,, , ,}]
\draw[gray!55,thin] (axis cs:0,0)--(axis cs:4,0);
\addplot[black,mark=*,mark size=1.2pt] coordinates {(0,1.2822538901145022) (1,0.7110656670600154) (2,0.3036974347173862) (3,0.1822176320983575) (4,0.2210732755513011)};
\addlegendentry{Single-task}
\addplot[only marks,forget plot,black,mark=o,mark size=1.55pt] coordinates {(0,1.4150032247759465) (1,0.7856720626002658) (2,0.30486123935070264) (3,0.22131673069401347) (4,0.2243752904270583)};
\addplot[black!65,mark=square*,mark size=1.2pt] coordinates {(0,0.6932003453755142) (1,0.32985354074198703) (2,0.10711686303433558) (3,0.07602404608243546) (4,0.07575562598385634)};
\addlegendentry{317M}
\addplot[only marks,forget plot,black!65,mark=square,mark size=1.55pt] coordinates {(0,0.7888953766473951) (1,0.3795763024097339) (2,0.11180591641010823) (3,0.050891494113863456) (4,0.08883195235019145)};
\addplot[black!40,mark=triangle*,mark size=1.2pt] coordinates {(0,2.3270364394741527) (1,1.7872031427776958) (2,1.603872528856774) (3,1.6586664657000163) (4,1.9685606191581133)};
\addlegendentry{critic}
\addplot[only marks,forget plot,black!40,mark=triangle,mark size=1.55pt] coordinates {(0,2.4732945857320185) (1,1.7416465568404496) (2,1.5950077614256404) (3,1.8337950851638327) (4,2.0650009840829107)};
\addplot[dashed,black,mark=diamond*,mark size=1.3pt] coordinates {(0,0.5475039190784067) (1,0.24170405751086965) (2,0.10403759284866394) (3,0.06906839172508777) (4,0.1496005966421497)};
\addlegendentry{D-pol}
\addplot[dashed,black!60,mark=pentagon*,mark size=1.3pt] coordinates {(0,0.8050735026712204) (1,0.42110532908714804) (2,0.2567106608422976) (3,0.298088051775646) (4,0.4863552524624087)};
\addlegendentry{D-mix}
\nextgroupplot[title={Pendulum},ymin=0,xticklabels={,, , ,}]
\draw[gray!55,thin] (axis cs:0,0)--(axis cs:4,0);
\addplot[black,mark=*,mark size=1.2pt] coordinates {(0,2.225188183948904) (1,1.7437900578457346) (2,0.9360607195094006) (3,0.5586008912099254) (4,0.4131874545155301)};
\addplot[only marks,forget plot,black,mark=o,mark size=1.55pt] coordinates {(0,2.1643774719773075) (1,1.7589160190232929) (2,0.9703853564600347) (3,0.5321519646677145) (4,0.40989792878307824)};
\addplot[black!65,mark=square*,mark size=1.2pt] coordinates {(0,2.2127725995139125) (1,1.6148692891857606) (2,0.5308527558605207) (3,0.5387203565384899) (4,0.6516607640446019)};
\addplot[only marks,forget plot,black!65,mark=square,mark size=1.55pt] coordinates {(0,2.1504832067301667) (1,1.619606051706194) (2,0.5867332422781538) (3,0.5630646688444211) (4,0.6475530113290628)};
\addplot[black!40,mark=triangle*,mark size=1.2pt] coordinates {(0,3.9702442830851554) (1,3.5526428244856256) (2,2.77829256334197) (3,2.8429290652881622) (4,2.7979144132729696)};
\addplot[only marks,forget plot,black!40,mark=triangle,mark size=1.55pt] coordinates {(0,3.924309771837157) (1,3.5326177826079896) (2,2.855631825546274) (3,2.8511797899342057) (4,2.79901954832327)};
\addplot[dashed,black,mark=diamond*,mark size=1.3pt] coordinates {(0,2.915319336281897) (1,2.4536798185839244) (2,1.597721624744767) (3,1.8160528156691167) (4,1.9951590645692383)};
\addplot[dashed,black!60,mark=pentagon*,mark size=1.3pt] coordinates {(0,3.2886721351441905) (1,2.768608835562916) (2,1.9653770844507936) (3,2.262833833647591) (4,2.2494474476934725)};
\nextgroupplot[title={Cart-pole balance},ymin=-0.4611084925260138,xlabel={$q$},ylabel={Nats/origin}]
\draw[gray!55,thin] (axis cs:0,0)--(axis cs:4,0);
\addplot[black,mark=*,mark size=1.2pt] coordinates {(0,2.935193152521826) (1,2.779450115543157) (2,2.3811226688344114) (3,0.16501289120815743) (4,-0.4611084925260138)};
\addplot[only marks,forget plot,black,mark=o,mark size=1.55pt] coordinates {(0,2.931756835345287) (1,2.7766299290986485) (2,2.3928340176376794) (3,0.2767805513007757) (4,-0.44953411363291507)};
\addplot[black!65,mark=square*,mark size=1.2pt] coordinates {(0,2.295760606136803) (1,2.2611280699436307) (2,2.1452159250195844) (3,0.5845867665422853) (4,0.26356722163998036)};
\addplot[only marks,forget plot,black!65,mark=square,mark size=1.55pt] coordinates {(0,2.296032367607522) (1,2.258040527938779) (2,2.125314493496268) (3,0.7681370849907938) (4,0.2526695232997765)};
\addplot[black!40,mark=triangle*,mark size=1.2pt] coordinates {(0,4.695807860692414) (1,4.6817280194932405) (2,4.643455729904598) (3,3.4262943559661525) (4,3.538573351068637)};
\addplot[only marks,forget plot,black!40,mark=triangle,mark size=1.55pt] coordinates {(0,4.690766272910048) (1,4.681281272953636) (2,4.638469916305945) (3,3.624352467925976) (4,3.4899561372991874)};
\addplot[dashed,black,mark=diamond*,mark size=1.3pt] coordinates {(0,0.16437954906845523) (1,0.16245197556080404) (2,0.15990703454438188) (3,-0.01734895437675686) (4,-0.12705542282637483)};
\addplot[dashed,black!60,mark=pentagon*,mark size=1.3pt] coordinates {(0,2.251126799093451) (1,2.2405811107946274) (2,2.213525409168682) (3,1.0918069519358533) (4,0.8468928614167699)};
\nextgroupplot[title={Cart-pole swingup},ymin=-0.33255964593895665,xlabel={$q$}]
\draw[gray!55,thin] (axis cs:0,0)--(axis cs:4,0);
\addplot[black,mark=*,mark size=1.2pt] coordinates {(0,2.4458788496610895) (1,1.7122757770441523) (2,0.9812464293241361) (3,0.21386576293910942) (4,-0.33255964593895665)};
\addplot[only marks,forget plot,black,mark=o,mark size=1.55pt] coordinates {(0,2.4545904466497452) (1,1.7055396598451926) (2,0.9754571476002011) (3,0.16989695833028284) (4,-0.3299363714107243)};
\addplot[black!65,mark=square*,mark size=1.2pt] coordinates {(0,2.098625961681359) (1,1.5609240465883352) (2,1.0974796661628807) (3,0.6004371590889039) (4,0.3585950933181501)};
\addplot[only marks,forget plot,black!65,mark=square,mark size=1.55pt] coordinates {(0,2.0846093375623704) (1,1.5676546178553727) (2,1.0643356684904326) (3,0.5634850092547663) (4,0.3759286002354415)};
\addplot[black!40,mark=triangle*,mark size=1.2pt] coordinates {(0,4.407309810049039) (1,4.211313489176737) (2,3.9642455494580426) (3,3.591459313713373) (4,3.274325719893828)};
\addplot[only marks,forget plot,black!40,mark=triangle,mark size=1.55pt] coordinates {(0,4.421229282816523) (1,4.174799943293937) (2,3.9626207756624794) (3,3.545999538108957) (4,3.3249924333417673)};
\addplot[dashed,black,mark=diamond*,mark size=1.3pt] coordinates {(0,1.5091781548189676) (1,1.3376843084914027) (2,1.163760277905291) (3,0.90606887931137) (4,0.7347566824324769)};
\addplot[dashed,black!60,mark=pentagon*,mark size=1.3pt] coordinates {(0,2.01863512769133) (1,1.81293550100828) (2,1.587611434702873) (3,1.3119267171131517) (4,1.1445566069200408)};
\end{groupplot}
\node[anchor=south] at ([yshift=1pt]current bounding box.north) {\pgfplotslegendfromname{qlegend}};
\end{tikzpicture}
}
\caption{Evidence at policy-centred steps as more of the trajectory's steps are drawn uniformly. The tested action's conditional law, each scorer's configuration and the reference are held fixed across levels. Hollow markers: E6. Filled markers and lines: E7, fresh units with the E6 procedure frozen. Dashed: forward models trained by us on the policy's trajectories only (D-pol) or on all levels (D-mix).}\label{fig:v5-occupancy}
\end{figure}
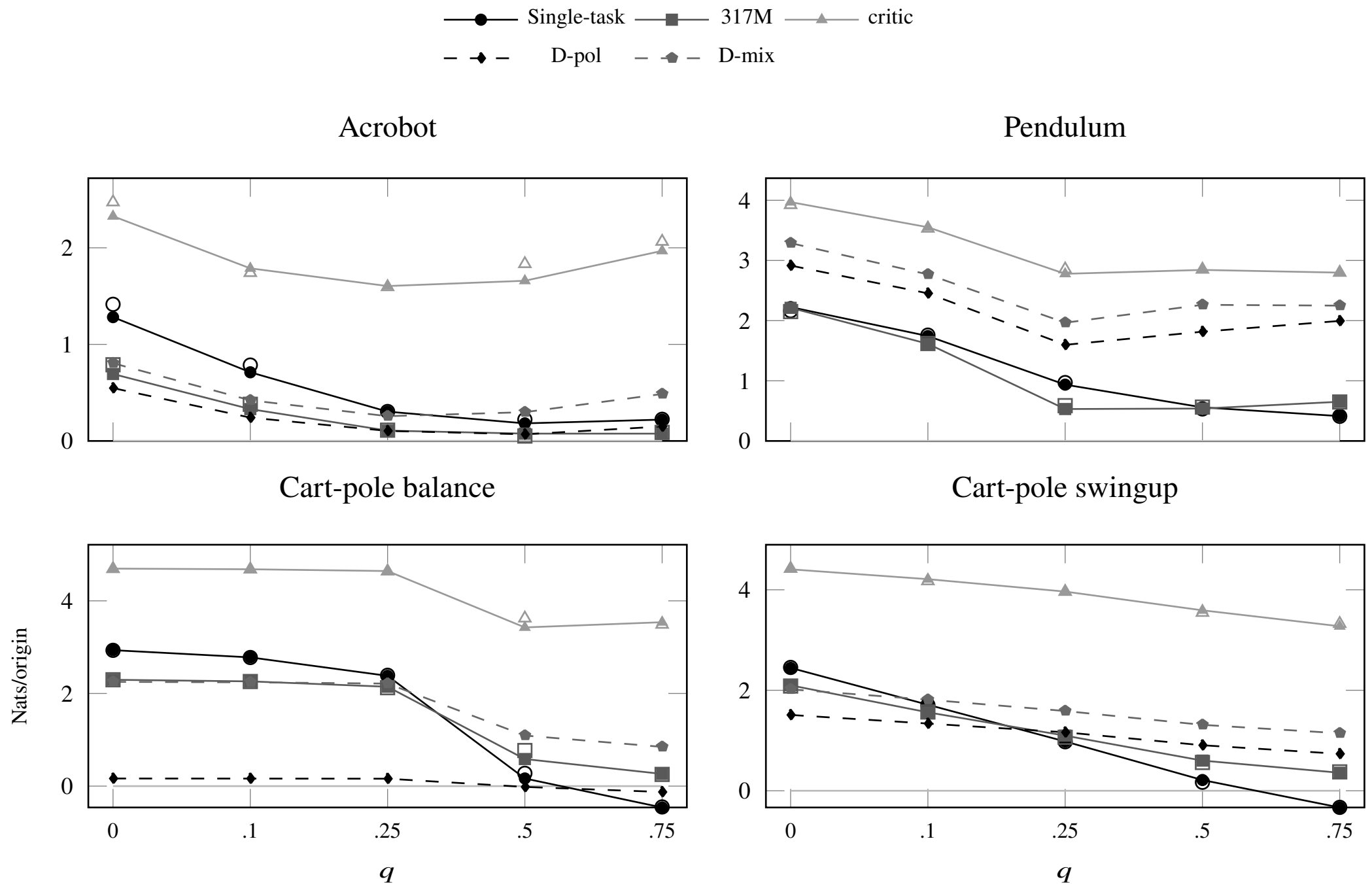

%% file: sections/09_discussion.tex
\section{Discussion and limitations}\label{sec:discussion}

\paragraph{Across settings.} Table 13 summarizes the earlier tempering comparisons in \S{}\S{}6–12. Under the collapse definition used here, none of the fixed-grid mixtures collapsed in those comparisons. The uniform mixture missed several power bars that validation tuning met. Each statistic that relies on one factor collapsed---its median log e-value fell far below zero while another valid statistic rejected in most units---in at least one setting: the pooled score in most of them; Kish tempering in the pixel world, on DIAMOND, on IRIS's decoded pixels and on four TD-MPC2 tasks; contrast tempering on DIAMOND, on IRIS's decoded pixels and on two TD-MPC2 tasks; and validation tuning on the ERK test fields, where a site-count-adaptive form removed the collapse but still lost most of the evidence for one model (\S\ref{sec:erkrho}). In the IRIS and TD-MPC2 study the weighted mixture reached the bar in exactly the games and tasks in which validation tuning did, and it kept the uniform mixture's protection on the ERK test fields.

\paragraph{Where the pooled score collapses.} Collapse occurred with per-site Gaussian scores on pixels and latents and on the ERK fields. In the earlier partial-execution study at $p=0.5$, the pooled statistic rejected in none of the 32 units in each of nine sets, while W(3/4) reached at least 24 rejections in seven sets. For the three named IRIS games at full execution and for the robot working-law analysis, pooled token scores did not meet our collapse definition. The pooled token score had a negative median log e-value in 14 of 26 IRIS games. We cannot tell in advance which case a new model falls in. Where the pooled score worked, the weighted mixture cost nothing in rejections, and it bounded the loss in the units where the pooled score failed.

\input{appendix/tables/V3_all_settings.tex}

\paragraph{What to do in practice.} For anyone asking whether a world model trained on experimental data responds to actions, our results suggest five habits. First, replacement diagnostics can report an effect through a stand-in. Second, use the conditional assignment law for resampling. A simpler resampler can reintroduce the trap if it changes that law. Third, for a known open-loop assignment law and fixed analysis schedule, compute the design bound before opening outcomes. Fourth, the mixture bounds the log-e loss relative to the best member of its fixed grid, and validation weighting bounds the loss relative to the selected grid member. Under partial execution, compare bounded stakes in the stated design (\S{}13). Fifth, give every negative control a statistic that has power, or an invalid resampler can go unnoticed.

\paragraph{The confirmatory evidence is narrow.} On biological data, the confirmatory analysis used one field of one dataset at one horizon. The repaired statistic rejected in this positive-control field under the reconstructed assignment law and the stated measurement and frozen-training assumptions. The thirty-minute horizon and the CcaS data were not part of that test. The registered tests on published world models (\S\ref{sec:published}--\S\ref{sec:p5}) cover four model families, 26 Atari games, six control tasks and one robot dataset, at a horizon of one step; the fourth registered study (\S\ref{sec:v5-stakes}) tested horizons of up to eight steps on one latent model. Two of them use simulators with random actions. The third uses recorded data, and its arm with the recorded commands is conditional on a design law that we reconstructed from the command records, because the configuration of the original collection could not be established; its null arm is exact. The registered simulator studies document pooling amplification and repair by mixtures on released models. On recorded video, the constructed-null checks used a known assignment law, and the 155 of 159 W(3/4) count with recorded commands is a sensitivity result under a reconstructed law and undocumented timing. They do not validate those models' counterfactual predictions, they do not test the stand-in trap, and they reach weak effects only through partial execution. No registered test of the mixture on biological data exists.

\paragraph{The analysis was not blind, and part of it is post hoc.} The datasets had been analyzed before in our own earlier study, so M4 is a registered re-analysis rather than a blind test. The pooled statistic registered for M4 failed its positive controls; the diagnosis, the repair and the exploratory check X all followed that failure, and only M5 is confirmatory for the repair. The repair was selected on synthetic data under a pre-committed rule whose tie-resolution detail was implemented only after the evaluation; on this evidence the Kish-tempered and field-mean statistics are interchangeable. The sealed challenge's designer had seen the source of the reference models. Every such departure is listed, with dates and hashes, in Appendix~\ref{app:disclosures}.

\paragraph{Validity rests on assumptions about the design and the measurement.} Every guarantee is conditional on the working design law; for ERK it is a reconstruction whose exact acquisition rule is not verified for the confirmatory field, and every CcaS rejection is conditional on a one-step delay, consistent with a shift at data export, that was found after the first design checks failed. We calibrated the working laws on the released command records, but no theorem licenses them (Proposition~\ref{prop:P3}). We also assume that the observation mask and segmentation do not respond to future actions. Designs whose logging policy reads past outcomes need the conditional product law of Assumption~\ref{ass:design-cl}; reading another unit's actions inside the block can violate that law, so a policy outside the sufficient construction of Proposition~\ref{prop:closedloop}(b) needs a separate argument for its conditional law. The own-channel analyses carry a vacuous eligibility bound and are descriptive only.

\paragraph{The repairs have limits of their own.} The tests are conservative and have little power for effects that are masked at the level of each unit or that are weak relative to the design's entropy. Contrast tempering restored power in the pixel world at the larger effect size, but even there without extracting the additional evidence that more pixels carry, so its e-values stayed far below the oracle's; on published world models it repaired neither DIAMOND nor IRIS's decoded pixels and failed on two of the six TD-MPC2 tasks, and we do not recommend it. The mixture can lose up to the logarithm of its grid size in log e-value relative to the best grid member, which can matter near the rejection threshold; it was weaker than contrast tempering and validation tuning at the smaller pixel effect, and it missed the power bar in two IRIS games and one TD-MPC2 task in which validation tuning reached it. The weighted mixture closes most of that gap, but trails validation tuning by up to log(4/3) when validation tuning is right. Where the pooled score does not collapse, as on robot video, tempering buys little in the typical unit; its value there is the bound on the worst units. The over-betting account (\S\ref{sec:overbet}) is exploratory. The earlier follow-the-leader temperature rules in \S{}12 had fewer rejections than the mixtures in the saved-score comparisons. Its growth guarantee across precision scales also requires grid coverage and the moment-domain conditions of Proposition~\ref{prop:growth}. Propositions~\ref{prop:growth}--\ref{prop:closedloop} were written after the registrations and checked in several sessions of AI review, each run by a language-model agent in its own session (Appendix~\ref{app:osf}), not by a human statistician. Proposition~\ref{prop:overbet} was added in the fourth version; its proof is short (Appendix~\ref{app:proofs:overbet}), and a later AI review of the statement led to the scope corrections of the fifth version. Separate automated sessions reviewed the conference antecedents of Propositions 14 and 16 and the finite-bag oracle underlying Proposition 15 in the exploratory stake protocol. Those records concern the versions they identify. The present statements and proofs are in \S{}\S{}13–14 and Appendix B.15. The pixel demonstration itself uses one simulator family, two model classes and one training run per cell, at reduced episode counts. Finally, a sharp-null rejection says that some site responds to some future action within the horizon; it does not say which, by how much, or through what mechanism. The use of AI tools is described in the statement before the references.

\paragraph{Outlook.} Three extensions would most strengthen the method: a registered test of the mixture on biological data with weak effects, which the ERK dataset analyzed here cannot provide (\S\ref{sec:erkxm})---randomized intervention experiments with high-dimensional readouts, in which one treatment label is shared by thousands of measured sites, have the same structure; statistics that recover the evidence pooling discards rather than merely discounting it---a joint Gaussian with a low-rank model of the residual covariance did not (\S\ref{sec:score-sampling}); and designs that are randomized with these tests in mind, since a known open-loop law and fixed analysis schedule allow the design bound to be evaluated before model training. On data that cannot be replayed, a known randomized assignment law can support a test of the specified sharp null under Proposition 1. The robot-data study also shows what such use requires: for data collected years ago, the design law had to be reconstructed from the command records. Datasets that record how their actions were randomized, and when each action took effect, would let an analyst use the recorded design law directly, subject to the score, timing and eligibility conditions of Proposition~\ref{prop:P1}.

\subsection{Limitations of Sections 13 to 15}
The stake result rests on one latent model family under a constructed regime in which the model always receives a label that is executed only part of the time; partial execution weakens the effect and makes the model wrong at the same time, and we have no result for a small, correctly specified effect or for sticky actions. Condition-specific validation used labels generated at the same known $p$, which a logged application may not have. In the exploratory IRIS-token comparison the partial-execution net gain was four of 288 for SW over T and three for TS over T. The results of Sections 13 to 15 other than the registered counts are exploratory. E3–E7 use simulated tasks at horizon one, principally TD-MPC2, with learned inverse-dynamics, critic and forward-model comparators. E1 also reuses token, pixel and robot scores and includes horizons 2, 4 and 8 on TD-MPC2. E2 uses TD-MPC2 and inverse-dynamics fits, with longer horizons reported descriptively. The 30 tasks share bodies and dynamics and are not 30 independent replications, and counts are not prevalence estimates. Four single-task checkpoints were chosen because the earlier seeds did not load with our loader, so comparisons involving single-task models describe these checkpoints. The references are a few networks at a few budgets, the weaker one was used throughout E4, and E5's comparisons cannot isolate which change made its references stronger. The tasks of E5 to E7 were chosen after earlier results, and the four pole tasks share dynamics in pairs. The reference of E6 and E7 is one critic trained on all levels, not an upper bound; E7's forward models are one small class with one fit each, and its repetition of E6 shares tasks, checkpoints and simulator with E6. E8 reports descriptive pointwise trajectory-bootstrap intervals for evidence rates and contrasts at P-selected steps in E6 and E7, including midpoint changes and retention ratios. These intervals quantify trajectory sampling conditional on the evaluated design, fitted models, configurations and tasks. Longer horizons were tested only to eight steps on one model with deterministic latent dynamics. A rejection concerns the system's sharp null; an evidence rate concerns a scorer on a given distribution of trajectories.

%% file: appendix/tables/V3_all_settings.tex
\begin{table}[htbp]\centering\small
\caption{Earlier comparisons of tempering statistics (\S{}\S{}6–12). Pixel world: rejected episodes at the larger effect (of 192, two-model average). ERK-like: alternative cells with all 16 replicates rejected (of 12, $E\ge100$). ERK test fields: four-model $\log E$ at 10 and 30 minutes (exploratory; X and the ERK validation-tuning check). DIAMOND: rejected ACTIVE units (of 96). IRIS tokens and TD-MPC2: games or tasks at or above the power bar (of 26 and 6). IRIS pixels: rejected units at $p=1$ (of 96). Robot video: rejections with the recorded commands under L-emp, a working-law sensitivity analysis (of 159). DIAMOND W(3/4): computed post hoc from saved scores (\S\ref{sec:published:scope}). A dash: not defined.}\label{tab:all-settings}
\resizebox{\linewidth}{!}{\begin{tabular}{@{}lrrrrrrrr@{}}\toprule
Statistic & Pixel world & ERK-like & \shortstack{ERK test\\fields} & DIAMOND & \shortstack{IRIS\\tokens} & \shortstack{IRIS\\pixels} & TD-MPC2 & \shortstack{Robot\\video} \\
\midrule
Pooled (C0) & 2 & 1 & \ensuremath{-}9,576 / \ensuremath{-}2,975 & 0 & 10 & 0 & 1 & 143 \\
Kish (C1) & 54 & 12 & 349 / 218 & 0 & -- & 0 & 2 & -- \\
Contrast (C1c) & 180 & -- & -- & 1 & -- & 20 & 4 & -- \\
Validation-tuned (V) & 185 & -- & 210 / \ensuremath{-}101 & 96 & 24 & 96 & 6 & 155 \\
Mixture (M) & 178 & 12 & 347 / 218 & 96 & 22 & 96 & 5 & 154 \\
W(3/4) & 185 & -- & 345 / 217 & 96 & 24 & 96 & 6 & 155 \\
\bottomrule\end{tabular}}\end{table}

%% file: sections/10_statements.tex
\enlargethispage{2\baselineskip}
\phantomsection
\section*{Author Contributions}
\label{sec:author-contributions}

\textbf{Bowen Xu:} Conceptualization, Methodology, Software, Investigation,
Formal analysis, Visualization, Writing -- original draft,
Writing -- review \& editing, Project administration.

\textbf{Boyu Chen:} Conceptualization, Methodology, Validation, Writing -- review \& editing.

\textbf{Jiaying Li:} Writing -- review \& editing.

\textbf{Haiyi Li:} Writing -- review \& editing.

%% file: appendix/A_biowm_eval.tex
\section{BioWM-Eval}\label{app:biowm-eval}

Both datasets of \S\ref{sec:real} were analyzed earlier in our BioWM-Eval study. This appendix states what that study measured, how its results relate to the tests of this paper, and what its records do and do not establish about its own analysis history. The quotations and figures below are taken from the committed revision package of that study (Appendix~\ref{app:osf} lists the files and hashes).

\paragraph{What it measured.} BioWM-Eval defined the utility of future actions as the relative reduction in the best achievable squared prediction error when the future actions are added to the state and the clock, $U^{*}(h)=[R_0^{*}(h)-R_1^{*}(h)]/R_0^{*}(h)$, and estimated it by comparing held-out risks of matched fitted models with and without the future actions. In its own words, this quantity depends on the information set and does not estimate a causal effect. For ERK it compared matched one-step ridge models with and without the next programmed light pulse, both using the same past values and clock terms, and forecast thirty minutes ahead; for CcaS/CcaR it used seven leave-one-date-out comparisons at the same horizon. The ERK gain on the original minimum-gap fields was \BiowmEvalForecastGainPooledErkHthirty; the CcaS gain was \BiowmEvalForecastGainPooledCcasHthirty, below the study's \BiowmEvalPracticalGainThresholdPooled{} practical cutoff, which was an internal decision rule fixed before the biological analyses and not a public preregistration.

\paragraph{How it relates to this paper.} A with-versus-without-actions forecast comparison is the kind of diagnostic whose apparent action effects \S\ref{sec:trap} shows can arise from stand-in effects. Our rejections therefore neither validate the size of the ERK gain nor say anything about whether the CcaS gain is practically meaningful: the sharp-null tests ask whether any site responds to the future actions, not how much prediction improves. Where the two studies are mentioned together, the juxtaposition is descriptive. The original ERK and CcaS/CcaR studies had already demonstrated that these systems can be predicted or controlled; BioWM-Eval did not claim those biological findings as its own, and neither do we.

\paragraph{What its records establish.} BioWM-Eval's biological analyses were sequential and exploratory; unlike its synthetic challenges, none had sealed inputs or a sealed evaluator. Before the clock-controlled ERK comparison, an earlier comparison without the matched state-and-clock baseline had already been run on the same two held-out experiments, with gains of \BiowmEvalEarlierErkGainExpOne{} and \BiowmEvalEarlierErkGainExpTwo{}. A rule-freeze record precedes the first clock-controlled result by less than an hour, but it does not name that comparison as primary, and the original project kept no version history; whether the ERK comparison was designated primary before or after its first result therefore cannot be established. The study's synthetic challenges were sealed and revealed in a recorded order, but the challenge author and the evaluator were operated by the same agent, so its ``blind'' evaluations reflect an internal sealed-input order, not independent human blinding or public preregistration. These facts are why we treat M4 as a re-analysis (\S\ref{sec:real:data}) and why no BioWM-Eval figure is used as evidence in this paper.

%% file: appendix/B_proofs.tex
\section{Proofs}\label{app:proofs}

\subsection{Sources and changes relative to the frozen documents}\label{app:proofs:sources}

The obligations P1--P4 of the registered specification are proved in two theory documents that are part of the registered code archive and were therefore fixed before the registration: a core document (P1(a)--(d), P1b, P2(a)--(e), P3, P4) and a calibration supplement (P1(e) and the template-avoidance recursion). A third note in the archive covers averaging across models. Table~\ref{tab:theory-sources} lists them with their hashes; the files used here are byte-identical to the archived ones.

\begin{table}[h]
\centering
\small
\caption{Frozen theory documents. Hashes are SHA-256 of the bytes in the registered code archive.}
\label{tab:theory-sources}
\begin{tabular}{@{}l l p{6.3cm}@{}}
\toprule
Document & Covers & SHA-256 \\
\midrule
Core theory & P1(a)--(d), P1b, P2, P3, P4 & \texttt{c726f679\dots c2c8c} \\
Calibration supplement & P1(e), template HMM & \texttt{0a35b031\dots 57081} \\
Pilot note & model averaging, ceilings & \texttt{81f39548\dots 8c452a} \\
\bottomrule
\end{tabular}
\end{table}

The proofs below follow these documents. Writing them for the paper, we made the following changes, none of which affects the registered method: [Added] the selection, training-independence and Monte Carlo hypotheses (i)--(iii) of Proposition~\ref{prop:P1}, which the frozen proofs use implicitly, together with the frozen training and design information in the conditioning fields and the grid filtration; [Tightened] the conditioning field under $H_0^{\mathrm{own}}$ is written out explicitly, with the targets of units outside the independent set excluded; [Added] the hypotheses of Proposition~\ref{prop:elig}(b) on the template list and on positive conditioning probability, and the derivation of the product path law from the per-origin design assumption; [Tightened] the approximate-law bound (Proposition~\ref{prop:P3}), replacing a coupling argument by a direct total-variation bound conditional on the frozen baseline and defining $\Delta$ on the whole command path; [Tightened] the range of density ratios in Proposition~\ref{prop:P1b} ($r\ge1$, with the reversed case noted) and the conditioning on frozen training information in Proposition~\ref{prop:P2}(c); [Added] for Proposition~\ref{prop:P4}, the reading as independent and identically distributed copies at a fixed clock, the restriction of the frozen construction that the predictor's window contains recent commands and constant outcomes but not the stratum, now stated in the proposition, and a written argument for its probability-one step for ideal predictors. Proposition~\ref{prop:kish} is new and is proved in \S\ref{sec:theory:tempering}.

Propositions~\ref{prop:growth}--\ref{prop:closedloop} have no frozen counterpart: they were written for the earlier theory extension after the registrations and checked by Codex, a model family different from the drafting model, in an independent review, a delta review of the revision, a check of the trap experiment's logging policy, and a final check of the integrated text with a delta check of its revisions; the counterexamples are archived with the reviews (Table~\ref{tab:e5-reviews}). Their proofs are in \S\ref{app:proofs:growth}--\S\ref{app:proofs:closedloop}.

\subsection{Validity (Proposition~\ref{prop:P1})}\label{app:proofs:P1}

\paragraph{Conditioning fields.} Let $\mathcal{P}$ be the $\sigma$-field generated by all potential-outcome functions, including exogenous noise and the measurement process, $\mathcal{T}$ the frozen training information and $\mathcal{S}$ the structural design information (strata, layout and fixed allocations); when Proposition~\ref{prop:elig}(b) is used, let $J$ be the eligibility pattern of all units. Put $\mathcal{C}^{G}_t=\sigma(\mathcal{F}_t,\mathcal{P},\mathcal{T},\mathcal{S},J)$ and, for an independent set $I$ of $\mathcal{N}$, $\mathcal{C}^{I}_t=\sigma(\mathcal{F}_t,\mathcal{P},\mathcal{T},\mathcal{S},J,A^{F}_{j,t}:j\notin I)$, omitting $J$ when no eligibility conditioning is used. [Added] By hypothesis (ii), the design law of the joint vector of tested blocks keeps its product form after $\mathcal{T}$ and $\mathcal{S}$ are adjoined.

\paragraph{Targets are measurable.} Under no anticipation and $H_0^{G}(h)$, $Y_{i,c,t+h}=Y_{i,c,t+h}(A)$ depends on $A$ only through $A_{\cdot,\le t}$, so it is measurable with respect to $\sigma(\mathcal{P},A_{\cdot,\le t})\subseteq\mathcal{C}^{G}_t$. Under $H_0^{\mathrm{own}}(h;\mathcal{N})$ and for $i\in I$: the own block has no effect; the blocks of non-neighbours have no effect; and the neighbours of $i$ lie outside $I$ because $I$ is independent. Changing these blocks one at a time, each invariance holding for every value of the other coordinates, shows that $Y_{i,c,t+h}$ is $\mathcal{C}^{I}_t$-measurable. [Tightened] A target of a unit $j\notin I$ need not be: it may depend on $A^{F}_{i,t}$ for a neighbour $i\in I$, which is why the conditioning in Proposition~\ref{prop:P1}(b) includes only the targets of units in $I$.

\paragraph{The design law is unchanged.} Assumption~\ref{ass:design} gives $\mathrm{Law}(A^{F}_{\cdot,t}\mid\mathcal{F}_t,\mathcal{P})=\prod_iQ_{i,t}$. By hypothesis (ii), adjoining $\mathcal{T}$ and $\mathcal{S}$ leaves this unchanged; under eligibility conditioning, $Q_{i,t}$ is replaced by its conditioned version (Proposition~\ref{prop:elig}(b)). Adjoining the blocks $A^{F}_{j,t}$, $j\notin I$, leaves $\prod_{i\in I}Q_{i,t}$ as the conditional law of the blocks in $I$, by independence across units. Since the targets in question are measurable with respect to the enlarged field, conditioning on them changes nothing further.

\paragraph{Mean one.} By hypotheses (i)--(ii), each $T_{i,t}(\cdot)$ is a $\mathcal{C}_t$-measurable random function of the candidate block, where $\mathcal{C}_t$ is $\mathcal{C}^{G}_t$ or $\mathcal{C}^{I}_t$. Hence
\[
\mathbb{E}\bigl[e^{\mathrm{ex}}_{i,t}\mid\mathcal{C}_t\bigr]=\frac{\sum_aQ_{i,t}(a)\,e^{T_{i,t}(a)}}{\sum_aQ_{i,t}(a)\,e^{T_{i,t}(a)}}=1 .
\]
For the MC form, by hypothesis (iii) the observed block and the $B$ draws are independent and identically distributed with law $Q_{i,t}$ given $\mathcal{C}_t$. Their normalised shares $e^{T(\cdot)}/\sum e^{T(\cdot)}$ sum to one and are exchangeable, so the observed share has conditional mean $1/(B+1)$ and $e^{\mathrm{mc}}_{i,t}$ has conditional mean one; ties are immaterial because every score is strictly positive. Conditional independence across units of (blocks, draws) gives the product statements of (a) and (b).

\paragraph{Grid martingale.} Fix a residue $r$ and its grid $t_0<t_1<\dots$, $t_{k+1}=t_k+h$. Let $\mathcal{G}_k=\sigma(\mathcal{F}_{t_k},\mathcal{T},\mathcal{S},J,U_{t_0},\dots,U_{t_{k-1}})$, where $U_{t_j}$ are the independent auxiliary seeds that generate the Monte Carlo draws at $t_j$. The factor $X_k$ at $t_k$ is a function of $\mathcal{F}_{t_k}$-measurable covariates, of the block $A_{\cdot,t_k+1:t_k+h}$ and the targets at $t_k+h$ (both $\mathcal{F}_{t_{k+1}}$-measurable), and of the draws at $t_k$; so $X_k$ is $\mathcal{G}_{k+1}$-measurable. [Tightened] Given the history, the earlier draws are functions of that history and of the earlier seeds, which are independent of the assignments and of $\mathcal{T}$ and the potential outcomes; adjoining them leaves the law of the future blocks unchanged, so
\[
\mathbb{E}[X_k\mid\mathcal{G}_k]=\mathbb{E}\bigl\{\mathbb{E}[X_k\mid\mathcal{C}_{t_k},\text{earlier seeds}]\big|\mathcal{G}_k\bigr\}=1 .
\]
Thus $\prod_{j<k}X_j$ is a non-negative $(\mathcal{G}_k)$-martingale with initial value one. Ville's inequality gives $\Pr(\sup_k\prod_{j<k}X_j\ge1/\alpha)\le\alpha$, and $\mathbb{E}[\prod_{j<\tau}X_j]\le1$ for every $(\mathcal{G}_k)$-stopping time $\tau$, by localization and Fatou's lemma, with the product on $\{\tau=\infty\}$ read as the almost-sure limit of the non-negative martingale.

\paragraph{Averages and the batch test (part (c)).} A prechosen arithmetic average of non-negative random variables with expectation at most one has expectation at most one, without any assumption on their dependence; Markov's inequality then gives $\Pr(E\ge1/\alpha)\le\alpha$. Averages over independent sets or models of processes adapted to the same filtration are again martingales; processes of different residue classes are adapted to different grids, and on a common clock a class factor is not conditionally mean one once part of its block is revealed, so the class average is a fixed-time e-value only. For the batch test at one common origin, the observed joint block and the joint draws are exchangeable given $\mathcal{C}_t$ (all units under $H_0^{G}$, one independent set under $H_0^{\mathrm{own}}$), and the observed-inclusive rank p-value that counts ties as exceedances is super-uniform.

\subsection{Eligibility (Proposition~\ref{prop:elig})}\label{app:proofs:elig}

\paragraph{Part (a).} Let $\Pi^{*}$ be the pipeline that uses the structural eligibility events and $\hat\Pi$ the pipeline that uses command-based classifiers, run with the same statistics and Monte Carlo draws. Let $D$ be the event that any classifier decision differs from its structural event. On the complement of $D$ the two pipelines are identical. Hence, unconditionally,
\[
\Pr(\hat\Pi\text{ rejects})\le\Pr(\Pi^{*}\text{ rejects})+\Pr(D)\le\alpha+\delta,
\]
capped at one; distinct classifier errors may be combined by a union bound. No independence between classifier errors and assignments is needed. The bound does not hold conditionally on an observed selected set.

\paragraph{Part (b): from the per-origin law to the path law.} [Added] Let $K_{i,s}(\cdot\mid A_{i,\le s})$ be the one-step kernel of the design, with the fixed stratum suppressed. Assumption~\ref{ass:design} at every clock gives the product of these kernels conditionally on the full history, the potential outcomes and the frozen information. Taking conditional expectations onto $\sigma(\mathcal{F}_t,\mathcal{P},\mathcal{T},\mathcal{S},A_{\cdot,t+1:s})$ preserves the product, because each kernel is a function only of its own action prefix and of fixed design information, and iterating gives $\mathrm{Law}(A_{\cdot,>t}\mid\mathcal{F}_t,\mathcal{P},\mathcal{T},\mathcal{S})=\prod_iQ^{\mathrm{path}}_{i}(\cdot\mid A_{i,\le t})$: future paths are conditionally independent across units and follow the open-loop design. For an eligibility pattern $j=(j_i)$ of positive probability, put $\mathcal{E}_i^{(1)}=\mathcal{E}_i$ and $\mathcal{E}_i^{(0)}=\mathcal{E}_i^{c}$. Conditioning on $\bigcap_i\mathcal{E}_i^{(j_i)}$, each event a function of path $i$ only, preserves the product; an analysed unit has path law $Q^{\mathrm{path}}_i(\cdot\mid A_{i,\le t},\mathcal{E}_i)$, and an excluded unit contributes the factor one. Its one-block conditionals are
\[
Q_{i,t}(a\mid\mathcal{E}_i,A_{i,\le t})=\frac{Q_{i,t}(a)\,\Pr(\mathcal{E}_i\mid A_{i,\le t},a)}{\Pr(\mathcal{E}_i\mid A_{i,\le t})},
\]
well defined because the denominator is positive by hypothesis. Under the null the targets remain measurable as in \S\ref{app:proofs:P1}, so the argument there applies verbatim under the conditional law given the eligibility pattern ($J$ in $\mathcal{C}_t$). The hypothesis that every unit on which $\mathcal{E}_i$ holds follows $Q_i$ is what makes the conditioned law the law of the analysed units; a template list that missed a fixed design would violate it.

\paragraph{Part (b): exact computation for a hidden-Markov design.} Index the emissions by $0,\dots,T-1$, so that $T$ is the terminal clock after the last emission. Let $H_t$ be the hidden state just before the emission at clock $t$, and let $M_t$ be the bit mask of which fixed templates still match the emitted prefix (two templates give four mask values). A transition branch $j$ from state $s$ at clock $t$ has destination $d_t(s,j)$, emission $b_t(s,j)$ and probability $w_t(s,j)$, and updates the mask to $m\land\mathrm{match}_t(b_t(s,j))$; the eligibility event holds iff the final mask is zero. Define $\beta_T(s,m)=\mathbf{1}\{m=0\}$ and
\[
\beta_t(s,m)=\sum_jw_t(s,j)\,\beta_{t+1}\bigl(d_t(s,j),\,m\land\mathrm{match}_t(b_t(s,j))\bigr).
\]
By induction, $\beta_t(s,m)$ is the probability that the remaining path avoids every template, given state $s$ and mask $m$. With $\alpha_{t+1}$ the ordinary forward filter conditional on the observed prefix (the current mask $m$ is determined by that prefix), the eligibility probability is $Z=\sum_s\alpha_{t+1}(s)\,\beta_{t+1}(s,m)$. For a candidate block $a$, propagate the unnormalised filter through $a$, update the mask, and take the inner product with $\beta_{t+h+1}$; dividing by $Z$ gives $Q_{i,t}(a\mid\mathcal{E}_i,A_{i,\le t})$. Summing over all blocks recovers $Z$ by the backward recursion, which proves normalisation. For exact sampling, draw the initial state with mass proportional to $\alpha\beta$ and each branch with mass proportional to $w_t\,\beta_{t+1}$ at its updated destination and mask; the $\beta$ factors telescope, leaving the full-path mass times $\mathbf{1}\{\mathcal{E}_i\}/Z$. Once the mask is zero, $\beta\equiv1$ and the ordinary kernel applies. The sampling transition is used only at states with positive $\beta$; states with $\beta=0$ have probability zero under the conditioned sampler.

\subsection{Distributional nulls do not compose (Proposition~\ref{prop:P1b})}\label{app:proofs:P1b}

With design law $(\tfrac12,\tfrac12)$ and a score whose density ratio between the matching and the non-matching bit is $r$, the exact factor is $2r/(r+1)$ for a match and $2/(r+1)$ otherwise; $r=9$ gives $9/5$ and $1/5$. In system (ii), unit $i$ matches iff $U_k=0$, so at each origin both units match or both fail, each with probability $\tfrac12$. In system (i), unit 1 matches iff $A_{1,k}=U_k$ and unit 2 matches iff $A_{2,k}=U_k\oplus A_{1,k}\oplus A_{2,k}$, that is iff $A_{1,k}=U_k$; again both match or both fail with probability $\tfrac12$. The per-origin product therefore has mean $\tfrac12\bigl[(9/5)^2+(1/5)^2\bigr]=1.64$, and the product over three independent origins has mean $1.64^3\approx4.41$. Because $(9/5)^6\ge20$ while any product containing a factor $1/5$ is below $20$, rejection at $\alpha=0.05$ occurs iff both units match at all three origins, with probability $1/8$. [Tightened] Among ratios $r\ge1$ the same holds exactly when $(2r/(r+1))^6\ge20$, that is $r\ge20^{1/6}/(2-20^{1/6})\approx4.675$, and for $1\le r<4.675$ the rejection probability is zero: a product that mixes matching and non-matching origins is at most $64r^4/(1+r)^6\le1024/729<20$. For $r<1$ the roles reverse, and for $r\le(2-20^{1/6})/20^{1/6}$ rejection occurs when all three origins mismatch, again with probability $1/8$. Each unit-wise null holds because $U_k$ is a fresh fair bit: in (ii), $Y_{i,k}=A_{i,k}\oplus U_k$ is independent of $A_{i,k}$; in (i), $Y_{1,k}=U_k$ and $Y_{2,k}$ is a fair bit independent of $A_{2,k}$ given $A_{1,k}$. If $U$ were held fixed across origins, earlier outcomes would reveal it, and after the first origin the unit-wise null would fail for unit 2 in system (i) and for both units in system (ii). Both systems violate the sharp nulls, because the outcomes depend on the actions within the horizon.

\subsection{Design limits (Proposition~\ref{prop:P2})}\label{app:proofs:P2}

\paragraph{Parts (a) and (b).} The exact denominator includes the term $Q_{i,t}(A^{F}_{i,t})\,e^{T(A^{F}_{i,t})}$, so $e^{\mathrm{ex}}_{i,t}\le1/Q_{i,t}(A^{F}_{i,t})$; a point-mass law gives $e\equiv1$. If $K$ of the $B$ draws equal the observed block, the MC denominator contains at least $(1+K)\,e^{T(A^{F}_{i,t})}$, so $e^{\mathrm{mc}}_{i,t}\le(B+1)/(1+K)$, and given the commands $K\sim\mathrm{Binomial}(B,Q_{i,t}(A^{F}_{i,t}))$. Within a product component, $\log E$ is therefore at most the sum of the corresponding per-factor log bounds. An arithmetic average of components can reach $1/\alpha$ only if some component does. Under the open-loop assumption of part (b), integrating this necessary event over coherent command paths drawn from $Q$ (and, for MC, over independent draws) bounds the rejection probability. Because averaging models adds no design entropy, the same command-only bound applies to the model average; for the Monte Carlo form this presumes that the models share their candidate bags, and otherwise the maximum runs over models with their own collision counts. The MC ceiling differs from the exact one: for a fair bit with $B=20$ and a matching-score ratio of $10^{6}$, the MC test rejects only when no draw matches, with probability $2^{-20}$, while the exact-form ceiling at $\alpha=0.05$ is zero. The exact-form ceiling therefore bounds exact e-values only; for a Monte Carlo analysis it is a comparator.

\paragraph{Part (c).} [Tightened] Fix $\mathcal{F}_t$ and the frozen training information $\mathcal{T}$, on which the score $s_y$ depends; let $P_y=\mathrm{Law}(A^{F}\mid Y=y,\mathcal{F}_t,\mathcal{T})$, $Z_y=\sum_aQ(a)s_y(a)$ and $R_y(a)=Q(a)s_y(a)/Z_y$. Then $\log e^{\mathrm{ex}}=\log R_Y(A^{F})-\log Q(A^{F})$, and taking expectations over $A^{F}\sim P_y$ and then over $Y$,
\[
\mathbb{E}\bigl[\log e^{\mathrm{ex}}\mid\mathcal{F}_t,\mathcal{T}\bigr]=\mathbb{E}_Y\bigl[\mathrm{KL}(P_Y\Vert Q)-\mathrm{KL}(P_Y\Vert R_Y)\bigr]\le\mathbb{E}_Y\,\mathrm{KL}(P_Y\Vert Q),
\]
and $\mathbb{E}_Y\mathrm{KL}(P_Y\Vert Q)=I(Y;A^{F}\mid\mathcal{F}_t,\mathcal{T})\le H_Q(A^{F}\mid\mathcal{F}_t)$. The equality uses that $A^{F}\sim Q$ given $(\mathcal{F}_t,\mathcal{T})$ under Assumption~\ref{ass:design} and hypothesis (ii), whatever the alternative. Without $\mathcal{T}$ in the conditioning the bound can fail, since a score fitted on training data may carry information about the targets that $\mathcal{F}_t$ does not. The floor guarantees that $R_y$ charges the support of $Q$. This is a growth statement, not a power threshold.

\paragraph{Part (d).} The exact weighted-tail p-value includes the observed joint block's mass, so $p\ge Q^{\mathrm{joint}}(A^{F}_{\mathrm{obs}})$; in the add-one MC form, $K$ matching draws give $p\ge(1+K)/(B+1)$.

\subsection{Approximate design laws (Proposition~\ref{prop:P3})}\label{app:proofs:P3}

[Tightened] Let $\mathcal{B}$ contain the pre-test history and the frozen design, training and working-law information. Conditionally on $\mathcal{B}=b$, let $q_b$ and $\hat q_b$ be the laws of the command path under the true design $Q$ and under the working law $\hat Q$, and let $\nu_b$ be the common conditional law of the potential outcomes and the Monte Carlo seeds, which under Assumption~\ref{ass:design} are independent of the path under both laws. Let $\mathbb{P}_0$ be the reference experiment in which the path follows $\hat Q$ and everything else is unchanged. Because $\hat Q$ is fixed independently of the tested blocks and satisfies the requirements of Assumption~\ref{ass:design}, Proposition~\ref{prop:P1} applies to $\mathbb{P}_0$ conditionally on $\mathcal{B}$, so $\mathbb{P}_0(E\ge1/\alpha\mid b)\le\alpha$ and $\mathbb{E}_0[E\mid b]\le1$. The same measurable analysis map produces $E$ in both experiments, and $d_{\mathrm{TV}}(q_b\otimes\nu_b,\hat q_b\otimes\nu_b)=\Delta(b)$. Hence $\mathbb{P}(E\ge1/\alpha\mid b)\le\alpha+\Delta(b)$ and, for $0\le E\le M$ with a constant $M$, $\mathbb{E}[E\mid b]\le1+M\Delta(b)$. Integrating over $\mathcal{B}$ gives the stated bounds with $\mathbb{E}[\Delta]$; a baseline-dependent bound $M(\mathcal{B})$ gives $1+\mathbb{E}[M(\mathcal{B})\Delta(\mathcal{B})]$ instead. For sequential kernels,
\[
\Delta(b)\le\min\Bigl\{1,\;\mathbb{E}_Q\Bigl[\textstyle\sum_s d_{\mathrm{TV}}\bigl(K_s(\cdot\mid A_{<s},b),\hat K_s(\cdot\mid A_{<s},b)\bigr)\,\Big|\,\mathcal{B}=b\Bigr]\Bigr\},
\]
and for product kernels each one-step distance is at most the sum of the unit distances. The frozen document derived the moment bound and, separately, the event bound via a coupling; the direct argument here gives the event bound without a coupling and without a bound on $E$.

\subsection{Failure of replacement diagnostics (Proposition~\ref{prop:P4})}\label{app:proofs:P4}

\paragraph{Positivity of the gaps.} [Added] Write $Y=\theta+\eta$ with $\eta$ independent centred noise of finite variance, so $\mu(W,F)=\mathbb{E}[\theta\mid W,F]$ and $\mu_0(W)=\mathbb{E}[\theta\mid W]$. As in the frozen construction, $W$ consists of finitely many recent commands, the recent outcomes, which are constant, and the clock, and $\theta$ enters only an earlier outcome outside the window. [Tightened] This restriction is needed: if the recent outcomes revealed $\theta$, then $\mu(W,F)=\mu_0(W)=\theta$ and all three gaps would be zero. Under it, every finite command window has positive probability under both strata, because the chain is stationary with transition probabilities in $(0,1)$, so $0<\Pr(\theta=1\mid W)<1$. Given the last command in $W$, the first future command has stratum-specific probabilities that differ by $(1-\rho)(p_1-p_0)\ne0$, so the likelihood ratio of $F$ given $W$ between $\theta=1$ and $\theta=0$ is non-constant and $\mu(W,\cdot)$ is non-constant on a set of windows of positive probability. (i) Hence $\mathbb{E}[(\mu-\mu_0)^2]>0$; if the window's commands are removed as well, the retrained predictor is the constant $\Pr(\theta=1)$ and the gap is $\mathrm{Var}(\mu)>0$. (ii) Every future block, including the zero block, has positive conditional probability given $W$, so $\mu(W,0)$ is defined and $\mathbb{E}[(\mu(W,F)-\mu(W,0))^2]>0$. (iii) With $F'$ an independent unit's block, $\mathbb{E}[(Y-\mu(W,F'))^2]-\mathbb{E}[(Y-\mu(W,F))^2]=\mathbb{E}[(\mu(W,F)-\mu(W,F'))^2]>0$; the cross term vanishes because $\mathbb{E}[Y-\mu(W,F)\mid W,F,F']=0$ when $F'$ is independent of $(Y,W,F)$. Under squared-error loss, each population gap equals the corresponding expected squared difference of predictors.

\paragraph{Probability tending to one.} [Added] For the ideal predictors the per-unit loss differences $D_i=(Y_i-g_i)^2-(Y_i-\mu_i)^2$ are independent and identically distributed (over fixed pairs for the swap diagnostic), with finite variance because $\mu,g\in[0,1]$ and the noise has finite variance, and with mean equal to the population gap $\gamma>0$. By the law of large numbers $\bar D_N\to\gamma$ in probability, and by consistency of the nonparametric bootstrap for the mean the lower bootstrap quantile is $\bar D_N+O_P(N^{-1/2})$. Hence the probability that the lower interval bound exceeds zero tends to one. As in the frozen document, this is a statement about the ideal predictors. It is not claimed for fitted predictors, which would need, in addition, fitting on independent units and convergence in $L^2$ under both the observed and the replacement input laws.

\paragraph{Validity of the randomization test.} The design law $Q(\cdot\mid A_{i,t},\theta_i)$ is known to the design, the outcome does not depend on any action, and the statistic uses only admissible inputs, so Proposition~\ref{prop:P1} applies: resampling from the design law preserves the dependence between the future block and $\theta_i$, and the size is at most $\alpha$.

\subsection{Growth bound (Proposition~\ref{prop:growth})}\label{app:proofs:growth}

Write $K(\lambda)=\log\mathbb{E}_{Q}e^{-\lambda\Delta S}$ for one origin, dropping indices, so that $\log e^{\mathrm{ex}}(\lambda)=-K(\lambda)$. $K$ is a log Laplace transform, hence convex on the interval where it is finite; integrability of $\Delta S$ and a finite right exponential moment give $K'(0+)=-\mu$. Centering, $K(\lambda)=-\lambda\mu+\log\mathbb{E}_{Q}e^{-\lambda(\Delta S-\mu)}\le-\lambda\mu+\lambda^{2}\nu/2$ on $[0,\Lambda]$, which is the per-origin bound; summing over the origins of one component gives $f$, and completing the square gives the optimizer when $\mathcal{V}>0$ (if $\mathcal{V}=0$ the bound is linear). Hoeffding's lemma applies to bounded $\Delta S$. For an MC factor, $\log e^{\mathrm{mc}}(\lambda)=-\log\mathbb{E}_{\hat Q}e^{-\lambda\Delta S}$ with $\hat Q$ the empirical law of the observed-inclusive bag, so the same algebra holds with $\hat Q$-moments; a $Q$-based bound does not transfer to every sampled denominator. Under the null, Proposition~\ref{prop:P1} gives $\mathbb{E}[e\mid\mathcal{C}_t]=1$, so Jensen gives $\mathbb{E}[\log e\mid\mathcal{C}_t]\le0$; exchangeability gives $\mathbb{E}[\mu\mid\mathcal{C}_t]=0$ when integrable, and taking expectations of the per-origin bound, within a common deterministic moment domain, gives the lower bound $-\lambda^{2}\mathbb{E}[\nu]/2$. On a single path the term $\lambda\mu$ need not vanish: for a fair binary design with scores $\pm1$, the low-score path has $\log e=-\lambda-\log\cosh\lambda$. A positive expected lower bound needs $\mathbb{E}[\mathcal{M}]>0$, $0<\mathbb{E}[\mathcal{V}]<\infty$ and $0<\lambda<2\mathbb{E}[\mathcal{M}]/\mathbb{E}[\mathcal{V}]$ within the domain; if $\mathbb{E}[\mathcal{V}]=0$ the linear bound $\lambda\mathbb{E}[\mathcal{M}]$ applies.

\subsection{Precision rescaling (Proposition~\ref{prop:blind})}\label{app:proofs:blind}

For an unfloored Gaussian site score, $\ell(a)=-\tfrac12\log(2\pi v/\gamma)-\gamma(Y-\mu(a))^{2}/(2v)$, so the candidate-dependent part scales by $\gamma$ and the rest, $C(\gamma)$, is common to all candidates and cancels in the exact and the same-bag MC normalizations. Hence $\lambda S^{(\gamma)}=(\gamma\lambda)S^{(1)}+\lambda C(\gamma)$ and (a) follows; if an unrestricted finite maximizer of the actual curve exists, the maximizer set scales by $1/\gamma$ and the maximal value is unchanged. For (b), $z=rD/v$ becomes $\gamma z$ (and $\kappa z$ under $D\mapsto\kappa D$), which leaves the ratio \eqref{eq:contrast} and fixed clipping endpoints unchanged; correlation coefficients of nonconstant series are scale-free. (c) combines (a) with \eqref{eq:loge-bound}.

\paragraph{Floored scores.} With the floor $\ell_{\min}=\log10^{-12}$ write $L_{\gamma,c}(a)=C_{\gamma,c}-\gamma R_c(a)$, $C_{\gamma,c}=-\tfrac12\log(2\pi v_c/\gamma)$, $R_c(a)=(Y_c-\mu_c(a))^{2}/(2v_c)$, and $d_\gamma(a)=\sum_c\{[\ell_{\min}-L_{\gamma,c}(a)]_+-\gamma[\ell_{\min}-L_{1,c}(a)]_+\}$. Then $S^{\mathrm{floor}}_\gamma(a)=\gamma S^{\mathrm{floor}}_1(a)+C_\gamma+d_\gamma(a)$ with $C_\gamma=\sum_c(C_{\gamma,c}-\gamma C_{1,c})$ independent of $a$, and bounding the weighted exponential average of $d_\gamma$ between its extremes gives $|\log e^{\mathrm{floor}}_\gamma(\lambda)-\log e^{\mathrm{floor}}_1(\gamma\lambda)|\le\lambda\operatorname{osc}(d_\gamma)$, the oscillation taken over the relevant support including the observed block (or over the observed-inclusive MC bag); sum along a component. The bound reduces to (a) when no floor is active at either scale. At each site the floored score range is at most $\max(0,C_{\gamma,c}-\ell_{\min})$. For a finite candidate set containing the observed block in which every candidate has a nonzero residual at every scored site, all site scores eventually reach the floor together and the factor equals one for all large $\gamma$; this rules out an unconditional collapse linear in $\gamma$ for floored scores, but it does not give a universal limit $e\to1$ (a candidate with a zero residual does not floor).

\subsection{Tempering mixture (Proposition~\ref{prop:mixture})}\label{app:proofs:mixture}

(a) is a fixed-weight average of e-values. (b) keeps one term of the sum. For (c), if $\lambda_{\mathrm{quad}}$ is a grid point use it; otherwise it lies strictly between adjacent grid points $\lambda_g<\lambda_{\mathrm{quad}}<\lambda_{g-1}$, and with $c=\lambda_g/\lambda_{\mathrm{quad}}\in(1/2,1)$, $f(c\lambda_{\mathrm{quad}})=(2c-c^{2})f(\lambda_{\mathrm{quad}})$ and $f(2c\lambda_{\mathrm{quad}})=(4c-4c^{2})f(\lambda_{\mathrm{quad}})$, whose maximum is at least $8/9$ of $f(\lambda_{\mathrm{quad}})$, with equality at $c=2/3$. If $\mathcal{M}/\mathcal{V}>\lambda_0$, then $\lambda_0\mathcal{V}<\mathcal{M}$ and $f(\lambda_0)\ge\lambda_0\mathcal{M}/2$. For (d), mixing and averaging over classes are both linear, so $E^{\mathrm{M}}=R^{-1}\sum_rE^{\mathrm{M}}_r$, and the logarithm of a sum of positive terms gives the bound; a class of fixed weight $w_r$ contributes $\log w_r$. Alternatively, if every classwise quadratic bound holds on the grid, Jensen gives $\log E(\lambda)\ge\lambda\bar{\mathcal{M}}-\lambda^{2}\bar{\mathcal{V}}/2$ with the class averages, to which (c) applies; summed $\mathcal{M}$ and $\mathcal{V}$ must not be used as if all origins formed one product. In the setting of Proposition~\ref{prop:blind}, the scaled proxies give the bound of Proposition~\ref{prop:growth} on $[0,\Lambda/\gamma]$, so (c) holds for every $\gamma\in[\lambda_{\mathrm{quad}}/\lambda_0,\lambda_{\mathrm{quad}}/\lambda_{G-1}]$ for which the comparison grid points lie in that domain (for instance when $\gamma\lambda_0\le\Lambda$). The null statements follow from (b) and Proposition~\ref{prop:growth} at $\lambda_{G-1}$.

\subsection{Rank calibrators (Proposition~\ref{prop:rank})}\label{app:proofs:rank}

Jensen gives $c_g\le\log(N_{\mathrm{cand}}^{-1}\sum_kg(k))=0$, strictly unless $g$ is constant. For any function $h$ with range $r$, $|\mathbb{E}_ph-\mathbb{E}_uh|\le\mathrm{TV}(p,u)\,r$; apply it to $h=\log g$ and sum expectations over origins. \emph{Tie averaging.} If the observed score occupies the positions $J$ of the sorted bag, the implemented factor is $\bar g_J=|J|^{-1}\sum_{k\in J}g(k)$. For a fixed bag, assign each candidate label the average of $g$ over the positions of its tie block; these values sum to $\sum_kg(k)=N_{\mathrm{cand}}$, so conditional exchangeability of the observed label gives mean one regardless of ties, and products follow as in Proposition~\ref{prop:P1} or~\ref{prop:closedloop}. Conditioning on $\mathcal{F}_t$ and the unordered bag $\mathcal{B}$, with $h_{\mathcal{B}}(k)=\log\bar g_{J(k)}$, $c_{\mathcal{B}}=N_{\mathrm{cand}}^{-1}\sum_kh_{\mathcal{B}}(k)$ and $r_{\mathcal{B}}$ the range of $h_{\mathcal{B}}$, the same argument gives $\mathbb{E}[\log e^{\mathrm{R}}\mid\mathcal{F}_t,\mathcal{B}]\le c_{\mathcal{B}}+\mathrm{TV}(p_{\mathcal{B}},u)\,r_{\mathcal{B}}$, where $p_{\mathcal{B}}$ is the law of the observed position with uniform randomization inside its tie block; under the null $p_{\mathcal{B}}$ is uniform, and Jensen gives $c_g\le c_{\mathcal{B}}\le0$, with $c_{\mathcal{B}}=0$ when all candidates tie.

\subsection{Delayed feedback (Proposition~\ref{prop:closedloop})}\label{app:proofs:closedloop}

(b) Since $s-L\le t$ for every $s\in\{t+1,\dots,t+h\}$, every history input of the policies is fixed given $\mathcal{F}_t$, and recursion on $s$ makes each $A^{F}_{i,t}$ a function of that history and of unit~$i$'s own innovations $U_{i,t+1:t+h}$. Independence of these innovation vectors, and their independence of $\mathcal{F}_t$ and the potential outcomes, gives the product law of Assumption~\ref{ass:design-cl} given $\mathcal{F}_t$ and $\mathcal{P}$; because no policy reads another unit's actions inside the block, the same product holds for the tested set given the outside blocks. (a) Under $H_0^{G}$ the targets are functions of $\mathcal{P}$ and the pre-block actions, and under $H_0^{\mathrm{own}}$ with the conditional factorization the targets of the tested set are fixed given $\mathcal{C}^{I}_t$, as in \S\ref{app:proofs:P1}; conditioning on them leaves the law unchanged, and the denominator identity, Monte Carlo exchangeability and grid measurability arguments of \S\ref{app:proofs:P1} apply verbatim. Eligibility conditioning must first be shown to preserve the asserted conditional law; the open-loop path factorization of Proposition~\ref{prop:elig}(b) is not inferred from the delayed-policy construction. Two counterexamples in the review archive show the conditions are needed: when one unit's policy copies another unit's earlier action inside the block, the marginal factors each have mean one but their product has mean $1.64$; and when an outside policy carries a tested unit's within-block action into its target through a spillover channel, an autonomous tested policy is rejected with probability one under a true own-unit null.

\subsection{Over-betting (Proposition~\ref{prop:overbet})}\label{app:proofs:overbet}

(a) For fixed numbers $s_0, \dots, s_B$, the function $\kappa(\lambda) = \log\{(B+1)^{-1}\sum_b e^{\lambda s_b}\}$ is convex by Hölder's inequality, and $\kappa(0) = 0$. Each summand $\lambda S_t(0) - \kappa_t(\lambda)$ is therefore concave and zero at zero, and so is their sum. (b) By concavity, $L(\lambda) = L\{(\lambda/\lambda')\lambda' + (1 - \lambda/\lambda')\cdot 0\} \ge (\lambda/\lambda')L(\lambda') + (1 - \lambda/\lambda')L(0)$. (c) Apply (b) to the pair $\lambda \le \lambda''$ and rearrange; the second statement is (b) with $L(\lambda') \ge 0$. (d) Some grid point satisfies $\lambda^\ast/2 < \lambda_g \le \lambda^\ast$, so by (b) $L(\lambda_g) \ge L(\lambda^\ast)/2$; then use Proposition~\ref{prop:mixture}(b). (e) Write $P_{g,t} = \prod_{s \le t} e_s(\lambda_g)$ and $P_{g,0} = 1$. Then $\sum_g P_{g,t} = \sum_g P_{g,t-1}\,e_t(\lambda_g) = \big(\sum_g P_{g,t-1}\big)\sum_g w_{g,t}\,e_t(\lambda_g)$, and the product over $t$ telescopes to $\sum_g P_{g,T} / G$. The scores are fixed numbers once computed, so per-site floors inside the score do not affect the argument.

\subsection{Numerical checks}\label{app:proofs:checks}

Two theory reviews (Appendix~\ref{app:osf}) recomputed by exact enumeration the exact values stated in Propositions~\ref{prop:P1b} and~\ref{prop:P2}, the regression cases of the specification (exact sizes of the naive and set-averaged products under mutual spillover, shared noise and a path graph; the design ceilings for $Q(\mathrm{ON})=0.01$ and for a fair bit with $B=20$), and the forward--backward recursion and exact sampler of Proposition~\ref{prop:elig}(b); the values agreed with those of the frozen documents. The first review also recomputed the exposure-clock cases. Both checked the identity of Proposition~\ref{prop:kish}, which has no frozen counterpart, on random arrays. For Propositions~\ref{prop:growth}--\ref{prop:closedloop}, scripts checked the inequalities on random finite designs, the $8/9$ grid constant, the rank constants and a closed-loop toy, and the independent review added ten counterexamples (heavy tails, shrinking competitor mass, floors, residue classes, null paths, ties, cross-unit copying, spillover feedback, grid coverage and sampled denominators), each of which the final statements respect; these checks are not exhaustive. The scripts and their outputs are archived with their hashes (Table~\ref{tab:e5-reviews}).

%% file: appendix/B_v5_proofs.tex
\subsection{Proofs of Propositions~\ref{prop:stake} to~\ref{prop:information}}\label{app:proofs:v5-execution}

\subsubsection{Proof of Proposition~\ref{prop:stake}}
\emph{Validity.} Condition on the applicable field of Proposition 1, including the history, frozen information and the null-invariant target at $t+h$. Then $\mathbb E[1-f+f e_t(\lambda)\mid\text{that field}]=1$. The product within a residue class is a nonnegative martingale for the grid filtration of Proposition 1. A fixed convex average of the class products is an e-value at the analysis time.

\emph{(a)} $e_t(\lambda) \ge 0$, so each factor is at least $1 - f$.

\emph{(b)} For $u \ge -1/2$, $\log(1 + u) \ge u - u^2$. Indeed $g(u) = \log(1+u) - u + u^2$ has $g(0) = 0$ and $g'(u) = u(1 + 2u)/(1 + u)$, which is non-positive on $[-1/2, 0]$ and non-negative on $[0, \infty)$. With $u = f x_t$, where $x_t = e_t(\lambda) - 1 \ge -1$ and $f \le 1/2$, summing over the origins gives (b). $\square$

\emph{Remark.} Write $m$ and $v$ for the means of $x_t$ and $x_t^2$ over the origins. If $m > 0$, part (b) gives positive log wealth for $0 < f < \min\{1/2, m/v\}$, and the stake $f = m/(2v)$ gives $\log E(\lambda, f) \ge T m^2/(4v)$ when $m \le v$. This stake is a pathwise existence calculation: inferential validity requires the stake to be chosen from permitted information or included in a mixture fixed in advance, and optimizing it on the tested unit is not justified by this bound. At general horizons apply these inequalities to each residue-class product, with $T$ the number of factors in that class; if class sizes differ, part (a) gives $E(\lambda, f) \ge h^{-1} \sum_r (1-f)^{T_r}$.

\subsubsection{Proposition~\ref{prop:v5-stake-oracle} in full}
Setting, one origin. Condition on the completed past h and on any independently fitted, frozen score information. Let $Q_h$ be the design law and $K=B+1$. Represent the unordered candidate bag as $K$ labelled slots $a^{(0)}, \dots, a^{(B)}$ drawn independently from $Q_h$, with multiplicities retained, and let J be the uniformly selected recorded slot. This is the random-permutation representation of one recorded action and B independent decoys. Slot labels and any ordering of the remaining decoys carry no information about the execution mechanism. Write $L_j(y)$ for the true conditional density of the next observation if $a^{(j)}$ is executed, with respect to a common dominating measure, and define $m=\int L(y\mid h,a)\,Q_h(da)$ and $\bar L=K^{-1}\sum_jL_j(y)$. The relevant densities and denominators are assumed finite; formulas apply on their probability-one support. Ties and atoms are allowed: probabilities are assigned to slots, not to distinct action values.

Alternative $H_p$, $0<p\le 1$. Conditional on h, the current execution indicator has probability p, is independent of the current candidate draws, recorded-slot selection and environmental innovation, and selects $a^{(J)}$ when it is one. Otherwise the executed action is an independent fresh draw from $Q_h$ without using the bag, allowing repeated action values. The null comparison has a uniform recorded slot conditional on h, the labelled bag and y. In particular, the reference null $y\sim m$ independently of the bag has this property. An arbitrary verbal assertion that y does not reference the recorded action is insufficient unless the required joint conditional randomization property holds.

(a) The comparison class consists of nonnegative factors $g_J(h,\boldsymbol a,y,U)$ satisfying $K^{-1}\sum_j \mathbb E_U[g_j\mid h,\boldsymbol a,y]\le 1$ almost surely, where U is permitted auxiliary randomization independent of J conditional on $h$, $\boldsymbol a$ and $y$ under both hypotheses. This condition is necessary and sufficient for conditional e-validity given $h$, $\boldsymbol a$ and $y$ under the uniform-slot null. It is sufficient, but not necessary, for the weaker requirement $\mathbb E_0[g\mid h]\le 1$. Deterministic, permutation-equivariant factors have the familiar position-average condition. Dependence on completed history is allowed; dependence on future observations or unavailable future history is not allowed in a sequential factor.

(b) Under $H_p$ the conditional probability of recorded slot j is

\[\pi_j=\frac{(1-p)m+pL_j(y)}{K\{(1-p)m+p\bar L\}}.\]

Consequently its likelihood ratio against the uniform-slot null is $R_j=K\pi_j$. When $\bar L>0$, put $e^\ast_j=L_j/\bar L$ and $w=p\bar L/\{(1-p)m+p\bar L\}$. Then $R_j=1-w+we^\ast_j$. Here w is the posterior probability of the executed branch given h, the bag and y. The non-executed branch itself leaves J uniform, but conditioning changes its mixture weight. In general w is not p. The expression $1-p+pe^\ast_j$ is exact when $\bar L=m$, or $p=1$, and is also harmless in the neutral case where all $e^\ast_j$ equal one. For $0<p<1$ with $m>0$ and $\bar L=0$, $R_j=1$; define $e^\ast_j=1$ and $w=0$ on this event. Zero-denominator events having zero alternative probability need no value beyond an arbitrary measurable convention.

(c) For the comparison class in (a), conditional expected log growth satisfies $\mathbb E_{\pi,U}[\log g_J\mid h,\boldsymbol a,y]\le\sum_j\pi_j\log(K\pi_j)$, attained by $g_j=R_j$. This is the log-score/KL inequality. With positive $\pi_j$, equality holds if and only if $g_j=R_j$ for every slot and almost every U. More generally equality is understood on the alternative support, with $0\log0=0$ and $\log 0=-\infty$ where the alternative has positive probability. Integrating gives the corresponding unconditional inequality. Thus the finite-bag oracle optimum is R, using the bag- and observation-dependent w, not generally the fixed stake p. Even the best constant stake in $1-f+fe^\ast$ need not be p. For $0<p<1$ on the stated support, the unstaked oracle $e^\ast$ is strictly suboptimal whenever it differs from R on a set of positive alternative probability; it is neutral when all candidate likelihoods coincide.

(d) At each attempted origin t, apply (a)--(c) conditional on the completed past $h_t$, using the current conditional execution probability $p_t$ and current conditional kernel and design law. The same local conditional randomization and execution assumptions must hold at every origin. A fixed p and mutually independent indicators are not required: $p_t$ may depend predictably on $h_t$. Mere marginal execution probabilities or marginal independence statements do not replace these conditional assumptions. Products of the local factors $R_t$ are null e-processes. For a fixed cap and a native stopping time, include the transition that causes termination and use factor one thereafter; do not select away factors according to the current outcome. Among products of factors in the local comparison classes, evaluated on the same externally generated histories and the same bounded stopping rule, the product of $R_t$ maximizes expected log wealth by the conditional inequality and the tower property. Wagers must not alter that history or stopping rule. This is a sequential conditional-validity and log-growth statement. It is not a claim that this product is the likelihood ratio conditional on all of the unit's bags and observations at once: future observations, histories and stopping information can reveal earlier recorded positions. Such a whole-unit conditional-likelihood claim needs an additional joint factorization argument and is not asserted here.

(e) Scope and the different likelihood ratio. For the full joint one-origin experiment against the reference null $y\sim m$, the unnormalized oracle likelihood ratio $\ell=L_J/m$ does satisfy $dP_p/dP_0=1-p+p\ell$. This is distinct from the finite-bag-normalized $e^\ast$. The full joint ratio generally does not satisfy the pointwise bag-average constraint in (a). None of these results identifies p with the optimizer of a temperature-and-stake fit using a misspecified model score. A finite validation sample, non-identification, the prescribed numerical search, and changes of scorer calibration across conditions also affect that fit. Whether the stored fitted stake is near p is a descriptive empirical question, not a conclusion of this proposition.

Relation to known results. Mixture likelihood ratios, Bayes conditioning of mixture weights, uniform smoothing of a classifier, and log-optimal likelihood-ratio betting under a simple null are standard results. The log-growth inequality is the standard KL/log-score argument and has the usual numeraire interpretation. The corrected finite-bag calculation is an application of these results to this conditional-randomization experiment, not a new general theorem identifying a fitted stake with the execution rate. The potentially useful contribution is the design-specific exposition and the descriptive behavior of the frozen fits. No claim of literature priority is made on the basis of an offline review.

The large-bag limit and the distinction between this oracle and fitted scores are described in \S\ref{sec:v5-estimates}.

\emph{Remark (what it does not say).} The proposition concerns the factor computed from the system's true densities. With a model-supplied score and a fitted temperature, $e_t(\lambda)$ is not $e^\ast$, and nothing here identifies the stake returned by TS with $p$.

\subsubsection{Proof of Proposition~\ref{prop:information}}
\emph{Proof.} Write $K = B + 1$. Since $e_t \le K$, the expectation on the left is well defined in $[-\infty, \log K]$. If $\mathcal I_t(h) = \infty$ there is nothing to prove. Assume it is finite. Then $P$ has a density $r(a, y)$ with respect to $q \otimes P_Y$, and $\mathcal I_t(h) = \mathbb E[\log r(A_t, Y_{t+1})]$.

For a tuple $a = (a^{(0)}, \dots, a^{(B)})$ and an observation $y$ put $\pi_b = e^{T(a^{(b)})} / \sum_c e^{T(a^{(c)})}$, where $T = \lambda S_t$ is a fixed function of the candidate given $y$, $h$ and $\mathcal D$, and $\rho_b = r(a^{(b)}, y) / \sum_c r(a^{(c)}, y)$. On the set where the denominator defining $\rho$ is zero, define $\rho$ arbitrarily; this set has probability zero under $\bar P$ defined below. Then $e_t = K \pi_0$.

Let $J$ be uniform on $\{0, \dots, B\}$ and consider the experiment in which position $J$ of the tuple holds the recorded action and the other positions hold independent draws from $q$. Let $P_b$ be the law of $(y, a)$ given $J = b$, let $\bar P = K^{-1} \sum_b P_b$, and let $R = P_Y \otimes q^{\otimes K}$. The density of $P_b$ with respect to $R$ is $r(a^{(b)}, y)$, so the density of $\bar P$ is $Z = K^{-1} \sum_c r(a^{(c)}, y)$, and given $(y, a)$ the probability that $J = b$ is $\rho_b$. Because the same score is applied to every position, $\mathbb E[\log \pi_0]$ in the original experiment equals $\mathbb E[\log \pi_J]$ in this one. Given $(y, a)$, $\pi$ is a probability vector and $J$ has law $\rho$, so Gibbs' inequality gives $\mathbb E[\log \pi_J] \le \mathbb E[\log \rho_J]$. Hence
\[\mathbb E[\log e_t] \;\le\; \mathbb E[\log (K \rho_J)] \;=\; \mathbb E[\log r(a^{(J)}, y)] - \mathbb E_{\bar P}[\log Z].\]
The first term is $\mathcal I_t(h)$, because $(a^{(J)}, y)$ has law $P$. The second is $D_{\rm KL}(\bar P \,\|\, R) \ge 0$. The split is legitimate because both terms are finite: relative entropy is convex, so $D_{\rm KL}(\bar P \,\|\, R) \le K^{-1} \sum_b D_{\rm KL}(P_b \,\|\, R) = \mathcal I_t(h) < \infty$. Therefore $\mathbb E[\log e_t] \le \mathcal I_t(h) - D_{\rm KL}(\bar P \,\|\, R) \le \mathcal I_t(h)$. $\square$

\paragraph{Remark.} Write $\mathcal I_t(h, d) = D_{\rm KL}(P_{A_t, Y_{t+1} \mid h, d} \,\|\, Q_t(h) \otimes P_{Y_{t+1} \mid h, d})$. Averaging over histories conditionally on $\mathcal D$ gives $\mathbb E[\log e_t \mid \mathcal D] \le \mathbb E[\mathcal I_t(\mathcal F_t, \mathcal D) \mid \mathcal D]$ almost surely. When the conditional action law equals $Q_t(h)$, averaging also over $\mathcal D$ gives $\mathbb E[\log e_t] \le I(A_t; Y_{t+1} \mid \mathcal F_t, \mathcal D)$, the usual scalar conditional mutual information. The conditioning on $\mathcal D$ matters: if $A$ and $U$ are independent fair bits, $\mathcal D = U$ and $Y = A \oplus U$, then $I(A; Y) = 0$ but a score that uses $\mathcal D$ has positive expected log e-value. The reported means $G$ are sample summaries of these log e-values; they are not lower confidence bounds on information.

%% file: appendix/C_assignment_laws.tex
\section{Design laws}\label{app:assignment}

Every validity statement of the paper is conditional on a working design law. This appendix states each law exactly, reports how it was checked against the released command records, and lists what remains unverified. The defining documents were committed before the M4 registration; their hashes are in Appendix~\ref{app:osf}.

\subsection{ERK}\label{app:assignment:erk}

\paragraph{Laws.} In each field the light program is a train of pulses whose inter-pulse intervals were randomized by the acquisition protocol. All working laws hold the first and last pulse and the leading and trailing zero-command periods at each field's observed values. In fields with a rest of \DesignLawErkRestIntervalMinutes{} minutes between pulses, the variants differ in how they treat it: V1 fixes the rest's rank among the intervals; V2 inserts the rest after the first pulse at or beyond a threshold time taken from a stated range, sampling full arrangements uniformly by rejection; and V3, the law used for every primary analysis, shuffles the non-rest intervals before and after the rest separately. Fields without a rest reduce, under every variant, to a uniform permutation of the multiset of intervals. Counts of arrangements are counts of distinct paths, so repeated gap lengths are not double-counted. At an origin, the conditional sampler keeps the completed intervals, conditions on the time elapsed since the last pulse by rejecting a too-short ongoing gap, and returns the future pulse train. Fields run under a binary encoding policy are deterministic and contribute $e\equiv1$.

\paragraph{Checks.} All field-by-variant combinations passed unconditional checks of multiset, rank and segment structure, and simulated combinations passed conditional checks of the future train at several origins. Before any outcome was used, each field's observed interval sequence was compared with uniform permutations of its own non-rest multiset (\DesignLawErkPermutationCount{} permutations, fixed seed) on three statistics: the Spearman correlation of interval length with position, the number of runs up and down, and the lag-one autocorrelation of intervals. A field was to be flagged if any statistic had two-sided $p<0.01/39$, a Bonferroni correction over thirteen fields and three statistics. No field was flagged; the smallest per-field p-value was \DesignLawErkMinPerFieldP{}, and the Fisher-combined p-values were \DesignLawErkFisherCombinedSpearmanP, \DesignLawErkFisherCombinedRunsP{} and \DesignLawErkFisherCombinedLagOneP{} for the three statistics. The same procedure gated the M5 field (\S\ref{sec:real:m5}).

\paragraph{What remains unverified.} The acquisition protocol's finite list of interval values, the uniformity of its shuffle, the rule that triggers the rest and the treatment of the series' edges and time budget were not established from released code. An exchangeability check cannot certify an exact randomization kernel; it can only fail to find evidence against the working law.

\subsection{CcaS/CcaR}\label{app:assignment:ccas}

\paragraph{The source law.} The released acquisition code draws, for each lineage, a pair of bounds and a hidden integer that starts at the midpoint and moves up or down by one with equal probability, clipped at the bounds. For the first thirty-six steps the command is forced off while the hidden walk keeps evolving; afterwards the command is on whenever the hidden value is non-negative. The exact forward filter of this law matched an independent brute-force enumeration to rounding error.

\paragraph{The source law fails, and so does a fitted alternative.} Under goodness-of-fit criteria committed before any statistic was computed, which compare on- and off-run-length distributions and lag correlations of the released commands with simulations from the law, the source law was rejected: every one of the seven dates failed at least the off-run total-variation criterion. A fitted order-$k$ Markov model of the commands (K4) also failed the frozen gate on two dates, so no fitted kernel was admissible.

\paragraph{The working law D1 and when it was found.} A diagnostic recorded during that check showed that the released commands match the source law delayed by one step, consistent with a one-step shift introduced at data export. Under the delayed law (D1), the run-length total-variation distances were at most \DesignLawCcasDOneAprilOneThreeOnRunTV{} across dates and run types, against a frozen threshold of \DesignLawGofRunTVThreshold{}. The delay was not one of the candidates evaluated in advance, and its documentation does not overturn the rejection of the source law or establish the actual export transformation; D1 is therefore a working law, and every CcaS result in the paper is conditional on it. Its first numerical record is dated 2026-10-06, and the D1 implementation was committed the same day, both before the M4 registration on 2026-10-08; it was thus found after the first design-law check failed, but before any registered analysis.

\paragraph{Fixed templates and eligibility.} On each date, most lineages follow the random walk, but some repeat one of two fixed command templates; across dates, \DesignLawCcasFixedTemplateRowsPooled{} rows belong to \DesignLawCcasTemplateGroupsPooled{} template groups. The templates were identified as the command paths duplicated more than ten times on a date, and frozen by hash before calibration. Template rows have zero design entropy and contribute $e\equiv1$; the random-walk rows are tested under D1 conditioned on avoiding both templates of their date, computed exactly by the forward--backward recursion of Proposition~\ref{prop:elig}(b). That conditioning is exact only if the templates are genuinely fixed designs; because they were inferred from released duplicates rather than released by the experimenters, this is an assumption.

\paragraph{Neighbour exposure.} The released neighbour-exposure covariate is aligned one step ahead of the neighbours' commands and would let a score use other lineages' future actions. Every CcaS score uses its lag instead (\S\ref{sec:setup:nulls}); the sensitivity analysis restricted to lineages whose lagged exposure could be verified from commands is reported in \S\ref{sec:real:m4}.

\subsection{Status of the working laws}\label{app:assignment:status}

Proposition~\ref{prop:P3} would bound the cost of a wrong working law if the law had been fixed independently of the tested commands and its distance from the true law were known. Neither holds here: the fitted law was estimated from the released commands, and D1 was chosen after the source law failed. We therefore claim no theorem-level guarantee for the working laws and rely on calibration with the released command records (\S\ref{sec:synthetic:calib}), which bounds size empirically under the commands that were actually issued.

%% file: appendix/D_tables_disclosures.tex
\section{Tables and disclosures}\label{app:disclosures}

\subsection{Synthetic validation tables}\label{app:disclosures:tables}

Table~\ref{tab:d1-dev} gives the development grid of \S\ref{sec:synthetic:dev} by family, test and baseline; Table~\ref{tab:d2-m3} the sealed challenge of \S\ref{sec:synthetic:m3} by family, with the family descriptions of the challenge report; and Table~\ref{tab:d3-fix} the selection grid of \S\ref{sec:pooling:fix} cell by cell. Each table is generated from the committed result files named in its caption.

\begin{figure}[t]
\centering
\includegraphics[width=\linewidth]{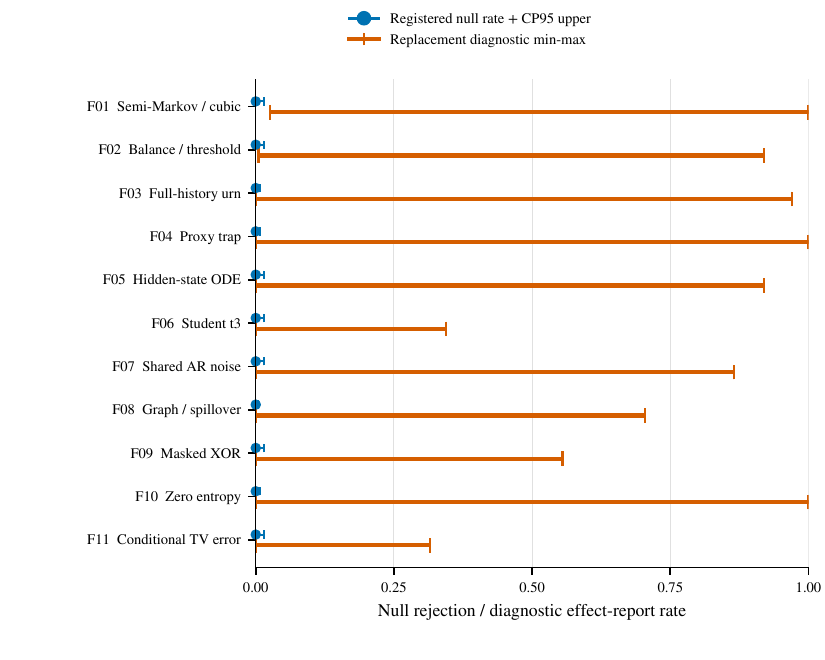}
\caption{Sealed challenge by family. For each of the eleven families, the null rejection rate of the pooled statistic registered for M4 with its one-sided $95\%$ Clopper--Pearson upper bound, and the range of effect-report rates of the replacement diagnostics across the family's null cells and models. Family names are abbreviated from the challenge report (Table~\ref{tab:d2-m3}).}
\label{fig:m3diag}
\end{figure}

\input{appendix/tables/D1_dev_grid}
\input{appendix/tables/D2_m3_families}
\input{appendix/tables/D3_fix_cells}

\subsection{Corrections to registered texts}\label{app:disclosures:corrections}

Registered documents cannot be edited. Where a registered text is inaccurate or ambiguous, we correct it here.
\begin{enumerate}
\item \emph{Selection of the repaired statistic.} The M5 registration states that the statistic was selected on synthetic data under a pre-committed rule. The rule was committed in advance, but it did not specify how to break a tie between equally hard cells, and the script that resolved the tie by grid order was first committed together with the selection result. Kish tempering would have been selected under most alternative tie resolutions (\S\ref{sec:pooling:fix}, Table~\ref{tab:d3-fix}).
\item \emph{Status line of the M5 plan.} The registered plan is byte-identical to the file committed before the registration, whose header still reads that it is a draft and not registered. Its registered status is established by the OSF record and by the registration addendum, which states that it supplements the byte-identical plan (Appendix~\ref{app:osf}).
\item \emph{Name of the M4 operational plan.} The frozen specification refers to the operational plan as \path{M4_ANALYSIS_PLAN_FINAL.md}; the registered, hash-bound plan is \path{M4_ANALYSIS_PLAN_v6.md}, and the registration's summary field stated this correspondence at registration. An older file named \path{M4_ANALYSIS_PLAN_FINAL.md} exists with different content and is a superseded predecessor.
\item \emph{Public versus embargoed registration.} The M4 registration text and plan call for a public registration before training; zxar3 and qezbx were created with an embargo, which the governing specification permits. Both embargoes were lifted on 28 October 2026, before this paper was posted.
\item \emph{Development settings.} The development simulations used Monte Carlo e-values with $B=63$ and horizon three; the frozen method uses $B=127$.
\item \emph{Rehearsal threshold.} The single-field rehearsal for M5 used the threshold $E\ge40$ of an unregistered two-hypothesis draft, not the registered $E\ge20$ (\S\ref{sec:synthetic:p3}).
\end{enumerate}

\subsection{Disclosures by stage}\label{app:disclosures:stages}

\paragraph{Before M4.} M4 is a registered re-analysis of data analyzed earlier in BioWM-Eval (Appendix~\ref{app:biowm-eval}), not a blind test. During the power pilot, a documentation search incidentally displayed earlier aggregate results; no setting changed on that basis. The designer of the sealed challenge had seen the source of the reference models and one correction to the challenge design. The CcaS working law D1 was adopted after the source law failed its goodness-of-fit check, before the registration (Appendix~\ref{app:assignment}); every CcaS rejection is conditional on it. The own-channel eligibility bound is vacuous. Observation-process assumptions (finite masks, segmentation) are stated, not tested. Power is low for weak or masked effects, and the e-value tests are conservative. The variance floor of the published CcaS model was active. The pilot gave each synthetic ERK field one trajectory, and the end-to-end rehearsal used a simple generator; neither could reveal pooling amplification.

\paragraph{M4 and its aftermath.} The M4 registration was submitted, and its content fixed, before any outcome was opened. OSF's activity log for the associated project, public since the embargoes were lifted, records its approval about two days after submission, after the M4 training and test outcomes had been accessed (Table~\ref{tab:e2-m4-timeline}); the log does not show whether the approval was given by hand or by OSF's automatic approval, which falls due 48 hours after submission. Until approval the registration could have been cancelled; it was not, and its content is fixed by the hashes in Table~\ref{tab:e1-registrations}. Neither ERK positive control was rejected in the registered analysis. The diagnosis of pooling amplification, the selection of the repair and the exploratory analysis X all followed that failure; X was run on test fields opened in M4. The dated note that fixes interpretation rules for the positive controls was written after the M4 report had been read, and its rules only restrict claims. Two operational scripts (a launcher and a report formatter) were added during the test stage and are not in the freeze manifest; their hashes are in Appendix~\ref{app:osf}. M4 did not save per-field, per-factor records, so the decomposition of its log e-values by field is not available. The repaired statistic's correlation parameter was estimated on training fields only.

\paragraph{M5.} The M5 registration followed X; the plan had been committed before X and was not changed afterwards, and the registration disclosed the X results. The confirmatory field's thirty-minute intervals lie outside the ten-to-twenty-five-minute training range. The single-field rehearsal used the linear-Gaussian model only. The command-law gate inherited the multi-field threshold of M1, which is lenient for one field. Permission to retry the data download was added in the execution instruction and implemented in a separate input--output adapter; it was not used. Origins at which every cell was masked were not tallied separately. The acquisition law of the confirmatory field is unverified, so validity is conditional on V3. OSF's activity log records the approval of the M5 registration minutes after its submission and before the first data access (Table~\ref{tab:e3-m5-timeline}); a self-reported browser visit shortly after the logged approval is also archived. Five preparation files carry modification times later than X but are byte-identical to their earlier commits.

\paragraph{Pixel demonstration.} All results are simulated, from one simulator family and two model classes, with one training run per cell and a reduced compute budget. Kish tempering, registered for M5, did not repair pooling amplification there; contrast tempering, which restored power at the larger effect size, is exploratory and has no confirmatory test. Two expectations of the protocol were wrong: the oracle did not collapse under gain noise, and the fitted models also collapsed under offset noise. The trap experiment could not show the diagnostics' growing confidence with data, and its frozen negative control was uninformative; the informative negative control was added post hoc. Three dated deviation notes record a reclassified pre-freeze audit flag (blocked standard-library probes, no data access), a reconstructed counter in the negative-control code, and a refresh of version-control index metadata in another repository.

\paragraph{ERK training diagnostic.} Command records were copied and hashed before the access marker was created; they are design information, not outcomes. The access log merges internal hash reads into logical open events, and the new repository's version-control identity was read from an existing repository with non-modifying commands. The agreement between the two tempering factors is partly structural (Proposition~\ref{prop:kish}).

%% file: appendix/E_osf_hashes.tex
\section{Registrations and hashes}\label{app:osf}

This appendix records registrations, data access and review archives for the earlier studies. Appendix M and Appendices N–U give the fourth-study and exploratory records added in version 5. All times are UTC.

\paragraph{Registrations.} Table~\ref{tab:e1-registrations} lists the five OSF registrations: zxar3 (M4), qezbx (M5), ctfxv (the DIAMOND study), j75g2 (the IRIS and TD-MPC2 study) and 4wj7k (the robot-data study). All five are public. The registered component of 4wj7k (2ksxa) also holds the results note, the decision file, the beacon report and the record of the stop before the run. The registered component of j75g2 (7mx6p) also holds the results update, the decision file and the deviation note posted after the run.

\input{appendix/tables/E1_registrations}

\paragraph{Order of real-data access.} Tables~\ref{tab:e2-m4-timeline} and~\ref{tab:e3-m5-timeline} give the time-ordered events of the two registered analyses: registration, creation and consumption of the one-time marker, the first opening of any outcome member, the command-law gate and the first outcome statistic, together with the number of member reads. The exploratory analysis X and the ERK training diagnostic ran under separate authorizations and markers, also listed.

\input{appendix/tables/E2_m4_timeline}
\input{appendix/tables/E3_m5_timeline}

\paragraph{Repositories.} Table~\ref{tab:e4-repositories} lists each stage's repository, its final commit and the role of the stage; the operational scripts added during the M4 test stage are listed with their hashes in Table~\ref{tab:e2-m4-timeline}. The paper's own repository records, for every number in the text, the source file, commit, hash and key from which it was generated.

\input{appendix/tables/E4_repositories}

\paragraph{Reviews.} In the archives for the earlier studies, the theory of \S\ref{sec:theory} and Appendix~\ref{app:proofs} was checked in two AI-assisted review sessions with read-only access to the project, distinct from the drafting sessions and from each other: the first by Claude Code, the same model family as the drafting model, and the second by Codex, a different model family. According to the project records, the frozen theory documents had themselves been written in an earlier Codex session, so neither review is independent of all the authorship involved in the way a human statistician's would be. Every bibliography entry was checked in a separate Claude Code session, which retrieved each source and checked each attributed claim against its text. Both theory reviews recomputed the stated exact values by enumeration. Neither found an error in the frozen proofs under their stated premises, and the bibliography review found no nonexistent reference. Both theory reviews found statements in our drafts that were stronger than what had been proved; the second also found that the hypotheses added to Proposition~\ref{prop:P4} after the first review did not exclude a predictor window that reveals the stratum. The revised text narrows these statements; the review archives record each finding and how it was addressed (Table~\ref{tab:e5-reviews}). For the earlier theory extension, Propositions~\ref{prop:growth}--\ref{prop:closedloop}, which were drafted by Claude, and every sentence that uses them were checked by Codex in five further read-only sessions: an independent review with ten runnable counterexamples, a delta review of the revised draft, a check of the trap experiment's logging policy against Proposition~\ref{prop:closedloop}(b), a final check of the integrated text, and a delta check of its revisions. The final check also found that the duplicate-score rule of the baseline study does not by itself establish the exchangeability that the rank statistics need (Appendix~\ref{app:baselines}). The references added in the earlier theory extension were verified twice, in the drafting session and in a separate Claude Code session. The reviews check mathematics and citations; they do not certify the actual acquisition laws, the implementation of every named statistic, or the chronology of the archives.

\input{appendix/tables/E5_reviews}

%% file: appendix/F_erk_field_diag.tex
\section{ERK per-field diagnostics}\label{app:erk-field}

Table~\ref{tab:f1-erk-field} reports, for each training field, reference model and horizon, the quantities behind \S\ref{sec:real:diag}: the frozen median-pairwise residual correlation $\hat\rho$, the mean pairwise residual correlation, the correlation $\rho_c$ implied by the contrast-tempering factor, the two ratios $\rho_c/\hat\rho$ and $\rho_c/\rho_{\mathrm{mean}}$, and the number of validation cells per contributing origin (mean, minimum and maximum). Pooled rows sum the numerators and denominators of the contrast factor across fields, as the pre-declared protocol specifies. The diagnostic used training fields only and computed no e-value. Ratios below the pre-declared band in individual fields are retained; the reading rule concerns the pooled rows.

\input{appendix/tables/F1_erk_field_diag}

%% file: appendix/G_pixel_full_grid.tex
\section{Pixel demonstration: full grid}\label{app:pixel-grid}

The protocol, the dated predictions and every result file of the pixel demonstration are frozen and hashed (Appendix~\ref{app:osf}). Compute limits fixed the episode length at the reduced level of the protocol's pre-declared ladder, with sixteen test episodes per alternative cell, thirty-two per null cell of the main grid and sixty-four per trap cell; intervals for alternative cells are correspondingly wide. Every statistic is the average of the two learned models unless marked otherwise.

Table~\ref{tab:g1-null} gives the null cells of both grids: rejection counts and mean e-values for every statistic, the basis of the validity criterion. Table~\ref{tab:g2-alt} gives the alternative cells of the main grid by noise condition, frame size and effect size: median $\log E$ and rejection rate for the pooled, Kish-tempered, contrast-tempered, field-mean, site-average and fractional-Kelly statistics and for the oracles. Table~\ref{tab:g3-diag} gives the effect-report rates of the three common diagnostics in every null cell. Table~\ref{tab:g4-temper} gives the tempering quantities of every model and cell: the median-pairwise correlation, the Kish and contrast factors and the two design effects. For the oracles, whose residuals are independent pixel noise, both design effects are close to one and their ordering is not informative; the rows in which the ordering differs from that of the learned models are all oracle rows and are marked. Table~\ref{tab:g5-nc} gives the post hoc negative control with contrast tempering, by cell and model, next to the frozen negative control.

\input{appendix/tables/G1_null}
\input{appendix/tables/G2_alt}
\input{appendix/tables/G3_diag}
\input{appendix/tables/G4_temper}
\input{appendix/tables/G5_nc}

%% file: appendix/H_baselines.tex
\section{Baseline study}\label{app:baselines}

This appendix documents the exploratory baseline study of \S\ref{sec:pooling:baselines}. It used only simulated data and the saved artifacts of the pixel demonstration and of the selection grid of \S\ref{sec:pooling:fix}; no real data were copied, opened or used. It was not registered on OSF. Its protocol, with the statistics, the decision rules and six dated predictions, was committed to the study's repository with its hash recorded before any value of a new statistic was computed. Table~\ref{tab:h1-records} lists the commits and hashes of the protocol, its amendment and every result record cited below.

\subsection{Protocol}\label{app:baselines:protocol}

\paragraph{Statistics.} The study added five statistics to those of the original runs: the rank e-values R and R$_{1/2}$, the tempering mixture M over the grid $\lambda_g=2^{-g}$, $g=0,\dots,12$, validation-tuned tempering V (\S\ref{sec:method:robust}), and ORACLE\_R, the rank e-value of the oracle's pooled score, as a reference for what a rank statistic can reach. V was set for each cell and learned model from the ten validation episodes and the validation bags of $31$ draws per origin that the contrast factor uses; every selected factor was written to a per-cell file whose hash was logged before any test score was computed. V is not defined on the ERK-like grid, which has no contrast design. Every statistic, old and new, was computed on the same test episodes and the same Monte Carlo draws ($B=127$).

\paragraph{Ties.} In the pixel world the scores of the candidate blocks are computed in batched floating-point arithmetic, so identical action blocks in different columns of a bag can receive scores that differ in the last bits. For the rank statistics each column's score was therefore replaced, before ranking, by the score of the first column with the same action block; distinct blocks were tied only when those scores were exactly equal, and the factor averaged $g$ over tied positions. This makes the scores of duplicate blocks identical within a bag. Exact mean-one validity additionally requires the scored bag to be exchangeable under relabelling of the candidates, and the first-occurrence rule alone does not establish that when numerical errors depend on a candidate's position; we did not verify it for this run. The registered study of \S\ref{sec:published} computed one score per distinct action and does not rely on this rule. M and V use the scores unchanged, as contrast tempering does. On the ERK-like grid identical blocks already share one score.

\paragraph{Scenarios.} In the pixel world the study used the main grid of the demonstration: $24$ alternative cells with $16$ test episodes each and $12$ null cells with $32$, one episode being one decision unit, with rejection at $E\ge20$ for the two-model average, as frozen. The trap grid of \S\ref{sec:trap} was not part of the study. On the ERK-like grid it used the $12$ alternative cells with $16$ replicates each and the $2$ null cells with $32$, one replicate being one decision unit, with the grid's threshold $E\ge100$.

\paragraph{Decision rules.} The primary comparison was the larger effect in the pixel world, $192$ decisions. A new statistic \emph{matched} contrast tempering if its total rejections were at least the contrast-tempering total minus $19$ and it was at most $6$ of $16$ below contrast tempering in every cell; it was \emph{clearly weaker} if its total was at most the contrast-tempering total minus $48$, and \emph{intermediate} otherwise. At the smaller effect a statistic was clearly stronger at $19$ or more rejections above contrast tempering and clearly weaker at $10$ or more below. On the ERK-like grid a statistic with two or more alternative cells below $14$ of $16$ was \emph{weaker than Kish tempering}. A new statistic with two or more rejections in any null cell would have been stopped as a suspected implementation error. The protocol fixed, in order, four outcomes and what each would mean for this version of the paper: (1) R or M matches contrast tempering and is not weaker than Kish tempering on ERK-like data, in which case the repair is reframed around statistics robust to score magnitude and the better of R and M is recommended for new studies as an exploratory finding, its confirmatory use needing its own registration; (2) V matches but R and M are clearly weaker; (3) R, M and V are all clearly weaker; (4) anything else. In every case the role of Kish tempering in the registered analysis M5 was to stay unchanged.

\subsection{Recomputation and the reproduction gate}\label{app:baselines:gate}

The study reused the saved trained models, episodes and frozen parameters and reran only the test stage, and the validation scoring for V, on a different machine; nothing was retrained. Monte Carlo bags were regenerated from the frozen seed keys, and the convolutional model ran on a CPU with the frozen determinism settings. Before any new statistic was computed, a reproduction gate compared, for every test episode or replicate, the recomputed per-origin log factors and episode $\log E$ of every existing statistic with the saved records. A cell was EXACT if every difference was at most $10^{-9}\max(1,|x|)$; CLOSE if it was not exact but every rejection decision was identical and the episode $\log E$ differences were small; and FAIL otherwise, which would have stopped the study.

The gate compared \PoneGateCells{} cells, and in \PoneGateDecisionCellsIdentical{} of them every rejection decision was identical to the frozen record. As first written, the CLOSE tier bounded the relative difference of episode $\log E$ by $10^{-4}$. That is ill-posed for reference values near zero, and \PoneGateFailVoneCells{} pixel cells were classified FAIL only because their references were close to zero. An amendment, written after these gate results and before any new statistic was computed, bounded instead the scaled difference $|\Delta|/\max(1,|\log E_{\mathrm{ref}}|)$ by $10^{-4}$, the scaling the EXACT tier already used. It changed no statistic, grid, threshold or prediction. Under the amended rule \PoneGateFailAmendedCells{} cells failed, and the largest scaled difference in any cell was \PoneGateMaxScaledDiffDisp. The differences arose only in arms that involve the convolutional model. The amendment also recorded the tie rule above and a mechanical reading of the decision rules, before any rank value was computed; the gate records keep both classifications (Table~\ref{tab:h1-records}).

\subsection{Results}\label{app:baselines:results}

Table~\ref{tab:h2-decisions} gives the mechanical reading of the decision rules and the status of each dated prediction; Tables~\ref{tab:h3-pixel} and~\ref{tab:h4-fix} give every cell. The rules returned outcome~\PoneOutcome, and \S\ref{sec:pooling:baselines} follows the reading that the protocol attached to it. The prediction that R would match contrast tempering failed; \S\ref{sec:pooling:baselines} discusses a possible mechanism suggested by Proposition~\ref{prop:rank}, whose baseline depends on tie handling and does not by itself explain the observed counts. The mixture became the primary statistic of the registered study of \S\ref{sec:published}, which is the confirmatory test; this study is not.

\input{appendix/tables/H1_records}
\input{appendix/tables/H2_decisions}
\input{appendix/tables/H3_pixel}
\input{appendix/tables/H4_fix}

%% file: appendix/I_published_model.tex
\section{Registered test on a published world model}\label{app:published}

This appendix documents the registered study of \S\ref{sec:published}: its registration and the order of events, the design in full, the two deviations and every result. Table~\ref{tab:i1-records} gives the identities, hashes and times on which the account rests; all times are UTC.

\subsection{Registration and order of events}\label{app:published:registration}

The protocol, its draft with the dated predictions, the pipeline and a pilot were committed to the study's repository before registration. The confirmatory code was then frozen in one commit, after unit tests on toy inputs and an exact replay of the pilot, and the protocol, identified by its hash, and the frozen commit were registered on OSF (ctfxv). A note filed with the registration states that the dates written in the protocol text, including the date attached to the predictions, are labels assigned in advance, and that the authoritative times are the commit times and the registration time in Table~\ref{tab:i1-records}; by those, the predictions were fixed in the draft protocol and carried unchanged into the registered version.

The confirmatory seeds could not exist before the registration. The protocol named a public randomness beacon, the drand mainnet chain of the League of Entropy~\citep{drand}, whose rounds are scheduled every $30$ seconds, and the round used was the first beacon round scheduled at least two minutes after the start of the minute in which the registration was recorded. Every seed of the confirmatory run was derived by hashing a fixed label together with a master value, itself the hash of the protocol's hash and the round's randomness. The runner recorded the round's signature and checked its format but did not verify it cryptographically; the record can be verified against the chain's public key. Generation and analysis ran once from the frozen commit, behind a one-time marker with an access log, apart from the two restarts described in \S\ref{app:published:deviations}.

The pilot used separate seeds and never entered the confirmatory analysis. It ran the full pipeline on null units only ($4$ validation and $8$ test units per game, $200$ steps), where the largest number of rejections by any statistic in any game was \PtwoPilotNullMaxRejections{} of $8$. On active pilot units it computed predictions, variances and the two estimated factors, but no e-value, rank, mixture, validation-tuned factor, p-value, rejection or replacement diagnostic.

\subsection{Design}\label{app:published:design}

\paragraph{Model and emulator.} DIAMOND's code was used at a fixed commit with its upstream files unchanged, and its released checkpoints for Boxing, Pong and Breakout at a fixed revision, with the hashes in Table~\ref{tab:i1-records}. The model card states no licence for the weights; they were used for research inference only and are not redistributed. Only the denoiser was used, with DIAMOND's three-step Euler sampler conditioned on the last four observations and actions. The emulator and preprocessing were those of DIAMOND's training: the NoFrameskip-v4 environments of ALE~0.9.0~\citep{bellemare2013arcade}, sticky-action probability zero, four frames per action with the maximum of the last two, $64\times64$ RGB frames by area interpolation, $1$ to $30$ no-op actions at reset and the minimal action sets of $18$, $6$ and $4$ actions. Everything ran on one laptop GPU, in an environment whose exact specification is part of the frozen commit.

\paragraph{Units and eligibility.} A unit is one segment: one reset, then up to $500$ agent steps, ending earlier at the first termination or truncation. A life loss does not reset the emulator but makes the origins around it ineligible: an origin $t$ is eligible if the observations $t-3,\dots,t$ and the transition to $t+1$ lie in one life. Each unit is tested on its own. A unit without eligible origins has $E=1$ for every statistic and counts as a non-rejection; Table~\ref{tab:i2-results} lists such units.

\paragraph{Prediction and scores.} The candidate at origin $t$ is the label $a_t$; the model conditions on the observations $t-3,\dots,t$ and on the recorded labels $a_{t-3},\dots,a_{t-1}$, also in the NULL condition, and the target is the real observation $Y_{t+1}$ ($h=1$, one residue class). The bag has $128$ columns: the recorded label and $127$ independent uniform draws. At each origin one Gaussian noise image was drawn and the model predicted once per distinct action in the bag, the validation contrast draws and the replacement draw, with that image; predictions were expanded back to the columns, so identical actions have identical predictions and scores. Predictions were stored as one of $256$ intensity levels, and a storage check stopped the run if a sampler output differed from its stored level by more than a set limit (\S\ref{app:published:deviations}). The per-site variance is the mean squared residual of the recorded-label prediction over all eligible origins of the validation units of the same game and condition, floored at $(2/255)^{2}$, and each site score is floored at $\log10^{-12}$.

\paragraph{Statistics.} The pooled score, the mixtures, the rank e-values and the rank tie rule are those of \S\ref{sec:method:robust} and Appendix~\ref{app:baselines}, computed with the study code of the baseline study, copied byte for byte, and $h=1$. Kish tempering uses the correlation estimate of \S\ref{sec:method:tempering} from the standardized validation residuals of the recorded-label prediction, clipped to $[0,1]$. Contrast tempering uses $31$ validation draws per origin and is clipped to $[1/n,1]$. Validation-tuned tempering maximizes the median validation $\log E$ over the $16$ validation units on the fifteen-value grid, ties going to the larger factor. The replacement diagnostic is, per unit, a one-sided one-sample t-test that replacing the recorded label by the replacement draw increases the mean squared prediction error, with an effect reported at $p<0.05$; a unit with fewer than two origins or with all differences zero has no p-value and reports no effect.

\paragraph{Criteria, predictions and interpretation.} The criteria are those of \S\ref{sec:published:design}. For validity the count governs; the accompanying one-sided Clopper--Pearson bound, written in the protocol as $0.095$, is $0.0951217$ for $2$ of $64$. The five dated predictions were: validity in all three games ($95\%$); collapse in at least two games ($75\%$); repair in at least two games ($60\%$); Kish tempering rejecting in fewer than $24$ of $32$ units in at least one game ($60\%$); and the rank e-value R rejecting less often than the mixture in at least two games ($70\%$). The protocol also fixed the interpretation: overall confirmation supports the claim that, on a published world model, pooling collapses the pooled test and the pre-specified mixture restores power while keeping validity; if collapse held but repair failed, the paper would say that tempering prevents collapse without guaranteeing power; and a validity failure would be treated as an implementation error, because validity holds by construction.

\subsection{Deviations}\label{app:published:deviations}

Two integrity checks of the frozen code stopped the confirmatory run. Each was documented in a dated note, published on OSF before the run resumed (Table~\ref{tab:i1-records}); neither changed generation, prediction, scoring, statistics, decision rules or gates, and both restarts reused the same beacon round and seeds.

\paragraph{Deviation 1: access log.} After the first unit, a null validation unit of Boxing, had been generated, stored and its hash verified, the runner's check of the access log's hash chain stopped the run. In two calls the unit's file hash had been passed under the key that the log also uses for each row's chain digest, so the digest overwrote it and could no longer be reproduced. No score or statistic had been computed for any confirmatory unit and no ACTIVE unit had been generated. The fix renamed the key, made the logging function refuse reserved keys, and added regression tests; the full test suite and the exact pilot replay passed. The aborted run was kept unchanged.

\paragraph{Deviation 2: storage tolerance.} The restarted run reproduced the first unit exactly and stopped after \PtwoDevTwoUnitsCompleted{} completed units, in an ACTIVE validation unit of Boxing, because one sampler output differed from its stored intensity level by \PtwoDevTwoMaxQuantDiffDisp, above the registered limit of two float32 epsilons. The sampler already rounds its output to an intensity level and returns it in float32, so the check measured float32 rounding only, while a level is unambiguous for any difference below $1/255$. The registered limit left too little margin: the largest difference in the pilot had been \PtwoPilotMaxQuantDiffDisp, and this should have been noticed before registration. The limit became $10^{-5}$, still far below half the spacing of the levels, so an output off the intensity lattice still stops the run; the change was one line, and no stored value changed. At this stop the runner had already analysed the Boxing NULL group, and ACTIVE validation units of Boxing had been generated; no ACTIVE test unit existed, and no score, e-value, rank, mixture, validation-tuned factor, p-value or rejection had been computed on any ACTIVE unit. The stopped run was kept unchanged.

\paragraph{Reruns.} The final run regenerated every unit of the stopped runs from the same seeds. Of \PtwoRerunUnitsCompared{} units compared array by array, \PtwoRerunUnitsIdentical{} were identical, and the results of the Boxing NULL group were reproduced exactly.

\subsection{Results}\label{app:published:results}

Table~\ref{tab:i2-results} gives every statistic in both conditions of each game, with the estimated and selected factors and the replacement diagnostic; Table~\ref{tab:i3-criteria} gives the criteria, the overall decision and the status of each prediction, as written by the analysis code. \S\ref{sec:published:results} summarizes them.

\subsection{A post hoc illustration of the factors}\label{app:published:illustration}

The reading at the end of \S\ref{sec:published:results} rests on Table~\ref{tab:i4-illustration}, a descriptive and post hoc calculation on a saved subset of the ACTIVE test units, the first four of each game, with no new simulation or sampling. For each unit the tempered $\log E$ was evaluated over the factors $2^{-k/4}$, $k=0,\dots,71$, with the saved observed-inclusive bags of $128$ scores, and the median of the four unit-wise maximizers is reported as the hindsight peak; this grid is used only here and is not a selection rule of the registered analysis. The margin column is the median across units of each unit's median positive maximum competitor margin in its bags. In this subset the validation-selected factor lies near the hindsight peak and the contrast factor is an order of magnitude above it, with larger margin penalties, which is consistent with a possible pooling-amplification mechanism. By the bound \eqref{eq:loge-bound}, a sampled competitor of multiplicity $k$ limits the log factor by $\log(128/k)$ minus the tempered margin, so a positive margin alone does not make the factor negative. The calculation does not isolate a cause, does not test the unfloored identity of Proposition~\ref{prop:blind}, and four units per game do not verify the counts of the full run.

\input{appendix/tables/I1_records}
\input{appendix/tables/I2_results}
\input{appendix/tables/I3_criteria}
\input{appendix/tables/I4_illustration}

%% file: appendix/J_two_models.tex
\begingroup\sloppy
\section{The IRIS and TD-MPC2 study}\label{app:p4}

\subsection{Registration and order of events}

Protocol v0 was dated 8 November 2026, with SHA-256 \texttt{5912af3d}\allowbreak \texttt{d3991d01}\allowbreak \texttt{90c54bf3}\allowbreak \texttt{8cfb155f}\allowbreak \texttt{958adc4c}\allowbreak \texttt{8a9330b4}\allowbreak \texttt{4e36ac3f}\allowbreak \texttt{f9cd9bb3}. Protocol v1, dated 8 November 2026, has SHA-256 \texttt{49c3857d}\allowbreak \texttt{329a04c7}\allowbreak \texttt{a0b366cd}\allowbreak \texttt{aa9d2a6f}\allowbreak \texttt{8aed6e0d}\allowbreak \texttt{18d87d85}\allowbreak \texttt{57b12adc}\allowbreak \texttt{25807d77}. Version 1 resolves the build pilot's open issues, freezes the game and task sets, adds the storage policy and the pilot record, and leaves the primary statistic, conditions, sample sizes, score definitions, criteria and predictions unchanged.

\input{appendix/tables/V3_J_timeline}
The beacon randomness was \texttt{6fe3f3aa}\allowbreak \texttt{56c76f8d}\allowbreak \texttt{5b18a0ee}\allowbreak \texttt{651b874a}\allowbreak \texttt{0eac91ef}\allowbreak \texttt{5b20acd5}\allowbreak \texttt{1730176c}\allowbreak \texttt{fe0698c0}. The master seed was \texttt{aaab1aa8}\allowbreak \texttt{e56ecac4}\allowbreak \texttt{df6b85e6}\allowbreak \texttt{f5fe74f4}\allowbreak \texttt{1f06f689}\allowbreak \texttt{abf8ff93}\allowbreak \texttt{96614302}\allowbreak \texttt{00a19207}. The beacon record verifies saved-response integrity, with no BLS signature verification.

CONF-STOP-01 preceded generation: the run instruction asked for a per-unit disk check that the protocol did not require. The instruction was withdrawn; no unit was generated in that attempt. CONF-STOP-02 occurred at 2026-11-09T13:13:26.240048Z, after all 5392 units and 97 group summaries were stored, because the decision writer could not serialize a NumPy boolean. DEV-P4-01 changed only conversion of NumPy scalar values to built-in types. Three serialization tests and 22 unchanged tests passed; seven tests were skipped, four because they replayed the historical pilot and three because they wrote toy archives or a synthetic audit artifact. Before and after finalization, all 14,015 manifested original inputs and five further frozen metadata inputs matched their hashes. An independent standard-library comparison reproduced all 1279 decision scalar values. No unit or scientific statistic was recomputed.

On 10 November 2026, between 02:58 and 03:04 UTC, the authors posted to component 7mx6p the results update P4\_\allowbreak{}RESULTS\_\allowbreak{}UPDATE\_\allowbreak{}20261109.md (SHA-256 \texttt{6a2c2127}\allowbreak \texttt{a254964c}\allowbreak \texttt{8dd55090}\allowbreak \texttt{97a8e509}\allowbreak \texttt{4b655709}\allowbreak \texttt{7370cafd}\allowbreak \texttt{244d2ea1}\allowbreak \texttt{59025ef6}), the decision file, the deviation note and the run summary, and made the component public. The times are those shown in OSF's file list, as recorded by the authors. The posted results update differs in wording from the draft OSF\_\allowbreak{}UPDATE\_\allowbreak{}P4.md kept with the run records.

\subsection{Design details}

The IRIS code revision was \texttt{24326aaa}\allowbreak \texttt{a283c527}\allowbreak \texttt{f42b89b4}\allowbreak \texttt{4cfdecf2}\allowbreak \texttt{665a7a16}, and its checkpoint revision was \texttt{c0539e99}\allowbreak \texttt{adf924d4}\allowbreak \texttt{4be0f19b}\allowbreak \texttt{12de0063}\allowbreak \texttt{9daa5550}. The TD-MPC2 code revision was \texttt{332f19e4}\allowbreak \texttt{aba4bffe}\allowbreak \texttt{2dc736f4}\allowbreak \texttt{5ea7f7bd}\allowbreak \texttt{c585903a}, and its checkpoint revision was \texttt{8fb2a82e}\allowbreak \texttt{fb3bae96}\allowbreak \texttt{941da440}\allowbreak \texttt{128fe133}\allowbreak \texttt{2e4394fd}. The MyoSuite wrapper patch, SHA-256 \texttt{b9b09b33}\allowbreak \texttt{7b02f205}\allowbreak \texttt{3beff265}\allowbreak \texttt{1c439850}\allowbreak \texttt{1bcdb9b9}\allowbreak \texttt{5725e2d0}\allowbreak \texttt{46b7f5d2}\allowbreak \texttt{ad598b51}, removes the read-only Timeout assignment and passes the seed to the simulator constructor; it changes no dynamics or assets.

\input{appendix/tables/V3_J_td_checkpoints}
\input{appendix/tables/V3_J_iris_checkpoints}
IRIS uses four emulator frames per action, 64 by 64 RGB frames resized with bilinear interpolation, and the maximum of the last two resized frames. Sticky actions are disabled, minimal action sets are used, and resets include 1--30 no-op actions. A life loss makes the affected origins ineligible without a pseudo-reset. An origin and its four-frame context must lie in one life. The token score is teacher-forced joint next-frame token likelihood over 16 tokens; the secondary decoded-pixel score uses the Gaussian over 12,288 sites in Boxing, Pong and Breakout. TD-MPC2 uses a Gaussian over 512 latent coordinates, with variances fixed on validation units and action repeat two for DeepMind Control Suite. Each unit ends at the first native termination, truncation or cap, retaining the terminal observation as its last target. The caps are 500 steps for the three DeepMind tasks, 100 for myo-pose and myo-reach, and 75 for myo-obj-hold.

Each condition uses 16 validation units, 32 test units under partial or full execution, and 64 under NULL. Bags have 127 design draws plus the observed action; validation contrast bags have 31 draws plus the observed action. Seeds are the first four big-endian bytes of the SHA-256 namespace consisting of the study identifier, phase, arm, task, condition, split, unit and stream. The confirmatory phase includes the master seed. PCG64 generates external non-noise streams; simulator reset seeds are passed to the frozen native simulator, and CUDA generators supply the pixel-sampling noise. Design draws are stored losslessly as float64 and converted to float32 at required model and simulator boundaries; identical converted inputs share predictions.

Validation units retain targets, observed-action predictions, the 31 contrast-draw predictions and replacement predictions. Test units retain all 128 bag-column scores in float64, the observed-action and replacement predictions, and targets. Other bag predictions are computed and scored but not stored; the token arm retains per-token target log probabilities of every distinct candidate. Decoded pixel means are clipped and rounded to nearest-even uint8, then dequantized by $2k/255-1$ for scoring. Gaussian validation summaries are fixed before test generation. The protocol stores no model weights in its registration package.

\subsection{Choice of the primary statistic}

The regime map applied its pre-committed rule and selected V. The worst-cell rates were 0.8125 for V and 0.78125 for W(3/4), within the rule's 1/32 tolerance; V won on mean rate. Amendment 1 left V and the weighted mixtures undefined for the saved ERK-like cells, because validation scores and residuals were missing. The later ERK validation-tuning check gave V a four-model log e-value of -101 at thirty minutes, with all four models negative, and two negative models at ten minutes; M gave 347 and 218 and W(3/4) gave 345 and 217. Protocol Section 2.1 registered W(3/4) before any confirmatory data existed. This departs from the regime map's protocol, which said that its selection rule would fix the primary statistic.

\subsection{Full results}

The following tables copy every rejection count, median log e-value and factor from the pinned result tables. Rounding affects presentation only.

\input{appendix/tables/V3_J_iris_0}
\input{appendix/tables/V3_J_iris_1}
\input{appendix/tables/V3_J_iris_2}
\input{appendix/tables/V3_J_pixels_0}
\input{appendix/tables/V3_J_pixels_1}
\input{appendix/tables/V3_J_pixels_2}
\input{appendix/tables/V3_J_tdmpc2_0}
\input{appendix/tables/V3_J_tdmpc2_1}
\input{appendix/tables/V3_J_tdmpc2_2}
\input{appendix/tables/V3_J_factors_iris}
\input{appendix/tables/V3_J_factors_pixels}
\input{appendix/tables/V3_J_factors_tdmpc2}

\subsection{Predictions}

\input{appendix/tables/V3_J_predictions}

\subsection{Replacement diagnostic}

\input{appendix/tables/V3_J_replacement}

\subsection{Compute}

The synchronized GPU-stage hours were 2.53421079 for IRIS tokens, 2.86836434 for the secondary decoded pixels, and 0.0810389767 for TD-MPC2. The secondary pixel units are shared with IRIS tokens and are not additional distinct units. Overall wall time was 8.27198099 hours. Other processes shared the GPU during the run; this affects timing only. GPU-event hours were not emitted by the frozen code.
\endgroup

%% file: appendix/K_exploratory.tex
\begingroup\sloppy
\section{Exploratory studies after the second version}\label{app:exploratory}

\subsection{Score-rule study}

The score-rule study was exploratory, under the protocol dated 7 November 2026, SHA-256 \texttt{3e772433}\allowbreak \texttt{628581b0}\allowbreak \texttt{d600a80f}\allowbreak \texttt{dbdce415}\allowbreak \texttt{3b8f0f6b}\allowbreak \texttt{3f6bf36f}\allowbreak \texttt{e459bd2d}\allowbreak \texttt{cfce3c79}. It rescored saved DIAMOND predictions, observations and bags without generating new units. The reproduction gate recovered all rejection decisions of the registered run before any alternative score rule. The ten alternatives were the two higher variance floors, one common variance, Student-t and Laplace densities, three block sizes, a 256-dimensional random projection and a rank-16-plus-diagonal joint Gaussian. Kish and contrast factors are undefined for the joint Gaussian. No scientific method deviation was recorded; an initial file-name discovery traversed broader directories, after which access was restricted to the authorized inputs.

\input{appendix/tables/V3_K_score}
\input{appendix/tables/V3_K_sitecount}
\input{appendix/tables/V3_K_score_predictions}

\subsection{Sampling study}

The sampling study used the same protocol, with four predictions averaged at each eligible origin, and was exploratory. Its reproduction gate recovered the registered validation unit bit for bit, including all bag, contrast and replacement predictions. The initial disk rule stopped the study before generation. A dated amendment replaced it with a 100 GiB rule; the reproduction and 20-origin timing gates passed before generation. All 288 planned units completed. The saved readout records 12.925915 GPU-event hours and 13.08265 hours elapsed wall time. The two predictions dated 7 November 2026 were that collapse persists (probability 0.65) and that M repairs it where it persists (0.80); both were supported.

\input{appendix/tables/V3_K_sampling}
\input{appendix/tables/V3_K_sampling_factors}

\subsection{Regime map}

The regime map was exploratory, under the protocol dated 8 November 2026, SHA-256 \texttt{f9ab5aee}\allowbreak \texttt{d2659270}\allowbreak \texttt{4b5a71b0}\allowbreak \texttt{daa6df7d}\allowbreak \texttt{f826a00a}\allowbreak \texttt{cdb64468}\allowbreak \texttt{daae241f}\allowbreak \texttt{4c078eeb}, and Amendment 1, SHA-256 \texttt{9ea8ec9b}\allowbreak \texttt{1d1566da}\allowbreak \texttt{23f7e3e7}\allowbreak \texttt{8e55313f}\allowbreak \texttt{96b48f7d}\allowbreak \texttt{b94f485b}\allowbreak \texttt{d9d67172}\allowbreak \texttt{6847e983}. The simulator and crossed grid are described in Section~\ref{sec:regime}. Oracle-only calibration fixed the weak, medium and strong effects at 0.04296875, 0.078125 and 0.140625, targeting median log e-values of 3, 10 and 30 and attaining 3.23815054017, 10.328241766 and 30.9656204875. Calibration stopped within ten percent of target; its values were committed before the other new statistics were computed.

Amendment 1 followed the initial stop on missing ERK-like validation quantities. It leaves V, contrast tempering and W undefined on those cells; uses the 15-value mixture for selection while reproducing the baseline study's native 13-value mixture in the gate; restricts decision-relevant cells to common defined candidates; and excludes candidates using only the simulator's null cells. It fixes 31 validation draws and the calibration stopping rule. The restarted baseline reproduction gate passed on all 51 cells, with 10,824 comparisons, zero log-e-value differences and zero rejection mismatches. The baseline study's pixel-world Kish factors and residual correlations were not saved; their existing C1 results were retained without back-calculating parameters.

\input{appendix/tables/V3_K_regime_R5}
\input{appendix/tables/V3_K_regime_R6}
Selection step 1 excluded no candidate: every simulator null maximum was zero. Step 2 gave the highest worst-cell rate, 0.8125, to V. Step 3 treated W(3/4), at 0.78125, as tied within 1/32; mean simulator rates 0.634837962963 and 0.625192901235 selected V. Step 4 was not needed. There were 98 common decision-relevant cells. Q2 found a paired median V-minus-M deficit of at least half log 15 in 28 of 54 weak-effect cells, a fraction of 0.518518518519.

\input{appendix/tables/V3_K_regime_predictions}
The main driver stopped on an analytic-reference lookup after all other 230 cells were saved. The corrected reference handler used the saved per-replicate residual correlations, reused C0 when the Kish factor equalled one, and completed the reference statistics; all earlier outputs stayed unchanged. Exact total main CPU time was not captured. The original stop records were preserved.

\subsection{ERK validation-tuning check}

Protocol v1, dated 8 November 2026, SHA-256 \texttt{f48d4ac7}\allowbreak \texttt{460886e1}\allowbreak \texttt{04720e05}\allowbreak \texttt{2d3b9d3c}\allowbreak \texttt{fb4b6c69}\allowbreak \texttt{5a861602}\allowbreak \texttt{8082c161}\allowbreak \texttt{bce67e79}, stopped because the frozen M4 assets were absent on the new machine. Version 2, SHA-256 \texttt{baf5090a}\allowbreak \texttt{bb2b4c45}\allowbreak \texttt{78056f42}\allowbreak \texttt{ba1efbee}\allowbreak \texttt{558f5d48}\allowbreak \texttt{62296573}\allowbreak \texttt{35b5c933}\allowbreak \texttt{7985fb2d}, ran on the old machine, defined validation bags and added weighted mixtures without changing the four predictions. The reproduction gate recomputed C0 at ten and thirty minutes within 0.01 of M4 and reproduced the rounded C1 values 349 and 218 from X.

\input{appendix/tables/V3_K_erk_10}
\input{appendix/tables/V3_K_erk_30}
\input{appendix/tables/V3_K_erk_factors}
The dated predictions and outcomes were: M reaches E at least 100 at both horizons (0.90), held; V does so (0.75), failed; R stays below 100 at ten minutes (0.65), failed; and M remains within log 15 of the best constant-grid member at both horizons (1.00), held. The gaps were 2.70805020020578 and 2.70802211364344, below log 15 = 2.70805020110221.

DEVIATION\_01 changed a new accumulator from float32 to explicit float64, after the first attempt had passed the reproduction gate. The deviation and fix were committed before reopening. A formatting-only readout step later converted completed-table booleans to native JSON; it reopened no field and recomputed no score or statistic. The cumulative log records 52 ERK file opens: 26 trajectory-member opens and 26 command-snapshot opens, including both attempts. There were zero action-registry opens, zero historical outcome or score-cache reads and no refitting.

\subsection{ERK field inventory}

The ERK field inventory used archive metadata and the projects' retained identity and access records, without opening a biological archive member. Declared TRAIN, TEST and M5 assignments were preserved. Remaining fields were classified OPENED-OR-UNKNOWN whenever their lifetime opening history was missing or ambiguous; absence of a matching record was not treated as proof of non-access. The result was 67 fields OPENED-OR-UNKNOWN and zero UNOPENED. During the inventory, a source preview showed at least one hard-coded historical ERK summary value; no outcome value was reproduced, and source previews were stopped. This does not affect M5, whose field was opened only under its registration.

\subsection{Weighted mixture on DIAMOND (post hoc)}

The registered run's saved per-unit records do not contain every unit's e-values at all 15 grid values, so the third version of this paper reported no value. The over-betting study recomputed them from the saved scores (\ref{app:a3}): W(3/4) rejected in 96 of 96 ACTIVE units and 0 of 192 NULL units. The score bags were recomputed from the saved predictions by the score-rule study's registered-rule scorer.

\subsection{Name map}

\input{appendix/tables/V3_K_names}
\subsection{Over-betting study}\label{app:a3}

The protocol is dated 9 November 2026, SHA-256 \texttt{40aa36ee}\allowbreak \texttt{493192be}\allowbreak \texttt{e2d6e07b}\allowbreak \texttt{e42ca955}\allowbreak \texttt{04826513}\allowbreak \texttt{8f80437c}\allowbreak \texttt{3916d897}\allowbreak \texttt{63f273e9}. Parts 1--3 computed temperature curves, the validation-weighted mixture on DIAMOND, and online tempering on saved units. The analysis is exploratory: every unit had already been generated and analyzed. IRIS and TD-MPC2 used stored 128-column score bags. DIAMOND score bags were recomputed from saved predictions by the score-rule study's registered-rule scorer. No new unit was generated. The reproduction gates made 40,730 comparisons on 4224 IRIS and TD-MPC2 test units and 384 DIAMOND validation and test units, with zero differences in log e-values and identical rejection decisions.

The fine grid is $2^{-k/4}$, $k=0,\dots,56$. Best-factor ties go to the smaller factor; break-even is the largest grid factor with positive log e-value, or undefined if none is positive. A collapse is $L(\lambda_f)<0$ with $L^*\ge\log20$; an over-bet has $\lambda_f>\lambda^\ast$. Checks (a), (b) and (c) passed in 4512 of 4512 units each. The original check (d) passed in 4430 and failed in 82. AMENDMENT\_01, dated 9 November 2026 and committed after these results, retained prediction 1 as contradicted and replaced (d) by (d1), the grid-member bound, and (d2), the half bound conditional on a nonnegative fine-grid maximum. The corrected checks passed in 4512 of 4512 and 3114 of 3114 applicable units. All 82 original failures had negative fine-grid maxima.

\input{appendix/tables/V4_K_checks}
\input{appendix/tables/V4_K_factors_diamond}
\input{appendix/tables/V4_K_factors_pixels}
\input{appendix/tables/V4_K_factors_tokens}
\input{appendix/tables/V4_K_factors_td}

The eligible group-factor pairs for prediction 2 were the twelve full-execution IRIS decoded-pixel or TD-MPC2 groups in which C1 or C1c rejected in fewer than 24 of 32 units while W(3/4) rejected in at least 24. The median factor-to-best ratio, over every unit of each eligible group, exceeded two in all twelve; its minimum was 6.2594373676413. The online-mixture identity passed in every unit, with largest absolute difference $5.00222085975\times10^{-12}$.

\input{appendix/tables/V4_K_online_diamond}
\input{appendix/tables/V4_K_online_pixels}
\input{appendix/tables/V4_K_online_tokens}
\input{appendix/tables/V4_K_online_td}
\input{appendix/tables/V4_K_predictions}

\subsection{ERK site-count check}\label{app:a3e}

Part E of the over-betting protocol used the nine M4 test fields and the four training fields at ten and thirty minutes, with the frozen models, bags, design law, masks, variances and aggregation. The first attempt stopped at its reproduction gate: six of eight comparisons failed, while V passed at both horizons. No adaptive statistic or prediction was evaluated. The cause was conversion of outcomes to float64 before normalization. The recovery removed that conversion and the explicit casts of the normalization constants; the pooled-score accumulator and subsequent log-factor accumulation retained float64. The fix and deviation were committed before reopening. RECOVERY\_01 is exploratory; the results are retained at commit 62998d8.

The recovery gate passed all eight comparisons: C0, C1, V and M at both horizons, within $\max(10^{-6},10^{-9}|L|)$. The selected $\rho$ maximizes the median validation-field log e-value per model and horizon, with ties to the larger $\rho$. The grid has 15 values $2^{-g}$, $g=0,\dots,14$. M$_\rho$ averages complete grid-member e-values; W$_\rho$(3/4) assigns 3/4 to the selected member and 1/56 to each other member. At zero-cell origins the original neutral-factor convention applies.

\input{appendix/tables/V4_K_rho_10}
\input{appendix/tables/V4_K_rho_30}
\input{appendix/tables/V4_K_rho_factors}

The predictions dated 9 November 2026 both held: prediction 7 (70\%) required the four-model V$_\rho$ e-value to reach 100 at both horizons, and prediction 8 (55\%) required positive V$_\rho$ log e-values for all four models at both horizons. The four-model values were 350.54383590664 and 206.07503526448; every per-model value was positive.

The recovery opened 26 ERK files: 13 outcome members and 13 command snapshots. Cumulative counts over the two attempts were 52 file opens, 26 outcome-member opens and 26 command-snapshot opens, over the same 13 fields. Historical outcome or score-cache reads, refits, network requests, installs, pushes, uploads and contacts were zero. No further recovery was authorized.

\endgroup

%% file: appendix/L_robot_data.tex
\begingroup\sloppy
\section{The robot-data study}\label{app:p5}

\subsection{Registration and order of events}

The feasibility check (commit 97ec197) opened images of official test trajectories 0--15 only, for a constructed-null smoke test. It established computational feasibility and a working law with unverified parts. No recorded-action/image score, e-value, rank, p-value, likelihood ratio or action-effect result was computed. Protocol v0 had SHA-256 \texttt{36f0fc6c}\allowbreak \texttt{c92a99e6}\allowbreak \texttt{327187e7}\allowbreak \texttt{a66a1a2d}\allowbreak \texttt{0d22eeb8}\allowbreak \texttt{c7f5bf44}\allowbreak \texttt{155a9c6e}\allowbreak \texttt{bd7b230f}; v1 had SHA-256 \texttt{db523561}\allowbreak \texttt{602736c7}\allowbreak \texttt{f0f3cea7}\allowbreak \texttt{82e77739}\allowbreak \texttt{26794d80}\allowbreak \texttt{307a2900}\allowbreak \texttt{345300c0}\allowbreak \texttt{7b128dfb}. Version 1 fixed the eligible set and count criteria, fixed the pool and disjoint stream draws, added NULL V and W(3/4) at the ACTIVE-selected factor, added the command-record gate, and stated training-pool and checkpoint-selection limits. The primary statistic, laws, origins, power criterion and six dated predictions were unchanged.

The build pilot (fec685d) ran the full NULL pipeline on 48 trajectories outside the study set, with sixteen validation and thirty-two test units. No ACTIVE e-value was computed. The command-record check compared 3346 study rows with 14,294 pool rows. Its four p-values were 0.20207805209385643, 0.3420444752342814, 0.16309206772899076 and 0.40300540275743, all above the gate's 0.001. Synthetic rehearsals covered confirmation true and false and interrupted resume without image decoding. The exact NULL pilot replay reproduced 294 files byte for byte. The frozen code commit was \texttt{68ad9b37}\allowbreak \texttt{75a3d3e1}\allowbreak \texttt{d8e53570}\allowbreak \texttt{b5f0f5aa}\allowbreak \texttt{ad2f7ce0}.

\input{appendix/tables/V4_L_timeline}

Registration 4wj7k was public with no embargo. The raw registration timestamp has no zone marker and is read as UTC. The beacon rule was the first beacon round scheduled at least two minutes after the start of the minute in which the registration was recorded. Round 6629230 supplied randomness \texttt{6a9930ca}\allowbreak \texttt{5dcc2195}\allowbreak \texttt{b3e5ce53}\allowbreak \texttt{0bfd144b}\allowbreak \texttt{49f3d0b6}\allowbreak \texttt{2577a697}\allowbreak \texttt{47fe52c4}\allowbreak \texttt{b135e962}. The master seed was \texttt{7901ecac}\allowbreak \texttt{4b253d54}\allowbreak \texttt{65f0db8f}\allowbreak \texttt{107ff7bc}\allowbreak \texttt{dd0d72be}\allowbreak \texttt{1747d112}\allowbreak \texttt{4c66f9a4}\allowbreak \texttt{cd8ac97d}, the SHA-256 of the ASCII protocol hash, a vertical bar and the randomness, without a newline.

CONF\_STOP\_01 occurred before any network request, permit change or study-image access. Our run instruction required an idle GPU and use of the frozen checker on confirmatory archives, while that checker supported synthetic rehearsals only. Nothing had been fetched, changed or opened. The later instruction resolved both requirements. The single run completed with no deviation, interruption or resume in 2.591607 hours. A separately written checker, after the results commit, reproduced 10,516 unit log e-values and all decision counts with zero disagreements. After the reports quoted here were written, the first author posted nine files to the registered component (2ksxa) on 10 November 2026 (OSF file times 22:37 UTC) and made the component public the same day: the results note \path{P5_RESULTS_UPDATE_20261110.md}, \path{P5_CONF_REPORT.md}, \path{SUMMARY_P5.tsv}, \path{DECISION.json}, \path{P5_BEACON_REPORT.md}, \path{REGISTERED_RUN_PERMIT.json}, \path{OUTPUT_MANIFEST.tsv}, \path{ACCESS_AUDIT.json} and \path{CONF_STOP_01.md}. The draft \path{OSF_UPDATE_P5.md} was not posted. The posting is reported by the author; no local record of it exists.

\input{appendix/tables/V4_L_posted_files.tex}

\subsection{Design details}

\input{appendix/tables/V4_L_data}

iVideoGPT code revision was \texttt{d601d5ca}\allowbreak \texttt{c9e96c6a}\allowbreak \texttt{a0c17cb3}\allowbreak \texttt{7ed6a7c7}\allowbreak \texttt{ca1fb210} and checkpoint revision \texttt{2863382a}\allowbreak \texttt{dad16f49}\allowbreak \texttt{a8a6bc8e}\allowbreak \texttt{61c08a03}\allowbreak \texttt{e98bc5e5}. Tokenizer and Transformer weights had SHA-256 \texttt{f29ac860}\allowbreak \texttt{b2ab8855}\allowbreak \texttt{2d504d94}\allowbreak \texttt{e1aac0c3}\allowbreak \texttt{9dcee45e}\allowbreak \texttt{a90856a7}\allowbreak \texttt{f44dc1a7}\allowbreak \texttt{e46dd1ec} and \texttt{c7ac81f4}\allowbreak \texttt{13f69788}\allowbreak \texttt{3a8f770d}\allowbreak \texttt{40d0b5b1}\allowbreak \texttt{09670d0b}\allowbreak \texttt{838d5cef}\allowbreak \texttt{35848778}\allowbreak \texttt{eeca7bb0}, respectively. The released action-conditioned checkpoint was unchanged. Eligibility required exactly equal commands at every even/odd frame pair. Of test trajectories 16--255, trajectory 182 was excluded, leaving 239; none was replaced. The split was 80 validation and 159 test trajectories, shared by both laws and arms. Each trajectory used origins $t=2,4,\dots,28$, with one context frame and the next frame as target.

The empirical pool contained 14,294 commands from 1021 eligible training trajectories; training trajectories 371, 29538 and 29627 were excluded. Its SHA-256 was \texttt{b08500e9}\allowbreak \texttt{b43e786e}\allowbreak \texttt{c852411b}\allowbreak \texttt{59fedd35}\allowbreak \texttt{ae19715f}\allowbreak \texttt{e01486f7}\allowbreak \texttt{ec916da9}\allowbreak \texttt{27bbfb9e}. The streams bag, tempc and repl each permuted the pool with its own seed, dropped entries used by preceding streams of the unit, and assigned consecutive blocks to origins. No pool entry was used twice within a unit, apart from shared column zero in the contrast bag. Disjointness was by identifier, not command value. L-par drew fresh candidates from the fitted law: independent planar uniforms on $(-0.07,0.07)$ and two discrete components with the recorded five-category probabilities. The fitted-law SHA-256 was \texttt{1fbfbd14}\allowbreak \texttt{256e0dc9}\allowbreak \texttt{06c36815}\allowbreak \texttt{d6655c43}\allowbreak \texttt{15af35e4}\allowbreak \texttt{18ed74eb}\allowbreak \texttt{11fda3c5}\allowbreak \texttt{aae6705b}.

The tokenizer used 256 context tokens, one separator and sixteen target tokens. Inference used bfloat16 weights, eager attention, deterministic algorithms, TF32 off and padded batches of eight. Log-softmax over the full vocabulary of 16,386 was float32; sixteen log-probabilities were cast to float64, floored at $\log10^{-12}$ and summed in float64. Commands identical at model input precision shared a score. Main bags had 128 columns; validation contrast bags had 32. Scores, site log-probabilities, commands, pool identifiers, aliases and token identifiers were stored; no images were copied.

Seeds were the first four big-endian bytes of SHA-256 of the ASCII namespace \path{P5|<PHASE>|<law>|<arm>|<split>|unit<NNNNN>|<stream>}; the split used \path{P5|<PHASE>|split}. PCG64 was used throughout; the confirmatory phase was \path{CONF|<master>}. Validation choices were locked before test-image access.

\subsection{Working assumptions and limits}

The recorded-command arm required three conditions: commands at origins are independent draws from one law independent of scene; under L-emp the pool's empirical law stands in for the law of the recorded command, which equal treatment of commands by the model does not establish, while L-par instead requires the fitted law to equal the conditional assignment law; and the model was fixed without the study units' image-command pairs. The collection software and configuration of 2017 could not be established. The pool commands were from training trajectories, and use of the official test set for checkpoint selection cannot be excluded.

Conditional on the realised training pool, L-emp decoys follow the pool's empirical law, with the remaining entries changing as draws are used. Even if the commands were independent draws from one scene-independent population, a fresh recorded command need not have that empirical conditional law, and equal treatment by the frozen scorer would not establish exchangeability. Design-based validity for the recorded-command arm would require a conditional-law argument covering the frozen score information, the realised history, timing, eligibility and sampling without replacement after earlier origins; L-par would require its fitted law to equal the relevant conditional assignment law. Random permutation of the pool does not restore the population law given the pool. The possible overlap of frames with the start of a command's motion and the possible use of the test set for checkpoint selection were not resolved. The ACTIVE counts are therefore working-law sensitivity results. The constructed NULL uses labels drawn under the stated resampling schemes and does not rely on these assumptions.

A context frame may already show the start of its command's motion. Origin zero was excluded because its reset timing was unverified; this did not establish timing at other origins. Under NULL timing does not matter. Under ACTIVE, a context frame that already reflects the current command can raise the rejection count, which is then not a clean measure of power; recording each frame before its command is applied would be a sufficient safeguard, and it was not verified here. The feasibility record supported the working marginals, two-frame repetition and low refresh-lag correlations, but did not verify the exact joint conditional law, exception process, timing or safety-conditioned retention. The constructed NULL was exact under either law and did not need the ACTIVE assumptions.

\subsection{Full results}

\input{appendix/tables/V4_L_null}
\input{appendix/tables/V4_L_active}
\input{appendix/tables/V4_L_validation}
\input{appendix/tables/V4_L_replacement}

\subsection{Predictions}

\input{appendix/tables/V4_L_predictions}

\subsection{Descriptions from the stored scores}

D1 and D2 were planned in protocol Section 7; D3--D5 and the scatter below are exploratory. The fine grid had 65 values $2^{-j/4}$, $j=-8,\dots,56$. Best-factor ties chose the largest factor; break-even was the largest grid factor with nonnegative log e-value, reported above four if profitable at four, or undefined if every grid value was negative. Quartiles used linear interpolation.

\input{appendix/tables/V4_L_best}
\input{appendix/tables/V4_L_breakeven}
\input{appendix/tables/V4_L_prefix_emp}
\input{appendix/tables/V4_L_prefix_par}

The pooled score did not reject in sixteen L-emp and seventeen L-par units; thirteen and twelve had negative log e-values. The median worst-origin share of negative net unit loss was 1.0812946 and 1.3009535, and the median share of the worst two origins was 1.6054236 and 1.8323682. Shares can exceed one because positive origins offset negative contributions. Command magnitudes below use recorded units, without an assumed physical-unit conversion.

\input{appendix/tables/V4_L_nonrejections}
\input{figures/v4/fig_scatter_float.tex}
\input{appendix/tables/V4_L_commands}

W(3/4) rejected under both laws in 155 units, under neither in three, under L-par alone in one and under L-emp alone in none. Pearson and Spearman correlations of pooled unit log e-values were 0.9939136 and 0.97335503; for W(3/4) they were 0.98242546 and 0.97434022. No correlation test was performed. Sums over units are descriptive: units share the candidate pool under L-emp, so the product is not a registered e-value.

\input{appendix/tables/V4_L_sums}

\subsection{Compute}

One NVIDIA GeForce RTX 5070 Laptop GPU; elapsed wrapper time 2.591607 hours. Peak sampled device memory was 3223 MiB, from one-second nvidia-smi samples, not an allocator peak. There was one invocation, no interruption and no resume.
\endgroup

%% file: figures/v4/fig_scatter_float.tex
\begin{center}\begin{minipage}{.85\linewidth}\centering
\includegraphics[width=\linewidth]{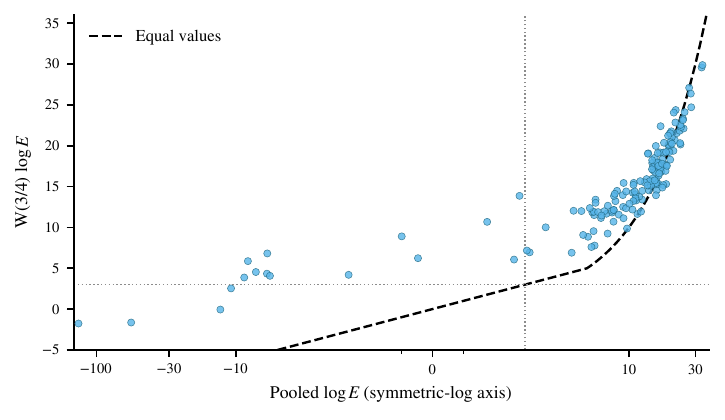}
\makeatletter\def\@captype{figure}\makeatother
\caption{Pooled versus weighted-mixture log e-value, ACTIVE L-emp (exploratory). Symmetric-log horizontal axis, equal-values line and $\log20$ reference lines.}\label{fig:p5-scatter}
\end{minipage}\end{center}

%% file: appendix/M_fourth_registration.tex
\section{TD-MPC2 at longer horizons and with partial execution}\label{app:p6}

This study is a preregistered replication of an exploratory finding.

\subsection{History, pilot and registration}

OSF registered the protocol and frozen code as 8mg9t on 12 November 2026 at 00:02:12.981096 UTC. The selected public randomness came from drand round 6633694 at 00:04:00 UTC. The frozen code revision was \texttt{f75b5b4928d9aa5aa90c2b49ae8b427141d2b0a0}.

A draft protocol fixed the design, criteria and eight dated predictions before any unit of this study was generated. A first version added partial execution, three comparator statistics and four predictions. It was not registered. The complete design was run once on separate seeds as an exploratory analysis before registration. Every criterion and prediction of the registered version was written after this exploratory run and was informed by it. No pilot or exploratory unit entered the confirmatory analysis.

The earlier stored-score study returned NO-GO: its G4 rule required both SW and UW to finish within two rejections of W(3/4) in every full-execution set and to reach the bar in at least 24 IRIS-token games and all six TD-MPC2 tasks. Both reached the bar in five TD-MPC2 tasks. SW lost five rejections in cartpole-balance (19 against 24), and UW lost seven there and seven in UpNDown. On the exploratory horizon run the SW loss did not recur and the partial-execution gain did. The registration then restricted the stake confirmation to TD-MPC2 and replaced the earlier rule by a total SW count of at least 97\% of W(3/4)'s total. A four-unit per-set allowance became a prediction. The registered SW maximum shortfall was one, within the earlier two-unit SW margin. The 58-unit registered gain margin followed exploratory gains of 84–109 units. The registration also replaced the per-task null threshold of two rejections in 64 with the pooled C-VAL threshold of 34 in 384 per statistic and horizon. It retained the earlier per-task threshold as a diagnostic forecast.

The earlier predictions and their outcomes on the exploratory run are listed in Table~\ref{tab:p6-earlier-predictions}. Eleven of the twelve protocol predictions and four of the five additional stake predictions held. The failed protocol prediction concerned smaller validation factors at four steps: this occurred in three tasks, where four were predicted. The failed stake prediction expected a full-execution loss in cartpole-balance; SW instead rejected in 29 units against 27 for W(3/4). The earlier predictions about factor shrinkage and declining pooled rejection counts disagreed, so their original rule drew no claim about growth of over-betting.

Under the draft protocol, the build pilot used 144 units: eight ACTIVE validation units and sixteen NULL test units per task, analysed at all four horizons. The NULL pipeline ran in full, with a largest rejection count of 1 of 16; ACTIVE units supplied predictions, variances and factors only. It showed native early termination in myo-reach and myo-obj-hold, with median lengths of 13 and 20 steps. At one step the new statistics code reproduced all 576 stored full-execution and NULL TD-MPC2 test units of the second study in 5760 comparisons, with no difference.

After the pilot and exploratory run, the registered protocol was implemented, unit-tested and checked against the pilot's stored NULL results before the confirmatory code was frozen. The protocol and code were registered on a public registry before the selected public randomness existed and before any confirmatory unit was generated. The beacon time was at least two minutes after the start of the registration minute; its randomness and the frozen protocol determined the confirmatory seeds. The saved beacon records agreed across two relays, but their cryptographic signature was not verified. The run completed in one invocation, with no interruption, resume or scientific deviation. An independently written checker subsequently reproduced every rejection, all selections, all summary results, the four criteria, the two confirmations and all twelve predictions, with zero disagreements.

\input{appendix/c_tables/horizon_earlier_predictions}

\subsection{Materials and design}

The study reused the released TD-MPC2 single-task checkpoints and the frozen simulator wrappers of the second study. The tasks were cartpole-balance, reacher-easy and walker-walk in DMControl, and myo-pose, myo-reach and myo-obj-hold in MyoSuite. Their unit caps were 500, 500, 500, 100, 100 and 75 model steps, respectively; a unit ended earlier at native termination or truncation and retained the terminal observation. DMControl used action repeat two. The latent dynamics are deterministic, so an $h$-step prediction is the exact $h$-fold composition of the one-step map. The local training configuration used horizon three, but the checkpoint metadata did not establish the training horizon of each checkpoint.

Each unit was analysed at horizons $h=1,2,4,8$. Labels were independent uniform draws from $[-1,1]^d$. At each step the label was executed with probability $p$ and an independent uniform action otherwise; the model always received the label. The conditions were full execution $p=1$, partial execution $p=0.5,0.2,0.05$, and NULL $p=0$. Each task and each non-null condition had sixteen validation and thirty-two test units; each task also had sixty-four NULL test units and no NULL validation units. There were 1536 units in all. Partial-execution fits used validation units of the same condition; NULL used the independent full-execution validation configuration of the same task and horizon.

Separate PCG64 streams supplied simulator resets, labels, executed actions, execution indicators, main bags and contrast bags at each horizon. Stream seeds were derived deterministically from the phase, task, condition, split, unit and stream, with a confirmatory master seed derived from the frozen protocol and public randomness. Draws were generated in float64 and converted to float32 at the model and simulator boundary; identical model inputs received identical predictions. Pilot, exploratory and confirmatory phases used separate seeds.

An origin was eligible when its context and its target $h$ steps later both lay inside the unit. Its main bag contained the recorded block and 127 independent draws from the uniform block design; validation contrast bags contained the recorded block and 31 draws. The target and predicted states had 512 latent coordinates. The score summed Gaussian log densities floored at $\log 10^{-12}$, with coordinate variances fitted from the recorded-block validation residuals at that horizon and floored at $10^{-8}$. Factors multiplied within each residue class modulo $h$, and the class products were averaged. Empty products equalled one; units without an eligible origin stayed in the denominator with e-value one. Validation configurations and contrast scores were stored before test units were scored. All e-values and decisions were computed from stored scores and configurations; large prediction and target arrays were not retained.

The sixteen statistics were pooled, Kish, contrast, V, M, W(1/4), W(1/2), W(3/4), R, R$_{1/2}$, T, N, RM, SM, SW and UW. W(3/4) was primary. Every statistic used the same bags at each horizon and rejected at $E\ge20$. Section~\ref{sec:v5-p6-config} gives the temperature objective, grids, stake pairs and tie rules. Factors and stake pairs were fitted with the 32-column validation contrast bags; test bags had 128 columns. A validation cell with no eligible origin would remain undefined, without imputation, and prevent any confirmation that depended on it.

\subsection{Criteria, confirmations and predictions}

C-VAL required each statistic at each horizon to reject in at most 34 of the 384 NULL test units pooled over tasks. C-POWER required W(3/4) to reject in at least 24 of 32 full-execution units in a task. S-GAIN required SW, at one step over the 576 partial-execution units, to gain at least 58 rejections over each of W(3/4), T, R and R$_{1/2}$, with an exact two-sided paired sign-test $p<0.01$ against each. S-NOHARM required the total SW rejections over the 768 full-execution unit--horizon pairs to be at least 97\% of W(3/4)'s total.

The horizon confirmation required C-VAL for every statistic and horizon and C-POWER in at least four tasks at two steps and at least four at four steps. The stake confirmation required C-VAL, S-GAIN and S-NOHARM; the two confirmations were evaluated separately. Both held. The largest pooled NULL count was 4 of 384, and five tasks reached the power bar at both two and four steps. Section~\ref{sec:v5-registered} gives the one-step and full-execution totals; R rejected in 228 partial-execution units at one step, and all four paired comparisons passed.

All twelve registered predictions were written after the exploratory run. Eleven of them held (Table~\ref{tab:p6-predictions}). The exception was the strict per-task NULL diagnostic: at eight steps in myo-reach, pooled, Kish and V each rejected in 3 of 64 units, above the predicted maximum of two. This did not fail the registered pooled C-VAL criterion. The confirmations and all prediction outcomes were applied mechanically to the frozen results.

\input{appendix/c_tables/horizon_predictions}

\subsection{Temperature and stake configuration}\label{sec:v5-p6-config}
At horizon $h$, replace each origin factor by $1-f+f e_t(\lambda)$, multiply these factors within each residue class, and average the $h$ class products:
\[
E_r(\lambda,f)=\prod_{t\equiv r\pmod h}\{1-f+f e_t(\lambda)\},\qquad
E(\lambda,f)=\frac1h\sum_{r=0}^{h-1}E_r(\lambda,f).
\]
Empty products equal one. With $f=1$ this is $E(\lambda)$. The temperature grid is $\lambda_g=2^{-g}$, $g=0,\ldots,14$, and the stake grid is $f_k=2^{-k}$, $k=0,\ldots,6$, giving 105 pairs.

SM is the uniform average of $E(\lambda_g,f_k)$ over these 105 pairs. SV selects the pair with the largest median unit log e-value over the validation units, computed with the 32-column contrast bags; ties go to the smaller stake, then to the larger temperature factor. SV is used only inside SW. SW puts weight $3/4$ on the selected pair and weight $1/(4\cdot104)$ on each of the other 104 pairs. The universal mixture is $\mathrm{UW}=\mathrm{SW}/2+\mathrm{R}/4+\mathrm{R}_{1/2}/4$.

For temperature scaling T, let $F(\lambda)$ be the sum of the origin log factors over all eligible origins of the validation units, computed with their 32-column contrast bags. The factor $\lambda_{\mathrm T}$ maximizes $F$ over $[2^{-14},4]$. Bounded scalar optimization is performed on $\log_2\lambda$ with SciPy's \texttt{minimize\_scalar}, method \texttt{bounded}, and \texttt{xatol} $10^{-6}$; the result is compared with both endpoints, and the smaller factor wins a tie. Then $\mathrm T=E(\lambda_{\mathrm T})$. The site-average temperature is $\mathrm N=E(1/n)$, with $n=512$ in the fourth study. The rank mixture with a neutral bet is $\mathrm{RM}=(1+\mathrm R+\mathrm R_{1/2})/3$.

The fourth study fits every validation-dependent parameter from the validation units of the same task, condition and horizon; NULL units use the full-execution configuration of the same task and horizon. V selects from the same temperature grid by median validation unit log e-value, with ties to the larger factor. The stake statistics and the other statistics use the same stored scores at each horizon.

\subsection{Full result tables}
The ancillary files \texttt{d-014.tsv} to \texttt{d-025.tsv} contain the fourth study tables, including unit and origin counts, rejections, paired counts and validation selections. File \texttt{e-029.tsv} compares the registered results with those of the earlier exploratory run on separate seeds. The exploratory horizon comparison is in Table~\ref{tab:v5-horizons}.

%% file: appendix/c_tables/horizon_earlier_predictions.tex
\begingroup\footnotesize\setlength{\tabcolsep}{3pt}
\begin{longtable}{@{}>{\raggedright\arraybackslash}p{.49\linewidth}>{\raggedright\arraybackslash}p{.11\linewidth}>{\raggedright\arraybackslash}p{.08\linewidth}>{\raggedright\arraybackslash}p{.26\linewidth}@{}}
\caption{Earlier protocol predictions (1--12) and additional stake predictions (S1--S5), with outcomes on the separate exploratory run. These outcomes are exploratory.}\label{tab:p6-earlier-predictions}\\
\toprule Prediction & Probability & Outcome & Record \\\midrule\endfirsthead
\toprule Prediction & Probability & Outcome & Record \\\midrule\endhead
\midrule\multicolumn{4}{r}{Continued on next page}\\\endfoot\bottomrule\endlastfoot
1. Validity in every task at every horizon & 90\% & held & As predicted. \\
2. Overall confirmation & 65\% & held & W(3/4) at the bar in 5 of 6 tasks at h = 2 and at h = 4 \\
3. At h = 8, power in at least 4 of 6 tasks & 50\% & held & 4 \\
4. At h = 1, W(3/4) at the bar in at least 5 tasks and pooled in at most 2 & 85\% & held & 6 and 1 \\
5. At h = 2 and h = 4, pooled's median log E below 0 in at least 3 tasks & 75\% & held & 5 and 5 \\
6. $\lambda_{\mathrm V}$ at h = 4 strictly smaller than at h = 1 in at least 4 tasks & 50\% & failed & 3 \\
7. pooled rejects in fewer ACTIVE units at h = 4 than at h = 1 & 60\% & held & 37 against 54 \\
8. W(3/4) rejects in at least pooled minus 2 in every task and horizon & 85\% & held & As predicted. \\
9. T and W(3/4) within 5\% over the 24 ACTIVE sets & 75\% & held & 655 and 666 \\
10. At h = 1, R$_{1/2}$ rejects in more partial-execution units than W(3/4) & 65\% & held & 280 against 256 \\
11. At h = 1, N at the bar in fewer sets than W(3/4) & 70\% & held & 10 against 13 \\
12. W(3/4) rejects in fewer partial-execution units at h = 4 than at h = 1 & 80\% & held & 150 against 256 \\
S1. At h = 1, SW rejects in more partial-execution units than W(3/4) and than R$_{1/2}$ & 75\% & held & 364 against 256 and 280 \\
S2. At h = 1 and p = 0.05, SW rejects in at least 24 of 32 units in reacher-easy & 65\% & held & 29 \\
S3. At h = 1 and p = 1, SW rejects in fewer units than W(3/4) in cartpole-balance & 70\% & failed & 29 against 27 \\
S4. SM, SW and UW each reject in at most 2 of 64 NULL units in every task and horizon & 90\% & held & As predicted. \\
S5. At h = 2 and h = 4, SW rejects in more partial-execution units than W(3/4) & 65\% & held & 264 against 209; 184 against 150 \\
\end{longtable}\endgroup

%% file: appendix/c_tables/horizon_predictions.tex
\begingroup\footnotesize\setlength{\tabcolsep}{3pt}
\begin{longtable}{@{}>{\raggedright\arraybackslash}p{.49\linewidth}>{\raggedright\arraybackslash}p{.11\linewidth}>{\raggedright\arraybackslash}p{.08\linewidth}>{\raggedright\arraybackslash}p{.26\linewidth}@{}}
\caption{Fourth study: twelve registered predictions, informed by the earlier exploratory run.}\label{tab:p6-predictions}\\
\toprule Prediction & Probability & Outcome & Record \\\midrule\endfirsthead
\toprule Prediction & Probability & Outcome & Record \\\midrule\endhead
\midrule\multicolumn{4}{r}{Continued on next page}\\\endfoot\bottomrule\endlastfoot
1. Horizon confirmation & 85\% & held & Both confirmations' component criteria held. \\
2. Stake confirmation & 80\% & held & C-VAL, S-GAIN and S-NOHARM held. \\
3. Strict null diagnostic: in every task and at every horizon, each statistic rejects in at most 2 of the 64 NULL units & 60\% & did not hold & Largest per-task count 3/64 at eight steps. \\
4. The pooled statistic collapses at every horizon: at each of h = 1, 2, 4 and 8, pooled's median log E over the ACTIVE test units is below 0 in at least 4 of the 6 tasks & 85\% & held & Negative medians in 4, 5, 5 and 5 tasks. \\
5. At h = 8, C-POWER holds in at least 4 of the 6 tasks & 60\% & held & 4 tasks at the bar. \\
6. Calibration matches mixing at full execution: summed over the 24 ACTIVE sets, the rejections of T and of W(3/4) differ by at most 5\% of the larger total & 85\% & held & T 630; W(3/4) 633. \\
7. Rank bets beat the weighted temperature mixture under partial execution: at h = 1, over the 18 partial-execution sets, R$_{1/2}$ rejects in more test units than W(3/4) & 75\% & held & 286 against 243. \\
8. Rank bets fail on few origins: at h = 8 in myo-reach, among the ACTIVE test units, R$_{1/2}$ rejects in at most 4 and W(3/4) in at least 24 & 80\% & held & 1 against 26. \\
9. At h = 1 and p = 0.05, SW rejects in at least 24 of the 32 test units of reacher-easy & 85\% & held & 30 of 32. \\
10. The stake leads at longer horizons too: at each of h = 2 and h = 4, over the 18 partial-execution sets, SW rejects in more test units than W(3/4) and in more than R$_{1/2}$ & 70\% & held & Two steps: 266 > 186 and 232; four steps: 169 > 147 and 144. \\
11. No set is hurt much: in every task and at every horizon, SW rejects in at least as many ACTIVE test units as W(3/4) minus 4 & 75\% & held & Largest shortfall: one unit. \\
12. The pooled statistic loses more per origin at longer horizons: the mean log factor per origin of pooled over the ACTIVE test units is lower at h = 8 than at h = 1 in at least 5 of the 6 tasks & 80\% & held & Lower in all six tasks. \\
\end{longtable}\endgroup

%% file: appendix/N_matched_comparators.tex
\section{Matched comparators on stored scores}\label{app:H}

The main comparison is in \S\ref{sec:v5-comparators} and \S\ref{sec:v5-horizons}.

Before execution, the protocol was examined by a separate automated reviewer, a language-model agent run in its own session, and revised until that reviewer passed it.

The protocol specified matched comparisons on registered scores, exploratory scores, and the other stored settings. The version used for the results was fixed before the new statistics were calculated. The 15 forecasts concerned a design that had already been examined. Fourteen held. The final validation passed after a correction of the median comparison tolerance.

Whole-unit PS, PZ and PR target a stronger global null. Their validity is certified only in the designated TD-MPC2 cells. On the token games their results are diagnostic, including the 198 of 288 rejections for PS. The task-reweighting intervals describe reweighting the observed tasks. A sampling interpretation requires exchangeable tasks.

The restart preserved earlier stopped-run records. Median validation passed after the authorized tolerance correction, with no change to an analysis statistic or forecast.

\paragraph{Prespecified reading rules (A6 protocol v3).} The h-series ancillary files give the per-task, paired and full-execution results, the interval direction, and the complete comparator coverage. A6's analytical deviations and final validation are recorded in A6\_REPORT.md and CORRECTION\_02.md.

Every conclusion is an observed, exploratory comparison on the specified stored units. All comparator outcomes, unavailable settings, paired reversals, task effects and uncertainty are reported. An unavailable or unvalidated comparator is not counted as evidence for SW.

Every validity and power comparison names its null. PS, PZ and PR, valid only for the global null, may be compared with the e-values in a panel explicitly interpreted as tests of the stronger whole-unit global null, under which both are valid. Such a comparison is not evidence that those p-values control size or replace the e-values for the weaker per-origin null. A claim about the latter uses only procedures valid under that null. "All evaluated valid comparators" always refers to the complete list valid for the null named in that sentence.

For each dataset, condition and horizon, name every comparator with more, equal or fewer rejections than SW, with counts and rate differences. Write "SW had the largest observed count among the evaluated valid comparators in this cell" only when it is true of the complete list; otherwise name the leading comparators and SW's losses. Do not infer superiority in population power, equivalence, or a general recommendation from an ordering of counts.

If a test that does not multiply factors and is valid for the null named (PRseq and PRhoef for the per-origin null; PS, PZ and PR for the global null where they have that status) matches or exceeds SW in D1 under partial execution, say so, even if it loses at full execution. Compare it separately with every e-value statistic, and report full-execution counts, paired retention of W(3/4)'s rejections and per-task losses. Say that it matches or exceeds every evaluated e-value's count only if that is true of every e-value in every cell the sentence names. A count ratio of 97\% is not called equivalence or "no cost".

If TS\_G or TS\_C is close to SW, say that SW's particular objective and mixing rule are not needed to reproduce a similar count in that setting. If staked ranks are close, say that score magnitudes were not needed to reproduce a similar count there. If either exceeds SW, report the excess. These findings alone do not establish that bounding the stake caused the gain.

"Close" is a display convention: an absolute difference of at most five percentage points in rejection rate (28 units of 576). It is not a non-inferiority test.

To assess fitting f while keeping the other design choices fixed, compare SV with V, TS\_G with T\_G, TS\_C with the original continuous T, RSV with RSV\_1, and RSW with RSW\_1, separately at K = 128 and K = 32. The fitting objective and the mixture structure of each pair are held fixed; the stake restriction is the changed feature. Report full-execution losses and selection changes as well as partial-execution gains. RM and the better observed count of R and R1/2 remain descriptive benchmarks, not matched stake ablations. A comparator chosen as "better" on the test set is an oracle count benchmark, not a new valid e-value procedure.

No result of A6 establishes that the test is useful for evaluating world models, that any rule transfers to new model families, or that the recorded-robot design is valid.

{\footnotesize\setlength{\parskip}{0pt}
The full table is in the ancillary files (file \texttt{h-001.tsv}). The full table is in the ancillary files (file \texttt{h-002.tsv}). The full table is in the ancillary files (file \texttt{h-003.tsv}). The full table is in the ancillary files (file \texttt{h-004.tsv}). The full table is in the ancillary files (file \texttt{h-005.tsv}). The full table is in the ancillary files (file \texttt{h-006.tsv}). The full table is in the ancillary files (file \texttt{h-007.tsv}). The full table is in the ancillary files (file \texttt{h-008.tsv}). The full table is in the ancillary files (file \texttt{h-009.tsv}). The full table is in the ancillary files (file \texttt{h-010.tsv}). 

The full table is in the ancillary files (file \texttt{h-011.tsv}). The full table is in the ancillary files (file \texttt{h-012.tsv}). The full table is in the ancillary files (file \texttt{h-013.tsv}). The full table is in the ancillary files (file \texttt{h-014.tsv}). The full table is in the ancillary files (file \texttt{h-015.tsv}). The full table is in the ancillary files (file \texttt{h-016.tsv}). The full table is in the ancillary files (file \texttt{h-017.tsv}). The full table is in the ancillary files (file \texttt{h-018.tsv}). The full table is in the ancillary files (file \texttt{h-019.tsv}). The full table is in the ancillary files (file \texttt{h-020.tsv}). 

The full table is in the ancillary files (file \texttt{h-021.tsv}). The full table is in the ancillary files (file \texttt{h-022.tsv}). The full table is in the ancillary files (file \texttt{h-023.tsv}). The full table is in the ancillary files (file \texttt{h-024.tsv}). The full table is in the ancillary files (file \texttt{h-025.tsv}). The full table is in the ancillary files (file \texttt{h-026.tsv}). The full table is in the ancillary files (file \texttt{h-027.tsv}). The full table is in the ancillary files (file \texttt{h-028.tsv}). The full table is in the ancillary files (file \texttt{h-029.tsv}). The full table is in the ancillary files (file \texttt{h-030.tsv}). 

The full table is in the ancillary files (file \texttt{h-031.tsv}). The full table is in the ancillary files (file \texttt{h-032.tsv}). The full table is in the ancillary files (file \texttt{h-033.tsv}). The full table is in the ancillary files (file \texttt{h-034.tsv}). The full table is in the ancillary files (file \texttt{h-035.tsv}). The full table is in the ancillary files (file \texttt{h-036.tsv}). The full table is in the ancillary files (file \texttt{h-037.tsv}). The full table is in the ancillary files (file \texttt{h-038.tsv}). The full table is in the ancillary files (file \texttt{h-039.tsv}). The full table is in the ancillary files (file \texttt{h-040.tsv}). 

The full table is in the ancillary files (file \texttt{h-041.tsv}). The full table is in the ancillary files (file \texttt{h-042.tsv}). The full table is in the ancillary files (file \texttt{h-043.tsv}). The full table is in the ancillary files (file \texttt{h-044.tsv}). The full table is in the ancillary files (file \texttt{h-045.tsv}). The full table is in the ancillary files (file \texttt{h-046.tsv}). The full table is in the ancillary files (file \texttt{h-047.tsv}). The full table is in the ancillary files (file \texttt{h-048.tsv}). The full table is in the ancillary files (file \texttt{h-049.tsv}). The full table is in the ancillary files (file \texttt{h-050.tsv}). 

The full table is in the ancillary files (file \texttt{h-051.tsv}). The full table is in the ancillary files (file \texttt{h-052.tsv}). The full table is in the ancillary files (file \texttt{h-053.tsv}). The full table is in the ancillary files (file \texttt{h-054.tsv}). The full table is in the ancillary files (file \texttt{h-055.tsv}). The full table is in the ancillary files (file \texttt{h-056.tsv}). The full table is in the ancillary files (file \texttt{h-057.tsv}). The full table is in the ancillary files (file \texttt{h-058.tsv}). The full table is in the ancillary files (file \texttt{h-059.tsv}). The full table is in the ancillary files (file \texttt{h-060.tsv}). 

The full table is in the ancillary files (file \texttt{h-061.tsv}). The full table is in the ancillary files (file \texttt{h-062.tsv}). The full table is in the ancillary files (file \texttt{h-063.tsv}). The full table is in the ancillary files (file \texttt{h-064.tsv}). The full table is in the ancillary files (file \texttt{h-065.tsv}). The full table is in the ancillary files (file \texttt{h-066.tsv}). The full table is in the ancillary files (file \texttt{h-067.tsv}). The full table is in the ancillary files (file \texttt{h-068.tsv}). The full table is in the ancillary files (file \texttt{h-069.tsv}). The full table is in the ancillary files (file \texttt{h-070.tsv}). 

The full table is in the ancillary files (file \texttt{h-071.tsv}). The full table is in the ancillary files (file \texttt{h-072.tsv}). The full table is in the ancillary files (file \texttt{h-073.tsv}). The full table is in the ancillary files (file \texttt{h-074.tsv}). The full table is in the ancillary files (file \texttt{h-075.tsv}). The full table is in the ancillary files (file \texttt{h-076.tsv}). The full table is in the ancillary files (file \texttt{h-077.tsv}). The full table is in the ancillary files (file \texttt{h-078.tsv}). The full table is in the ancillary files (file \texttt{h-079.tsv}). The full table is in the ancillary files (file \texttt{h-080.tsv}). 
\par}

\subsection{Dated forecasts and outcomes}
\begingroup\footnotesize\setlength{\tabcolsep}{2pt}
\begin{longtable}{@{}>{\raggedright\arraybackslash}p{0.085\linewidth}>{\raggedright\arraybackslash}p{0.665\linewidth}>{\raggedright\arraybackslash}p{0.11\linewidth}>{\raggedright\arraybackslash}p{0.09\linewidth}@{}}
\toprule
Forecast & Prespecified event & Probability & Outcome\\
\midrule\endfirsthead
\toprule
Forecast & Prespecified event & Probability & Outcome\\
\midrule\endhead
\bottomrule\endlastfoot
E1.1 & Every A in E1 has C\_A(Dd, NULL, h) \textless{}= 34 for d in \{1, 2\} and h in \{1, 2, 4, 8\}, and every A in Q has C\_A(Dd, NULL, 1) \textless{}= 34 for d in \{1, 2\}. A diagnostic event, not a validity criterion. & 95\% & held\\
E1.2 & C\_RSW(P) \textgreater{} C\_R1/2(P). & 85\% & held\\
E1.3 & 376 \textgreater{} C\_RSW(P). & 60\% & held\\
E1.4 & \textbar{}C\_TS\_G(P) - 376\textbar{} \textless{}= 38, a net count difference. & 65\% & held\\
E1.5 & 376 \textgreater{} C\_OS(P). & 70\% & held\\
E1.6 & 376 \textgreater{} max\{C\_PS(P), C\_PZ(P), C\_PR(P)\}. & 40\% & held\\
E1.7 & C\_PR(F) \textgreater{}= C\_W(3/4)(F). & 60\% & held\\
E1.8 & C\_PR(G) \textgreater{}= 90. & 80\% & held\\
E1.9 & C\_PS(G) \textgreater{}= 48. & 60\% & held\\
E1.10 & max over A in E1 union Q of C\_A(I) \textgreater{}= 193; a descriptive maximum. & 50\% & held\\
E1.11 & 376 \textgreater{} C\_PRseq(P). & 55\% & held\\
E1.12 & C\_PRseq(F) \textgreater{}= C\_W(3/4)(F). & 50\% & failed\\
E1.13 & 376 \textgreater{} C\_ON(P). & 60\% & held\\
E1.14 & \textbar{}C\_TS\_C(P) - 376\textbar{} \textless{}= 28. & 60\% & held\\
E1.15 & In P, the validation-selected staked rank family rejects in more units than fixed R1/2: C\_RSV(P) \textgreater{} C\_R1/2(P). This forecast changes the calibrator selection as well as allowing a stake and is not itself an isolated stake contrast. & 80\% & held\\
\end{longtable}
\endgroup

%% file: appendix/O_fitted_stakes.tex
\section{Fitted stakes and the execution rate}\label{app:I}

The main comparison is in \S\ref{sec:v5-fitted}.

Before execution, the protocol was examined by a separate automated reviewer, a language-model agent run in its own session, and revised until that reviewer passed it.

In the one-step exploratory analysis, a cell was classified as strong if its full-execution TS\_C log e-values summed over test units, divided by their total eligible-origin count, reached 0.5 nats per origin. Among the strong partial-execution cells the fitted stake was within a factor of two of $p$ in 29/31, 14/15 and 13/15 cells in the three data sets, with median $f/p$ of 0.96, 0.97 and 1.09. In the two data sets with three partial-execution rates, 11 of 12 task series were non-increasing as $p$ fell. T's factor was lower than its full-execution value in all 61 strong cells. Five of eight weaker cells were within the band. The registered-score data set includes reacher-easy, where an earlier grid-stake fit suggested the question. These are descriptive comparisons of fitted values with the known execution probability. Five of seven forecasts held. The separate checker reproduced 7,842 values.

The source-path audit initially stopped on a false positive. It continued after correction and left the numerical outputs byte-identical. The full-execution strength split was assigned before partial fits were read.

Forecast 6 met its factor-of-two temperature criterion in 28 of 61 strong cells, below the required half. Forecast 7 found 5 of 8 weaker cells in the stake band, above the predicted fewer than half. In the reading rule, $r=f_p/p$, $d=\log_2(\lambda_p/\lambda_1)$, and $d_T$ is the corresponding temperature-only log ratio (A7 protocol v3, \S{}\S{}3–6).

\paragraph{Prespecified reading rules (A7 protocol v3).}

- State only the repaired proposition and its specified comparison class. The exact finite-bag oracle weight is w, not generally p; the full joint likelihood statement using L/m\_H is a different result.
- For each of forecasts 1 and 3 report held, failed or unavailable, and give the r-band numerator and denominator separately for S1, S2 and S3. If both hold, say: ``The prespecified exploratory factor-of-two criterion was met in both the S1 strong-cell group and the pooled S2/S3 strong-cell group. The fitted stakes were within the band in N of M cells in each reported data set.'' Add: ``This is agreement with an empirical heuristic; it does not establish that the fitted stake estimates p.'' If either fails, identify the group whose criterion failed and report the counts without asserting that every fitted stake failed to track p. If either is unavailable, say that the combined criterion could not be assessed, while reporting every separately assessable group. Report all other forecasts with the same held/failed/unavailable distinction and their denominators.
- Report weak-cell counts regardless of outcome. Exactly neutral factors do not identify a stake; small signal may make identification weak, but membership in the weak group does not prove non-identification. Report the neutral-temperature and flat-series qualifications of Sections 5 and 6.
- Do not claim optimality of the fitted pair, estimation of p, a theorem predicting the empirical band counts, or extension beyond the specified TD-MPC2 and inverse-dynamics scorers, tasks and h=1. Higher horizons remain descriptive. Passing the broad band or monotonicity forecasts is not a calibration or identification test.
- Repeat the disclosure about the motivating observation wherever these results are reported. S2 contains the reacher-easy case that suggested the question and informed the forecasts, although the disclosed statistic was SV rather than TS\_C. Describe this as an informed exploratory check, not independent confirmation. No exclusion or change of the forecasts is made after reading the partial fits.

{\footnotesize\setlength{\parskip}{0pt}
The full table is in the ancillary files (file \texttt{i-001.tsv}). The full table is in the ancillary files (file \texttt{i-002.tsv}). The full table is in the ancillary files (file \texttt{i-003.tsv}). The full table is in the ancillary files (file \texttt{i-004.tsv}). 
\par}

\paragraph{Motivating observation.} E2 examined three sets of stored fits. The earlier reacher-easy observation used SV stakes $1/2$, $1/4$ and $1/16$ at $p=0.5$, $0.2$ and $0.05$ with temperature $1/2$; V chose $1/512$, $1/1024$ and $1/2048$. Set S2 contains that observation.

\subsection{Dated forecasts and outcomes}
\begingroup\footnotesize\setlength{\tabcolsep}{2pt}
\begin{longtable}{@{}>{\raggedright\arraybackslash}p{0.085\linewidth}>{\raggedright\arraybackslash}p{0.665\linewidth}>{\raggedright\arraybackslash}p{0.11\linewidth}>{\raggedright\arraybackslash}p{0.09\linewidth}@{}}
\toprule
Forecast & Prespecified event & Probability & Outcome\\
\midrule\endfirsthead
\toprule
Forecast & Prespecified event & Probability & Outcome\\
\midrule\endhead
\bottomrule\endlastfoot
E2.1 & S1, strong cells: at least two thirds have r in the band. & 55\% & held\\
E2.2 & S1, strong cells: the median r is in the band. & 70\% & held\\
E2.3 & S2 and S3 together, strong cells (at most 36): at least two thirds have r in the band. & 55\% & held\\
E2.4 & S2 and S3: in at least 9 of the 12 task-by-data-set series, f\_0.5 \textgreater{}= f\_0.2 \textgreater{}= f\_0.05. & 65\% & held\\
E2.5 & All data sets, strong cells: d\_T \textless{} 0 in at least 80\% of cells. & 75\% & held\\
E2.6 & All data sets, strong cells: \textbar{}d\textbar{} \textless{}= 1 in at least half of the cells. & 50\% & failed\\
E2.7 & Weak cells, all data sets: fewer than half have r in the band. & 60\% & failed\\
\end{longtable}
\endgroup

%% file: appendix/P_degraded_action.tex
\section{Degraded action input and inverse dynamics}\label{app:J}

The main comparison is in \S\ref{sec:v5-weakening}.

Before execution, the protocol was examined by a separate automated reviewer, a language-model agent run in its own session, and revised until that reviewer passed it.

The changes and clarifications before and during execution concerned eligibility and reset or checkpoint compatibility, stage ownership, resource admissions, and the reporting treatment of the reference scorer. The completed report preserves the stated limits and reports the affected records separately.

The study compared perturbations of the action input, released model sizes and inverse-dynamics scorers. It used six tasks for the perturbation and reference comparisons. Eight of ten forecasts held. The separate checker reproduced 71,852 values.

Masking a random fraction of the action input's coordinates had a similar effect (13.9, 8.5 and 5.7 at 3/4, 1/2 and 1/4 of the coordinates kept).

The three Myo observations contain 39 activation coordinates. The three DMControl observations contain no explicit action trace. The correlation of cartpole-balance cart velocity change with the action was 0.8812. Eight forecasts held, and forecasts 6 and 7 failed. The published runtime includes resource waits and clarification pauses.

The perturbations do not order true model quality. Gain zero is an implementation identity.

\paragraph{Prespecified reading rules (P7 protocol v3).}

Every statement names its arm, tasks, condition, horizon 1, and the exploratory status.

- Arm A. Arm A measures sensitivity to the specified action-input perturbations, separately for p=1 and p=0.2. The alpha=0 e-value identity is an implementation control, not empirical evidence of usefulness. Intermediate gains and masks are not assumed to order true model quality. For each condition, display every adjacent change in summed C and summed G along the gain sequence and the mask sequence, collapsing identical mask steps. Describe C as nonincreasing only if every adjacent increase is \textless{}=0; separately, for p=0.2 gain, report whether it meets forecast 1's three-unit tolerance. Describe G as nonincreasing only if every adjacent increase is \textless{}=1e-12 nats per origin. Do not apply count slack to G. A flat sequence is reported as flat, and a decline occurring only at alpha=0 is identified as such. Report error ratios and evidence changes in their own units; do not say which metric changed more across incompatible scales or infer that low sensitivity of total prediction error makes it an inferior metric. The sign-flip scorer's C, G and error metrics are reported beside the intact scorer's without a verdict.
- Arm B. Report which pairs were separated under the Holm rule in the E3 protocol, all raw and adjusted values, the order of G, the diagnostics, the parameter counts and the measured returns. Say only that observed G was non-decreasing across the five sizes if that is true in every planned task with all five results available. Identify entirely flat sequences; call an increase only when there is also a strict increase in each task named. No causal size effect, representation-quality ordering or control-quality ordering follows. If no pair is separated, say that the test did not separate these scorers on these units, and that this is not equivalence.
- Arm C. Report exact per-task and summed C, G and paired differences for WM-8, IDM-V, IDM-V-clean and IDM-L separately in each condition, with each scorer's training and configuration budget. Equal observed counts mean equal counts, not equivalence. If both scorers have zero rejections, state that neither detected the effect by this rule. If both reject every unit, state that C is saturated and use the prespecified G and paired displays without claiming equality of power. If IDM has nonzero rejections, state the number and denominator and that this fitted model-free scorer detected effects in those units. A WM-8 advantage is restricted to these fitted scorers, budgets and tasks; it does not establish that world models or their pretraining are necessary. Report disagreements between C and G explicitly. Report G\_WM-8/G\_IDM-L only when G\_IDM-L\textgreater{}0, together with its numerator and denominator; the ratio has no general efficiency or equivalence interpretation.
- Null check. For each prespecified scorer/configuration/statistic NULL panel with n completed units and c rejections, compute the upper tail Pr\{Binomial(n,0.05)\textgreater{}=c\}. A value below 0.001 triggers an implementation audit before substantive reporting, for PRseq as well as e-values. Report every count, actual denominator and audit outcome. This is an unadjusted diagnostic trigger across many dependent panels, not a familywise test or a certification of validity. Report any known implementation violation regardless of this threshold, and preserve failed or unavailable panels.
- Unavailable results. An unavailable scorer, cell or arm is reported as unavailable with its reason. No sentence of this rule is applied to a reduced set as if it were the planned set.
- Nothing in P7E is described as confirmed. If a registered confirmation follows, P7E is disclosed as the exploratory run that informed it.

{\footnotesize\setlength{\parskip}{0pt}
The full table is in the ancillary files (file \texttt{j-001.tsv}). The full table is in the ancillary files (file \texttt{j-002.tsv}). The full table is in the ancillary files (file \texttt{j-003.tsv}). The full table is in the ancillary files (file \texttt{j-004.tsv}). The full table is in the ancillary files (file \texttt{j-005.tsv}). The full table is in the ancillary files (file \texttt{j-006.tsv}). The full table is in the ancillary files (file \texttt{j-007.tsv}). The full table is in the ancillary files (file \texttt{j-008.tsv}). The full table is in the ancillary files (file \texttt{j-009.tsv}). The full table is in the ancillary files (file \texttt{j-010.tsv}). 

The full table is in the ancillary files (file \texttt{j-011.tsv}). The full table is in the ancillary files (file \texttt{j-012.tsv}). The full table is in the ancillary files (file \texttt{j-013.tsv}). The full table is in the ancillary files (file \texttt{j-014.tsv}). The full table is in the ancillary files (file \texttt{j-015.tsv}). The full table is in the ancillary files (file \texttt{j-016.tsv}). The full table is in the ancillary files (file \texttt{j-017.tsv}). The full table is in the ancillary files (file \texttt{j-018.tsv}). The full table is in the ancillary files (file \texttt{j-019.tsv}). The full table is in the ancillary files (file \texttt{j-020.tsv}). 

The full table is in the ancillary files (file \texttt{j-021.tsv}). The full table is in the ancillary files (file \texttt{j-022.tsv}). The full table is in the ancillary files (file \texttt{j-023.tsv}). The full table is in the ancillary files (file \texttt{j-024.tsv}). The full table is in the ancillary files (file \texttt{j-025.tsv}). The full table is in the ancillary files (file \texttt{j-026.tsv}). The full table is in the ancillary files (file \texttt{j-027.tsv}). The full table is in the ancillary files (file \texttt{j-028.tsv}). The full table is in the ancillary files (file \texttt{j-029.tsv}). The full table is in the ancillary files (file \texttt{j-030.tsv}). 

The full table is in the ancillary files (file \texttt{j-031.tsv}). The full table is in the ancillary files (file \texttt{j-032.tsv}). 
\par}

\paragraph{Sign reversal.} The sign-flipped scorers had fitted TS evidence rates of zero at full execution in the six tasks. Under partial execution the sum of the task evidence rates was slightly negative.

\subsection{Dated forecasts and outcomes}
\begingroup\footnotesize\setlength{\tabcolsep}{2pt}
\begin{longtable}{@{}>{\raggedright\arraybackslash}p{0.085\linewidth}>{\raggedright\arraybackslash}p{0.665\linewidth}>{\raggedright\arraybackslash}p{0.11\linewidth}>{\raggedright\arraybackslash}p{0.09\linewidth}@{}}
\toprule
Forecast & Prespecified event & Probability & Outcome\\
\midrule\endfirsthead
\toprule
Forecast & Prespecified event & Probability & Outcome\\
\midrule\endhead
\bottomrule\endlastfoot
E3.1 & Arm A, gain, p = 0.2, six tasks: summed C does not rise by more than 3 units at any step from alpha = 1 down through 0.5, 0.25, 0.1 and 0. & 65\% & held\\
E3.2 & Arm A, gain, p=1: summed G at alpha=1 is strictly positive and summed G at alpha=0.5 is at least half that value. This is a conjunction; a nonpositive intact baseline makes the forecast false, not evidence of successful retention. & 55\% & held\\
E3.3 & Arm A, mask, p=0.2, the five tasks reacher-easy, walker-walk, myo-pose, myo-reach and myo-obj-hold: summed C at r=0.25 is below summed C at alpha=1 by at least 20 units. & 70\% & held\\
E3.4 & Arm A, six tasks: the ratio M/M\_intact at alpha = 0 is less than 1.10 in at least 2 of the 6 tasks. & 50\% & held\\
E3.5 & Arm B, p = 1: G of mt30-317M exceeds G of mt30-1M in each of the three tasks. & 60\% & held\\
E3.6 & Arm B, p = 1: G is non-decreasing across the five sizes in at least 2 of the three tasks. & 25\% & failed\\
E3.7 & Arm B, p = 1: the single-task 5M model has a larger G than mt30-5M in at least 2 of the three tasks. & 65\% & failed\\
E3.8 & Arm C, p = 0.2, six tasks: summed C of IDM-L is at least summed C of WM-8. & 50\% & held\\
E3.9 & Arm C, p = 0.2, six tasks: summed C of IDM-V is below summed C of WM-8. & 75\% & held\\
E3.10 & Null: in every prespecified scorer/configuration panel over its declared tasks, each e-value statistic has at most six rejections, and PRseq's exact upper Binomial(n,0.05) tail at its observed count is at least 0.001. The denominator n is reported for every panel. This is a forecast of observed conservative behavior, not a consequence or criterion of validity. & 80\% & held\\
\end{longtable}
\endgroup

%% file: appendix/Q_thirty_tasks.tex
\section{Thirty tasks}\label{app:K}

The main comparison is in \S\ref{sec:v5-sizes}.

Before execution, the protocol was examined by a separate automated reviewer, a language-model agent run in its own session, and revised until that reviewer passed it.

The final protocol used the clarified single-task checkpoint-selection rule and the revised return record. The completed run retained its pause-and-continuation audit and the separate recomputation. The run history is given below.

The study compared five released multitask sizes, single-task checkpoints and an inverse-dynamics reference under two action laws. The separate checker reproduced 234,356 values.

For each task we used the lowest-numbered released single-task seed among 1, 2 and 3 whose file loaded strictly with our unchanged loader and needed no key mapping: seed 1 for 26 tasks; seed 3 for cartpole-balance and cartpole-swingup-sparse; seed 2 for cartpole-swingup and walker-walk-backwards. Selection is therefore conditional on compatibility with the loader, not a random sample of released seeds. Compatibility may be associated with training-code version or other release-history differences that also affect evidence, return or the policy-prior distribution; we did not establish whether such differences exist or their direction. The single-task comparisons, and all comparisons under P because the selected single-task checkpoint also sets the behaviour centre, describe these compatibility-selected checkpoints.

The run was paused once, at the operator's request, after the planner episodes of the 19M model had finished. At that point all 22,080 units, 60 reference fits, 420 configurations and 40,320 scores were complete, as were 450 of the 900 return episodes, and no numerical job was running when the dispatcher was stopped. Before continuing, every stored unit, fit, score, return file and worker log was checked against its receipt. The continuation used a driver written after the freeze. It called the frozen code unchanged for every numerical operation and ran only the groups that had not started; no completed group was rerun, and every numerical file committed before the pause (151,621 files) was byte-identical afterwards. Each return episode resets every random generator from its own identifiers, so the remaining episodes do not depend on state carried across the pause; we did not run a second, uninterrupted run to confirm that they are identical. The final validation and the independent recomputation covered results from both sides of the pause. Apart from the pause, the run waited 52 times for memory or disk to become available (492 failed admission checks at five-minute intervals); waiting changes no computation.

Phase 2 elapsed time was 101 hours and 17 minutes, including the pause and waits. The phase-only resource-admission record gives 492 five-minute failed checks, or 41 hours of recorded recheck allowances. That total is the count times the five-minute interval. The run's completion marker was written before the final publication step.

All seven forecasts held. One outcome per forecast does not validate its probability. The return is the five-episode mean of the whole planning agent.

\paragraph{Prespecified reading rules (P8 protocol v4).}

Every statement names the tasks counted, the action condition, h = 1, TD-MPC2, and the exploratory status, and says that the three P7 tasks suggested the question.

- Association (A1, A5). Report all task values, the positive/negative/zero/NA counts and their reasons, and the E4 protocol’s return sensitivity. Any statement of association refers to these observed five-episode returns and calibrated scores within the specified model family and task. If the positive-tau threshold is not met, state the count and denominator; do not translate it into a universal claim that evidence cannot order control performance. No count establishes that evidence measures control quality or that a size effect is caused by better dynamics.
- HRLE (A3). Report eligible, HRLE, unclassifiable and unavailable cells/tasks, by scorer and condition, with the prescribed sensitivity results. A permitted sentence is: ``In these specified tasks under U, N tasks had at least one model whose observed five-episode mean planner return was at least 90\% of the observed six-model best, with that best at least 100 raw reward units, and whose TS\_C mean log-evidence rate was at most one quarter of this fitted inverse-dynamics reference's rate.'' Give the resolved and unresolved task denominators. Use ``at most'', consistently with \textless{}=. Do not call this an estimate of prevalence, evidence of a poor dynamics model, or a confidence statement about expected return.
- Action conditions (A3, A4). Report forecast 3's result on its classifiable return-qualified ST tasks. Report forecast 5 on its own low-relative-evidence ST set B, identifying that this is not necessarily HRLE. If forecast 5 holds, say that at least half of the tasks in B crossed from rho\_U\textless{}=0.25 to rho\_P\textgreater{}=0.5 under the specified P law, with the exact numerator and denominator. If it fails, say that this threshold was not met; if unavailable, say why. Separately report how many U-HRLE cells remain low, cross the upper threshold, are intermediate or are unclassifiable in P, with their model identities. Always show numerator and reference changes and the ST behaviour/scorer asymmetry. Do not say that low evidence appeared off control-use states or disappeared near them: state occupancy, the action law and fitted reference/configuration all change, and the noisy prior is not the measured planner. Exactly one half is ``at least half'', not ``mostly''.
- Single-task against mt30 (A2). Report the separated pairs by direction, model and condition out of the 300, with adjusted values available in a table.
- Reference. Wherever rho is used, state the reference's budget and input, the number of cells with rho NA, and that the reference is not an upper bound.
- NULL. The E3 audit trigger and reporting rule (Appendix~\ref{app:J}).
- Nothing is described as confirmed. No sentence of this rule is applied to a reduced set as if it were the planned set.

Table~\ref{tab:v5-size} summarizes the model-level medians. The per-task and per-scorer rates, returns and tests are in the supplement.

{\footnotesize\setlength{\parskip}{0pt}
The full table is in the ancillary files (file \texttt{k-001.tsv}). The full table is in the ancillary files (file \texttt{k-002.tsv}). The full table is in the ancillary files (file \texttt{k-003.tsv}). The full table is in the ancillary files (file \texttt{k-004.tsv}). The full table is in the ancillary files (file \texttt{k-005.tsv}). The full table is in the ancillary files (file \texttt{k-006.tsv}). The full table is in the ancillary files (file \texttt{k-007.tsv}). The full table is in the ancillary files (file \texttt{k-008.tsv}). The full table is in the ancillary files (file \texttt{k-009.tsv}). The full table is in the ancillary files (file \texttt{k-010.tsv}). 

The full table is in the ancillary files (file \texttt{k-011.tsv}). The full table is in the ancillary files (file \texttt{k-012.tsv}). The full table is in the ancillary files (file \texttt{k-013.tsv}). The full table is in the ancillary files (file \texttt{k-014.tsv}). The full table is in the ancillary files (file \texttt{k-015.tsv}). The full table is in the ancillary files (file \texttt{k-016.tsv}). The full table is in the ancillary files (file \texttt{k-017.tsv}). The full table is in the ancillary files (file \texttt{k-018.tsv}). The full table is in the ancillary files (file \texttt{k-019.tsv}). The full table is in the ancillary files (file \texttt{k-020.tsv}). 

The full table is in the ancillary files (file \texttt{k-021.tsv}). The full table is in the ancillary files (file \texttt{k-022.tsv}). The full table is in the ancillary files (file \texttt{k-023.tsv}). The full table is in the ancillary files (file \texttt{k-024.tsv}). The full table is in the ancillary files (file \texttt{k-025.tsv}). The full table is in the ancillary files (file \texttt{k-026.tsv}). The full table is in the ancillary files (file \texttt{k-027.tsv}). The full table is in the ancillary files (file \texttt{k-028.tsv}). The full table is in the ancillary files (file \texttt{k-029.tsv}). 
\par}

\subsection{Dated forecasts and outcomes}
\begingroup\footnotesize\setlength{\tabcolsep}{2pt}
\begin{longtable}{@{}>{\raggedright\arraybackslash}p{0.085\linewidth}>{\raggedright\arraybackslash}p{0.665\linewidth}>{\raggedright\arraybackslash}p{0.11\linewidth}>{\raggedright\arraybackslash}p{0.09\linewidth}@{}}
\toprule
Forecast & Prespecified event & Probability & Outcome\\
\midrule\endfirsthead
\toprule
Forecast & Prespecified event & Probability & Outcome\\
\midrule\endhead
\bottomrule\endlastfoot
E4.1 & Condition U, A1: tau-b is positive in at least 18 of the 30 tasks. & 55\% & held\\
E4.2 & Condition U, A3: the ST scorer is HRLE in at least 5 tasks. & 55\% & held\\
E4.3 & A3: the ST scorer is HRLE in fewer tasks under P than under U. & 65\% & held\\
E4.4 & Condition U, A5: G of mt30-317M exceeds G of mt30-1M in at least 24 of the 30 tasks. & 65\% & held\\
E4.5 & A4: among the cells of the ST scorer with rho[\ensuremath{\cdot},U] \textless{}= 0.25 (at least one such cell being required), at least half have rho[\ensuremath{\cdot},P] \textgreater{}= 0.5. & 40\% & held\\
E4.6 & A2: in condition P the ST scorer is separated above mt30-5M (adjusted p \textless{} 0.01, ST larger) in at least 15 of the 30 tasks. & 50\% & held\\
E4.7 & NULL: in every scorer and condition panel pooled over its tasks, each e-value statistic's upper Binomial(n, 0.05) tail at its observed count is at least 0.001, and likewise PRseq's. & 80\% & held\\
\end{longtable}
\endgroup

%% file: appendix/R_references.tex
\section{References and mixed laws}\label{app:L}

The main comparison is in \S\ref{sec:v5-sizes}.

Before execution, the protocol was examined by a separate automated reviewer, a language-model agent run in its own session, and revised until that reviewer passed it.

The final protocol fixed the fresh and reused groups and the comparison budgets before evaluation. The reused-group comparison is descriptive. The run interruptions and the shared- and separate-configuration results are recorded below.

The study used fresh policy-centred units, descriptive comparisons on reused units, and fresh trajectories alternating between action laws. Five of seven forecasts held. The independent computation covered 34,752 score archives.

One process loss remained unexplained. A second loss followed accidental closure of the run console. An operator-requested pause and restart followed. Launch and verification scripts had errors before numerical work. No completed group was rerun, and the frozen definitions, settings and gates were unchanged.

The first reference comparison on fresh units is primary. Reused-unit comparisons are descriptive and conditional on the focal cases selected after E4. The Gaussian quadratic reweighting is heuristic. The mixed design conditions on the recorded type of the current draw, with earlier coins changing later histories.

\paragraph{Prespecified reading rules (P9 protocol v3).}

Every statement names the tasks, the law, h = 1, TD-MPC2 and the exploratory status, and says that P8E suggested the question.

- Q2, primary (Q2F). For each of the six focal tasks classify mt30-317M as strictly above the maximum G over all six references, tied at that maximum, or below it, on the fresh units. Report all six reference values and all maximizing identities, including ties; list unavailable tasks without removing them from the denominator. If strictly above in N tasks, the permitted sentence is: "Under the policy-centred law, on fresh units, in N of the six tasks the 317M model had a higher evidence rate than every reference we evaluated (tasks; references and budgets)." Add that this does not show that a world model is needed or that no reference could match it. Where a reference matched or exceeded it, say "a reference matched or exceeded the world model on fresh units" and name it; say "strictly exceeded" only for a strict inequality. Observed equality is not equivalence; failure to separate is not evidence of equal performance. A selected maximum is descriptive and has no selected e-value interpretation.
- Q2R. Reported beside Q2F with the E5 protocol’s qualification for reused units. If Q2F and Q2R classifications differ in a task, report both; Q2F is primary.
- Mechanism. The protocol’s motivating disclosure may be stated as the reason for R1, R2 and R3c. Results are stated only as counts: "Gaussian quadratic reweighting raised G in N of 9 tasks"; "the reference trained under U had a higher G on P units than the one trained under P in N of 9 tasks". No improvement is attributed causally to one changed component, and the hypothesis is not called established.
- Q1.
  \textgreater{} Permitted: "In exploratory TD-MPC2 horizon-one evaluation on mixed-law trajectories in [task], [scorer] had equal-unit TS\_C rates G\textasciicircum{}U=X and G\textasciicircum{}P=Y. The current law was randomized independently of the pre-action history at each clock. The realized state samples can differ, earlier laws affect later states, and the equal-unit rates use random stratum denominators. The result describes these mixed-law trajectories." Also report H, G\_origin, state balance and configuration differences. Do not say that the realized difference cannot be due in part to sampled states or their weighting.

  \textgreater{} Report primary and common-configuration G\textasciicircum{}U, G\textasciicircum{}P, Delta, H\textasciicircum{}U and H\textasciicircum{}P side by side for every scorer and task, with common minus primary differences. For G\textasciicircum{}U, G\textasciicircum{}P and Delta, explicitly identify every change among positive, zero and negative signs. Retain negative and zero values without clipping or ratio imputation. Do not use a factor-of-two disclosure trigger or select which analysis to emphasize by its result.

  Not permitted: that the model is wrong about off-policy actions in general; that the U-to-P change of P8E is explained by this result; any statement about planning or return.
- \textgreater{} Report the compatibility-selected ST seed wherever ST or its behaviour centre is used, and state that the ST centre and scorer share a checkpoint. Counts describe fixed tasks sharing bodies and dynamics, with no population-prevalence, planner-use, return, pretraining-necessity or general model-quality claim.
- NULL. The E3 audit trigger and reporting rule (Appendix~\ref{app:J}).
- Nothing is described as confirmed. No sentence is applied to a reduced set as if it were the planned set.

\begingroup\footnotesize\setlength{\tabcolsep}{2pt}
\begin{longtable}{@{}>{\raggedright\arraybackslash}p{0.43\linewidth}>{\raggedright\arraybackslash}p{0.16\linewidth}>{\raggedright\arraybackslash}p{0.17\linewidth}>{\raggedright\arraybackslash}p{0.16\linewidth}@{}}
\toprule
Focal task & 317M & R3 (larger Gaussian IDM) & R4 (critic)\\
\midrule\endfirsthead
\toprule
Focal task & 317M & R3 (larger Gaussian IDM) & R4 (critic)\\
\midrule\endhead
\bottomrule\endlastfoot
walker-stand & 1.03 & 1.23 & 2.07\\
walker-walk & 0.98 & 1.15 & 2.22\\
walker-run & 1.16 & 1.40 & 2.51\\
walker-walk-backwards & 1.47 & 1.61 & 2.70\\
walker-run-backwards & 0.83 & 1.00 & 2.04\\
reacher-three-easy & 3.60 & 4.68 & 4.65\\
\end{longtable}
\endgroup

{\footnotesize\setlength{\parskip}{0pt}
The full table is in the ancillary files (file \texttt{l-001.tsv}). The full table is in the ancillary files (file \texttt{l-002.tsv}). The full table is in the ancillary files (file \texttt{l-003.tsv}). The full table is in the ancillary files (file \texttt{l-004.tsv}). The full table is in the ancillary files (file \texttt{l-005.tsv}). The full table is in the ancillary files (file \texttt{l-006.tsv}). The full table is in the ancillary files (file \texttt{l-007.tsv}). The full table is in the ancillary files (file \texttt{l-008.tsv}). The full table is in the ancillary files (file \texttt{l-009.tsv}). The full table is in the ancillary files (file \texttt{l-010.tsv}). 

The full table is in the ancillary files (file \texttt{l-011.tsv}). The full table is in the ancillary files (file \texttt{l-012.tsv}). The full table is in the ancillary files (file \texttt{l-013.tsv}). The full table is in the ancillary files (file \texttt{l-014.tsv}). The full table is in the ancillary files (file \texttt{l-015.tsv}). The full table is in the ancillary files (file \texttt{l-016.tsv}). The full table is in the ancillary files (file \texttt{l-017.tsv}). 
\par}

\paragraph{Reference population.} The primary Q2F classification uses the maximum over all six references, R0, R1, R2, R3, R3c and R4. Forecast E5.3 uses the maximum over R1, R2, R3, R3c and R4. R3c shares R3's fit but uses a different score. See l-001.tsv for the values of all six references, including R0, the maximizing identity, and the fresh and reused comparisons. R4 maximized the fresh reference rate in each walker task and R3c maximized it in reacher-three-easy.

R0 reused E4’s policy-centred IDM. R1 used E4’s uniform-law fit without retraining. R2 matched R0’s small budget on newly generated P training data. R3 and R4 used 1,024 P units and 80,000 updates. R3c reused R3’s fit. See P9 protocol v3 \S{}3 for the optimization, held-out and calibration splits.

Q2F is the primary comparison on fresh P units. Q2R reuses E4's P units. The mixed-law Q1 analysis uses its own primary stratum-specific configurations and a common-configuration sensitivity.

\subsection{Dated forecasts and outcomes}
\begingroup\footnotesize\setlength{\tabcolsep}{2pt}
\begin{longtable}{@{}>{\raggedright\arraybackslash}p{0.085\linewidth}>{\raggedright\arraybackslash}p{0.665\linewidth}>{\raggedright\arraybackslash}p{0.11\linewidth}>{\raggedright\arraybackslash}p{0.09\linewidth}@{}}
\toprule
Forecast & Prespecified event & Probability & Outcome\\
\midrule\endfirsthead
\toprule
Forecast & Prespecified event & Probability & Outcome\\
\midrule\endhead
\bottomrule\endlastfoot
E5.1 & R4 has a higher G than R0 in at least 7 of the 9 tasks. & 80\% & held\\
E5.2 & R2 has a higher G than R0 in at least 5 of the 6 focal tasks. & 70\% & held\\
E5.3 & In at least 4 of the 6 focal tasks, the largest G among R1, R2, R3, R3c and R4 is at least mt30-317M's G. & 65\% & held\\
E5.4 & R3 has a G above mt30-317M's in at most 2 of the 6 focal tasks. & 60\% & failed\\
E5.5 & Q1: in at least 3 of acrobot-swingup, pendulum-swingup, cartpole-balance and cartpole-swingup, ST's G\textasciicircum{}U is at most one quarter of its G\textasciicircum{}P. & 60\% & failed\\
E5.6 & Q1: in at least 3 of those four tasks, CC(M)'s G\textasciicircum{}U is at least its G\textasciicircum{}P. & 55\% & held\\
E5.7 & NULL: no primary e-value NULL panel triggers the audit. & 90\% & held\\
\end{longtable}
\endgroup

%% file: appendix/S_occupancy.tex
\section{Occupancy levels}\label{app:M}

The main comparison is in \S\ref{sec:v5-trajectories}.

Before execution, the protocol was examined by a separate automated reviewer, a language-model agent run in its own session, and revised until that reviewer passed it.

Observations shifted with q in the locomotion control tasks as well.

There were three failed resource-admission checks at five-minute intervals. Corrections dealt with preflight vendor scope, complete prefreeze validation invariants, and a dispatch import before numerical work. The independent recomputation covered 21,888 score archives. Its largest absolute difference was 4.656612873077393e-10.

\paragraph{Prespecified reading rules (P10 protocol v3).}

Every statement names the tasks, h = 1, TD-MPC2, the exploratory status, and that P8E and P9 suggested the question.

- [R1-1] Permitted, per scorer and task: ``In this exploratory TD-MPC2 horizon-one evaluation, with P8E and P9 having suggested the question, the equal-unit TS\_C rates at P-selected origins were X0, X1, X2, X3, X4 nats per selected origin at nominal uniform-assignment probabilities q=0, 0.10, 0.25, 0.50, 0.75. The same conditional P kernel, fitted scorer configuration and reference were used across levels.'' Report G\_ref and rho, the U series, G\_origin, H, origin counts, realized fractions, the specific-configuration sensitivity and the state/action diagnostics alongside these values.
- [R1-1] If the unrounded G series is non-increasing with at least one strict decrease, it may be described as an observed non-increasing series with a decrease as nominal q increased. A constant series is described as constant. State that the population fixed-clock contrast with current law held fixed operates through prior history, while realized rate differences also reflect sampling variation and the equal-unit estimator's random-denominator weighting. Do not identify a responsible feature of history, an effect magnitude established by a test, general off-distribution model error, planning or return.
- [R1-3] Always report the complete G\_ref series beside each scorer's G and rho, including zero and negative reference rates and values below the rho floor. CC(pool) is one finite, mixture-trained and mixture-configured fitted discriminator, not an information bound or an equally accurate optimum at every level. Its approximation, optimization, selection and configuration quality can vary across the level/stratum populations. If both rates decline, describe both observed declines. Similar proportional declines may be described numerically when ratios are defined, but they do not establish a decline in intrinsic transition information, a common cause, or a world-model-specific cause or its absence. No ``roughly constant'' threshold controls disclosure.
- The P8E numbers may be shown beside L000 and L100 as historical context with their different reference named; equality or difference with them is not interpreted. P10 at q = 0.50 is not a reproduction of P9's mixed-law units and is not compared with them as one.
- NULL: the E3 audit rule (Appendix~\ref{app:J}). Nothing is described as confirmed. No sentence is applied to a reduced set as if it were the planned set.

{\footnotesize\setlength{\parskip}{0pt}
The full table is in the ancillary files (file \texttt{m-001.tsv}). The full table is in the ancillary files (file \texttt{m-002.tsv}). The full table is in the ancillary files (file \texttt{m-003.tsv}). The full table is in the ancillary files (file \texttt{m-004.tsv}). The full table is in the ancillary files (file \texttt{m-005.tsv}). The full table is in the ancillary files (file \texttt{m-006.tsv}). The full table is in the ancillary files (file \texttt{m-007.tsv}). The full table is in the ancillary files (file \texttt{m-008.tsv}). The full table is in the ancillary files (file \texttt{m-009.tsv}). The full table is in the ancillary files (file \texttt{m-010.tsv}). 

The full table is in the ancillary files (file \texttt{m-011.tsv}). The full table is in the ancillary files (file \texttt{m-012.tsv}). The full table is in the ancillary files (file \texttt{m-013.tsv}). The full table is in the ancillary files (file \texttt{m-014.tsv}). The full table is in the ancillary files (file \texttt{m-015.tsv}). The full table is in the ancillary files (file \texttt{m-016.tsv}). The full table is in the ancillary files (file \texttt{m-017.tsv}). 
\par}

\subsection{Dated forecasts and outcomes}
\begingroup\footnotesize\setlength{\tabcolsep}{2pt}
\begin{longtable}{@{}>{\raggedright\arraybackslash}p{0.085\linewidth}>{\raggedright\arraybackslash}p{0.665\linewidth}>{\raggedright\arraybackslash}p{0.11\linewidth}>{\raggedright\arraybackslash}p{0.09\linewidth}@{}}
\toprule
Forecast & Prespecified event & Probability & Outcome\\
\midrule\endfirsthead
\toprule
Forecast & Prespecified event & Probability & Outcome\\
\midrule\endhead
\bottomrule\endlastfoot
E6.1 & ST: G\textasciicircum{}P is non-increasing in q over the five levels in at least 3 of the 4 pole tasks. & 55\% & held\\
E6.2 & ST: G\textasciicircum{}P(L050) is positive and at most half of G\textasciicircum{}P(L000), with G\textasciicircum{}P(L000) \textgreater{} 0, in at least 3 of the 4 pole tasks. & 65\% & held\\
E6.3 & CC(pool): G\textasciicircum{}P(L050) is at least half of G\textasciicircum{}P(L000) in at least 3 of the 4 pole tasks. & 65\% & held\\
E6.4 & ST: rho\textasciicircum{}P(L000) is at least 0.5 in at least 2 of the 4 pole tasks. & 45\% & held\\
E6.5 & mt30-317M: G\textasciicircum{}P(L050) is at most half of G\textasciicircum{}P(L000) in at most 2 of the 4 pole tasks. & 50\% & failed\\
E6.6 & In the two control tasks, ST's ratio G\textasciicircum{}P(L050) / G\textasciicircum{}P(L000) is larger than its ratio in at least 3 of the 4 pole tasks, for both controls. & 45\% & held\\
E6.7 & No primary e-value NULL panel triggers the audit. & 90\% & held\\
\end{longtable}
\endgroup

%% file: appendix/T_fresh_trajectories.tex
\section{Fresh units and trained forward models}\label{app:N}

The main comparison is in \S\ref{sec:v5-trajectories}.

Before execution, the protocol was examined by a separate automated reviewer, a language-model agent run in its own session, and revised until that reviewer passed it.

The first forward-model fit stopped before its manifest was written. An authorized continuation redid that fit from the same seed. The selected weights and preprocessing matched the partial output byte for byte; the original selected step and held-out curve were unavailable for comparison. No completed group was rerun. The run recorded 29 failed resource-admission checks. Independent recomputation covered 40,704 score archives and 110 panel decisions. The largest absolute difference was 1.9786114080488915e-8.

The historical software check reproduced the stored E6 scores exactly on one fixed unit per task and scorer before generation. Two counted forecast cases have negative rates, D-pol in cartpole-balance at $q=0.5$ ($-0.017$) and D-uni in acrobot-swingup at $q=0$ ($-0.216$).

\paragraph{Prespecified reading rules (P11 protocol v3).}

Every interpretation names the tasks, h = 1, the exploratory status, and that P8E, P9 and P10 suggested the questions. Apply P10's descriptive reading rule to each complete within-stratum common-configuration series. With the current action kernel fixed, the fixed-clock population contrast across q operates through prior history. Realized equal-unit rate differences also reflect sampling variation and random-denominator weighting. Use TD-MPC2 language for ST and mt30-317M and forward-model language for the D arms. Report specific-configuration sensitivity only for the D arms, with its frozen primary reference. The following Q1 and Q2 rules govern the permitted summaries.

- [R1-5] For Q1, report P10 and P11 primary results and signed P11-minus-P10 differences for the same scorer, task, stratum and level, using the committed P10 decision and report. Both share tasks, checkpoints, simulator, policy centres and numerical procedure. If F1--3 hold, say ``The three prespecified P11 midpoint forecasts held on fresh units with the frozen P10 procedure,'' and give their numerical results and qualifying tasks. Describe positive retention or proportional loss only where both inputs are positive. Show both control tasks and all complete series with the same prominence. This repeats evaluation on new random draws from the same design. It does not replicate P10 training, calibration, specific sensitivity or other tasks, models or horizons.
- [R1-5] For Q2, report all four arms, both strata, the diagnostics and sign conventions together. If F4--6 hold, report ``D(pol) met the half-baseline decline criterion in [named tasks], D(mix) met the half-baseline lower bound in [named tasks], and D(mix) had the higher midpoint rate in all four pole tasks.'' Report slopes and ratios directly, with the overlap of qualifying tasks. Otherwise report the observed pattern with equal prominence. Every interpretation names the tasks, horizon one, exploratory status and prior-study motivation. The results do not identify the cause of the TD-MPC2 pattern, a responsible state feature, intrinsic transition information, comparative TD-MPC2 model quality, general prediction error, planning or return.

{\footnotesize\setlength{\parskip}{0pt}
The full table is in the ancillary files (file \texttt{n-001.tsv}). The full table is in the ancillary files (file \texttt{n-002.tsv}). The full table is in the ancillary files (file \texttt{n-003.tsv}). The full table is in the ancillary files (file \texttt{n-004.tsv}). The full table is in the ancillary files (file \texttt{n-005.tsv}). The full table is in the ancillary files (file \texttt{n-006.tsv}). The full table is in the ancillary files (file \texttt{n-007.tsv}). The full table is in the ancillary files (file \texttt{n-008.tsv}). The full table is in the ancillary files (file \texttt{n-009.tsv}). The full table is in the ancillary files (file \texttt{n-010.tsv}). 

The full table is in the ancillary files (file \texttt{n-011.tsv}). The full table is in the ancillary files (file \texttt{n-012.tsv}). The full table is in the ancillary files (file \texttt{n-013.tsv}). The full table is in the ancillary files (file \texttt{n-014.tsv}). The full table is in the ancillary files (file \texttt{n-015.tsv}). The full table is in the ancillary files (file \texttt{n-016.tsv}). The full table is in the ancillary files (file \texttt{n-017.tsv}). The full table is in the ancillary files (file \texttt{n-018.tsv}). The full table is in the ancillary files (file \texttt{n-019.tsv}). The full table is in the ancillary files (file \texttt{n-020.tsv}). 

The full table is in the ancillary files (file \texttt{n-021.tsv}). The full table is in the ancillary files (file \texttt{n-022.tsv}). The full table is in the ancillary files (file \texttt{n-023.tsv}). The full table is in the ancillary files (file \texttt{n-024.tsv}). The full table is in the ancillary files (file \texttt{n-025.tsv}). The full table is in the ancillary files (file \texttt{n-026.tsv}). The full table is in the ancillary files (file \texttt{n-027.tsv}). The full table is in the ancillary files (file \texttt{n-028.tsv}). The full table is in the ancillary files (file \texttt{n-029.tsv}). The full table is in the ancillary files (file \texttt{n-030.tsv}). 

The full table is in the ancillary files (file \texttt{n-031.tsv}). The full table is in the ancillary files (file \texttt{n-032.tsv}). 
\par}

\paragraph{Training budgets.} D-pol, D-uni and D-mix use 29,000 transitions per task and equal updates. D-all uses 174,000 transitions with the same update budget and is the sensitivity arm; see the ancillary files.

\subsection{Dated forecasts and outcomes}
\begingroup\footnotesize\setlength{\tabcolsep}{2pt}
\begin{longtable}{@{}>{\raggedright\arraybackslash}p{0.085\linewidth}>{\raggedright\arraybackslash}p{0.665\linewidth}>{\raggedright\arraybackslash}p{0.11\linewidth}>{\raggedright\arraybackslash}p{0.09\linewidth}@{}}
\toprule
Forecast & Prespecified event & Probability & Outcome\\
\midrule\endfirsthead
\toprule
Forecast & Prespecified event & Probability & Outcome\\
\midrule\endhead
\bottomrule\endlastfoot
E7.1 & ST: G\textasciicircum{}P(L050) is positive and at most half of G\textasciicircum{}P(L000), with G\textasciicircum{}P(L000) \textgreater{} 0, in at least 3 of the 4 pole tasks. & 85\% & held\\
E7.2 & CC(pool): G\textasciicircum{}P(L050) is at least half of G\textasciicircum{}P(L000) in all 4 pole tasks. & 80\% & held\\
E7.3 & mt30-317M: G\textasciicircum{}P(L050) is at most half of G\textasciicircum{}P(L000) in at least 3 of the 4 pole tasks. & 75\% & held\\
E7.4 & D(pol): G\textasciicircum{}P(L050) is at most half of G\textasciicircum{}P(L000), with G\textasciicircum{}P(L000) \textgreater{} 0, in at least 3 of the 4 pole tasks. & 60\% & failed\\
E7.5 & D(mix): G\textasciicircum{}P(L050) is at least half of G\textasciicircum{}P(L000), with G\textasciicircum{}P(L000) \textgreater{} 0, in at least 3 of the 4 pole tasks. & 50\% & failed\\
E7.6 & D(mix) has a larger G\textasciicircum{}P(L050) than D(pol) in all 4 pole tasks. & 65\% & held\\
E7.7 & D(pol) has a larger G\textasciicircum{}P(L000) than D(uni) in at least 3 of the 4 pole tasks. & 65\% & failed\\
E7.8 & No trigger in the 56 primary e-value NULL panels, each with its own statistic's rejection counts. & 88\% & held\\
\end{longtable}
\endgroup

%% file: appendix/U_trajectory_uncertainty.tex
\section{E8: trajectory-bootstrap intervals for E6 and E7}\label{app:v5-e8}

\paragraph{Record.} This CPU-only exploratory reanalysis used stored results from P10 at \texttt{a9150cd9} and P11 at \texttt{f83df188}. The protocol was \path{P13_PROTOCOL_v4.md}. Its SHA-256 digest was\newline\texttt{78f55695da09ebab3aed97998c43bfc1236a5713287550282c93f920cb8c0206}.\newline Design reviews v1, v2 and v3 requested fixes (\texttt{c71c40d6}, \texttt{50ed77e4} and \texttt{cd4fe23f}), and design review v4 passed before execution (\texttt{f8e2791e}). The freeze commit was \texttt{0f54d015} and the final commit was \texttt{72d6c282}. \path{CORRECTION_01.md} normalized equivalent missing-value codes in the verifier input layer only, with no scientific deviation.

\paragraph{Method.} Let $A_i$ be trajectory $i$'s sum of log factors at the P-selected origins, $N_i$ its count of such origins, and $n$ the number of planned trajectories in the cell. The equal-unit estimand $G$ averages $A_i/N_i$ over trajectories, with a realized empty trajectory contributing zero. The origin-weighted $G_o$ is $\sum_i A_i/\sum_i N_i$ when the denominator is positive. The fixed-clock $G_f$ is $\sum_i A_i/[n\cdot500(1-q)]$, where $1-q$ is the probability of a P step. It targets the equal-clock mean conditional P rate under the evaluated rollout design.

Whole trajectories were drawn with replacement independently within each source, task and level cell. Each cell used 10,000 replicates with indices shared across scorers on those units. The 2.5th and 97.5th percentiles give descriptive, pointwise 95\% intervals, with differences and ratios evaluated within each replicate. A ratio or difference-of-ratios interval is available only if its observed point and all replicates are defined and the lower endpoint of each relevant baseline interval is positive. Defined signed point estimates are retained even when their intervals are unavailable. A signed estimate and the direction of its interval are reported separately.

\paragraph{Dated forecasts.} These five E8 forecasts were dated 28 November 2026, after the committed E6 and E7 results were known. All five use E7 results.
\begin{enumerate}
\item All four pole-task ST G change intervals have upper endpoint $<$ 0. Probability 80\%. Outcome: held.
\item At least three pole-task ST minus CC(pool) G change intervals have upper endpoint $<$ 0. Probability 75\%. Outcome: held.
\item At least three pole-task mt30-317M G change intervals have upper endpoint $<$ 0. Probability 75\%. Outcome: held.
\item At least three control intervals for ST or mt30-317M G retention contain 1 or have lower endpoint $>$ 0.8. Probability 70\%. Outcome: held.
\item In every pole task the span of ST point retention across G, $G_o$ and $G_f$ is $<$ 0.1. Probability 70\%. Outcome: held.
\end{enumerate}

{\scriptsize
\setlength{\tabcolsep}{3pt}
\renewcommand{\arraystretch}{1.20}
\begin{longtable}{@{}lcccc@{}}
\caption{E7 equal-unit evidence rates, midpoint changes and retention ratios. Brackets enclose pointwise 95\% trajectory-bootstrap intervals. Rates and changes are in nats per origin.}\label{tab:v5-e8-endpoints}\\
\toprule
Scorer & $G$ at L000 & $G$ at L050 & L050 minus L000 & Retention ratio\\
\midrule\endfirsthead
\toprule
Scorer & $G$ at L000 & $G$ at L050 & L050 minus L000 & Retention ratio\\
\midrule\endhead
\midrule\multicolumn{5}{r@{}}{Continued on next page}\\\endfoot
\bottomrule
\multicolumn{5}{@{}l@{}}{\textsuperscript{a} Signed retention. The observed L050 rate is negative and its pointwise interval includes zero.}\\
\endlastfoot
\multicolumn{5}{@{}l}{\textbf{acrobot-swingup}}\\*
ST & $1.28\,[1.06,\,1.49]$ & $0.18\,[0.14,\,0.23]$ & $-1.10\,[-1.32,\,-0.88]$ & $0.14\,[0.10,\,0.19]$\\
mt30-317M & $0.69\,[0.55,\,0.83]$ & $0.08\,[0.05,\,0.10]$ & $-0.62\,[-0.76,\,-0.47]$ & $0.11\,[0.07,\,0.16]$\\
CC(pool) & $2.33\,[2.14,\,2.52]$ & $1.66\,[1.52,\,1.80]$ & $-0.67\,[-0.91,\,-0.43]$ & $0.71\,[0.63,\,0.80]$\\
D(pol) & $0.55\,[0.44,\,0.65]$ & $0.07\,[0.03,\,0.11]$ & $-0.48\,[-0.59,\,-0.37]$ & $0.13\,[0.05,\,0.22]$\\
D(mix) & $0.81\,[0.66,\,0.95]$ & $0.30\,[0.21,\,0.38]$ & $-0.51\,[-0.68,\,-0.34]$ & $0.37\,[0.25,\,0.51]$\\
\addlinespace[2pt]
\multicolumn{5}{@{}l}{\textbf{pendulum-swingup}}\\*
ST & $2.23\,[2.13,\,2.32]$ & $0.56\,[0.52,\,0.60]$ & $-1.67\,[-1.77,\,-1.56]$ & $0.25\,[0.23,\,0.27]$\\
mt30-317M & $2.21\,[2.10,\,2.32]$ & $0.54\,[0.51,\,0.57]$ & $-1.67\,[-1.79,\,-1.56]$ & $0.24\,[0.23,\,0.26]$\\
CC(pool) & $3.97\,[3.89,\,4.05]$ & $2.84\,[2.81,\,2.87]$ & $-1.13\,[-1.21,\,-1.04]$ & $0.72\,[0.70,\,0.73]$\\
D(pol) & $2.92\,[2.83,\,3.00]$ & $1.82\,[1.78,\,1.85]$ & $-1.10\,[-1.19,\,-1.01]$ & $0.62\,[0.60,\,0.65]$\\
D(mix) & $3.29\,[3.20,\,3.37]$ & $2.26\,[2.23,\,2.29]$ & $-1.03\,[-1.12,\,-0.93]$ & $0.69\,[0.67,\,0.71]$\\
\addlinespace[2pt]
\multicolumn{5}{@{}l}{\textbf{cartpole-balance}}\\*
ST & $2.94\,[2.92,\,2.95]$ & $0.17\,[-0.04,\,0.39]$ & $-2.77\,[-2.98,\,-2.54]$ & $0.06\,[-0.01,\,0.13]$\\
mt30-317M & $2.30\,[2.29,\,2.30]$ & $0.58\,[0.36,\,0.82]$ & $-1.71\,[-1.94,\,-1.47]$ & $0.25\,[0.15,\,0.36]$\\
CC(pool) & $4.70\,[4.69,\,4.70]$ & $3.43\,[3.14,\,3.70]$ & $-1.27\,[-1.56,\,-1.00]$ & $0.73\,[0.67,\,0.79]$\\
D(pol) & $0.16\,[0.16,\,0.17]$ & $-0.02\,[-0.05,\,0.01]$ & $-0.18\,[-0.21,\,-0.15]$ & $-0.11\,[-0.29,\,0.08]$\textsuperscript{a}\\
D(mix) & $2.25\,[2.24,\,2.26]$ & $1.09\,[0.87,\,1.31]$ & $-1.16\,[-1.38,\,-0.94]$ & $0.49\,[0.39,\,0.58]$\\
\addlinespace[2pt]
\multicolumn{5}{@{}l}{\textbf{cartpole-swingup}}\\*
ST & $2.45\,[2.30,\,2.53]$ & $0.21\,[0.01,\,0.43]$ & $-2.23\,[-2.47,\,-1.97]$ & $0.09\,[0.01,\,0.18]$\\
mt30-317M & $2.10\,[1.99,\,2.16]$ & $0.60\,[0.49,\,0.72]$ & $-1.50\,[-1.64,\,-1.34]$ & $0.29\,[0.23,\,0.35]$\\
CC(pool) & $4.41\,[4.36,\,4.44]$ & $3.59\,[3.50,\,3.68]$ & $-0.82\,[-0.91,\,-0.72]$ & $0.81\,[0.79,\,0.84]$\\
D(pol) & $1.51\,[1.47,\,1.53]$ & $0.91\,[0.83,\,0.98]$ & $-0.60\,[-0.68,\,-0.52]$ & $0.60\,[0.55,\,0.65]$\\
D(mix) & $2.02\,[1.98,\,2.04]$ & $1.31\,[1.23,\,1.39]$ & $-0.71\,[-0.79,\,-0.62]$ & $0.65\,[0.61,\,0.69]$\\
\addlinespace[2pt]
\multicolumn{5}{@{}l}{\textbf{walker-walk}}\\*
ST & $0.51\,[0.43,\,0.58]$ & $0.43\,[0.35,\,0.50]$ & $-0.08\,[-0.19,\,0.02]$ & $0.84\,[0.66,\,1.05]$\\
mt30-317M & $1.00\,[0.87,\,1.12]$ & $1.04\,[1.00,\,1.09]$ & $0.04\,[-0.08,\,0.18]$ & $1.04\,[0.93,\,1.20]$\\
CC(pool) & $1.44\,[1.32,\,1.55]$ & $1.73\,[1.62,\,1.83]$ & $0.28\,[0.13,\,0.45]$ & $1.20\,[1.08,\,1.33]$\\
D(pol) & $0.60\,[0.50,\,0.69]$ & $0.64\,[0.57,\,0.71]$ & $0.05\,[-0.07,\,0.17]$ & $1.08\,[0.90,\,1.32]$\\
D(mix) & $0.61\,[0.51,\,0.70]$ & $0.76\,[0.70,\,0.83]$ & $0.15\,[0.04,\,0.27]$ & $1.25\,[1.06,\,1.51]$\\
\addlinespace[2pt]
\multicolumn{5}{@{}l}{\textbf{cheetah-run}}\\*
ST & $1.67\,[1.37,\,1.94]$ & $2.27\,[2.20,\,2.34]$ & $0.61\,[0.32,\,0.91]$ & $1.36\,[1.17,\,1.66]$\\
mt30-317M & $1.37\,[1.22,\,1.51]$ & $1.88\,[1.83,\,1.94]$ & $0.51\,[0.36,\,0.67]$ & $1.37\,[1.24,\,1.55]$\\
CC(pool) & $4.37\,[4.20,\,4.47]$ & $4.74\,[4.73,\,4.75]$ & $0.37\,[0.27,\,0.54]$ & $1.08\,[1.06,\,1.13]$\\
D(pol) & $2.25\,[2.03,\,2.39]$ & $2.53\,[2.48,\,2.58]$ & $0.28\,[0.13,\,0.51]$ & $1.13\,[1.05,\,1.25]$\\
D(mix) & $2.01\,[1.71,\,2.26]$ & $3.23\,[3.18,\,3.28]$ & $1.22\,[0.97,\,1.53]$ & $1.61\,[1.43,\,1.89]$\\
\end{longtable}
}

\begin{figure}[!htbp]
\centering
\includegraphics[width=\linewidth]{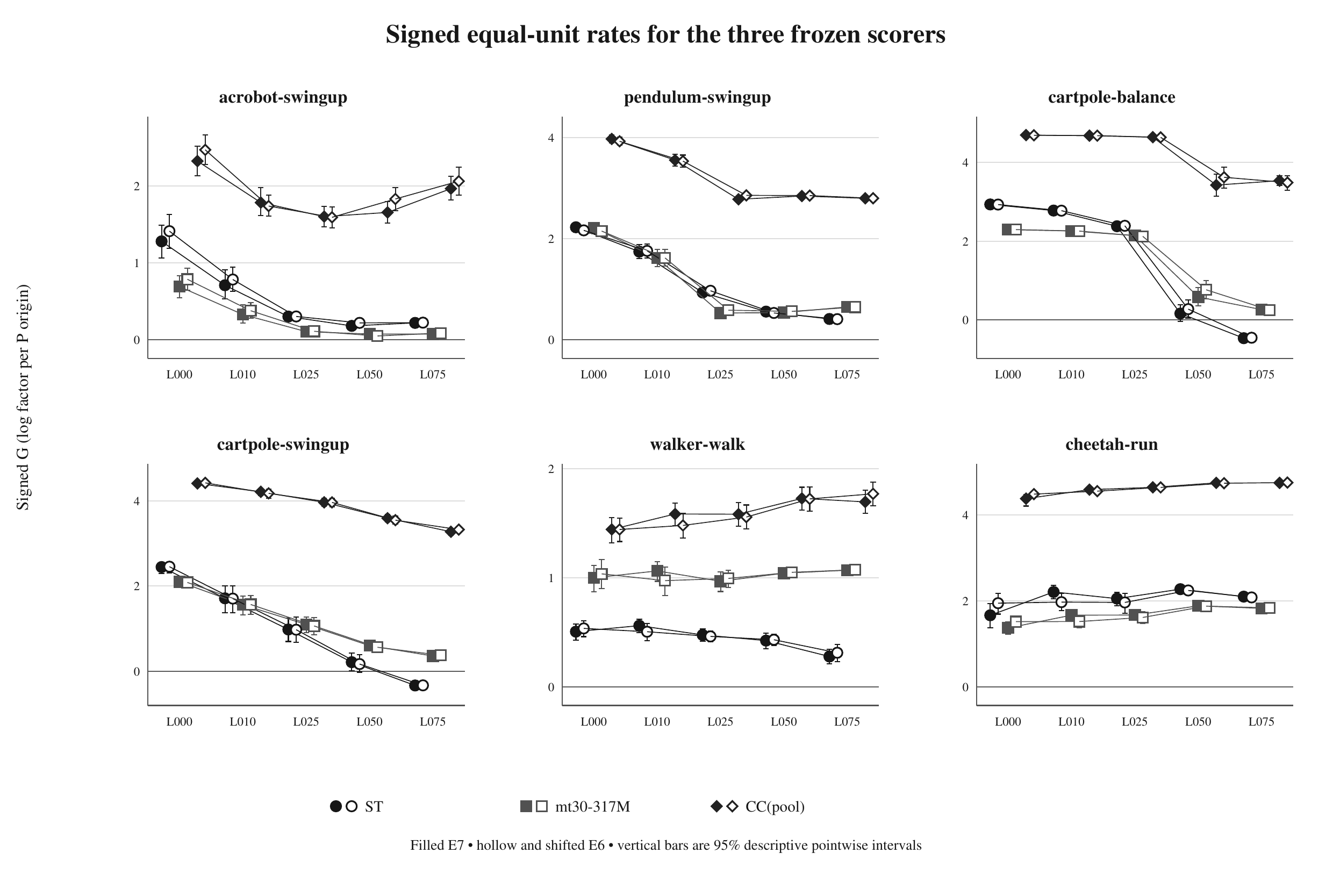}
\caption{Figure A of the P13 analysis. Signed equal-unit evidence rates and pointwise 95\% trajectory-bootstrap intervals for the three frozen scorers. Filled markers show E7 and horizontally offset hollow markers show E6. Each task has its own vertical scale and a zero reference line.}\label{fig:v5-e8-figure-a}
\end{figure}
\FloatBarrier

The ancillary files \texttt{u-001.tsv} through \texttt{u-005.tsv} contain the full rates, contrasts, forecast records and Figure A data.

%% file: sections/16_use_of_ai.tex
\section*{Use of AI tools}
We used Claude (Anthropic, through Claude Code and claude.ai) and Codex (OpenAI) during October and November 2026. Claude assisted in drafting protocols, forecasts and manuscript text. Codex assisted in implementing and running analyses, typesetting, and writing separate checks of reported results. Separate Codex sessions reviewed the E1–E8 protocols before the specified new analyses. Their reviews led to documented revisions in conditions for validity, comparator design and forecast wording. Earlier results informed some protocols, including analyses of stored results. The automated reviews assessed the protocol or draft identified in each record. The separate numerical checks assessed outputs from the corresponding analysis. The authors directed the research, authorized every study and every use of data, and take responsibility for the manuscript.